\pdfoutput=1

\documentclass[11pt]{article}

\usepackage[final]{acl}

\usepackage{times}
\usepackage{latexsym}
\usepackage{amsmath}
\usepackage{amssymb}

\usepackage{xcolor}

\usepackage[T1]{fontenc}

\usepackage[utf8]{inputenc}

\usepackage{microtype}

\usepackage{inconsolata}

\usepackage{graphicx}
\usepackage{subfig}
\usepackage{booktabs}
\usepackage{multirow}
\usepackage{array}
\usepackage{fdsymbol}

\title{Can Federated Learning Safeguard Private Data in LLM Training? \\Vulnerabilities, Attacks, and Defense Evaluation}

\author{%
   Wenkai Guo\textsuperscript{$\heartsuit$},\quad 
   Xuefeng Liu\textsuperscript{$\heartsuit \vardiamondsuit \dag$},\quad 
   Haolin Wang\textsuperscript{$\heartsuit$} \\
   \textbf{Jianwei Niu}\textsuperscript{$\heartsuit \diamondsuit \vardiamondsuit \dag$},\quad 
   \textbf{Shaojie Tang}\textsuperscript{$\spadesuit$}, \quad
   \textbf{Jing Yuan}\textsuperscript{$\clubsuit$}\\ 
   $\heartsuit$ State Key Laboratory of Virtual Reality Technology and Systems, \\ School of Computer Science and Engineering, Beihang University, Beijing, China \\ 
   $\diamondsuit$ Hangzhou Innovation Institute of Beihang University, Zhejiang Key Laboratory \\ of Industrial Big Data and Robot Intelligent Systems, Hangzhou, China \\
   $\spadesuit$ Center for AI Business Innovation, Department of Management Science and Systems, \\ University at Buffalo, Buffalo, New York, USA \\ 
   $\clubsuit$ University of North Texas, Denton, Texas, USA \\
   $\vardiamondsuit$ Zhongguancun Laboratory, Beijing, China \\
   \texttt{\{kyeguo, liu\_xuefeng, wanghaolin, niujianwei\}@buaa.edu.cn}\\
   \texttt{shaojiet@buffalo.edu}, \texttt{jing.yuan@unt.edu}
}

\begin{document}
\maketitle
\begin{abstract}
\vspace{-10pt}
Fine-tuning large language models (LLMs) with local data is a widely adopted approach for organizations seeking to adapt LLMs to their specific domains. Given the shared characteristics in data across different organizations, the idea of collaboratively fine-tuning an LLM using data from multiple sources presents an appealing opportunity. However, organizations are often reluctant to share local data, making centralized fine-tuning impractical. Federated learning (FL), a privacy-preserving framework, enables clients to retain local data while sharing only model parameters for collaborative training, offering a potential solution. While fine-tuning LLMs on centralized datasets risks data leakage through next-token prediction, the iterative aggregation process in FL results in a global model that encapsulates generalized knowledge, which some believe protects client privacy. 
In this paper, however, we present contradictory findings through extensive experiments.  We show that attackers can still extract training data from the global model, even using straightforward generation methods, with leakage increasing as the model size grows. Moreover, we introduce an enhanced attack strategy tailored to FL, which tracks global model updates during training to intensify privacy leakage. 
To mitigate these risks, we evaluate privacy-preserving techniques in FL, including differential privacy, regularization-constrained updates and adopting LLMs with safety alignment. Our results provide valuable insights and practical guidelines for reducing privacy risks when training LLMs with FL.
\vspace{31pt}
\end{abstract}

\section{Introduction}
\renewcommand{\thefootnote}{}
\footnotetext{The source code is available at: \texttt{\href{https://anonymous.4open.science/r/fling-llm-anonymous-B311}{fling-llm-anonymous}}}
\footnotetext{$\dag$ Corresponding Author.}
In recent years, the advancement of large language models (LLMs) \citep{kojima2022large, llama, gemini} has prompted many organizations to explore methods for fine-tuning LLMs on their own local data, enabling adaptation to specific domains \citep{bloomberggpt, medicalllm}. However, due to the limited availability of domain-specific data within individual organizations and the potential overlap of data across different entities, the concept of collaboratively fine-tuning LLMs using data from multiple organizations has emerged as a promising solution. Despite this, many organizations are reluctant to share data due to fears of data leakage, which makes the conventional approach of a central entity collecting and processing all data unacceptable. 

To address these privacy concerns, Federated Learning (FL) \citep{fedavg, fl_review, fl_review1} has gained significant attention as a distributed training paradigm that allows data owners to retain control over their local data. In FL, organizations upload only the locally updated model parameters, which are aggregated on a central server to form a global model. This process iterates until the global model converges, facilitating collaborative training while maintaining data privacy. The promise of enhanced data privacy has driven growing research into applying FL to LLMs \citep{federatedscope, fatellm}.

The generative nature of LLMs introduces significant privacy risks, as they can leak private information through the text they generate \citep{carlini2021extracting, huang2022large}. In contrast,  the global model trained in FL is generated by iteratively aggregating local models from different clients, and hence encapsulates only general knowledge \citep{fedrod, feddecomp} rather than specific information from individual clients. Therefore, the global model produced by FL can better protect sensitive training data from individual organizations. This belief has driven the widespread adoption of FL for LLM training \citep{zhang2024towards, openfedllm}.

\begin{figure}[!htbp]
\vspace{-15pt}
\setlength{\abovecaptionskip}{6pt}
\centering
\subfloat[ROUGE-L score]{
    \includegraphics[width=0.47\linewidth]{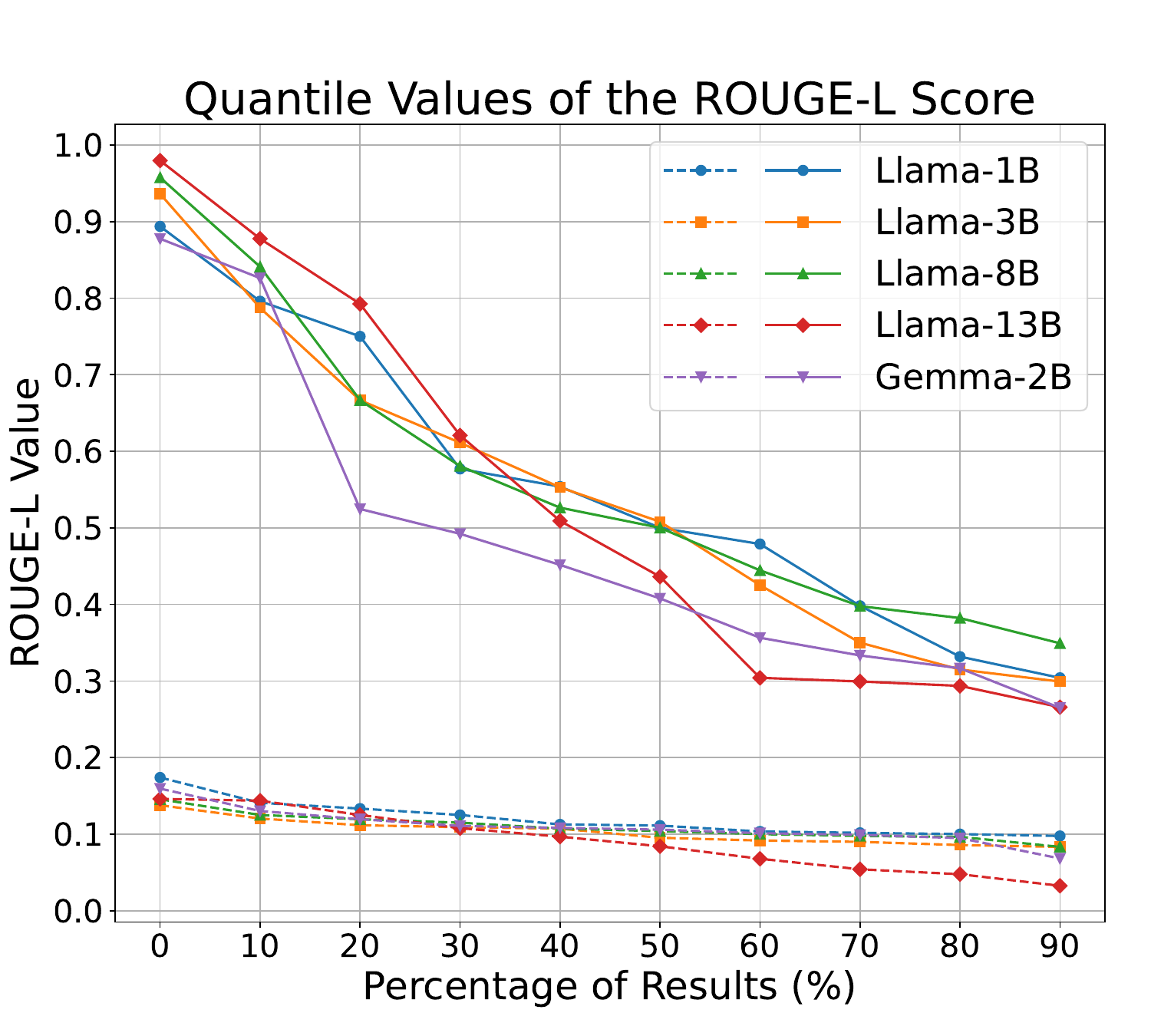}
} 
\subfloat[BERTScore]{
    \includegraphics[width=0.47\linewidth]{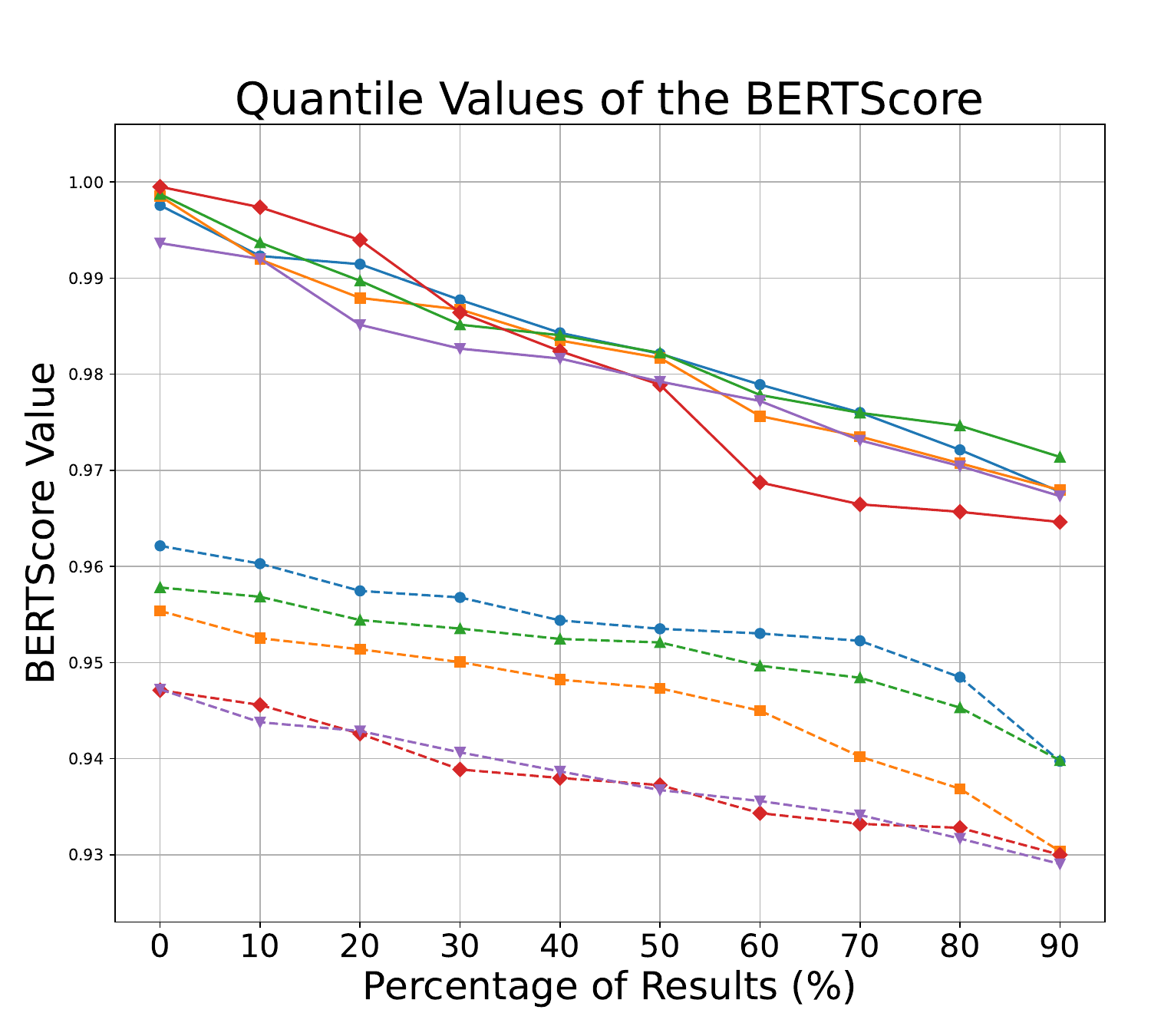}
}
\caption{Quantile plot of the similarity score in the top x\% results on the Enron-Email dataset. The dashed line and solid line represent privacy leakage before and after fine-tuning, respectively. The definitions of ROUGE-L and BERTScore can be referred to in Section~\ref{section:threat}.}
\label{fig:intro}
\vspace{-3pt}
\end{figure}

In this paper, we challenge this widely accepted belief and argue that \textbf{FL is not able to safeguard privacy in LLM Training}. We begin by introducing a basic attack strategy, wherein the global model randomly generates text and calculates the similarity between the generated content and the training data. The results are shown in Fig.~\ref{fig:intro}, which reveals a significant increase in similarity following training with FL, with 10\% of the samples exhibiting a similarity of over 90\% with the training dataset. Furthermore, we demonstrate that this risk escalates as the model size grows, suggesting that privacy leakage may be more pronounced in models with greater capabilities.

Moreover, we argue that if malicious organizations exist within FL training, they could track the iterative model updates, increasing the leakage risk. Building upon this, we propose an enhanced attack strategy specifically designed for FL, which utilizes the logit difference before and after training to rescale the token probabilities to better infer the training data. Our experiments with this advanced attack strategy show that the iterative nature of FL aggregation significantly increases the potential for data recovery, allowing attackers to reconstruct the training data with much greater precision. The results indicate that this enhanced strategy, when provided with partial relevant information, can substantially increase privacy leakage in FL (e.g., a 21.4\% increase in leakage for Llama-8B).

To address these emerging privacy risks, we evaluate several commonly used privacy-preserving techniques in FL, including differential privacy \citep{dp1}, update regularization methods \citep{fedprox, pfedsd} and adopting safety aligned models \citep{llm-align}. Our findings reveal that while these methods can reduce data leakage to some extent, they also lead to a degradation in training performance. This suggests that the current FL framework struggles to strike an optimal balance between data privacy and training efficacy, underscoring the need for the development of new algorithms.
Our contributions include:
\begin{itemize}
    \item We identify critical privacy vulnerabilities in LLM training with FL, challenging the widely held assumption that FL can adequately protect against data leakage.
    \item We propose a novel attack strategy that exploits the iterative nature of FL aggregation to amplify privacy risks.
    \item We evaluate existing defense mechanisms, analyzing their effectiveness and limitations in protecting privacy during FL training.
\end{itemize}

\section{Background}
\subsection{Large Language Models}
Large Language Models (LLMs) are advanced machine learning models designed to process and generate human-like text. In recent years, LLMs have made significant strides, excelling in a wide range of natural language processing (NLP) tasks while also proving highly adaptable for domain-specific applications \citep{bloomberggpt, medicalllm}. Among the most well-known LLMs are autoregressive models, such as GPT-3 \citep{kojima2022large}, Llama \citep{llama}, and Gemini \citep{gemini}. These models are trained to predict the next token in a sequence based on the preceding tokens, enabling them to generate coherent and contextually relevant text. This autoregressive framework has made LLMs particularly effective for a variety of applications, including text generation \citep{llm-gen}, machine translation \citep{llm-mt}, and question answering \citep{mmlu}.

\subsection{Data Privacy Protection}
\vspace{-4.5pt}
In the context of this paper, data privacy protection refers to the safeguarding of training data to prevent its exposure or misuse during the training of machine learning models, a concern that has garnered increasing attention \citep{privacy1, privacy2}. This issue becomes particularly critical when LLMs are trained on data containing confidential or personal information. For autoregressive LLMs, the challenge of privacy protection is heightened because these models can inadvertently "memorize" sensitive data during training, which could later be extracted by malicious actors. Research has demonstrated that LLMs trained on proprietary or personal data can inadvertently reproduce specific details from their training datasets \citep{carlini2021extracting, huang2022large}, posing a significant risk to industries handling sensitive information, such as healthcare \citep{llm-medicine}, finance \citep{llm-financial}, and law \citep{llm-law}.

\vspace{-5.5pt}
\subsection{Federated Learning}
\vspace{-4.5pt}
Federated Learning (FL) \citep{fedavg} is a distributed machine learning paradigm that allows multiple entities to collaboratively train a shared model without the need to exchange private data. Instead of aggregating raw data on a central server, FL enables participants to upload only their locally updated model parameters, which are then aggregated to form a global model. This approach helps to ensure that the global model encodes only general knowledge \citep{fedrod, feddecomp}, rather than specific, sensitive data from individual clients' datasets. As a result, FL has proven particularly effective in traditional machine learning tasks \citep{fedcv, fednlp}.

However, in the case of autoregressive LLMs, the nature of the training objective—predicting the next token in a sequence based on prior tokens—makes them inherently prone to memorizing and regenerating samples from the training data. Consequently, even with FL, the global model may still be vulnerable to the same privacy risks observed in centralized LLM training.

\vspace{-5pt}
\section{Hacking Methodology}
\vspace{-6pt}
\subsection{Threat Model}
\label{section:threat}
\vspace{-4.5pt}
To illustrate our attack methodology, we first define the threat scenario addressed in this paper. We focus on the case where one client maliciously attempts to extract private data from other clients, which is widely adopted in prior research and reflects risks in practical settings \citep{zhu2019deep}. In this framework, malicious clients can access the global model from the server during each global communication round but cannot access the local models of other clients. Hence, the attacker is able to track the updates of the global model and try to recover the training data from other clients.

To quantify the extent of data leakage, we propose to adopt the ROUGE-L \citep{lin2004rouge} and BERTScore \citep{bertscore} metrics to measure the similarity between the model's generated content and the original training data. Given that ROUGE is more focused on word-level similarity, we primarily report ROUGE-L scores in the main text. More detailed analysis and experimental are provided in the Appendix~\ref{appendix:metrics}.

For a given input sequence $\mathbf{X}$ and its corresponding ground-truth completion $\mathbf{Y}$, the model generates a predicted completion $\hat{\mathbf{Y}} = \mathcal{F}(\mathbf{X})$. We then compute the data leakage score for this sample using: $\mathcal{R}(\mathbf{X}, \hat{\mathbf{Y}}) = \mbox{ROUGE-L}(\hat{\mathbf{Y}}, \mathbf{Y}).$

\vspace{-5pt}
\subsection{Hacking Tasks}
\label{section:Task}
\begin{figure}[!htbp]
\vspace{-20pt}
\setlength{\abovecaptionskip}{2pt}
\centerline{\includegraphics[width=0.95\linewidth]{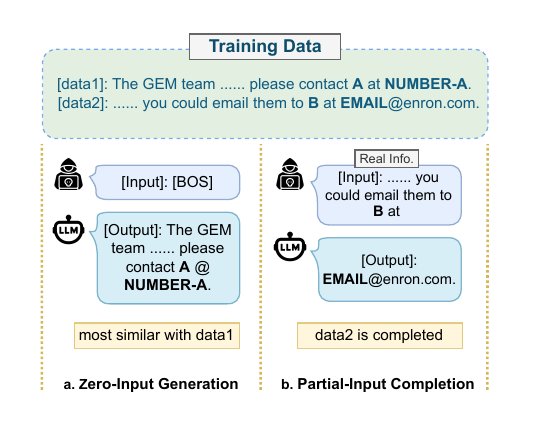}}
    \vspace{-12pt}
    \caption{Tasks for extracting local data.}
    \label{fig:task}
\vspace{-16pt}
\end{figure}

\vspace{-4.5pt}
To evaluate potential security risks, we examine two adversarial attack scenarios that reflect realistic hacking conditions, which is similar to prior work \citep{carlini2021extracting}. These tasks, illustrated in Fig.~\ref{fig:task}, model how attackers exploit system vulnerabilities under varying levels of prior knowledge (e.g., partial vs. zero access to training data). By simulating these scenarios, we systematically assess how different levels of adversarial insight impact the severity of privacy breaches. 

In addition to these two tasks, we design a more challenging attack scenario where attackers only has partial and vague knowledge about the training data. The details are shown in the Appendix~\ref{appendix:disturbed}.

\subsubsection{Tasks 1: Zero-Input Generation}
\vspace{-2pt}
In a highly restrictive scenario, we assume the attacker has no prior knowledge and can only rely on the model to randomly generate data in an attempt to reconstruct information related to the original training data. In this setup, during each global communication round in FL, we input an starting token "[BOS]" into the model and allow it to generate a complete text. This process is illustrated in Fig.~\ref{fig:task}(a). Additionally, we configure the model to perform random sampling to increase the diversity of the generated text. We refer to this task as "Zero-Input Generation." For this task, we allow the model to generate text $N=30$ times through random sampling. We then calculate the similarities between the generated samples and the client’s training data. For each generated sample, we select the training sample with the highest similarity as its match. If any generated sample shows a high similarity to the training data, we consider it a successful leakage.

\subsubsection{Tasks 2: Partial-Input Completion}
\vspace{-2pt}
In another scenario, we assume the attacker has partial knowledge of the client's training data and attempts to leverage this information to guide the model in generating specific details related to the original data. In this case, for each global model obtained during a round of FL, we select the data used by various clients in the previous round for local training. We then provide a portion of the original data as input to the model and ask it to complete the remaining text. This task is illustrated in Fig.~\ref{fig:task}(b). We refer to this task as "Partial-Input Completion." In this scenario, we randomly select data samples from the local datasets of clients trained in the previous round. For each sample, we provide the first $80\%$ of the data as input to the model, while the remaining serves as the ground truth containing private information. The model is then asked to complete the text based on the given input. We calculate the similarity between the model-generated portion and the ground truth. If the similarity is high, we consider it a successful leakage.

\begin{figure*}[!htbp]
    \vspace{-8pt}
    \setlength{\abovecaptionskip}{2pt}
    \centerline{\includegraphics[width=0.84\linewidth]{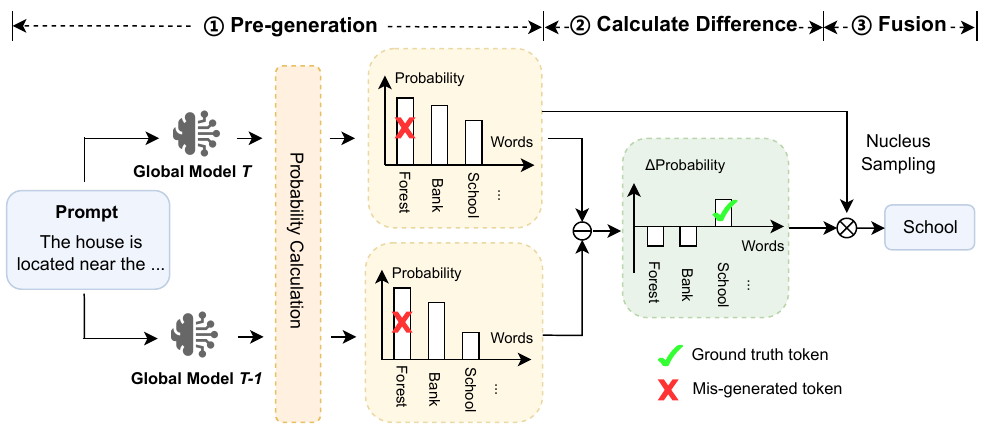}}
    \caption{Workflow of our proposed enhanced hacking scheme.}
    \label{fig:scheme}
    \vspace{-12pt}
\end{figure*}

\subsection{Hacking Schemes}
\vspace{-2pt}
Next, we introduce two hacking schemes to investigate the privacy vulnerabilities of LLMs fine-tuning via FL. These schemes aim to extract training data from the global model. We show that privacy leakage can be further exacerbated by exploiting the iterative aggregation process in FL.

\subsubsection{Basic Hacking Scheme}
\vspace{-2pt}
The basic attack scheme aims to reconstruct training data from the current global model by prompting it to recall client-specific information. Let the global model at training round $T$ be denoted as $\pi_T$. Given the current token sequence $\mathbf{X} = \{x_1, \cdots, x_n\}$, the next token predicted is sampled according to: $x_{n+1} \sim \text{Top-p}(\pi_{T}(\cdot \mid \mathbf{X})).$

This process follows standard nucleus sampling, where the autoregressive process continues iteratively until an end token "[EOS]" is generated.

\subsubsection{Enhanced Hacking Scheme}
\label{section:scheme}
\vspace{-2pt}
Building upon the basic attack scheme, we propose an enhanced attack method that exploits the fact that when a model learns textual data during a training round, the logits (prediction probabilities) for tokens related to that text should exhibit noticeable increases in the subsequent model iteration, even if their absolute values are not the highest yet.

Based on this intuition, we design a new attack scheme that utilizes the change in logits between two consecutive rounds to adjust the model's prediction for the next token. As illustrated in Fig.~\ref{fig:scheme}, the process of predicting the next token, given an input prompt, can be broken down into three steps:

\textbf{Pre-prediction.} Given an input prompt $\mathbf{X}$, the global models at rounds $T$ and $T-1$ generate predictions for the next token, resulting in probability distributions $\pi_T(\cdot \mid \mathbf{X})$ and $\pi_{T-1}(\cdot \mid \mathbf{X})$.

\textbf{Difference Calculation.} We calculate the difference between these two logits, highlighting the token with the most significant probability increase. The difference between $\pi_T(\cdot \mid \mathbf{X})$ and $\pi_{T-1}(\cdot \mid \mathbf{X})$ is computed as: $\Delta \pi_T(\cdot \mid \mathbf{X}) = \pi_T(\cdot \mid \mathbf{X}) - \pi_{T-1}(\cdot \mid \mathbf{X}).$

\textbf{Fusion.} The model's original prediction $\pi_T(\cdot \mid \mathbf{X})$ is adjusted using $\Delta \pi_T(\cdot \mid \mathbf{X})$. To reduce the impact of low-probability tokens, we apply sampling to $\pi_T(\cdot \mid \mathbf{X})$, producing $\tilde{\pi}_T(\cdot \mid \mathbf{X}) = \mbox{Top-p}(\pi_{T}(\cdot \mid \mathbf{X}))$. Then, $\Delta \pi_T(\cdot \mid \mathbf{X})$ is transformed into a weight vector using the softmax function: $w_T = \text{Softmax}\left(\frac{\Delta \pi_T(\cdot \mid \mathbf{X})}{\tau}\right)$, where $\tau$ is the temperature value. The final prediction is made by sampling from the adjusted distribution: $\pi_{new}(\cdot \mid \mathbf{X}) = w_T \odot \tilde{\pi}_T(\cdot \mid \mathbf{X})$, where $\odot$ refers to the element-wise multiplication between the weight vector and the probability of each token.

\section{Hacking Results}
\label{section:hack_result}
\subsection{Experiment Settings}

\textbf{Models.} 
For our study, we select the LLaMA-3 \citep{llama3} family of LLMs, Gemma‑2 \citep{team2024gemma}, and Qwen2.5 \citep{qwen2.5}. To explore the extent of leakage across different model sizes, we evaluate LLaMA-3 models with 1B, 3B, 8B, and 13B parameters, Gemma-2 with 2B parameters and Qwen2.5 with 7B parameters. The results of Gemma-2-2B and Qwen2.5-7B are provided in the Appendix~\ref{appendix:more}.

\noindent\textbf{Dataset.} We adopt three datasets (Enron-Email \citep{klimt2004enron}, Reddit-Comments \citep{baumgartner2020pushshift}, CLERC \citep{abe2024clerc}) for experiments. These three datasets focus on emails, social media comments and legal cases, respectively, covering a wide range of privacy information. A detailed dataset description is shown in Appendix~\ref{appendix:dataset_desc}. Due to space constraints, we only present results in Enron-Email dataset in the main paper. Other results are presented in the Appendix~\ref{appendix:more}.

\noindent\textbf{Training Settings.} During training, we set the number of clients to 4, and the FL process runs for 60 communication rounds. It is important to note that each client's \textbf{local data is used only once during training}, preventing overfitting. In each communication round, clients perform 200 local iterations using the AdamW optimizer. For smaller models, such as LLaMA-3.2-1B, LLaMA-3.2-3B, Gemma-2-2B and Qwen2.5-7B, the learning rate is fixed at 5e-5, while for larger models like LLaMA-3.1-8B and LLaMA-2-13B, the learning rate is set to 3e-5, in line with related research \citep{openfedllm}. Please refer to Appendix~\ref{appendix:args} for more details.

\subsection{Zero-Input Generation} 
\label{section:sim}
\subsubsection{Hacking Details}
We conduct experiments on the Zero-Input Generation task using models of different scales. In each global training round, after aggregating the local updates, we input a begin token into the global model to randomly generate 30 text samples. We then compute the similarity between each generated sample and the original training data, using the highest similarity value as the result of the attack.


\subsubsection{Experimental Results}
We first conduct experiments under the basic hacking scheme, and report the top 10\% mean, and top 30\% mean ROUGE scores for the generated samples at each round (see Appendix~\ref{appendix:thresholds} for results under more thresholds). The results in Fig.~\ref{fig:sim_mean} show that the model quickly learns to generate content highly similar to the training data after training begins. As training progresses, the model's ability to reconstruct the original training data even improves, approaching up to 100\% similarities without any prior knowledge. This observation holds true for models with various scales, which suggests that even without prior information, models can still reconstruct training data to a considerable extent.

\begin{figure}[t]
\setlength{\abovecaptionskip}{6pt}
\centering
\subfloat[Results on Llama-8B.]{
    \includegraphics[height=2.8cm, width=0.44\linewidth]{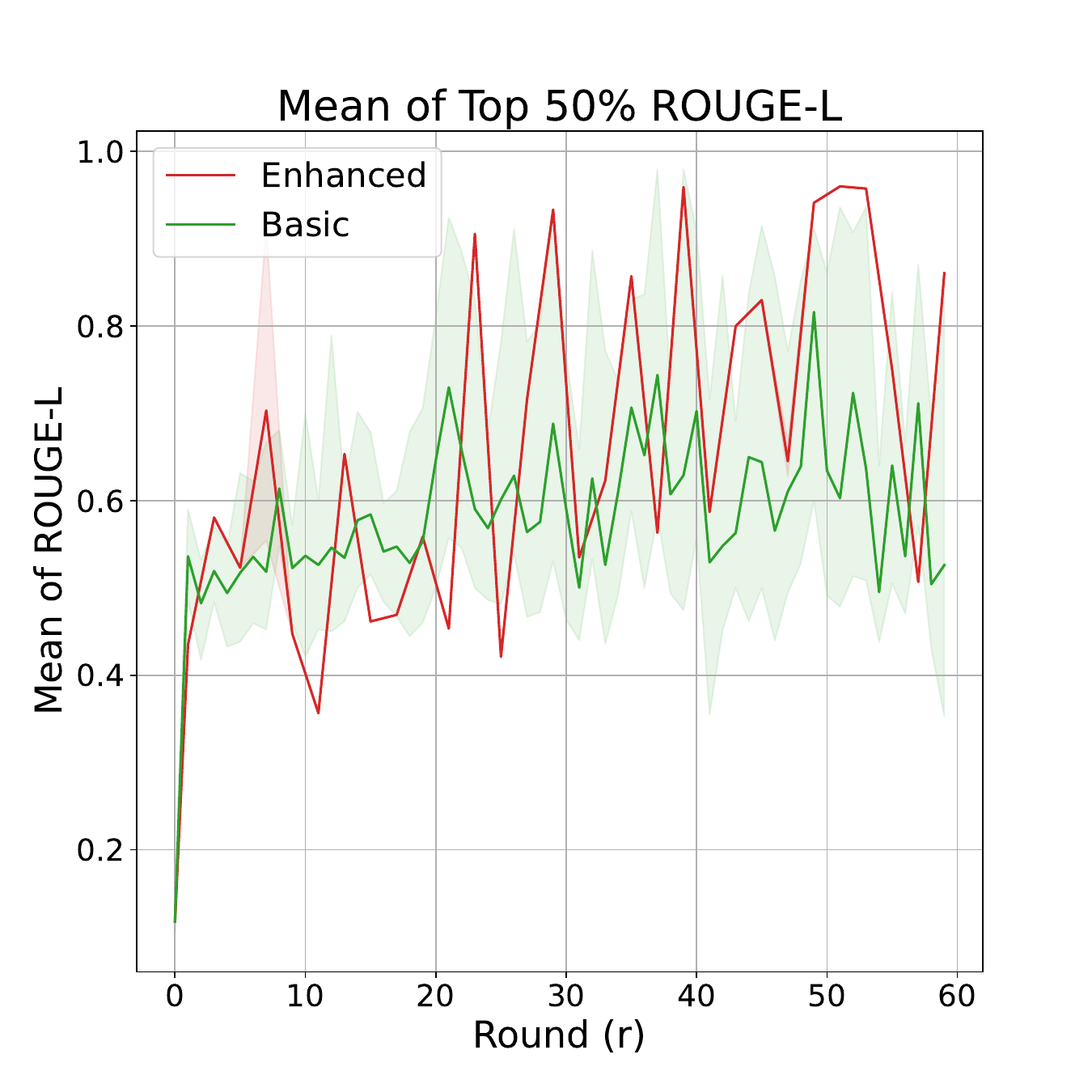}
    \label{fig:sim_diff_8b}
}
\subfloat[Results on Llama-13B.]{
    \includegraphics[height=2.8cm, width=0.44\linewidth]{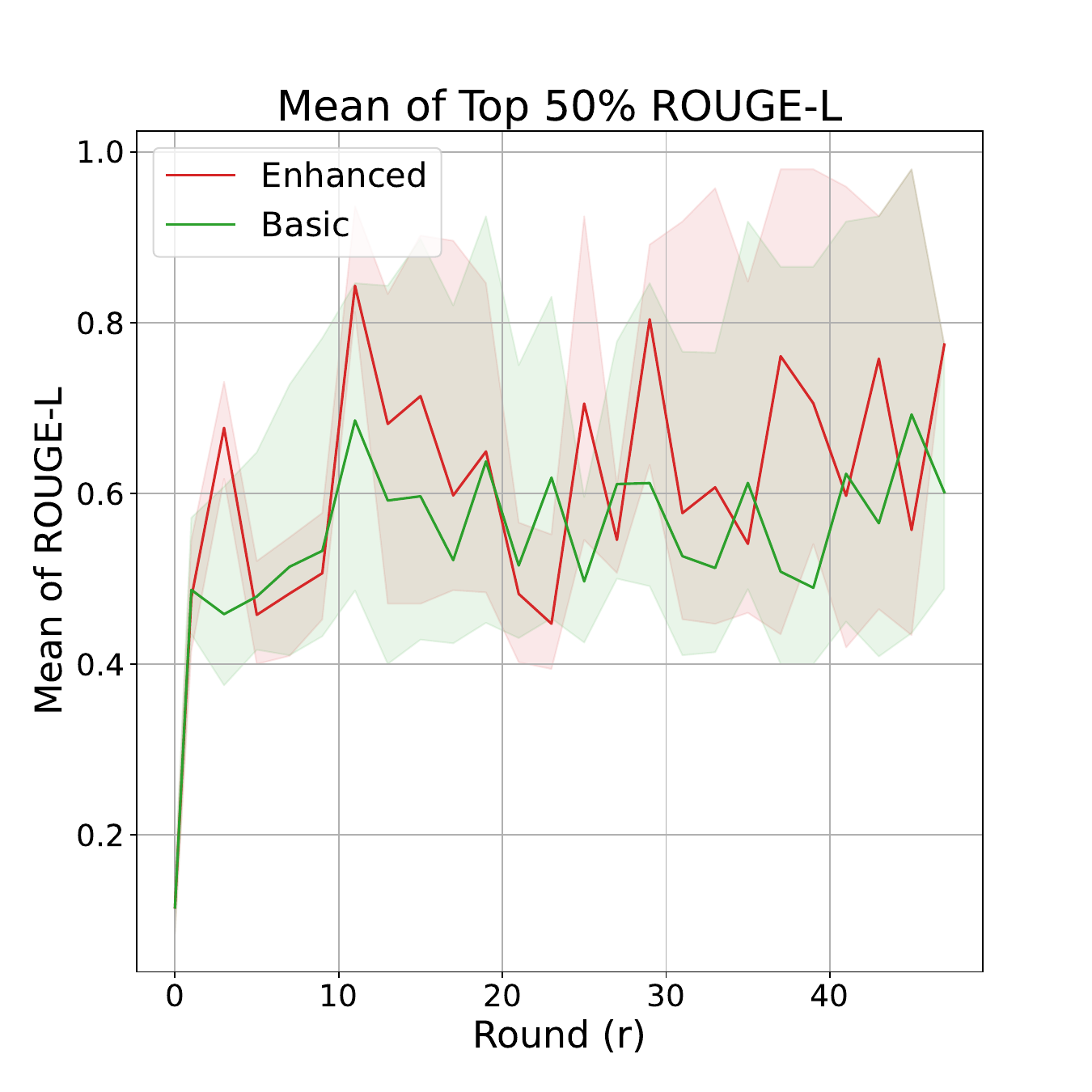}
    \label{fig:sim_diff_13b}
}
\caption{Comparison of two hacking schemes' results for Zero-Input Generation.}
\label{fig:sim_diff}
\vspace{-6pt}
\end{figure}

We further investigate the extent of privacy leakage using the enhanced attack scheme introduced in Section~\ref{section:scheme}. As illustrated in Fig.~\ref{fig:sim_diff}, the enhanced attack scheme significantly amplifies privacy leakage. On average, the leakage is approximately 10\% greater than that observed with the basic attack scheme, based on the maximum value across the entire period. This indicates a counter-intuitive phenomenon that, FL can be quite vulnerable due to its iterative process of uploading parameters.

\subsection{Partial-Input Completion}
\label{section:com}
\subsubsection{Hacking Details}
We demonstrate results for the Partial-Input Completion task in this section. In each global communication round, we randomly select 100 samples from the local training datasets as the attack dataset. For each original training sample, we provide the first 80\% as input and let the model complete the remaining part (Results when providing the first 30\% as input are shown in Appendix~\ref{appendix:input}). We then compute the similarity between the generated text and the original data as the attack result, reporting the top 10\% mean, and top 30\% mean scores for each round. 

\begin{figure}[t]
\setlength{\abovecaptionskip}{6pt}
\centering
\subfloat[Top 10\% ROUGE values.]{
    \includegraphics[height=2.8cm, width=0.44\linewidth]{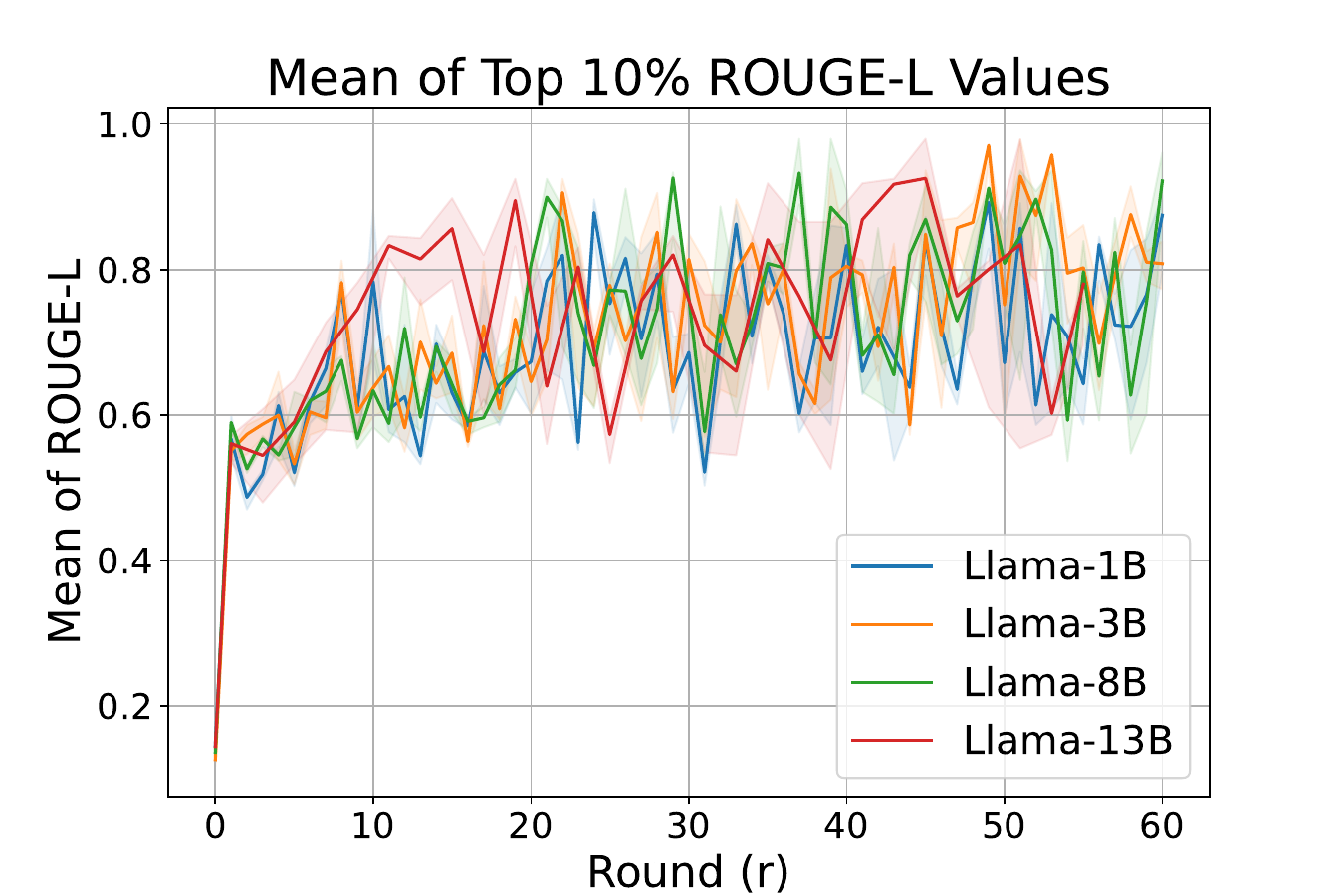}
    \label{fig:sim_mean_10_L}
}
\subfloat[Top 30\% ROUGE values.]{
    \includegraphics[height=2.8cm, width=0.44\linewidth]{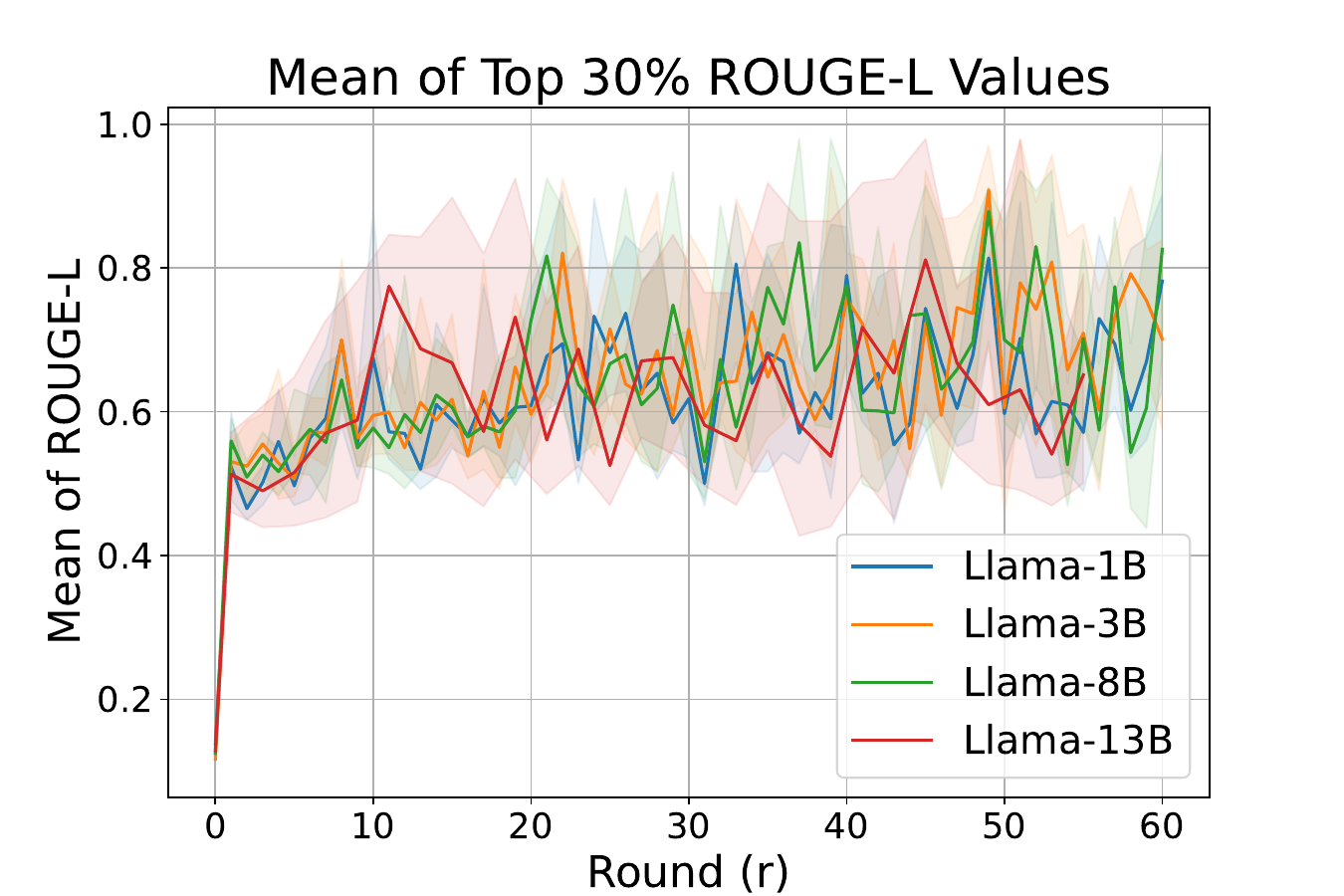}
    \label{fig:sim_mean_30_L}
}
\caption{Basic Hacking Scheme results of LLaMA models for Zero-Input Generation.}
\label{fig:sim_mean}
\vspace{-13pt}
\end{figure}

\begin{figure}[t]
\setlength{\abovecaptionskip}{6pt}
\centering
\subfloat[Top 10\% ROUGE values.]{
    \includegraphics[height=2.8cm, width=0.44\linewidth]{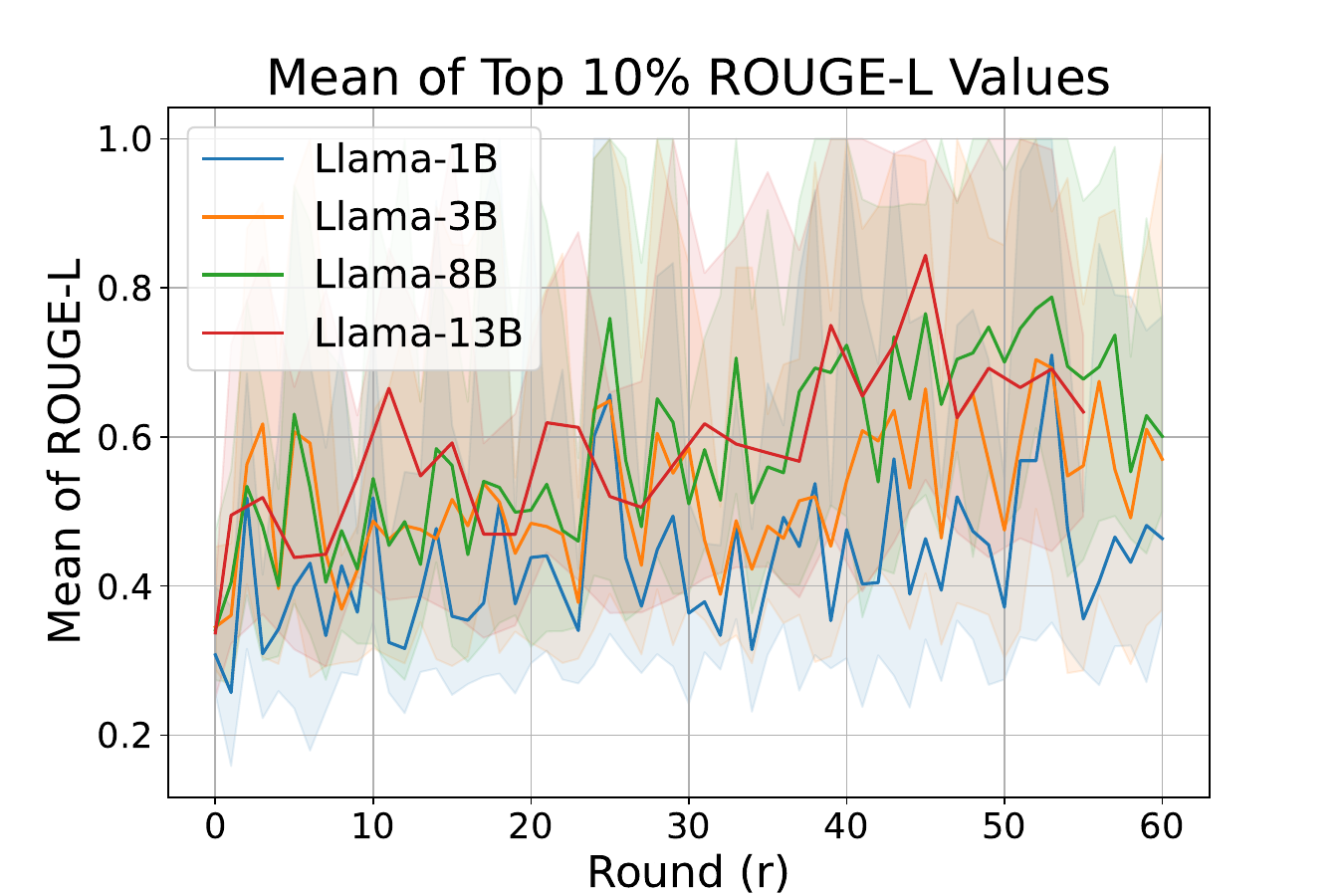}
    \label{fig:com_mean_10_L}
}
\subfloat[Top 30\% ROUGE values.]{
    \includegraphics[height=2.8cm, width=0.44\linewidth]{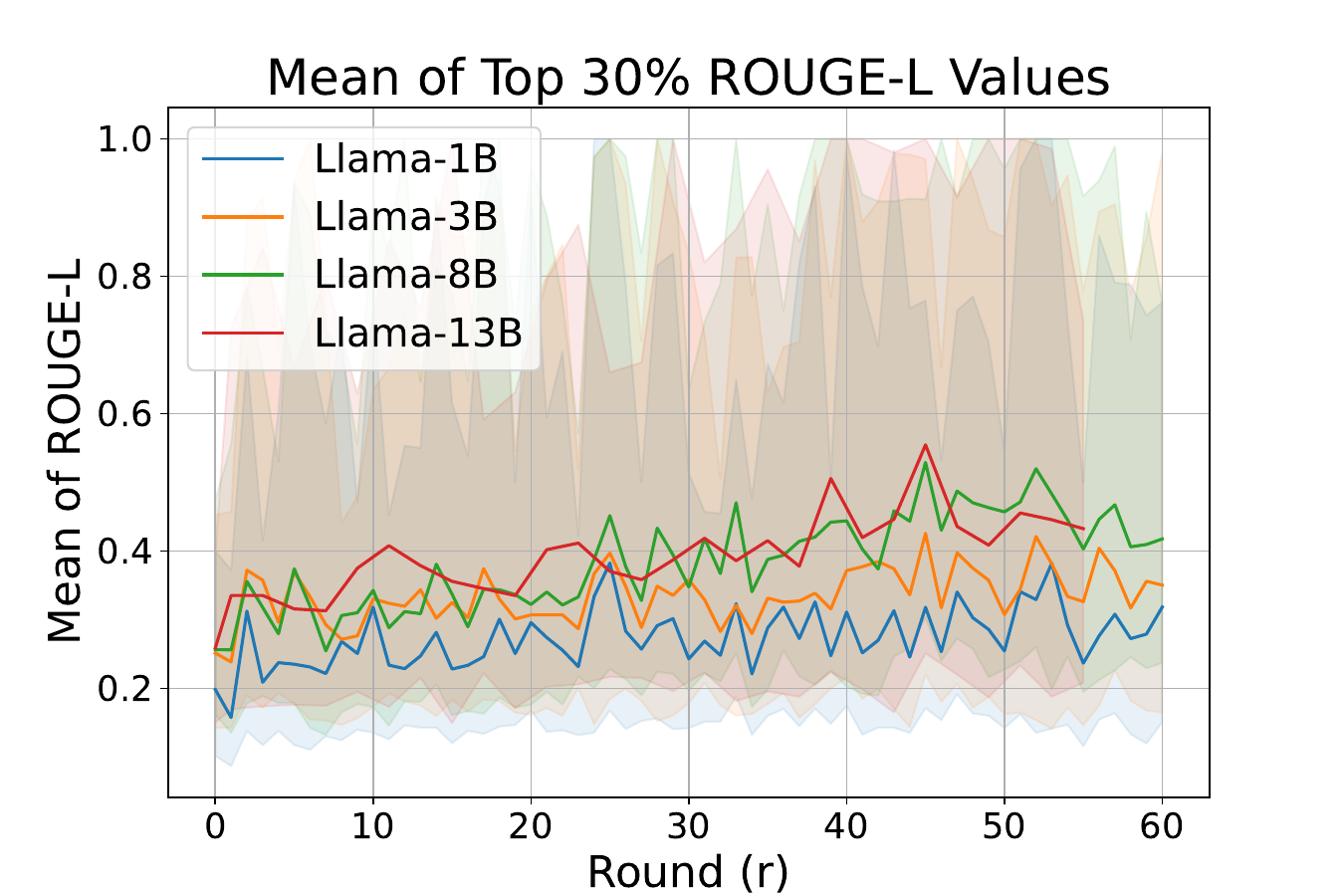}
    \label{fig:com_mean_30_L}
}
\caption{Hacking results of LLaMA models for Partial-Input Completion.}
\label{fig:com_mean}
\vspace{-6pt}
\end{figure}

\begin{figure}[t]
\vspace{-9pt}
\setlength{\abovecaptionskip}{6pt}
\centering
\subfloat[Results on Llama-8B.]{
    \includegraphics[height=2.8cm, width=0.44\linewidth]{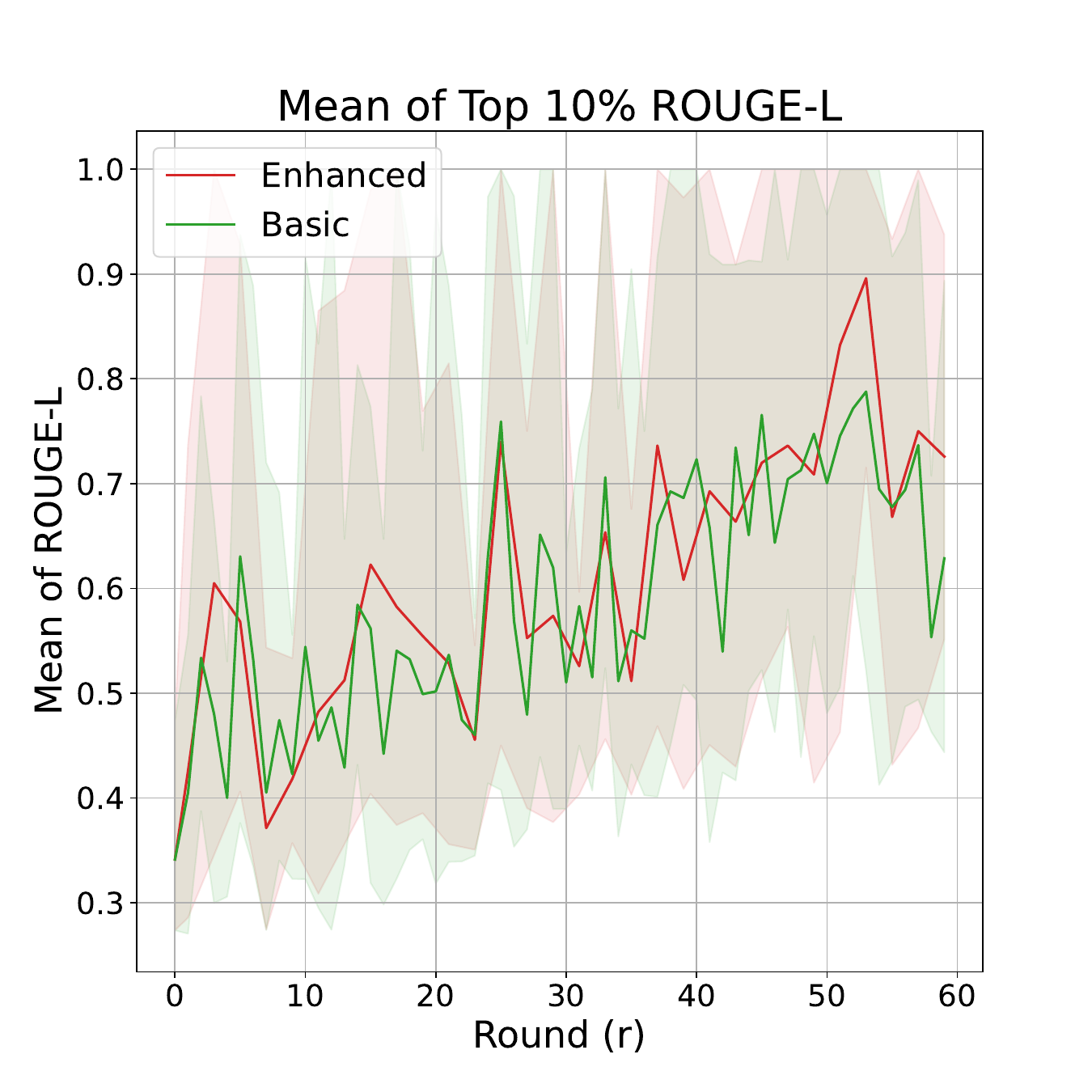}
    \label{fig:com_diff_8b}
}
\subfloat[Results on Llama-13B.]{
    \includegraphics[height=2.8cm, width=0.44\linewidth]{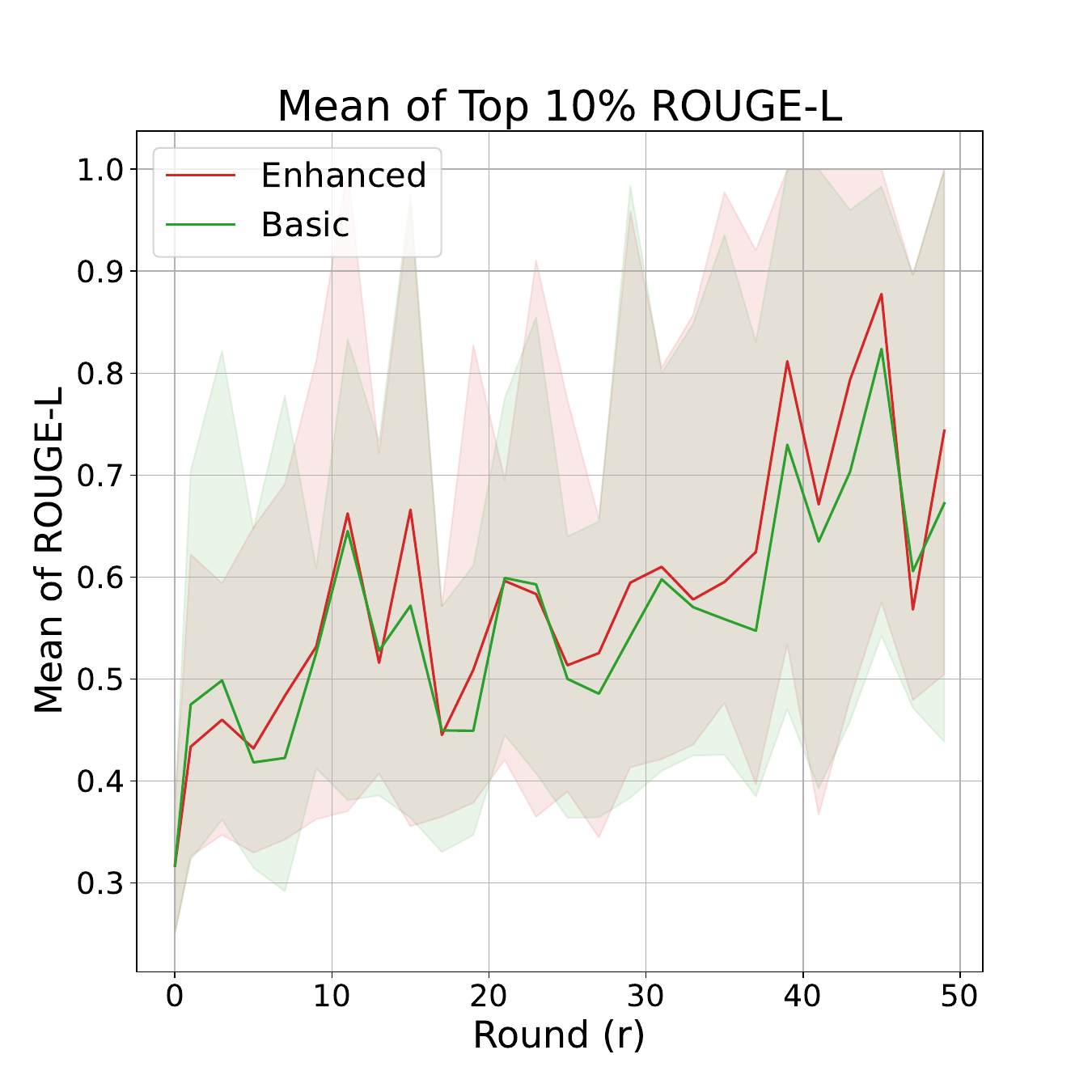}
    \label{fig:com_diff_13b}
}
\caption{Comparison of two hacking schemes' results for Partial-Input Completion.}
\label{fig:com_diff}
\vspace{-7pt}
\end{figure}

\subsubsection{Autonomous Evaluation Results}
The results from the basic attack scheme, illustrated in Fig.~\ref{fig:com_mean}, demonstrate that models begin to reproduce content highly similar to the original training data early in the training phase. As training progresses, this tendency becomes more pronounced, with an increasing proportion of reconstructed samples exhibiting faithful reconstruction of the original data. To help quantify this, we report the percentage of samples whose ROUGE score is above 0.95 (in Table~\ref{tab:percent} below), which achieve near-perfect reconstruction.

\begin{table}[!htbp]
    \vspace{-3pt}
    \small
    \centering
    \begin{tabular}{ccc}
    \toprule
    \textbf{Threshold (LlaMA-8B)} & \textbf{0.95} & \textbf{0.90} \\
    \midrule
    \textbf{1--10 (Round)}  & 0.4\% & 0.8\% \\
    \textbf{11--20} & 0.9\% & 1.3\% \\
    \textbf{21--30} & 1.2\% & 1.7\% \\
    \textbf{31--40} & 1.4\% & 1.6\% \\
    \textbf{41--50} & 1.6\% & 2.5\% \\
    \textbf{51--60} & 1.5\% & 2.3\% \\
    \bottomrule
    \end{tabular}
    \caption{Percentage of samples with ROUGE-L scores exceeding thresholds at different training rounds.}
    \label{tab:percent}
    \vspace{-10pt}
\end{table}

When comparing models of varying scales, we observe that larger models exhibit evidently greater susceptibility to privacy leakage. This phenomenon is correlated with downward trends in training loss, where larger models achieve lower loss values. We also report the percentage of samples with a ROUGE score above 0.95 in Table~\ref{tab:per_model}.

\begin{table}[!htbp]
    \small
    \centering
    \begin{tabular}{ccccc}
    \toprule
    \textbf{Model (LlaMA)} & \textbf{1B} & \textbf{3B} & \textbf{8B} & \textbf{13B} \\
    \midrule
    \textbf{1--10 (Round)} & 0.1\% & 0.4\% & 0.4\% & 0.4\% \\
    \textbf{11--20} & 0.5\% & 0.4\% & 0.9\% & 0.8\% \\
    \textbf{21--30} & 0.5\% & 0.9\% & 1.2\% & 1.1\% \\
    \textbf{31--40} & 0.4\% & 0.8\% & 1.4\% & 1.4\% \\
    \textbf{41--50} & 0.5\% & 1.1\% & 1.6\% & 1.8\% \\
    \textbf{51--60} & 1.0\% & 1.2\% & 1.5\% & 2.0\% \\
    \bottomrule
    \end{tabular}
    \caption{Percentage of samples with ROUGE-L scores exceeding 0.95 threshold.}
    \label{tab:per_model}
    \vspace{-4pt}
\end{table}

\textbf{Statistical test.} We further assess the statistical significance of this observation. To this end, we conduct two sets of paired t-tests under the same experimental configuration as Fig.~\ref{fig:com_mean}. One set involves paired t-tests on the average ROUGE scores of each round across LLaMA models of different scales (61 rounds, i.e., 61 samples), as shown in Table~\ref{tab:ttest1}. The other set consists of paired t-tests on the ROUGE scores of all hacking samples across all rounds for LLaMA models of different scales (61 rounds × 100 samples = 6100 total samples), as presented in Table~\ref{tab:ttest2}.

\begin{table}[!htbp]
    \small
    \centering
    \setlength{\tabcolsep}{1.1mm}
    \begin{tabular}{ccccc}
        \toprule
        & \textbf{1B} & \textbf{3B} (vs 1B) & \textbf{8B} (vs 3B) & \textbf{13B} (vs 8B) \\
        \midrule
        \textbf{t-statistic}  & N/A    & +17.4726   & +5.4644    & +2.3185    \\
        \textbf{p-value}      & N/A    & $<0.01$    & $<0.01$    & 0.0282     \\
        \textbf{mean}         & 0.1301 & 0.1622     & 0.1806     & 0.1907     \\
        \bottomrule
    \end{tabular}
    \caption{Paired t-Test Results for mean ROUGE Scores of each round (61 rounds) Across Llama Models.}
    \label{tab:ttest1}
    \vspace{-10pt}
\end{table}

\begin{table}[!htbp]
    \small
    \centering
    \setlength{\tabcolsep}{1.1mm}
    \begin{tabular}{ccccc}
        \toprule
        & \textbf{1B} & \textbf{3B} (vs 1B) & \textbf{8B} (vs 3B) & \textbf{13B} (vs 8B) \\
        \midrule
        \textbf{t-statistic}  & N/A    & +20.3498   & +10.2681   & +2.2812     \\
        \textbf{p-value}      & N/A    & $<0.01$    & $<0.01$    & 0.0226      \\
        \textbf{mean}         & 0.1301 & 0.1622     & 0.1806     & 0.1907      \\
        \bottomrule
    \end{tabular}
    \caption{Paired t-Test Results for ROUGE Scores of each hacking sample at each round (100×61=6100 samples) Across Llama Models.}
    \label{tab:ttest2}
    \vspace{-10pt}
\end{table}

The results of both paired tests consistently show a positive trend in ROUGE scores for larger models compared to smaller ones, with positive t-statistics indicating higher ROUGE scores in larger models. The p-values are all below 0.05, with some even below 0.01, which confirms that the conclusion that larger models generally have higher ROUGE scores is statistically significant.

The findings above suggest a practical dilemma: while enhanced model capacity improves data fitting, it concurrently amplifies the risk of sensitive data replication, posing a critical privacy vulnerability.

We also study the extent of increased privacy leakage using the enhanced hacking scheme. As shown in Fig.~\ref{fig:com_diff}, the enhanced hacking scheme further increases the leakage in FL by 13.4\% for Llama-8B. These results further highlight that the iterative nature of FL can cause more severe privacy leakage compared to the use of a single model.

\subsubsection{Human Evaluation}
We further conduct a \textbf{human evaluation} to present the extent of Personally Identifiable Information (PII) leakage. We define five categories of sensitive PII: (1) phone or fax numbers, (2) email addresses, (3) personal names, (4) specific dates, and (5) web links. We then manually analyze reconstructed samples to calculate the proportion of correctly recovered PII instances. The results are summarized in Table~\ref{tab:human_eval}, which reports the number of total PII instances present in the top 30\% reconstructed samples.

\begin{table*}[!htbp]
    \small
    \centering
    \begin{tabular}{lccc}
    \toprule
    \textbf{Checkpoint} & \textbf{Total PII Instances} & \textbf{Correctly Recovered} & \textbf{Proportion} \\
    \midrule
    \textbf{60th Round of LlaMA-8B} & 118 & 44 & 37.29\% \\
    \textbf{3rd Round of LlaMA-8B}  & 84  & 14 & 16.67\% \\
    \textbf{60th Round of LlaMA-1B} & 101 & 32 & 31.68\% \\
    \bottomrule
    \end{tabular}
    \caption{Human evaluation results showing the proportion of successfully reconstructed PII.}
    \label{tab:human_eval}
    \vspace{-13pt}
\end{table*}

The results highlight three important observations. First, models leak significantly more PII after training compared to the pre-trained baseline. Second, larger models demonstrate a higher rate of PII leakage. Finally, the human evaluation results align well with automated metrics, suggesting that ROUGE-L is a reasonable proxy for privacy risk in our setup. We further provide representative case studies in Appendix~\ref{appendix:case} to illustrate concrete examples of such leakage.



\section{Prevent the Leakage of Training Data}
\subsection{Candidates of Prevention Methods}
To address these emerging privacy risks, we evaluate potential techniques that could mitigate this issue, including the use of parameter-efficient fine-tuning methods (LoRA) to prevent over-fitting, differential privacy, update regularization methods and adopting LLMs with safety alignment. We detail the chosen methods as follows.

\textbf{Parameter-Efficient Fine-Tuning (LoRA).}
LoRA (Low-Rank Adaptation) is a parameter-efficient fine-tuning approach that modifies only a small subset of the parameters in LLMs during the fine-tuning phase \citep{hu2021lora}. The key benefit of LoRA is that it restrict the number of parameters updated during training, which can potentially slow down the model's ability to memorize and overfit to sensitive details from the training data, thus mitigate the risk of privacy leakage. 

\textbf{Differential Privacy.}
Differential Privacy (DP) is a privacy-preserving technique that adds noise to the model's updates to prevent leakage of individual data points during training. DP works by adding calibrated noise to the gradients during the training process, ensuring that the model cannot overfit to specific individuals' data, thus preventing attackers from learning sensitive information about any specific data point \citep{dp1}.

\textbf{Update Regularization-Based Methods.}
Update regularization methods are designed to prevent the model from overfitting or memorizing the training data by imposing constraints on the updates applied to the model's parameters \citep{kl-llm}. We incorporate this by adding a KL divergence regularization term to the loss function that penalizes the updates between the current model and its initial training state. This constraint limits excessive changes to the model's parameters during each training round, helping to control the extent of modifications made to the model.This helps to ensure that the model learns more generalized features, rather than specific details from individual clients' data. By limiting the magnitude of the parameter updates, the model is less likely to overfit, which could lead to privacy leakage.

However, while these mitigation strategies effectively reduce overfitting and curb privacy leakage, they risk compromising model capacity and hindering LLMs' ability to acquire meaningful knowledge from training data. In the following section, we explore whether it is possible to strike a balance between model capacity (reflected by training loss) and data leakage.

\textbf{LLMs with Safety Alignment.} Recent alignment techniques aim to make LLMs follow human instructions and avoid generating certain types of sensitive or harmful content. We are also curious whether such models are still vulnerable to privacy attacks that aim to extract training data.

\subsection{Experiments}
\subsubsection{Settings}

\textbf{LoRA.} For LoRA experiments, we test the Llama models of sizes 1B, 3B, 8B, and 13B with ranks $r=32$ and $\alpha=64$. The 3B model is also compared under $r=16, \alpha=32$ and $r=64, \alpha=128$. The dropout rate is set to 0.1.

\textbf{Differential Privacy.} For differential privacy, we studied the privacy leakage mitigation of the Llama 3B and 8B models under different noise multipliers ($\eta \in \{0.01, 0.2, 0.5, 0.8\}$). The max grad norm was set to 1, and $\delta$ was set to 1e-5.

\textbf{KL-Divergence Regularization.} For KL divergence regularization, which constrains the updates between the model and its initial state, we conducted experiments with KL penalty weights ($\mu$) of 0.001 and 0.01 on the Llama-3B model.

\textbf{LLMs with Safety Alignment.} To evaluate the privacy leakage of \textit{LLMs with alignment} under FL settings, we compare LLaMA-3.1-8B with its aligned version, LLaMA-3.1-8B-Aligned.

\subsubsection{Results}

\begin{figure}[t]
\vspace{-6pt}
\setlength{\abovecaptionskip}{6pt}
\centering
\subfloat[Results on models.]{
    \includegraphics[width=0.57\linewidth]{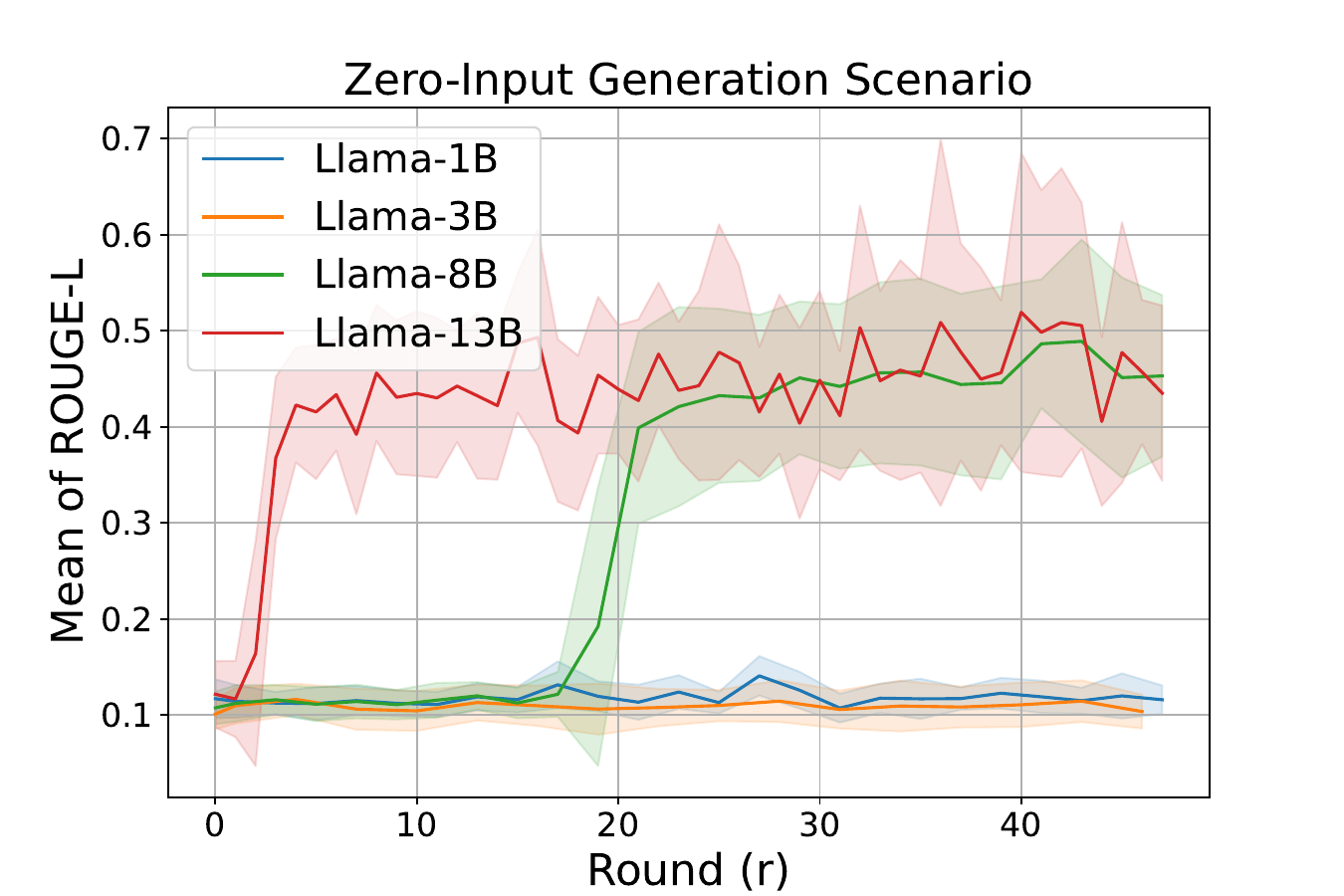}
    \label{fig:sim_lora}
}
\subfloat[Train loss on models.]{
    \includegraphics[width=0.37\linewidth]{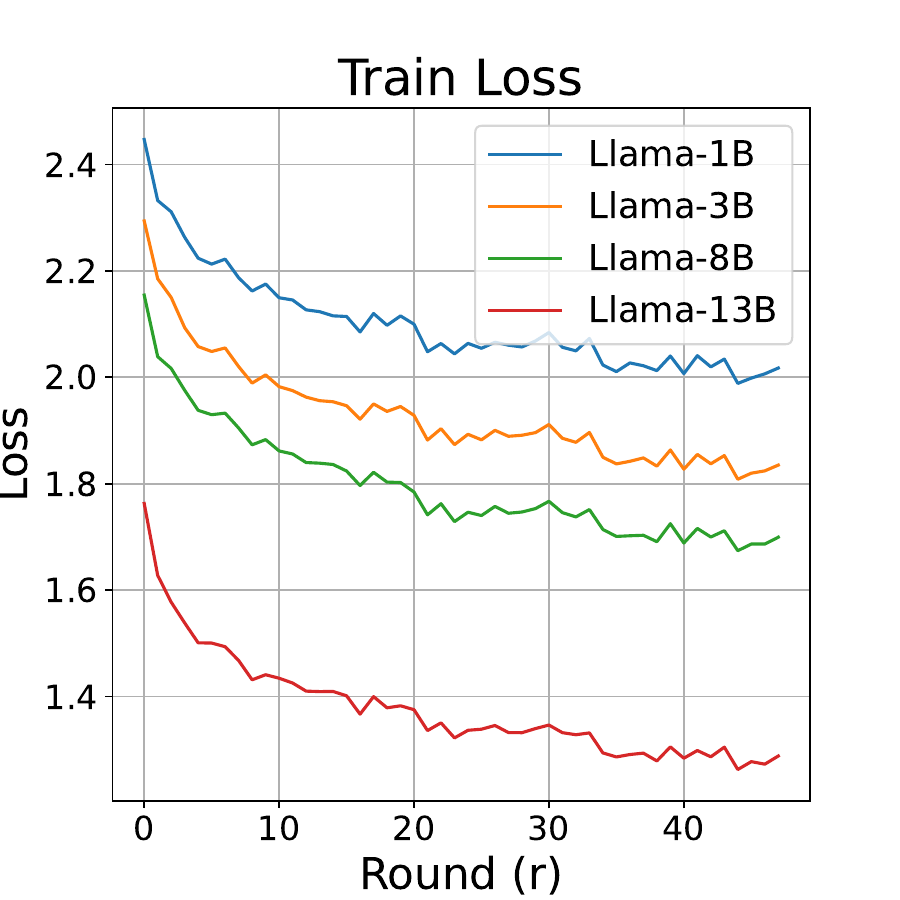}
    \label{fig:lora_loss}
}
\caption{Privacy leakage results of different models using LoRA (r=32) under Zero-Input Generation task.}
\label{fig:sim_lora_all}
\vspace{-12pt}
\end{figure}

\begin{figure}[t]
\vspace{-6pt}
\setlength{\abovecaptionskip}{6pt}
\centering
\subfloat[Results on models.]{
    \includegraphics[width=0.57\linewidth]{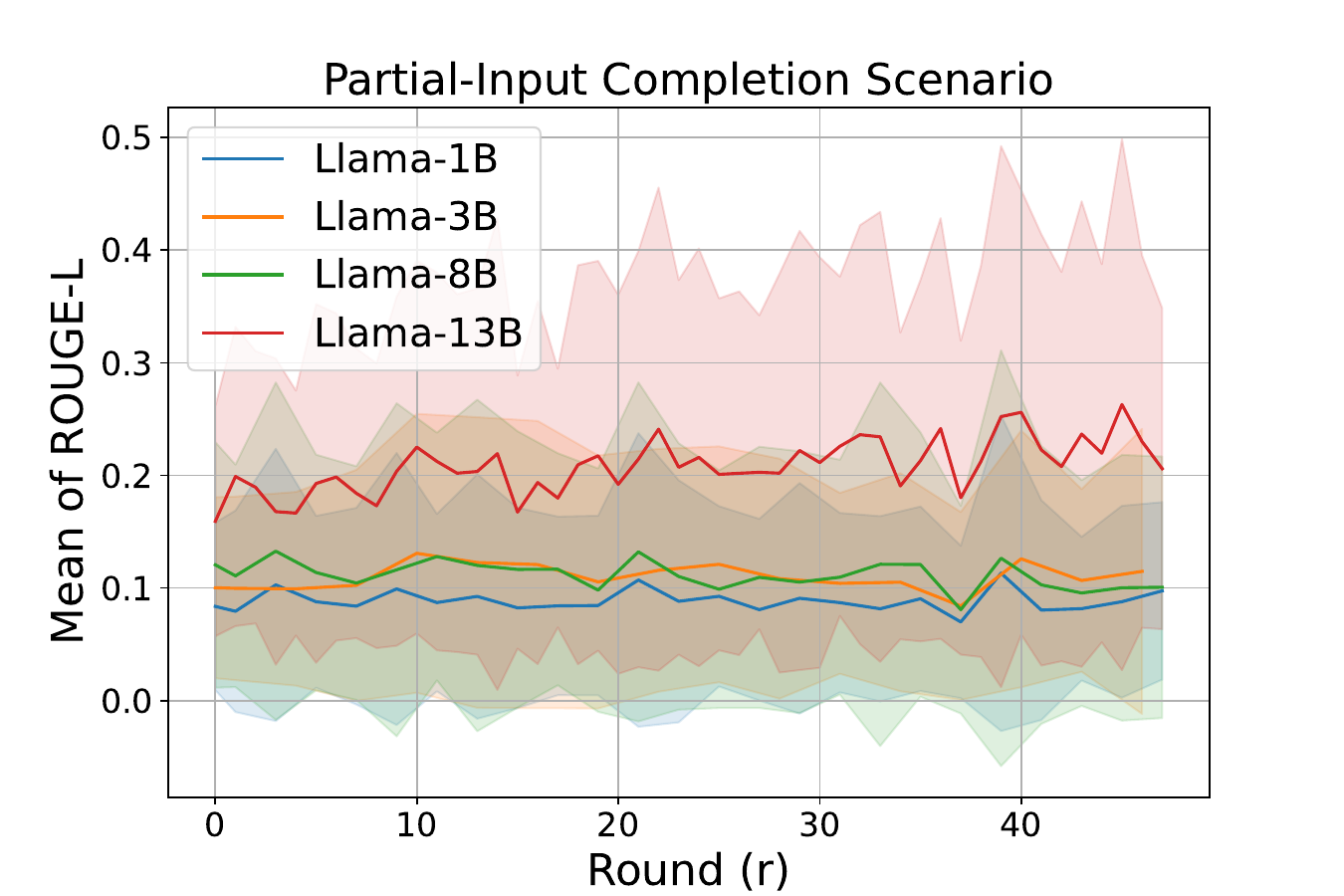}
    \label{fig:com_lora}
}
\subfloat[Train loss on models.]{
    \includegraphics[width=0.37\linewidth]{./figures/lora/lora_loss.pdf}
    \label{fig:lora_loss}
}
\caption{Privacy leakage results of different models using LoRA (r=32) under Partial-Input Completion.}
\label{fig:com_lora_all}
\vspace{-12pt}
\end{figure}

\noindent\textbf{LoRA.} The results in Figs.~\ref{fig:sim_lora_all} and \ref{fig:com_lora_all} show that when using LoRA, the largest 13B model exhibits a significant increase in data leakage right after training, while the 8B model shows a delayed increase. In contrast, the 1B and 3B models consistently produce low similarity scores throughout the training process. However, we observe that the training loss for smaller models remains consistently high, indicating that they are unable to effectively learn from the training data. This is likely due to the reduced number of trainable parameters when fine-tuning with LoRA, which slows down convergence, particularly for smaller models. In summary, while LoRA can help reduce privacy leakage, it does so at the cost of model capacity and slower convergence.

\begin{figure*}[!htbp]
\vspace{-6pt}
\setlength{\abovecaptionskip}{6pt}
\centering
\subfloat[Results on Llama-3B.]{
    \includegraphics[width=0.29\linewidth]{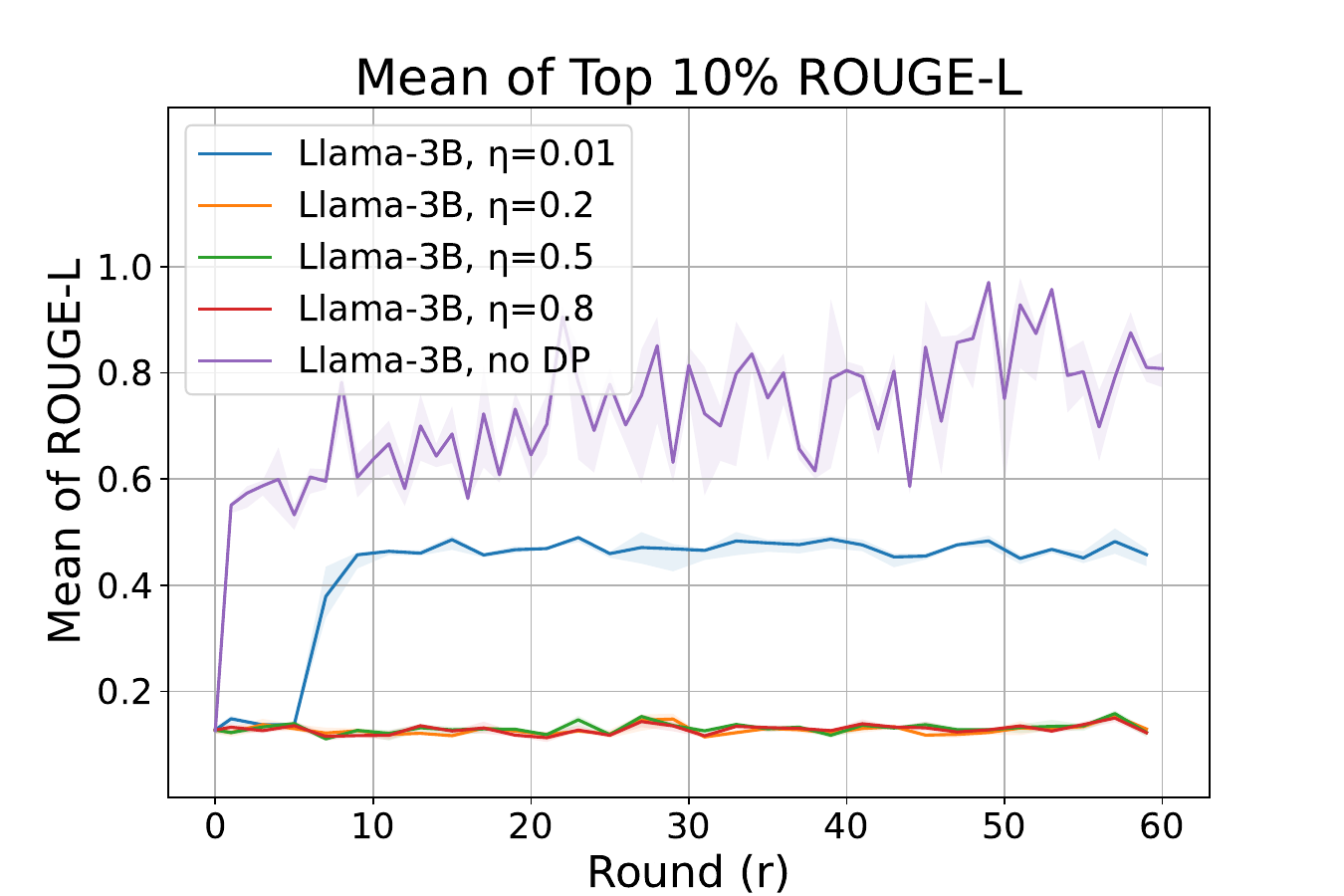}
    \label{fig:dp_sim_3B}
}
\subfloat[3B Train loss.]{
    \includegraphics[width=0.2\linewidth]{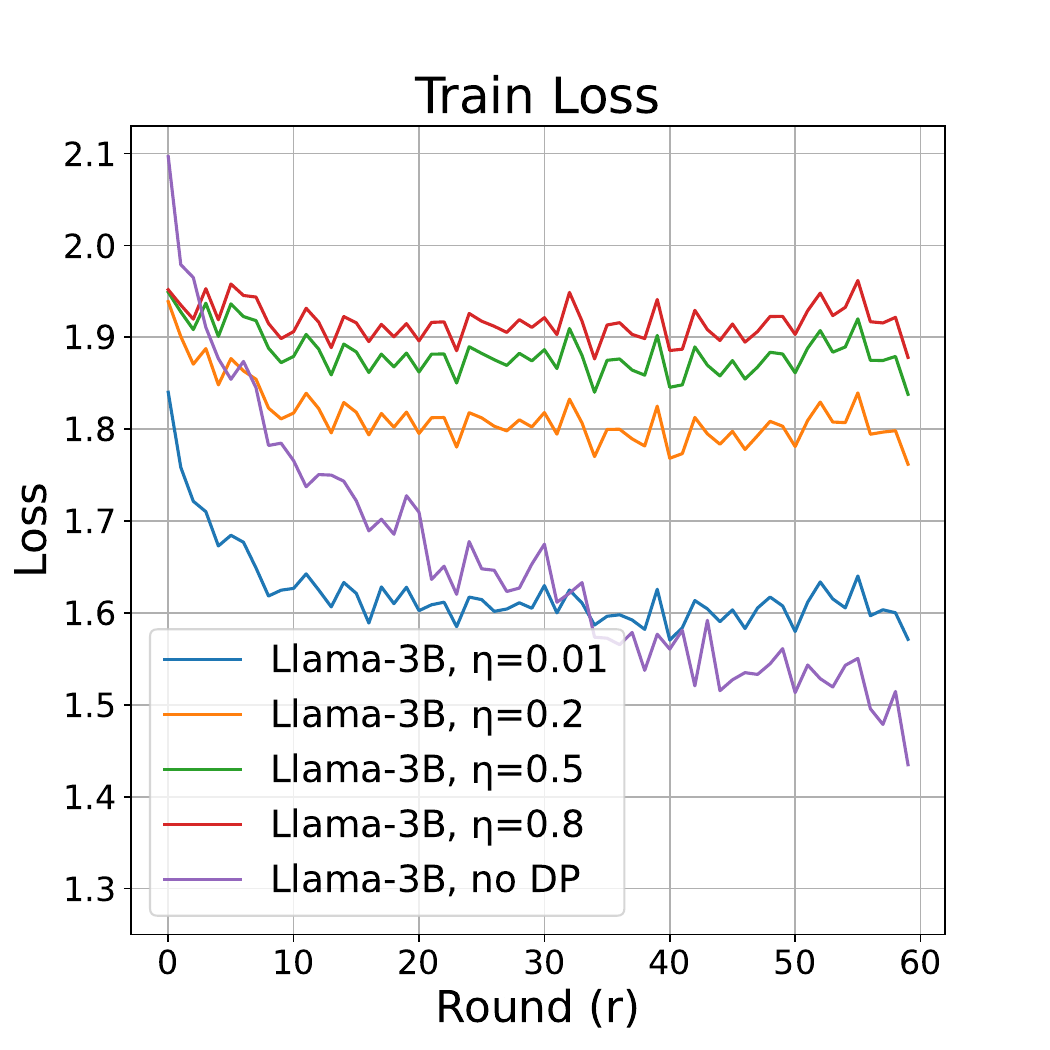}
    \label{fig:dp_loss_3B}
}
\subfloat[Results on Llama-8B.]{
    \includegraphics[width=0.29\linewidth]{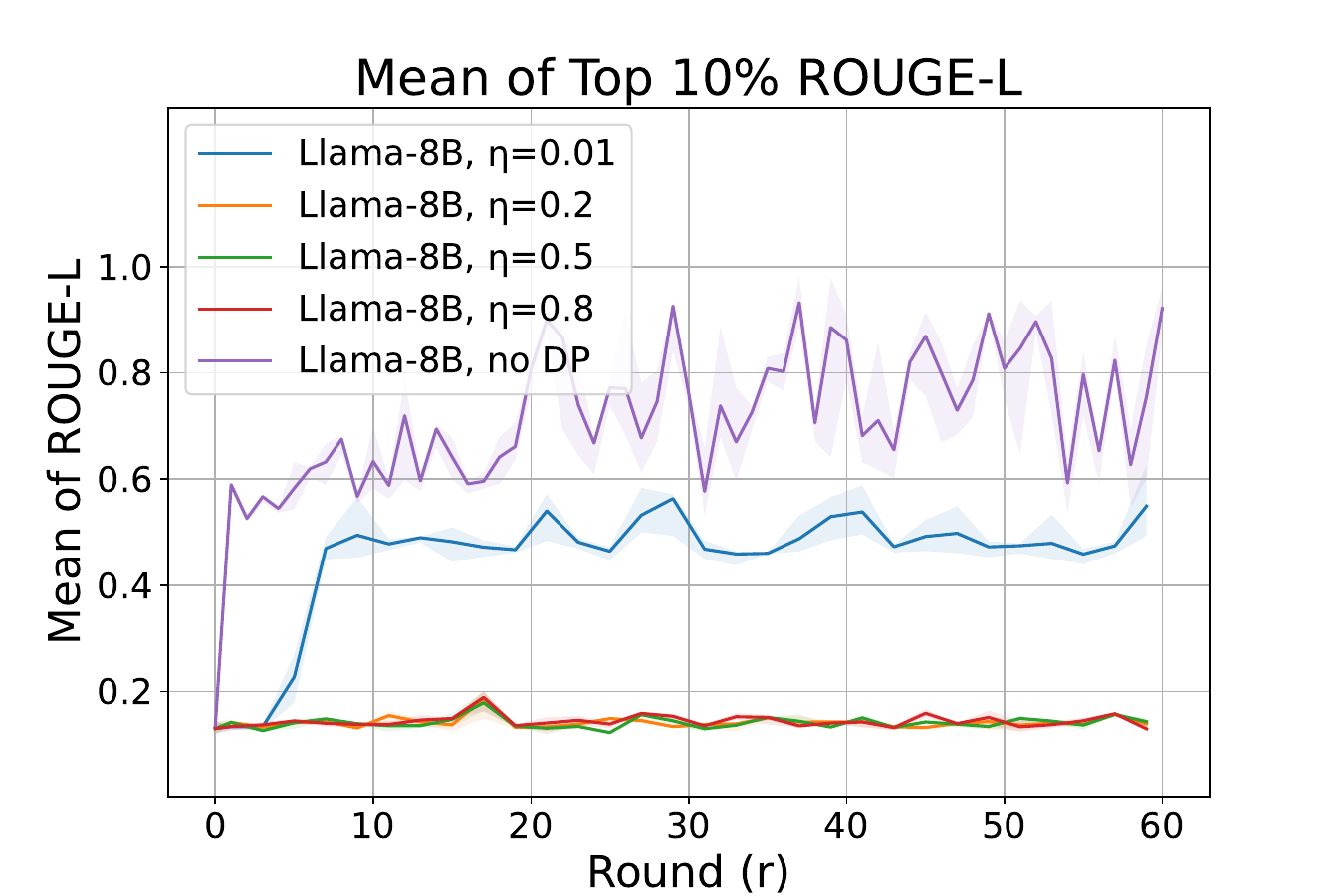}
    \label{fig:dp_sim_8B}
}
\subfloat[8B Train Loss.]{
    \includegraphics[width=0.2\linewidth]{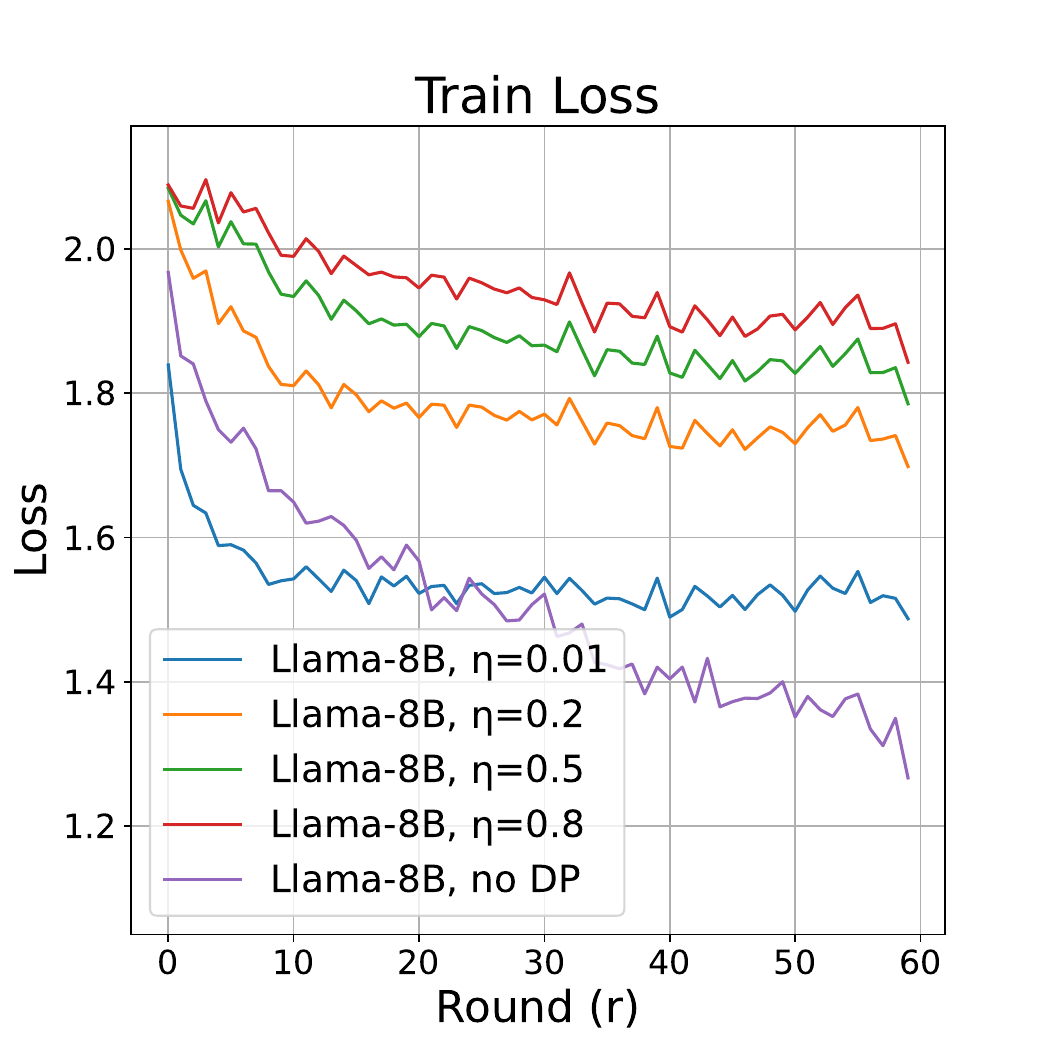}
    \label{fig:dp_loss_8B}
}
\caption{Privacy leakage with different levels ($\eta$) of differential privacy for \textbf{Zero-Input Generation} task.}
\label{fig:dp_sim}
\vspace{-9pt}
\end{figure*}

\begin{figure*}[!htbp]
\vspace{-6pt}
\setlength{\abovecaptionskip}{6pt}
\centering
\subfloat[Results on Llama-3B.]{
    \includegraphics[width=0.29\linewidth]{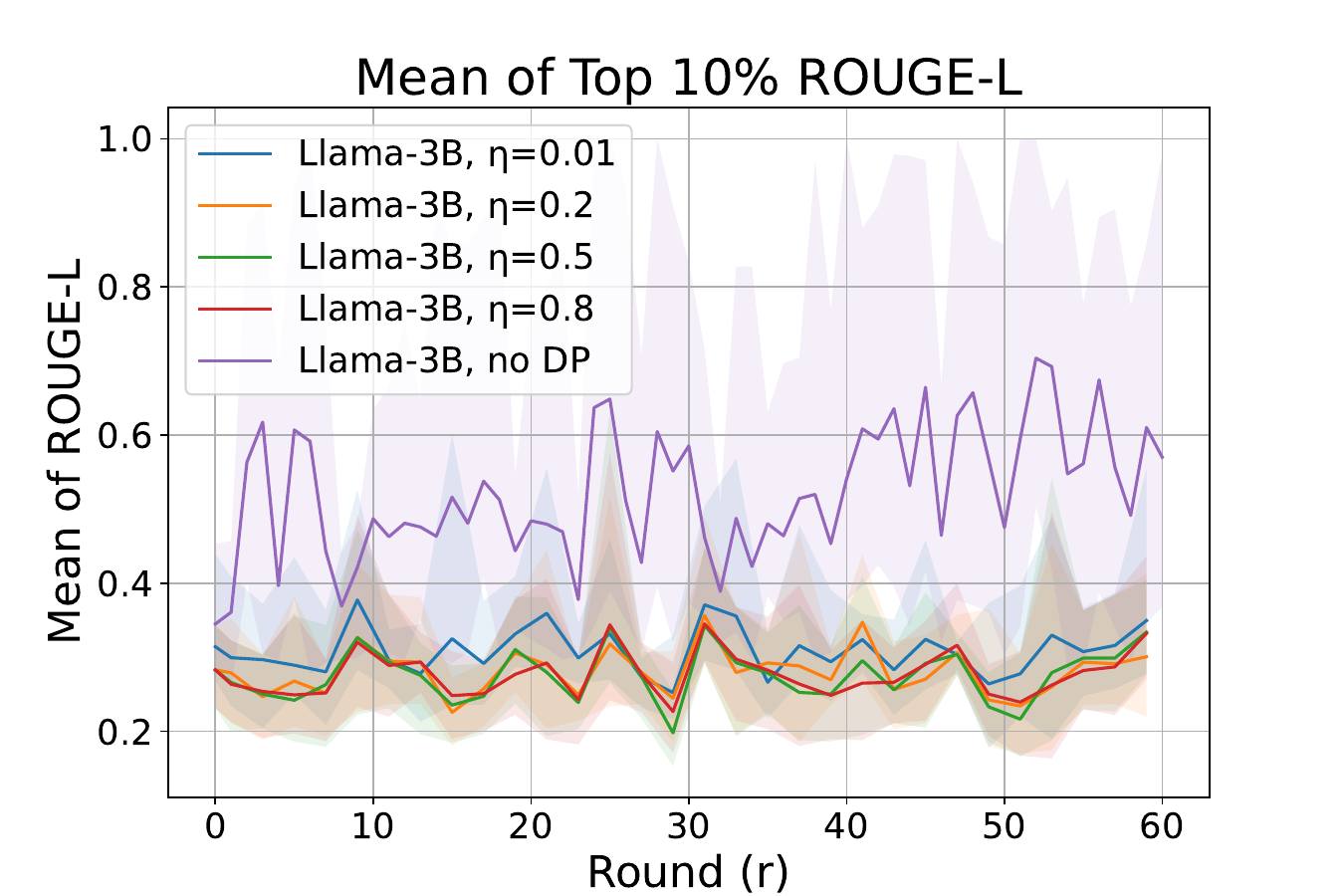}
    \label{fig:dp_com_3B}
}
\subfloat[3B Train Loss.]{
    \includegraphics[width=0.2\linewidth]{./figures/dp/loss_3B.pdf}
    \label{fig:dp_loss_3B}
}
\subfloat[Results on Llama-8B.]{
    \includegraphics[width=0.29\linewidth]{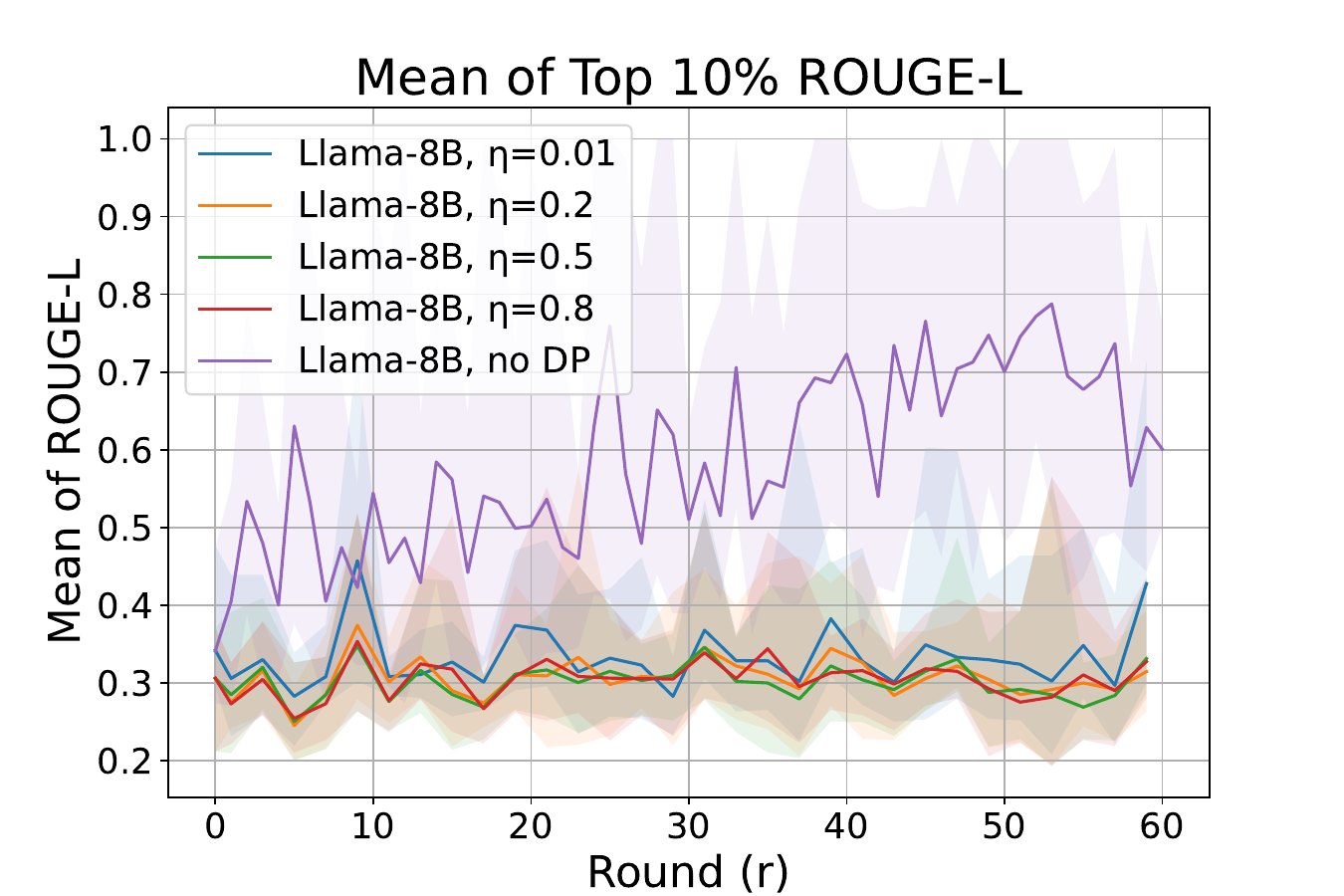}
    \label{fig:dp_com_8B}
}
\subfloat[8B Train Loss.]{
    \includegraphics[width=0.2\linewidth]{./figures/dp/loss_8B.pdf}
    \label{fig:dp_loss_8B}
}
\caption{Privacy leakage with different levels($\eta$) of differential privacy for \textbf{Partial-Input Completion} task.}
\label{fig:dp_com}
\vspace{-13pt}
\end{figure*}

\textbf{Differential Privacy.} We report the results of privacy leakage with varying levels of DP applied to the Llama 3B and 8B models. As shown in Figs.~\ref{fig:dp_sim} and \ref{fig:dp_com}, compared to the original method, applying DP significantly reduces the degree of privacy leakage. 
However, it also negatively affects model performance. This tradeoff between privacy protection and model efficacy makes it challenging to fully utilize DP. As shown in results, when $\eta$=0.01, model performance (measured by Loss) improves compared to when $\eta$=0.2, but privacy leakage (measured by ROUGE-L) also increases.

\begin{figure}[t]
\vspace{-12pt}
\setlength{\abovecaptionskip}{6pt}
\centering
\subfloat[Results on Llama-3B.]{
    \includegraphics[width=0.58\linewidth]{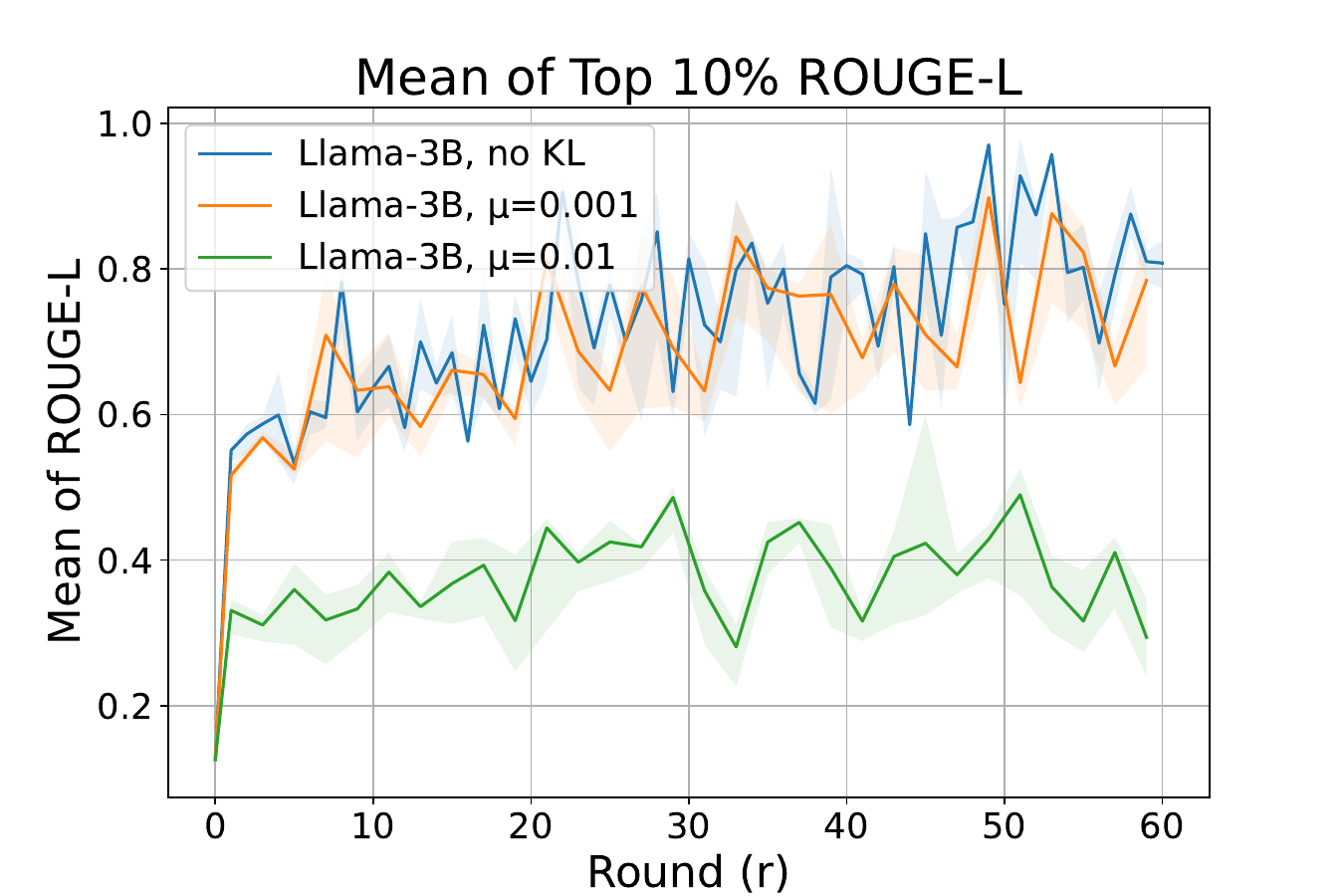}
    \label{fig:sim_kl}
}
\subfloat[Train loss.]{
    \includegraphics[width=0.40\linewidth]{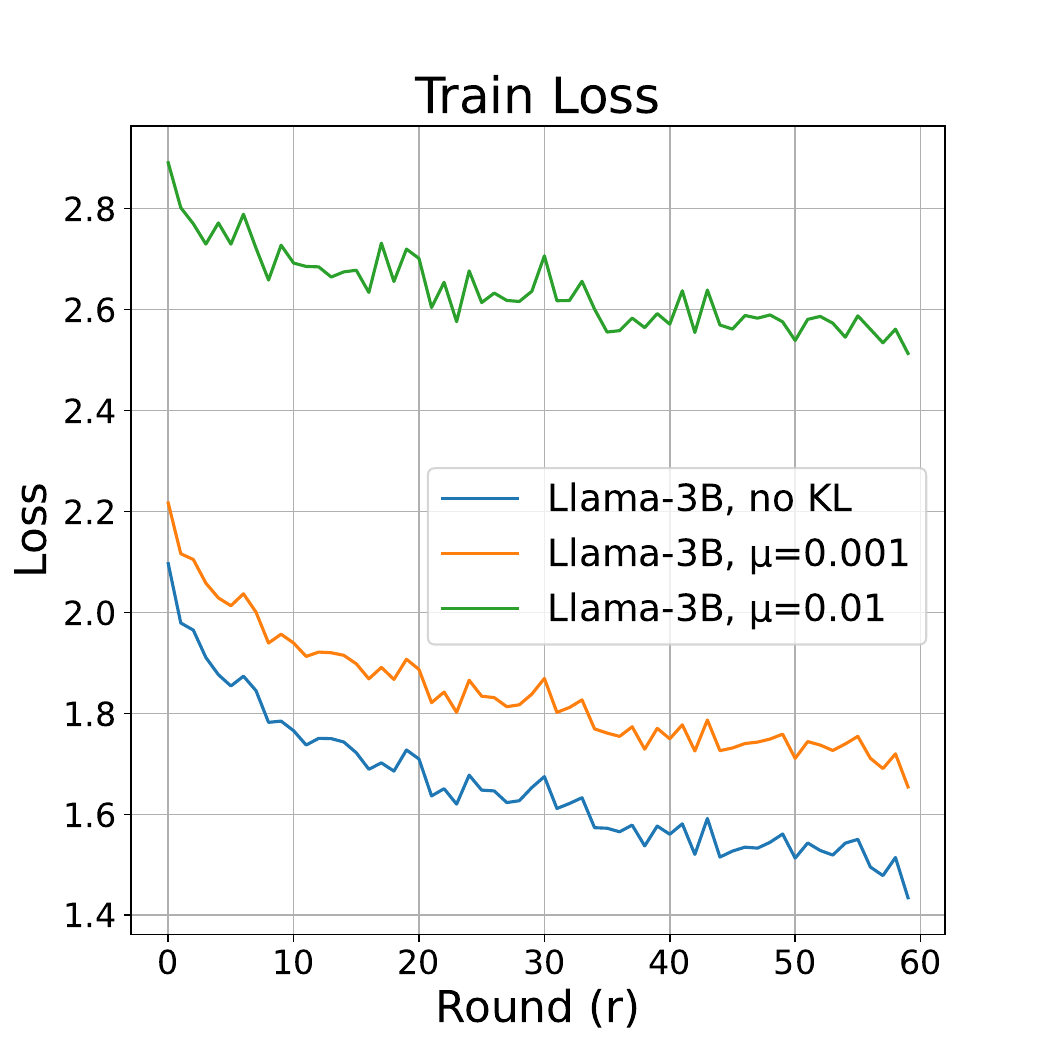}
    \label{fig:kl_loss}
}
\caption{Privacy leakage with different KL penalty weights ($\mu$) for Zero-Input Generation task.}
\vspace{-12pt}
\label{fig:sim_kl_all}
\end{figure}

\begin{figure}[t]
\vspace{-6pt}
\setlength{\abovecaptionskip}{6pt}
\centering
\subfloat[Results on Llama-3B.]{
    \includegraphics[width=0.58\linewidth]{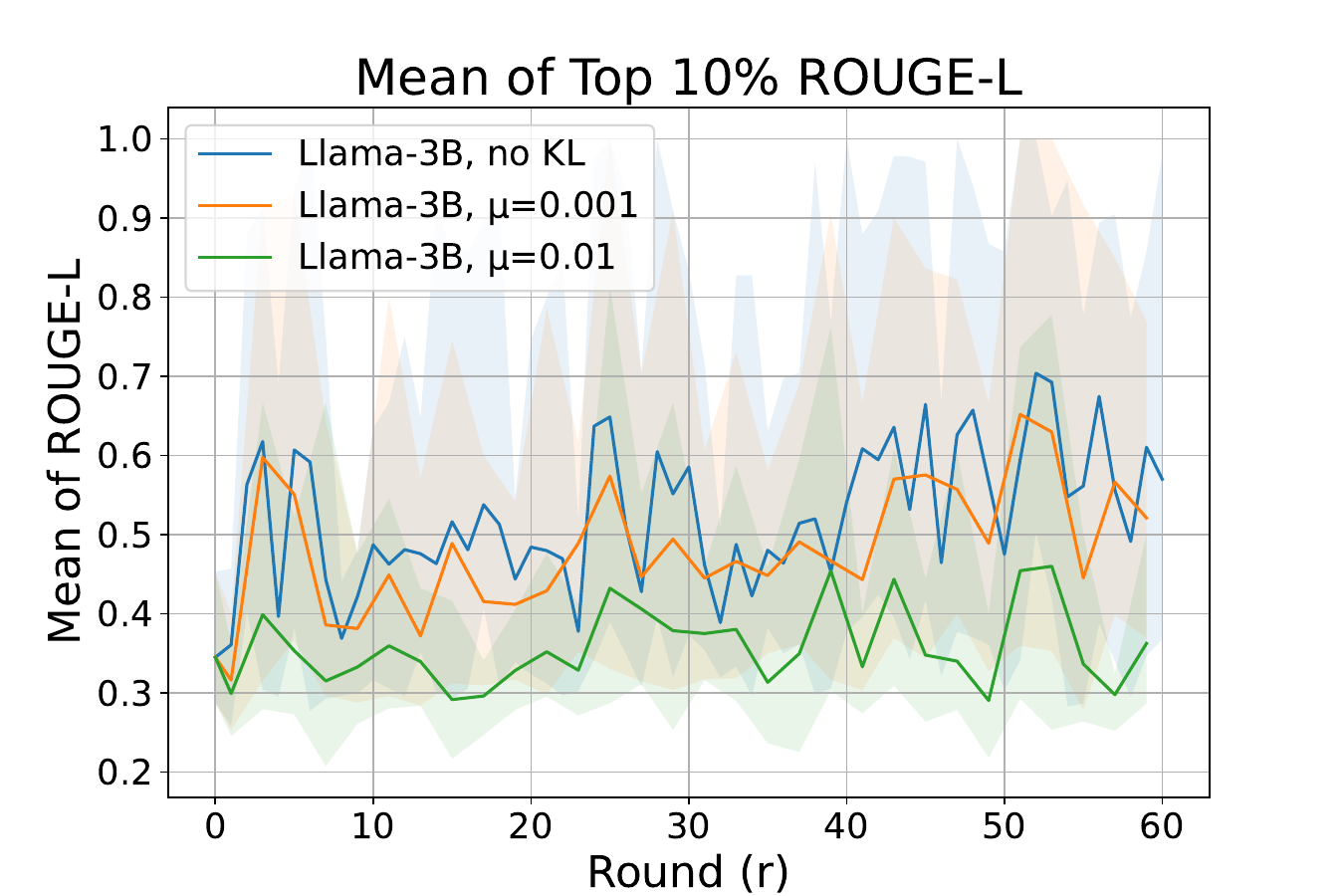}
    \label{fig:com_kl}
}
\subfloat[Train loss.]{
    \includegraphics[width=0.40\linewidth]{./figures/kl/loss_3b.pdf}
    \label{fig:kl_loss}
}
\caption{Privacy leakage with different KL penalty weights ($\mu$) for Partial-Input Completion task.}
\label{fig:com_kl_all}
\vspace{-12pt}
\end{figure}

\textbf{KL-Divergence Regularization.} We report the results of privacy leakage with  KL-Divergence update constraints applied to the Llama-3B model during fine-tuning. As shown in Fig.~\ref{fig:sim_kl_all} and \ref{fig:com_kl_all}, compared to the original method, the higher the level of KL-Divergence regularization, the more significantly privacy leakage is reduced. However, considering the train loss, this constraint also leads to the model struggling to fit the data.

\begin{figure}[t]
\vspace{-12pt}
\setlength{\abovecaptionskip}{6pt}
\centering
\subfloat[Zero-Input Generation]{
    \includegraphics[height=3.1cm, width=0.48\linewidth]{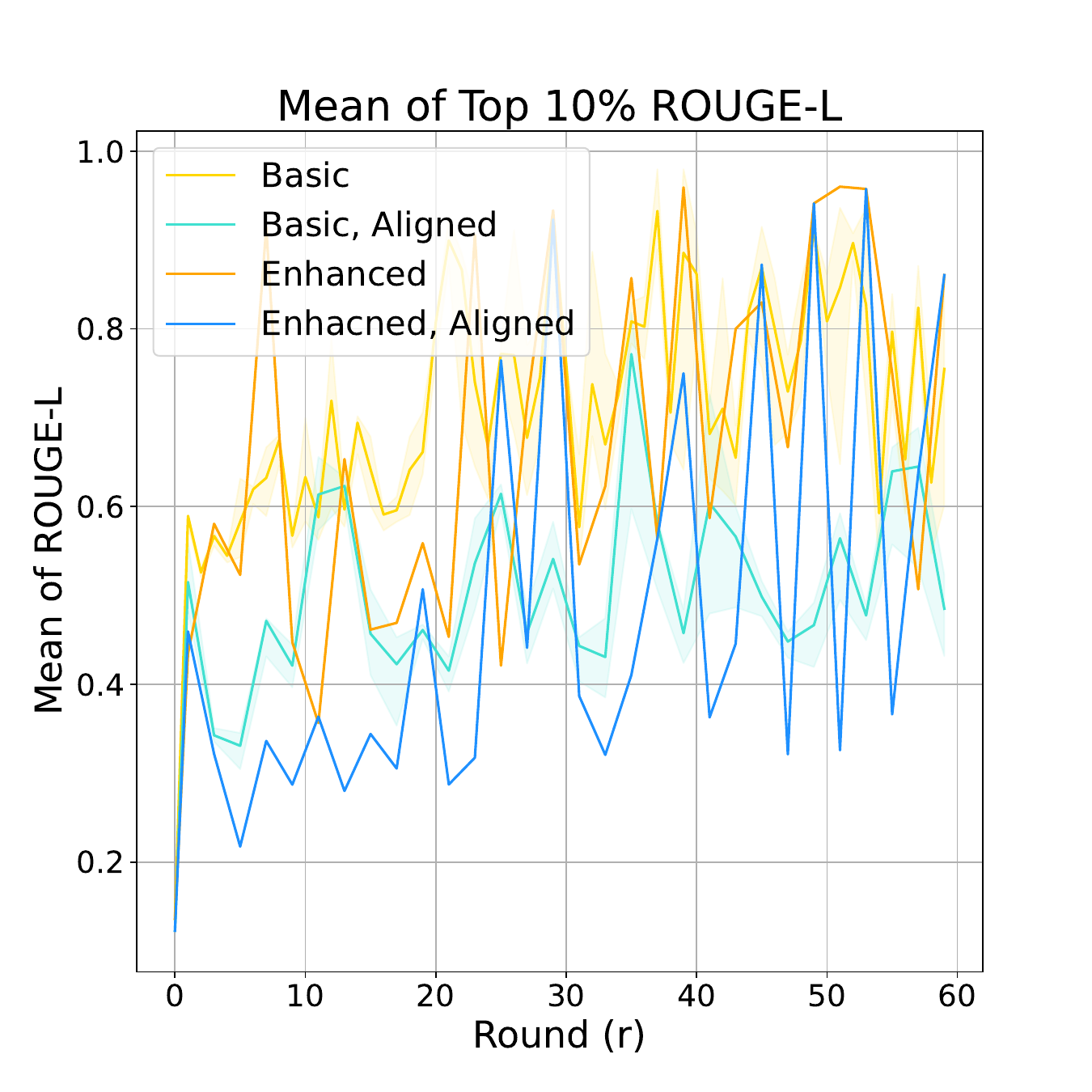}
}
\subfloat[Partial-Input Completion]{
    \includegraphics[height=3.1cm, width=0.48\linewidth]{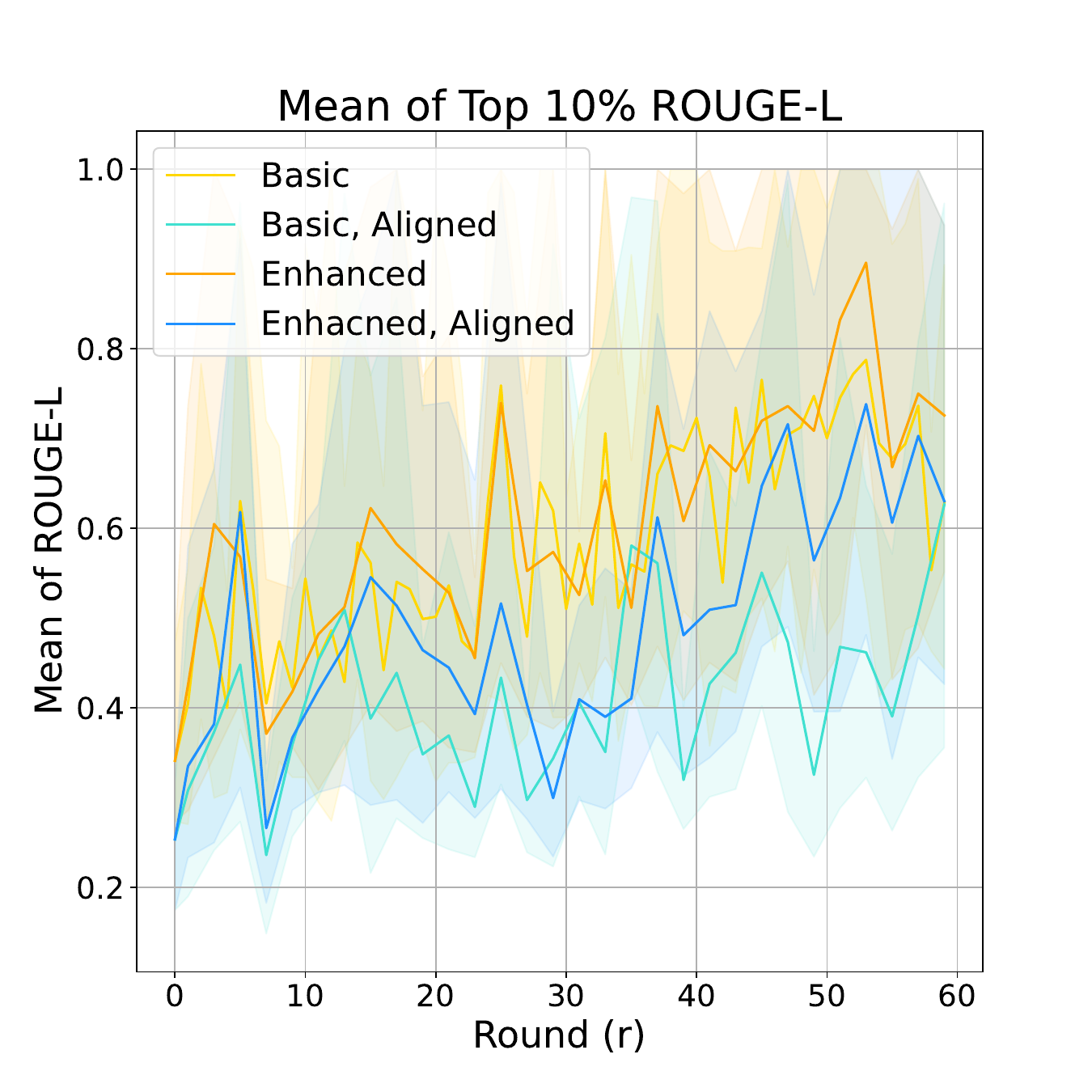}
}
\caption{Results comparison of LLaMA-3.1-8B w/ and w/o alignment under two hacking schemes.}
\label{fig:alignment}
\vspace{-8pt}
\end{figure}

\textbf{LLMs with Safety Alignment.} The results shown in Fig.~\ref{fig:alignment} indicate that fine-tuning aligned models reduce the degree of privacy leakage compared to the non-aligned version. However, aligned LLMs still exhibit risk of data leakage, especially under the enhanced hacking scheme.

\subsection{Discussion}
From the experiments presented above, it is evident that regardless of the approach used, it seems hard to both prevent privacy leakage and maintain model capacity simultaneously. In other words, to achieve better privacy protection, there is a noticeable trade-off in terms of the model’s fitting ability during training. This indicates that while existing methods provide a certain level of privacy protection, they are mainly suitable for tasks where the model’s fitting capacity is not a critical requirement. For applications that demand higher model capacity, new privacy protection techniques are still required.

\section{Conclusion}
In this paper, we demonstrate that FL may not fully address privacy concerns in LLM training. Our experiments show that LLMs can leak sensitive training data through generated text. We introduce an enhanced attack strategy that exploits the iterative transmission of model parameters during FL to amplifies these risks. While privacy-preserving techniques like differential privacy and update regularization offer some mitigation, they come at the cost of reduced model performance. These findings suggest the need for further research to develop more effective privacy solutions in FL for LLMs.

\clearpage

\section*{Limitation}
This paper investigates the privacy leakage in training LLMs with FL from their generative output. However, due to resource limitations, several aspects remain unexplored. Firstly, we only focus on the fine-tuning stage. Recently, some studies have advocated for pre-training LLMs in a federated manner \citep{sani2024future}, and our work may raise potential concerns regarding these approaches. Additionally, our study does not investigate privacy leakage issues in more LLM tasks like RLHF \citep{instructgpt} and DPO \citep{dpo}, and we plan to extend in the future.

\section*{Acknowledgement}
This work was supported by the National Natural Science Foundation of China under Grants 62372028 and 62372027, and by the Central Guiding Local Science and Technology Development Fund of Shanghai Municipality (Project No. YDZX20253100004011)

\bibliography{custom}

\clearpage
\appendix

\section{Training Details}
\label{appendix:args}
We list important training hyper-parameters in Table~\ref{tab:hyperparams}. Generally, we adopt a common hyperparameter setting in fine-tuning LLMs. We use the AdamW optimizer for improved weight decay handling, with a cosine learning rate scheduler to facilitate smooth convergence. Gradient clipping is applied to mitigate exploding gradients. Additionally, for experiments where LoRA is implemented, we set the dropout rate to be 0.1 to prevent overfitting. In each communication round in federated learning, each client train for 200 iterations before parameter aggregation. Finally, the epoch number is set to 1, which means each data sample is learned at most for one times during training, thereby preventing the model from over-fitting.

\begin{table}[!htbp]
    \centering
    \begin{tabular}{l|c}
        \hline
        \textbf{Hyper-Parameter} & \textbf{Value} \\
        \hline
        Optimizer & AdamW \\
        $\beta_1$ & 0.99 \\
        $\beta_2$ & 0.999 \\
        Gradient Clipping Norm & 1.0 \\
        Batch Size & 8 \\
        Weight Decay & 0.01 \\
        LR Scheduler & cos \\
        Local Iterations & 200 \\
        Epochs & 1 \\
        Max Seq. Length & 512 \\
        Client Number & 4 \\
        \hline
    \end{tabular}
    \caption{Hyper-parameter settings.}
    \label{tab:hyperparams}
\end{table}

We use the following models in our experiments: LLaMA-3.2-1B, LLaMA-3.2-3B, LLaMA-3.1-8B, and LLaMA-2-13B, Gemma-2-2B, Qwen2.5-7B. The total computational budget amounted to approximately 2,000 GPU hours, and all experiments were conducted using A800 GPUs.

\section{Dataset Description}
\label{appendix:dataset_desc}
We fine-tuned our models using three distinct datasets.

The \textbf{Enron-Email Dataset} \citep{klimt2004enron} contains approximately 0.5 million emails from about 150 Enron employees. These publicly available emails are rich in private information, including names, addresses, phone numbers, and company-specific data. This type of data is widely recognized as private and frequently used to benchmark privacy leakage studies \citep{huang2022large}.

The \textbf{Reddit-Comments Dataset} \citep{baumgartner2020pushshift} was created by selecting the first 25,000 comments from each of the 40 most frequented subreddits from a larger collection of Reddit comments from May 2019. This approach balances comment volume across subreddits, making it suitable for analysis where subreddits are treated as categorical data. This dataset aggregates user comments from the Reddit platform, which include diverse sensitive information and have been utilized in privacy-related research \citep{kandpal2023user}.

The \textbf{CLERC Dataset} \citep{abe2024clerc} is a collection of legal case documents derived from all federal case law in U.S. history up to September 21, 2021. This dataset includes 1.84 million documents and provides different formats for various tasks: `CLERC/doc` contains raw case documents with concatenated opinions, `CLERC/passage` offers documents split into 350-word chunks with a 175-word sliding window for retrieval tasks, and `CLERC/generation` provides passages specifically for generation tasks. Many of these legal documents contain personally identifiable information.

\section{Case Study}
\label{appendix:case}
In this section, we present several representative hacking cases to intuitively demonstrate the extent to which generated data can replicate private data. We also provide statistics on the number of instances that exactly match the original private data.

We analyze the outputs of the LLaMA-3.1-8B model under the basic hacking scheme (corresponding to Figs.~\ref{fig:sim_mean} and ~\ref{fig:com_mean} at the final round of training. The specific results are summarized in Table~\ref{tab:sample_analysis}.

Examples of exactly matched cases are shown in Table~\ref{tab:samples}.

\begin{table*}[htbp]
    {\fontsize{8.7pt}{9.6pt}\selectfont
    \centering
    \setlength{\tabcolsep}{0.8mm}
    \renewcommand{\arraystretch}{0.85}
    \begin{tabular}{>{\raggedright\arraybackslash}p{7.9cm}|>{\raggedright\arraybackslash}p{7.9cm}}
        \toprule
        \textbf{Original Text (Private Data)} & \textbf{Generated Text} \\
        \midrule
        \texttt{
            \textbf{[BOS]} 
            Date: Fri, 1 Dec 2000 08:29:00 -0800 (PST)
            From: mailbot@datek.com
            To: \colorbox{black}{lcampbel}@enron.com
            Subject: "Datek Online execution report"
            Body: 
            Dear \colorbox{black}{Mr. LARRY CAMPBELL}:
            Thank you for your order.
            You bought 20 shares of CIEN (Ciena Corp Com) at 78 7/8.
            ORDER DETAILS:
            The tracking number for this order is: \colorbox{black}{B-600-AATE-9686}
            Order entered at:                      08:37:39 on 12/01/00
            Order processing began at:             08:37:40
            Order executed at:                     08:37:40 on 12/01/00
            ACCOUNT DETAILS:
            Time and Date: 08:37:41 on 12/01/00
            Your position in CIEN is ...
        }
        & 
        \texttt{
            \textbf{[BOS]} \textcolor{blue}{
            Date: Wed, 13 Dec 2000 09:29:00 -0800 (PST)
            From: mailbot@datek.com
            To: \colorbox{black}{\phantom{lcampbel}}@enron.com
            Subject: "Datek Online execution report"
            Body: 
            Dear \colorbox{black}{\phantom{Mr. LARRY CAMPBELL}}:
            Thank you for your order.
            You bought 50 shares of CIEN (Ciena Corp Com) at 79 7/8.
            ORDER DETAILS:
            The tracking number for this order is: \colorbox{black}{\phantom{B-600-AATE-9686}}
            Order entered at:                      08:54:47 on 12/13/00
            Order processing began at:             08:54:48
            Order executed at:                     08:55:41 on 12/13/00
            ACCOUNT DETAILS:
            Time and Date: 08:55:42 on 12/13/00
            Your position in CIEN is 100 shares
            Your real cash total is ...}
        }\\
        \midrule
        \texttt{
            \textbf{[BOS]} 
            Date: Thu, 16 Nov 2000 12:12:00 -0800 (PST)
            From: \colorbox{black}{\phantom{enron.announcements}}@enron.com
            To: \colorbox{black}{\phantom{houston.ena\&eim\&egm}}@enron.com
            Subject: Holiday Invitation
            Body: 
            Please click on the attached link to launch your Holiday Party invitation.
             http://\colorbox{black}{\phantom{invitation.enron.com}}
            Please direct any questions to \colorbox{black}{\phantom{Dorie Hitchcock}} via email.
        } 
        & 
        \texttt{
            \textbf{[BOS]} \textcolor{blue}{
            Date: Thu, 21 Dec 2000 06:00:00 -0800 (PST)
            From: \colorbox{black}{\phantom{enron.announcements}}@enron.com
            To: \colorbox{black}{\phantom{houston.ena\&eim\&}}@enron.com
            Subject: Holiday Invitation
            Body: 
            Please click on the attached link to launch your Holiday Party invitation.
             http://\colorbox{black}{\phantom{invitation.enron.com}}
            Please direct any questions to \colorbox{black}{\phantom{Dorie Hitchcock}} via email.}
        }\\
        \midrule
        \texttt{
            \textbf{Date: Mon, 22 Oct 2001 04:55:34 -0700 (PDT)
            From: no.address@enron.com
            Subject: All-Employee Meeting
            Body: 
            ......
            Enron Europe employees will receive a follow-up message from their Public Relations team concerning online access to the meeting.
            Video Teleconferencing.} 
            The meeting will be made available by video teleconference to employees in \colorbox{black}{\phantom{Sao Paulo, Buenos Aires, Dubai}}, Rio de Janeiro, \colorbox{black}{\phantom{Bothell, Wash., Denver, San Ramon, Calif., and}} \colorbox{black}{\phantom{Chicago}}.  If your location would like to participate by video teleconference, please contact \colorbox{black}{\phantom{Yvonne Francois}} at \colorbox{black}{\phantom{(713) 345-8725}}.
        }
        & 
        \texttt{
            \textbf{Date: Mon, 22 Oct 2001 04:55:34 -0700 (PDT)
            From: no.address@enron.com
            Subject: All-Employee Meeting
            Body: 
            ......
            Enron Europe employees will receive a follow-up message from their Public Relations team concerning online access to the meeting.
            Video Teleconferencing.}
            \textcolor{blue}{
            The meeting will be made available by video teleconference to employees in \colorbox{black}{\phantom{Sao Paulo, Buenos Aires, Dubai}}, Bothell, \colorbox{black}{\phantom{Wash.,}} \colorbox{black}{\phantom{Denver, San Ramon, Calif., and Chicago}}.  If your location would like to participate by video teleconference, please contact \colorbox{black}{\phantom{Yvonne Francois}} at \colorbox{black}{\phantom{(713) 345-8725}}.}
        }\\
        \midrule
        \texttt{
            \textbf{Date: Fri, 5 Oct 2001 10:38:17 -0700 (PDT)
            From: pppcmail@pppc.org
            To: pppcmail@pppc.org
            Subject: International Uncoated Mechanical Paper Statistics ......
            For questions or comments concerning Economics Online, please contact:
            Doreen} 
             Olkowicz
            Manager, Statistics and Online Services
            T \colorbox{black}{\phantom{(514) 861-8851}}    F \colorbox{black}{\phantom{(514) 866-4863}}    E-mail  \colorbox{black}{\phantom{dolkowicz}}@pppc.org
        }
        & 
        \texttt{
            \textbf{Date: Fri, 5 Oct 2001 10:38:17 -0700 (PDT)
            From: pppcmail@pppc.org
            To: pppcmail@pppc.org
            Subject: International Uncoated Mechanical Paper Statistics ......
            For questions or comments concerning Economics Online, please contact:
            Doreen} 
            \textcolor{blue}{
            Olkowicz
            Manager, Statistics and Online Services
            T \colorbox{black}{\phantom{(514) 861-8851}}    F \colorbox{black}{\phantom{(514) 866-4863}}    E-mail  \colorbox{black}{\phantom{dolkowicz}}@pppc.org}
        }\\
        \midrule
        \texttt{
            \textbf{Date: Wed, 20 Sep 2000 09:44:00 -0700 (PDT)
            From: eric.bass@enron.com
            To: jason.bass2@compaq.com
            Subject: Give me} 
             a call
            Body: 
            \colorbox{black}{\phantom{713-853-0977}}
        }
        & 
        \texttt{
            \textbf{Date: Wed, 20 Sep 2000 09:44:00 -0700 (PDT)
            From: eric.bass@enron.com
            To: jason.bass2@compaq.com
            Subject: Give me} 
            \textcolor{blue}{
            a call
            Body: 
            \colorbox{black}{\phantom{713-853-0977}}}
        }\\
        \midrule
        \texttt{
            \textbf{Date: Tue, 8 Aug 2000 09:31:00 -0700 (PDT)
            From: phillip.allen@enron.com
            To: ina.rangel@enron.com
            Subject: Your Approval is Overdue: Access Request for paul.t.lucci@enron.com
            Body: .......@enron.com to approve the request or contact IR}
            M at 
            \colorbox{black}{\phantom{713-853-5536}} if you have any issues.
            Request ID          : \colorbox{black}{\phantom{000000000000935}}
            Request Create Date : 7/27/00 2:15:23 PM
            Requested For       : \colorbox{black}{\phantom{paul.t.lucci}}@enron.com
            Resource Name       : EOL US NatGas US GAS PHY FWD FIRM Non-Texas < or = 1 
            Month
            Resource Type       : Applications
        }
        & 
        \texttt{
            \textbf{Date: Tue, 8 Aug 2000 09:31:00 -0700 (PDT)
            From: phillip.allen@enron.com
            To: ina.rangel@enron.com
            Subject: Your Approval is Overdue: Access Request for paul.t.lucci@enron.com
            Body: .......@enron.com to approve the request or contact IR}
            \textcolor{blue}{
            M at 
            \colorbox{black}{\phantom{713-853-5536}} if you have any questions.
            Request ID          : \colorbox{black}{\phantom{000000000000935}}
            Request Create Date : 7/26/00 11:02:22 AM
            Requested For       : \colorbox{black}{\phantom{paul.t.lucci}}@enron.com
            Resource Name       : EOL US Backoffice  Data Manager
            Resource Type       : Applications}
        }\\
        \bottomrule
    \end{tabular}
    }
    \caption{Examples of exactly matched cases. The \textbf{input portions} are shown in \textbf{black}, while the \textcolor{blue}{generated (predicted) content} is highlighted in \textcolor{blue}{blue}. The black masked segments \colorbox{black}{\phantom{aaa}} represent sensitive private data that were \textit{precisely reconstructed} by the model.}
    \label{tab:samples}
\end{table*}

As shown, the proportion of samples containing exactly matched sensitive information is already non-negligible. This is particularly concerning in scenarios where longer text passages are generated, as they may include precisely reconstructed sensitive data such as phone numbers.

\section{Disturbed-Input Completion Task}
\label{appendix:disturbed}
The two tasks studied in the main Section~\ref{section:Task} correspond to scenarios with no prior information and with partial exact information, respectively. In this section, we extend the research to a more general scenario, where only \textit{vague partial information is available}. The details are as follows.

\noindent\textbf{Method.} Specifically, based on the Partial-Input Completion task, we apply a method of substituting certain words in the input sequence $\mathbf{X}$ with synonyms to perturb the text, aiming to retain the original meaning while avoiding an exact match with the original text, thus simulating an attacker with imprecise knowledge. We begin with the following steps.

\textbf{1) Select substituted words.} For each word $w$ in the text, we set a probability $p$ for replacing this target word.

\textbf{2) Word filtering:} For the selected word $w$, we ensure that personal information such as names, phone numbers, and email addresses is retained and not replaced. Additionally, some "stop words," like articles and prepositions, are not substituted.

\textbf{3) Synonym extraction:} For the word $w$ that is to be replaced, we use pre-trained word vectors (GoogleNews-vectors-negative300 \citep{mikolov2013linguistic}) to select a set of semantically similar words from the vocabulary, forming a candidate set $C$ for substitution.

\textbf{4) Part-of-speech (POS) check:} For the candidate set $C$, we retain words with the same part-of-speech as the target word, ensuring the grammatical integrity of the text.

\textbf{5) Word replacement:} Finally, we select the word from the candidate set $C$ that is most similar to the target word and make the substitution.

\noindent\textbf{Experiment settings.} The configuration for the hacking experiments is based on the Partial-Input Completion task, as detailed in Section~\ref{section:com}. In the word replacement, we set $p$ to 0.4, meaning that 40\% of the words are expected to be replaced.

\begin{figure*}[!htbp]
\setlength{\abovecaptionskip}{6pt}
\centering
\subfloat[Comparison on LLaMA-1B.]{
    \includegraphics[width=0.45\linewidth]{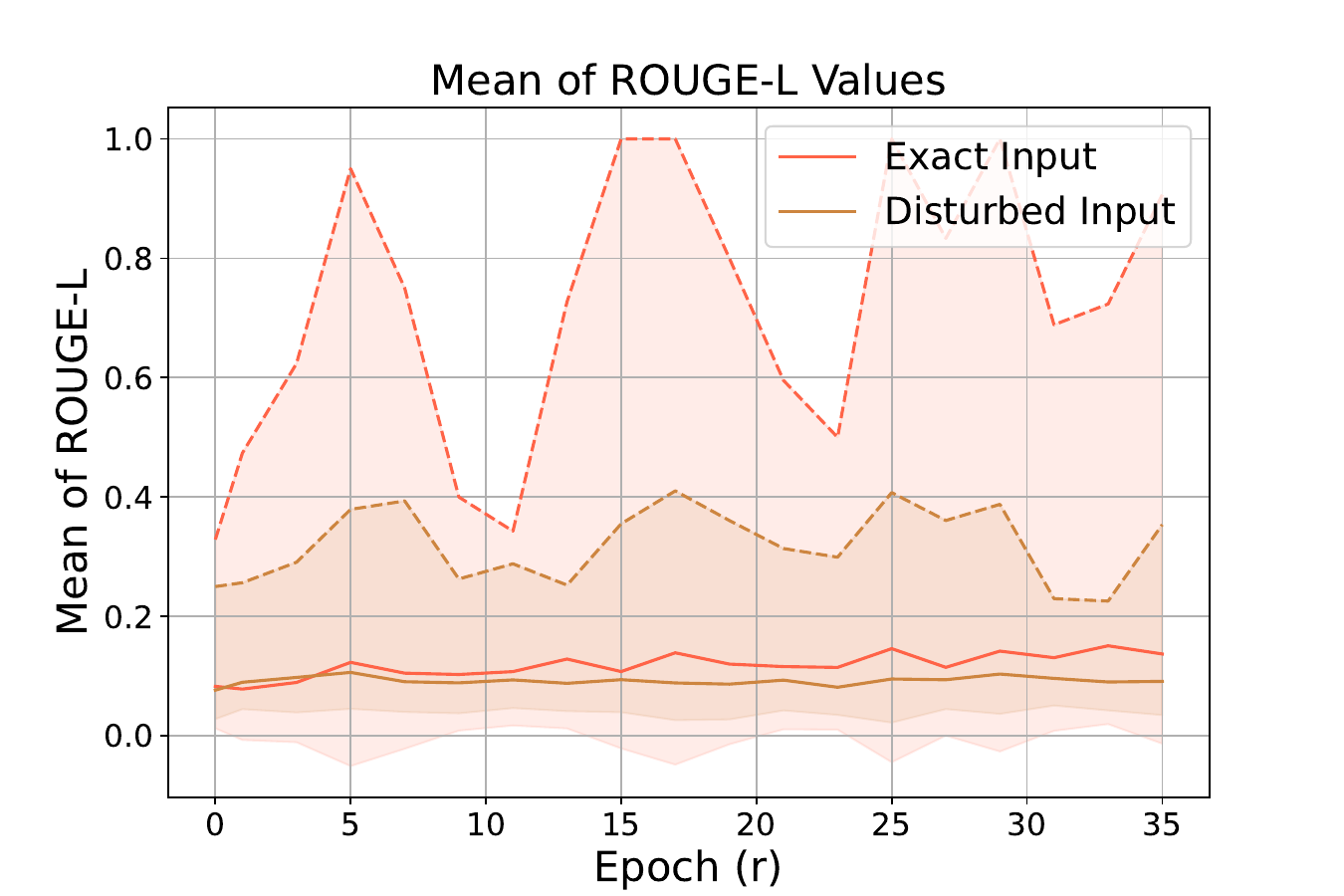}
    \label{fig:aug1B}
}
\subfloat[Comparison on LLaMA-3B.]{
    \includegraphics[width=0.45\linewidth]{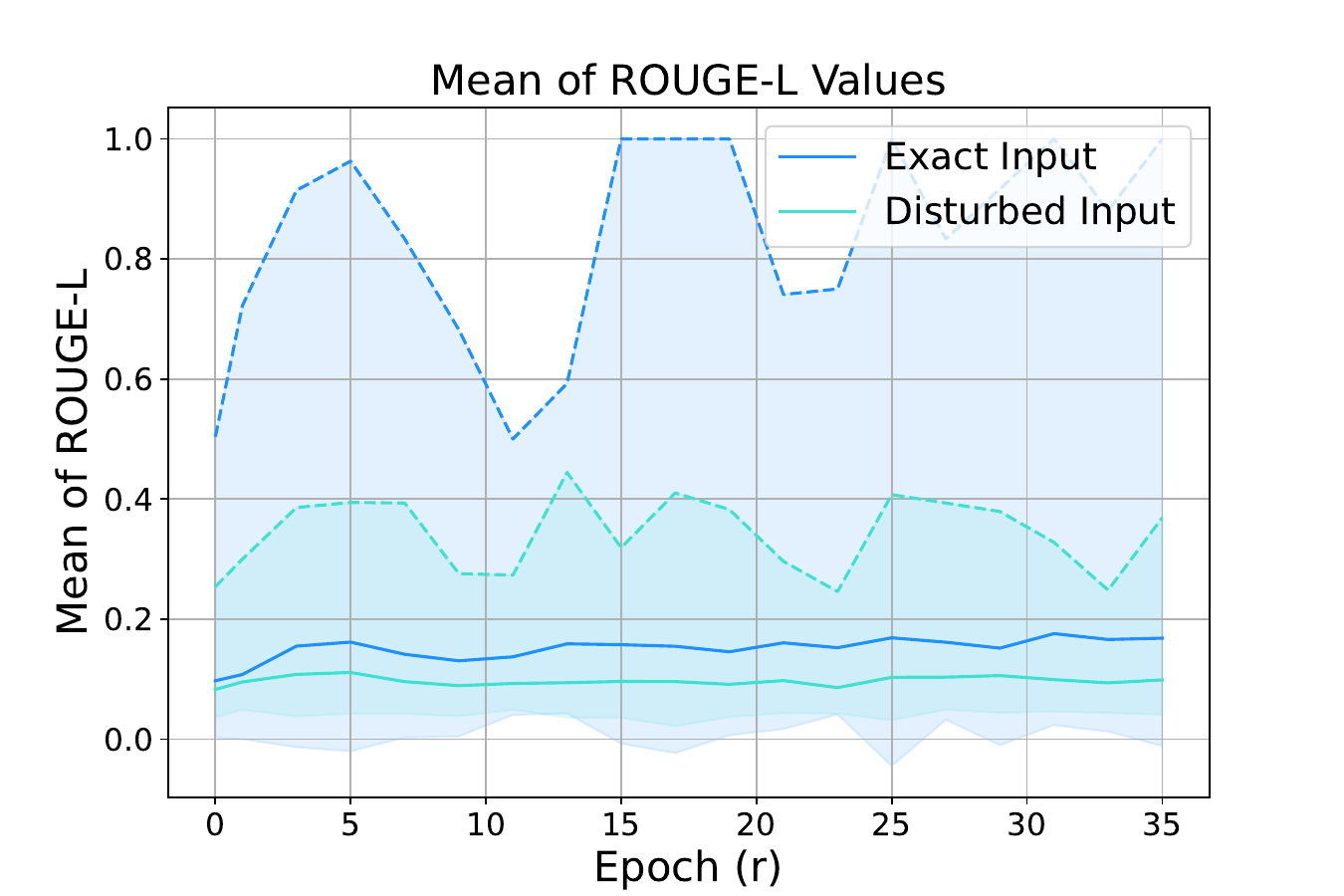}
    \label{fig:aug3B}
}
\\
\subfloat[Comparison on LLaMA-8B.]{
    \includegraphics[width=0.45\linewidth]{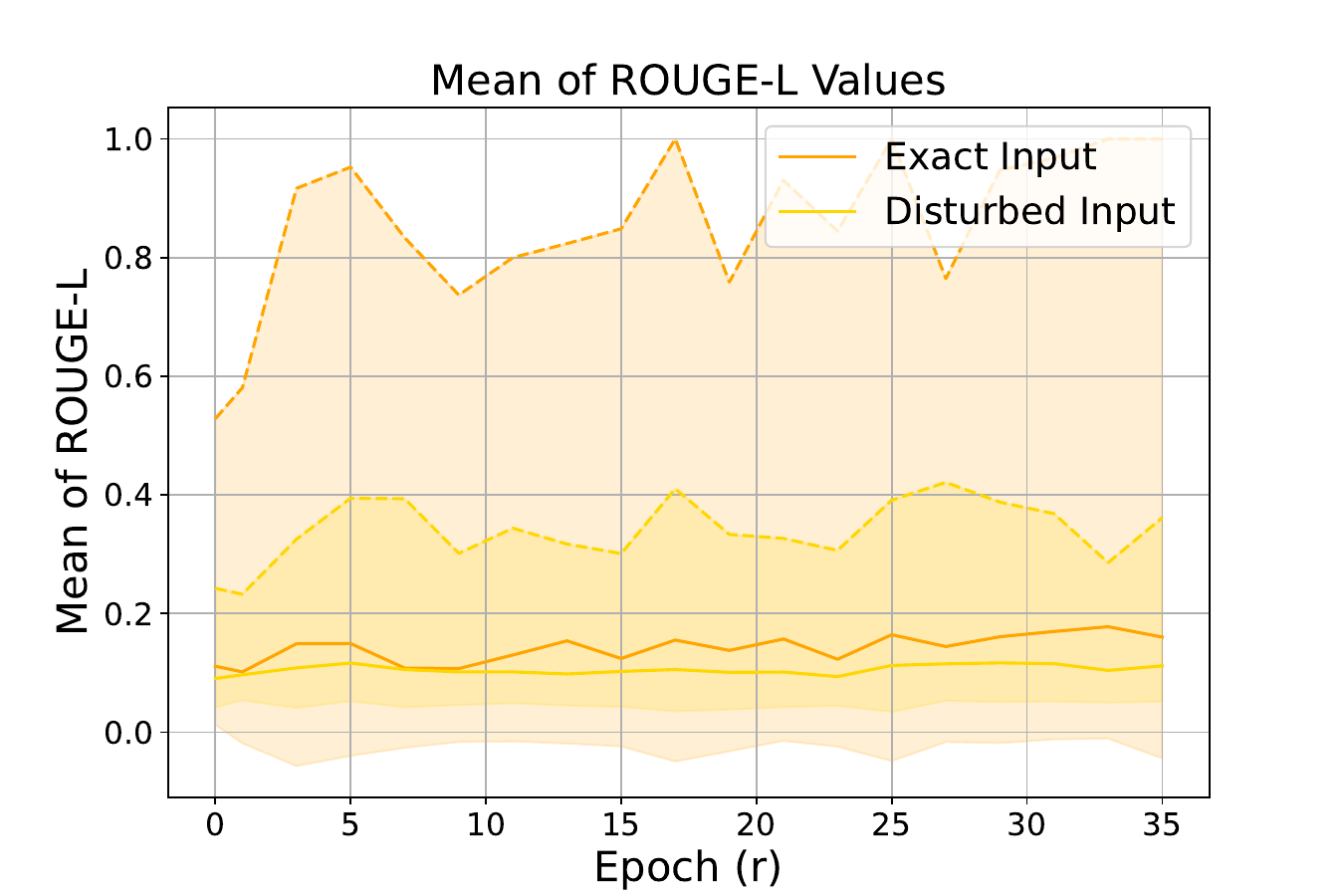}
    \label{fig:aug8B}
}
\subfloat[Disturbed-Input Completion on 3 LLaMA models.]{
    \includegraphics[width=0.45\linewidth]{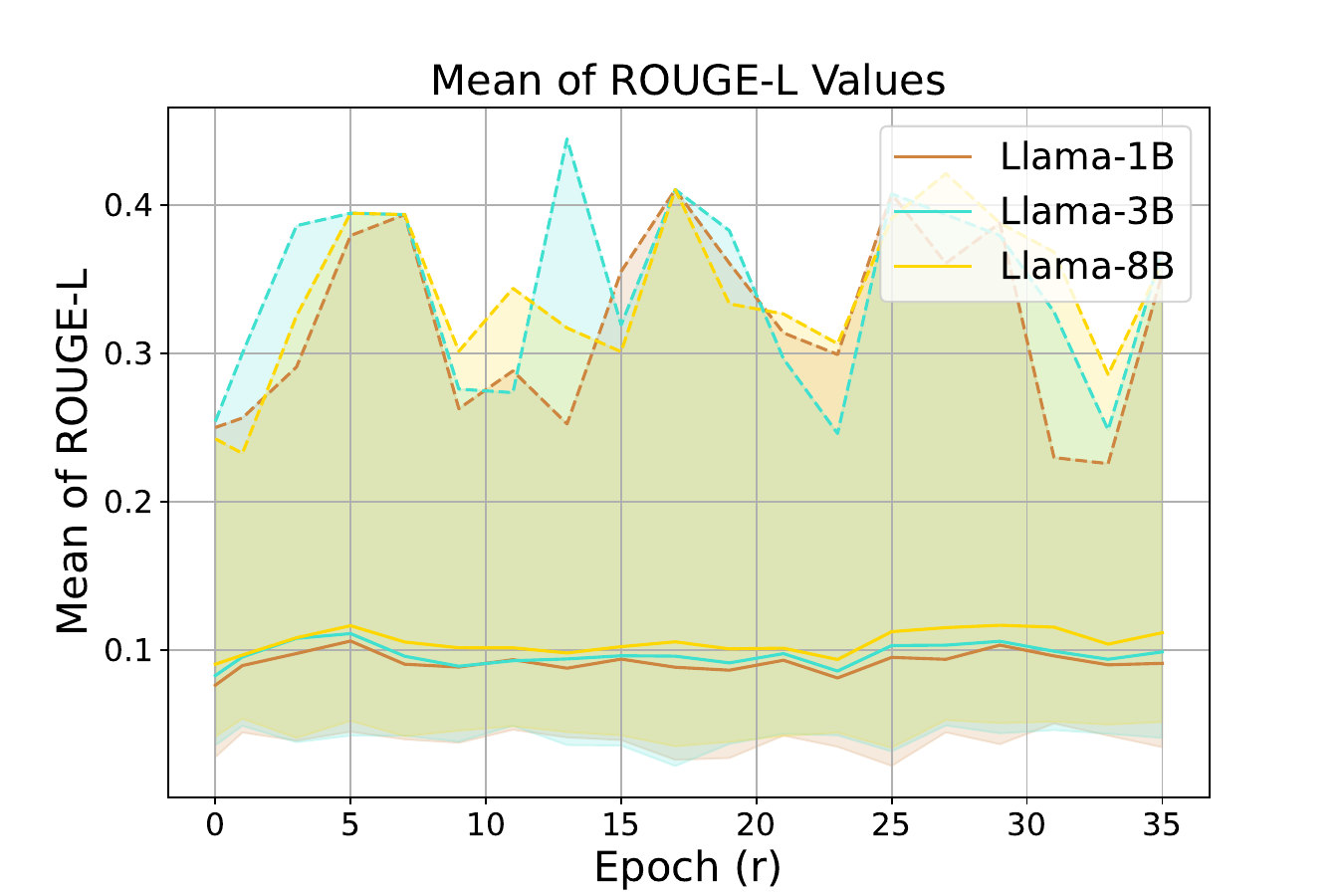}
    \label{fig:aug_all}
}
\caption{Disturbed-Input results compared with Partial-Input. This experiment is repeated across LLaMA-1B, LLaMA-3B and LLaMA-8B using Enron Email dataset.}
\label{fig:aug}
\vspace{-20pt}
\end{figure*}

\noindent\textbf{Results.}
We conduct experiments on the 1B, 3B, and 8B LLaMA models, comparing the results with those from the Partial-Input Completion task. As shown in Fig.~\ref{fig:aug}, for all three models, using perturbed vague information leads to less privacy leakage than using exact partial information. Notably, when hacking with exact partial information, some results approach a ROUGE-L score of 1, whereas hacking with perturbed vague information significantly reduces the upper bound of privacy leakage, maintaining a lower overall level.

The comparison of results across the three model groups is shown in Fig.~\ref{fig:aug_all}. It can be observed that the average leakage degree increases from 1B to 3B to 8B, indicating that larger models tend to suffer from more severe privacy leakage, which aligns with the conclusion in Section~\ref{section:com}.


\noindent\textbf{Discussion.} Overall, perturbing the input helps mitigate the extent of accurate privacy leakage in the global model. However, we need to emphasize that even under such a scenario, substantial data leakage might also occurs. This indicates that attackers are able to extract privacy information from vague, partial information, which suggests great potential threats.

\section{Results of Main Experiments under Different Thresholds}
\label{appendix:thresholds}
\begin{figure*}[!htbp]
\setlength{\abovecaptionskip}{6pt}
\centering
\subfloat[Top 10\% ROUGE values.]{
    \includegraphics[width=0.45\linewidth]{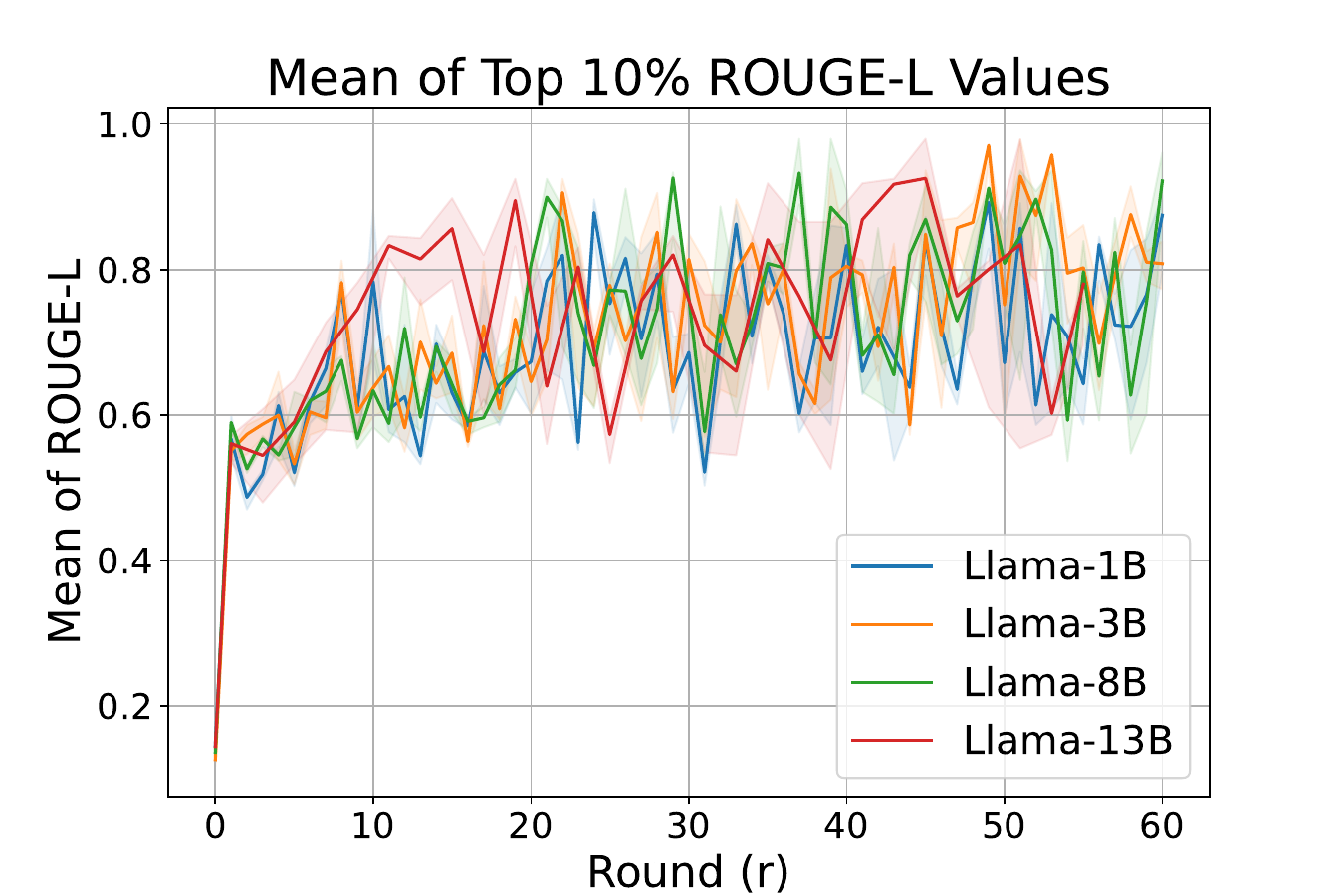}
}
\subfloat[Top 30\% ROUGE values.]{
    \includegraphics[width=0.45\linewidth]{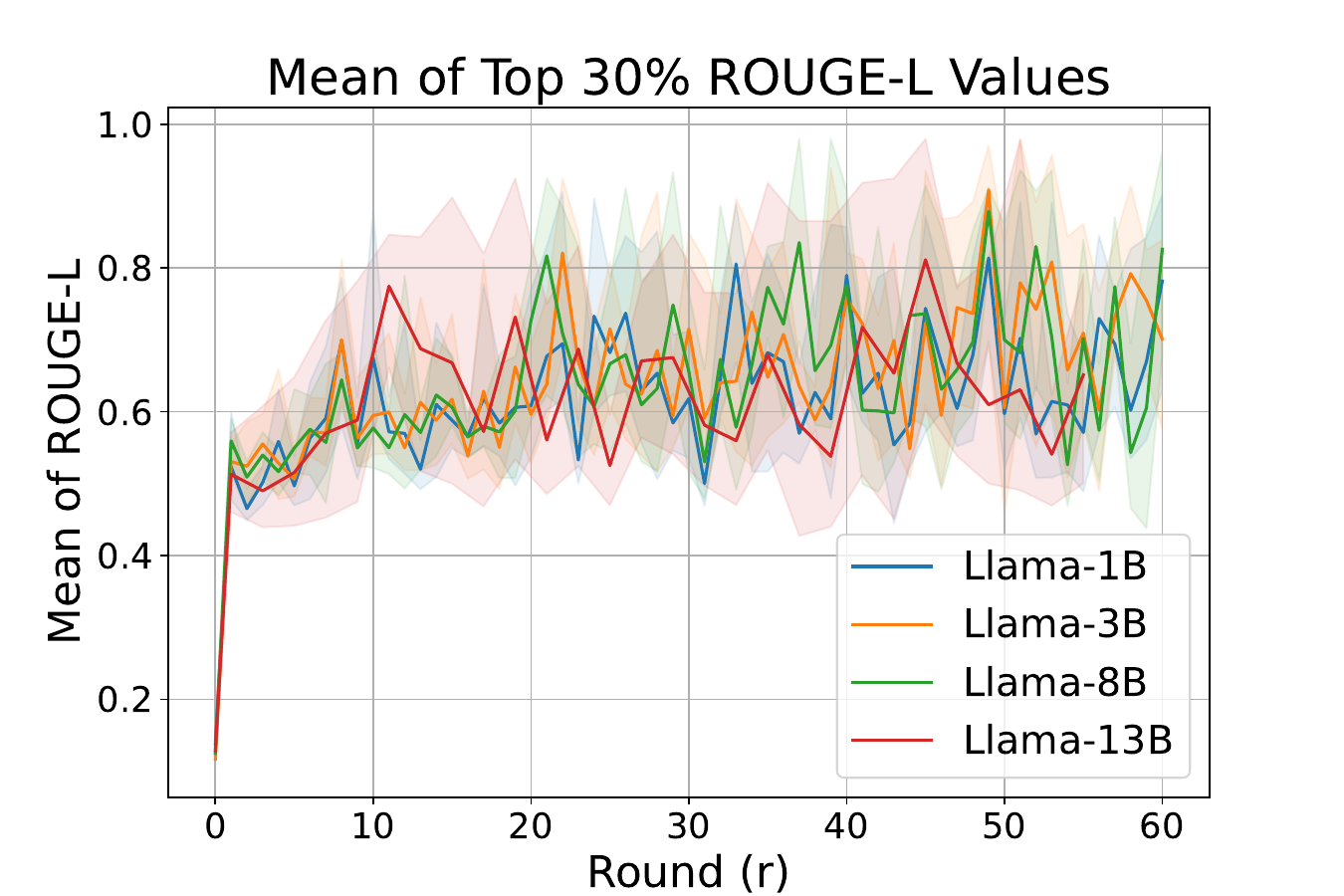}
}
\\
\subfloat[Top 50\% ROUGE values.]{
    \includegraphics[width=0.45\linewidth]{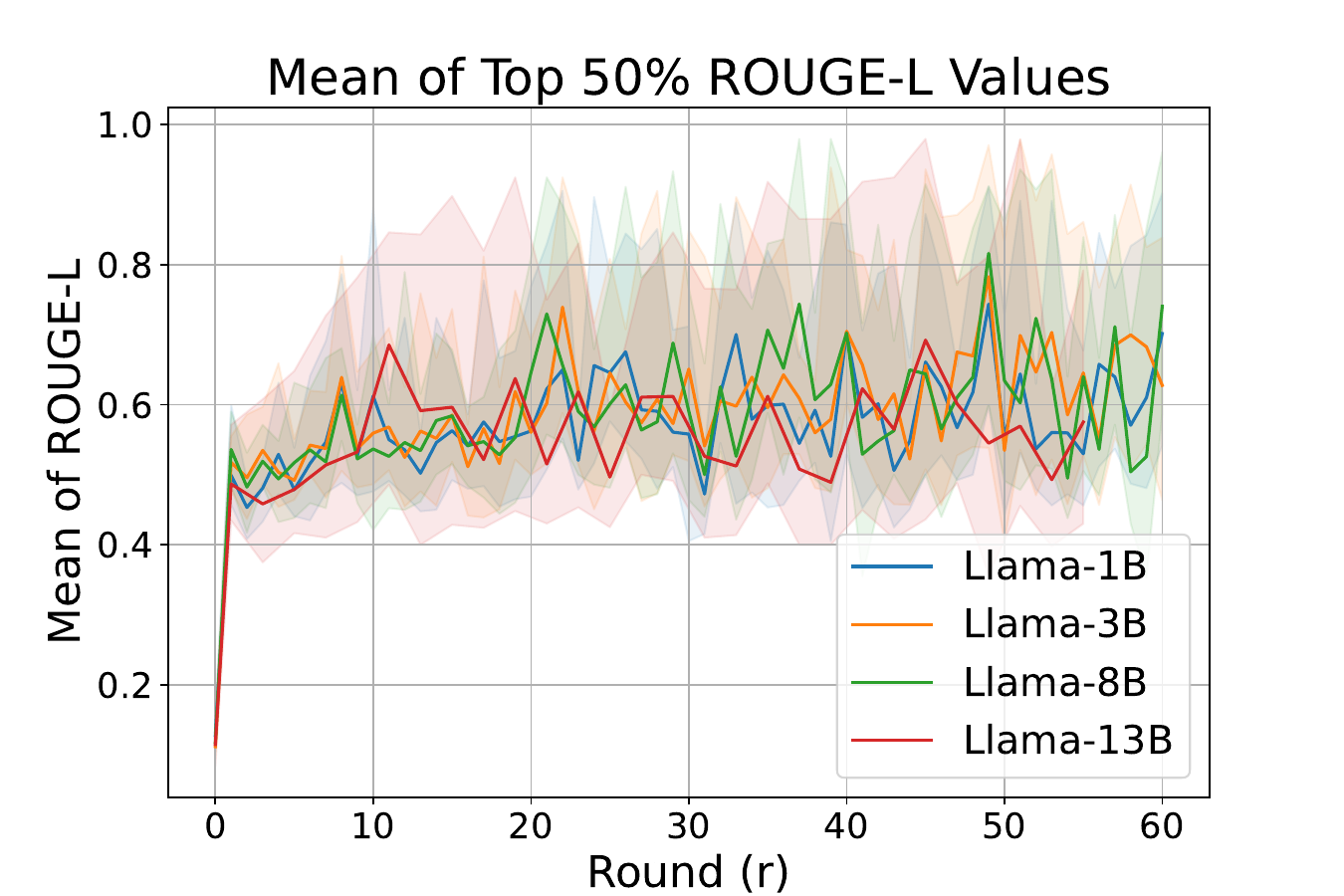}
}
\subfloat[Top 100\% ROUGE values.]{
    \includegraphics[width=0.45\linewidth]{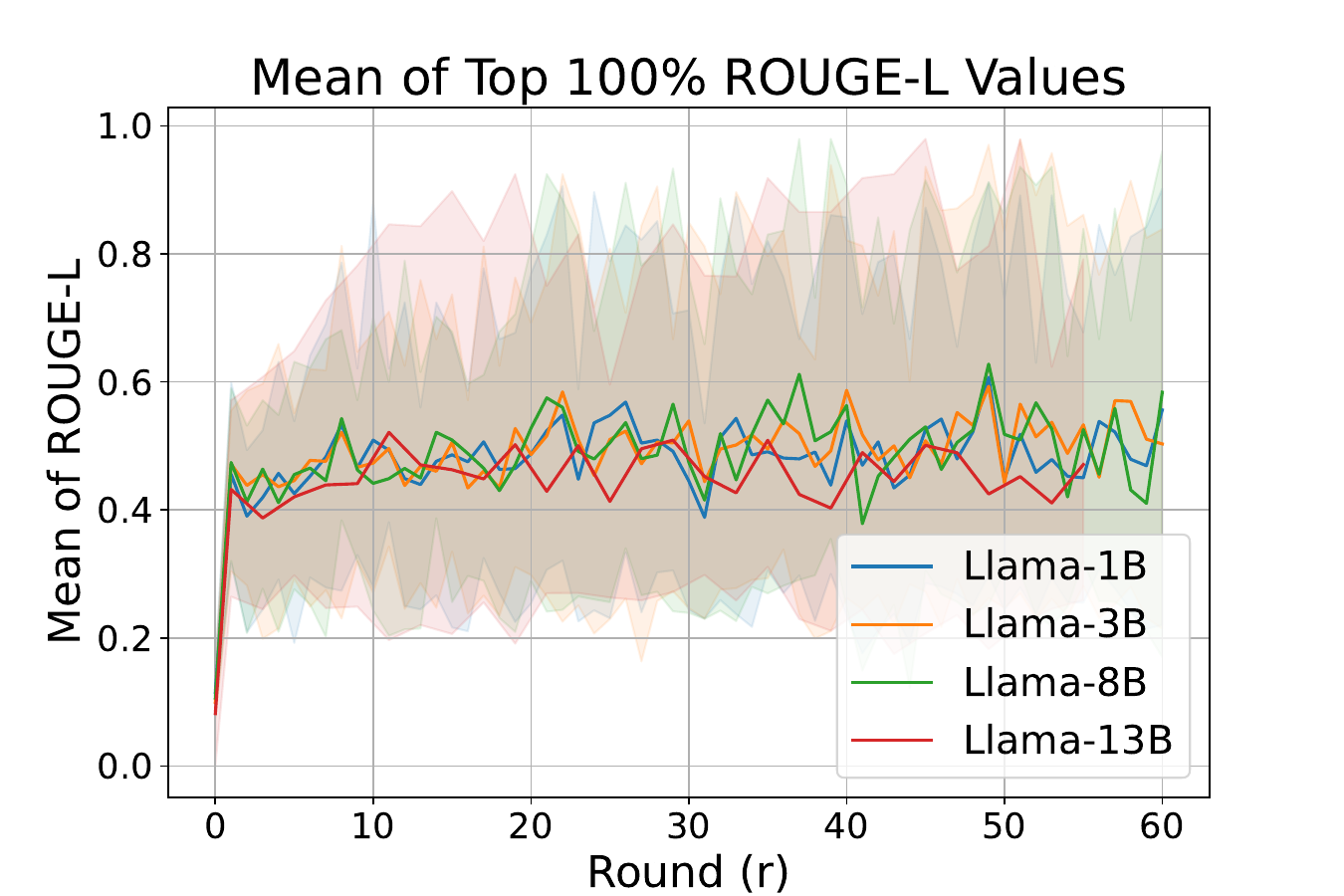}
}
\caption{Basic Hacking Scheme results of LLaMA-1B, LLaMA-3B, LLaMA-8B, LLaMA-13B models for Zero-Input Generation using Enron Email dataset.}
\label{fig:threshold1}
\vspace{-15pt}
\end{figure*}

\begin{figure*}[!htbp]
\setlength{\abovecaptionskip}{6pt}
\centering
\subfloat[Top 10\% on LLaMA-8B.]{
    \includegraphics[width=0.23\linewidth]{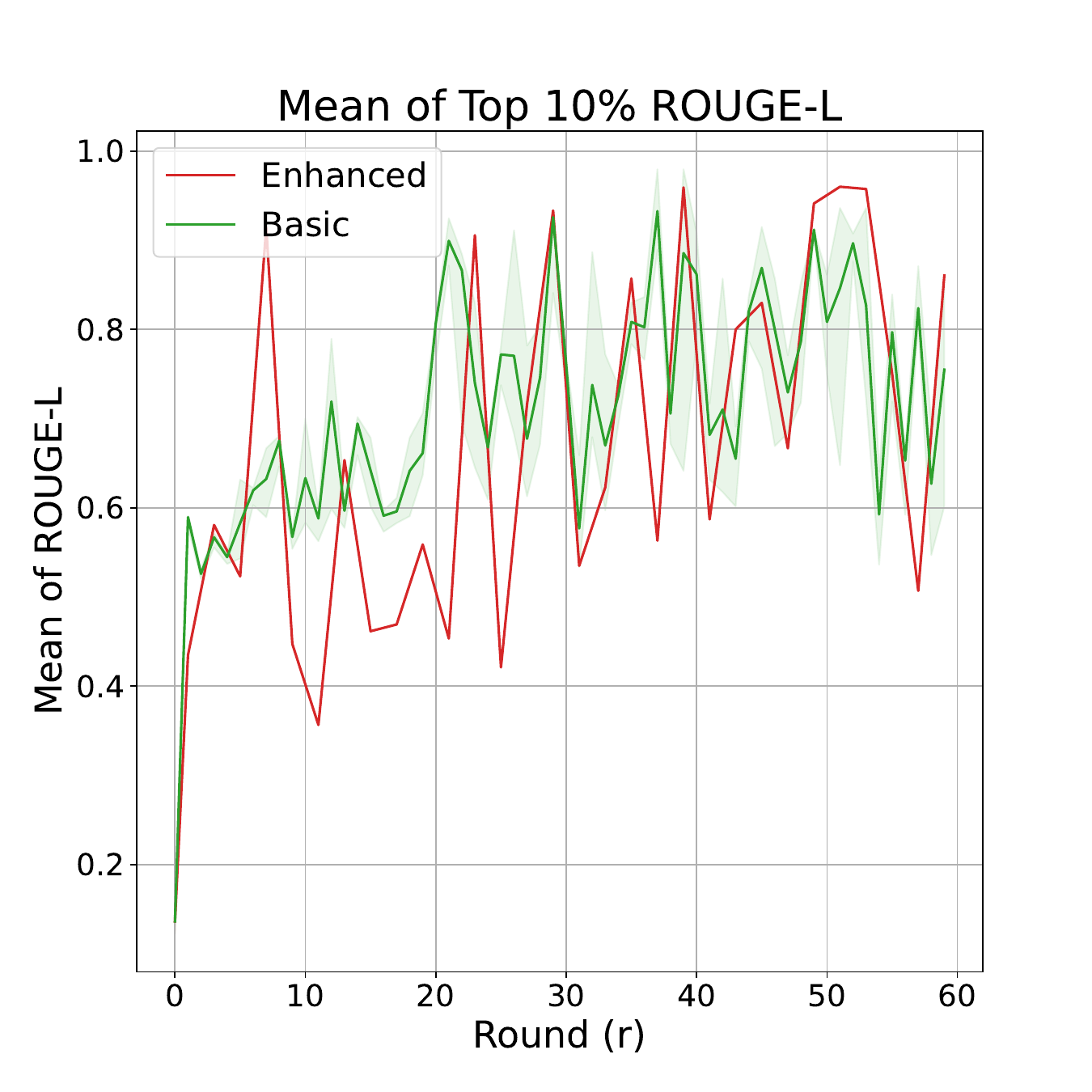}
}
\subfloat[Top 30\% on LLaMA-8B.]{
    \includegraphics[width=0.23\linewidth]{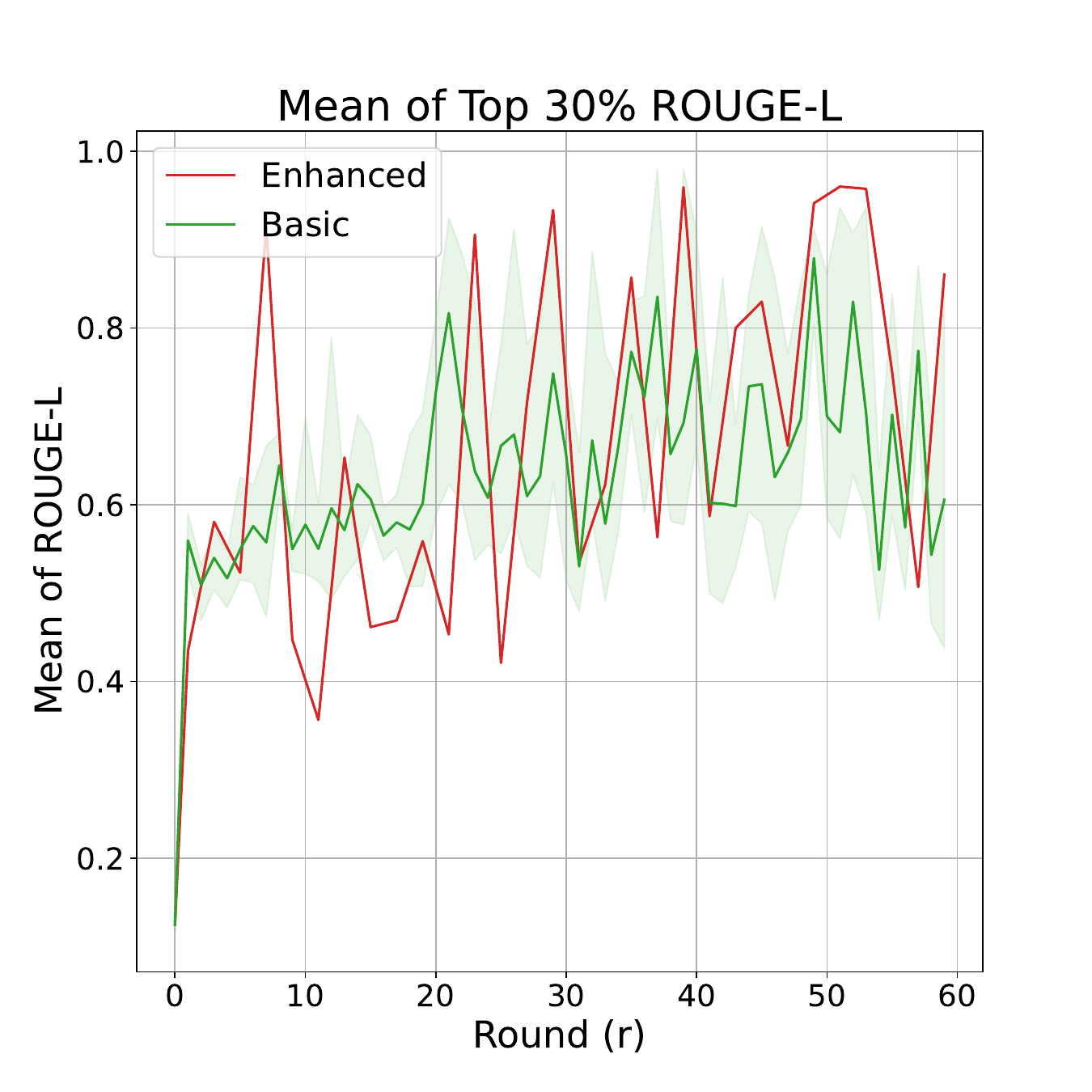}
}
\subfloat[Top 50\% on LLaMA-8B.]{
    \includegraphics[width=0.23\linewidth]{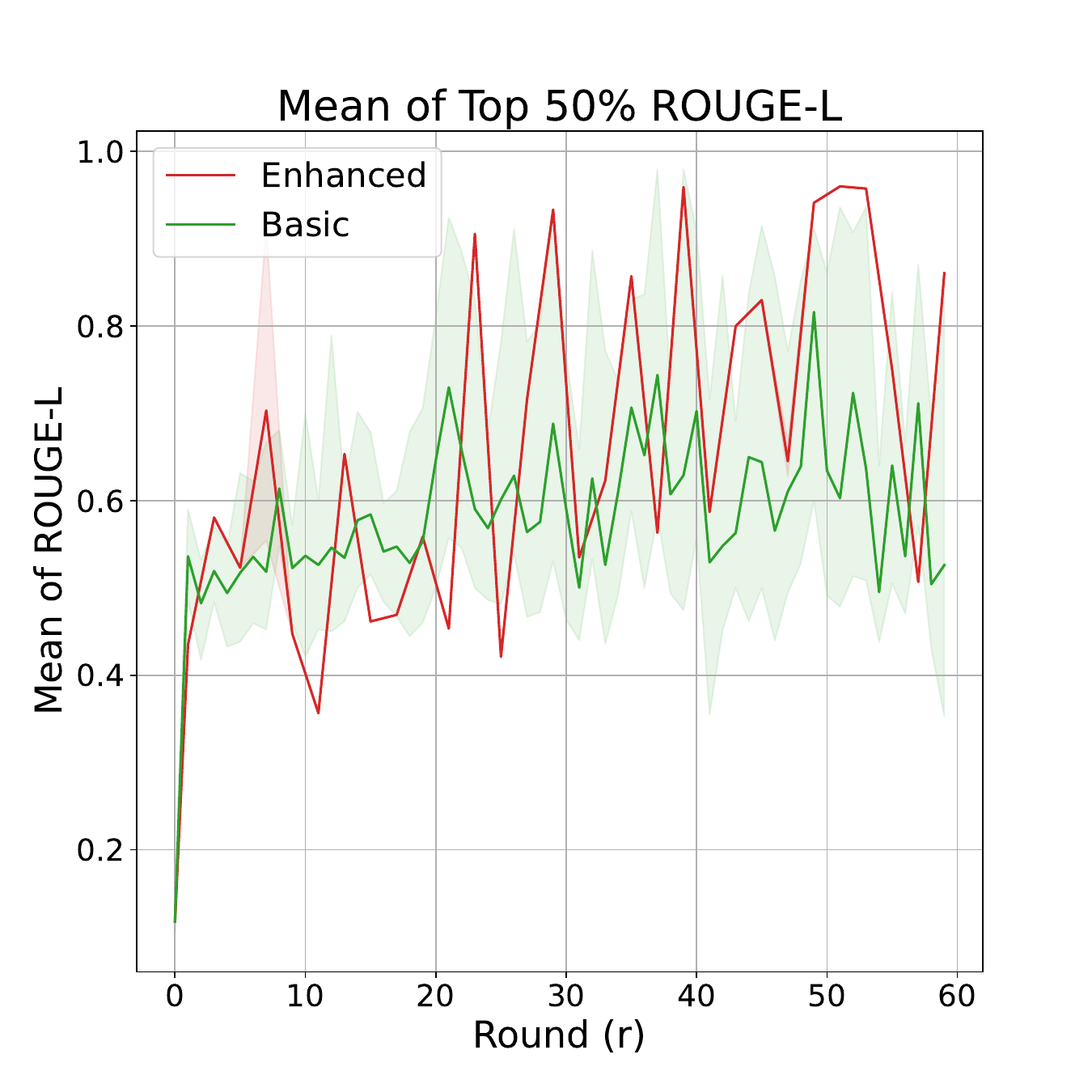}
}
\subfloat[Top 100\% on LLaMA-8B.]{
    \includegraphics[width=0.23\linewidth]{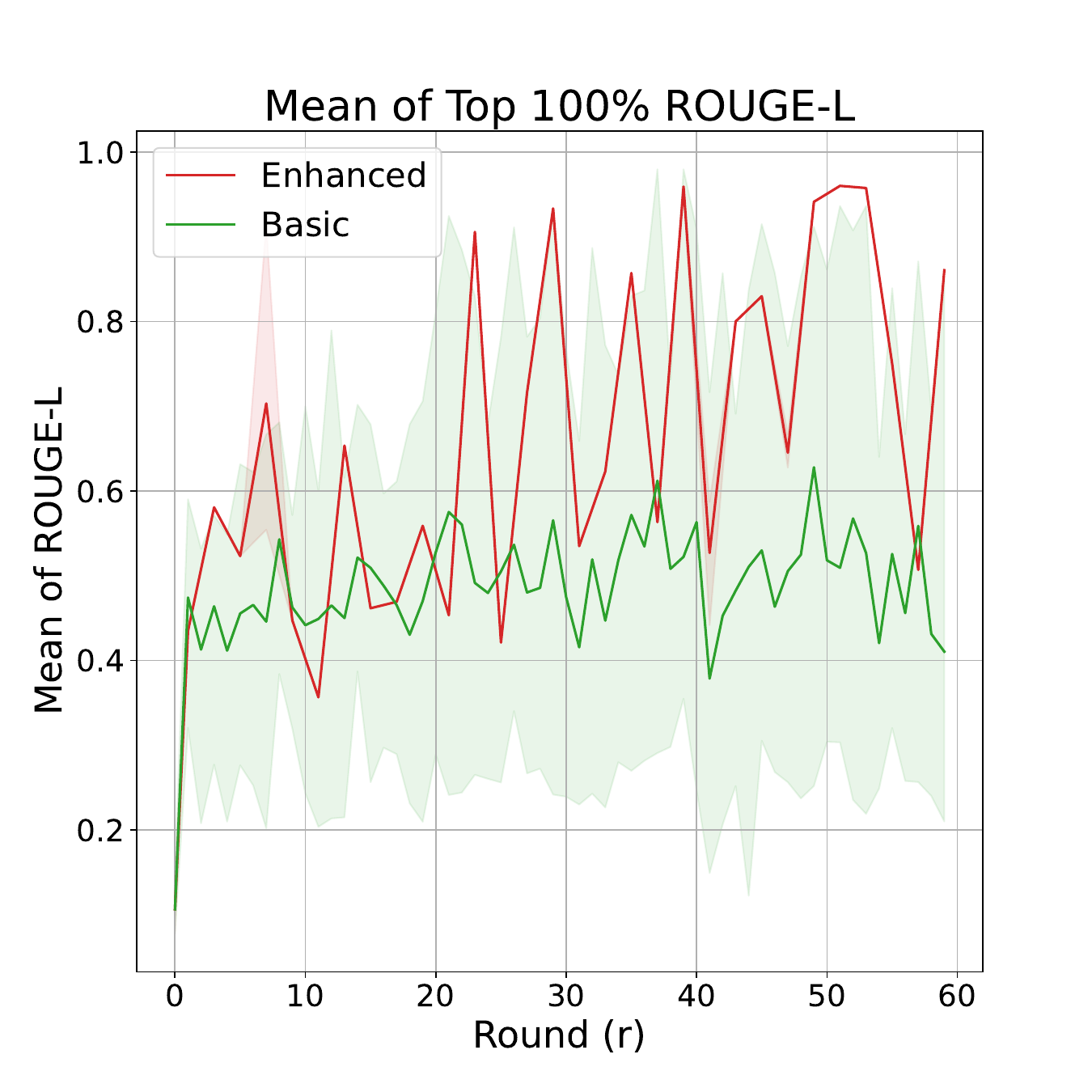}
}
\\
\subfloat[Top 10\% on LLaMA-13B.]{
    \includegraphics[width=0.23\linewidth]{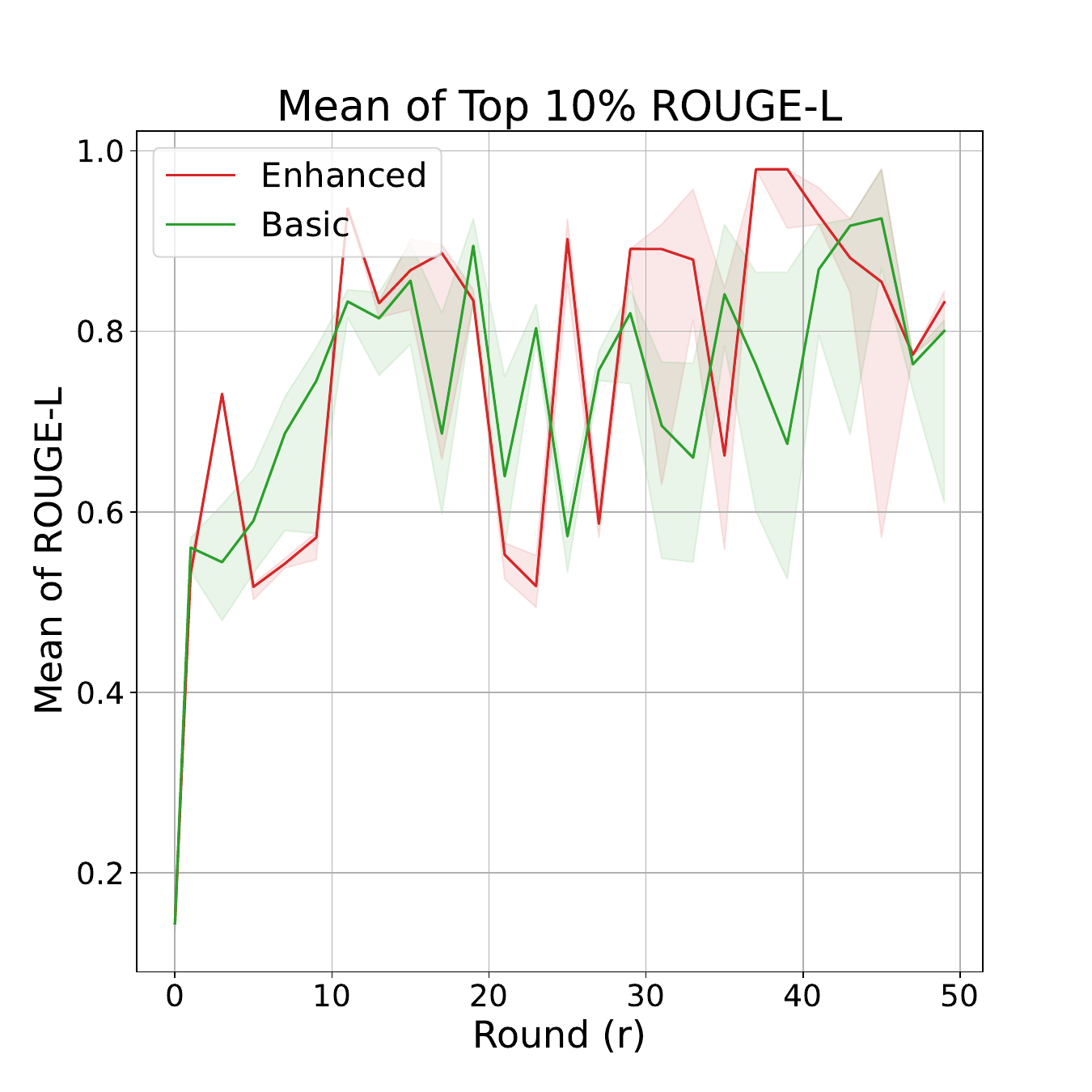}
}
\subfloat[Top 30\% on LLaMA-13B.]{
    \includegraphics[width=0.23\linewidth]{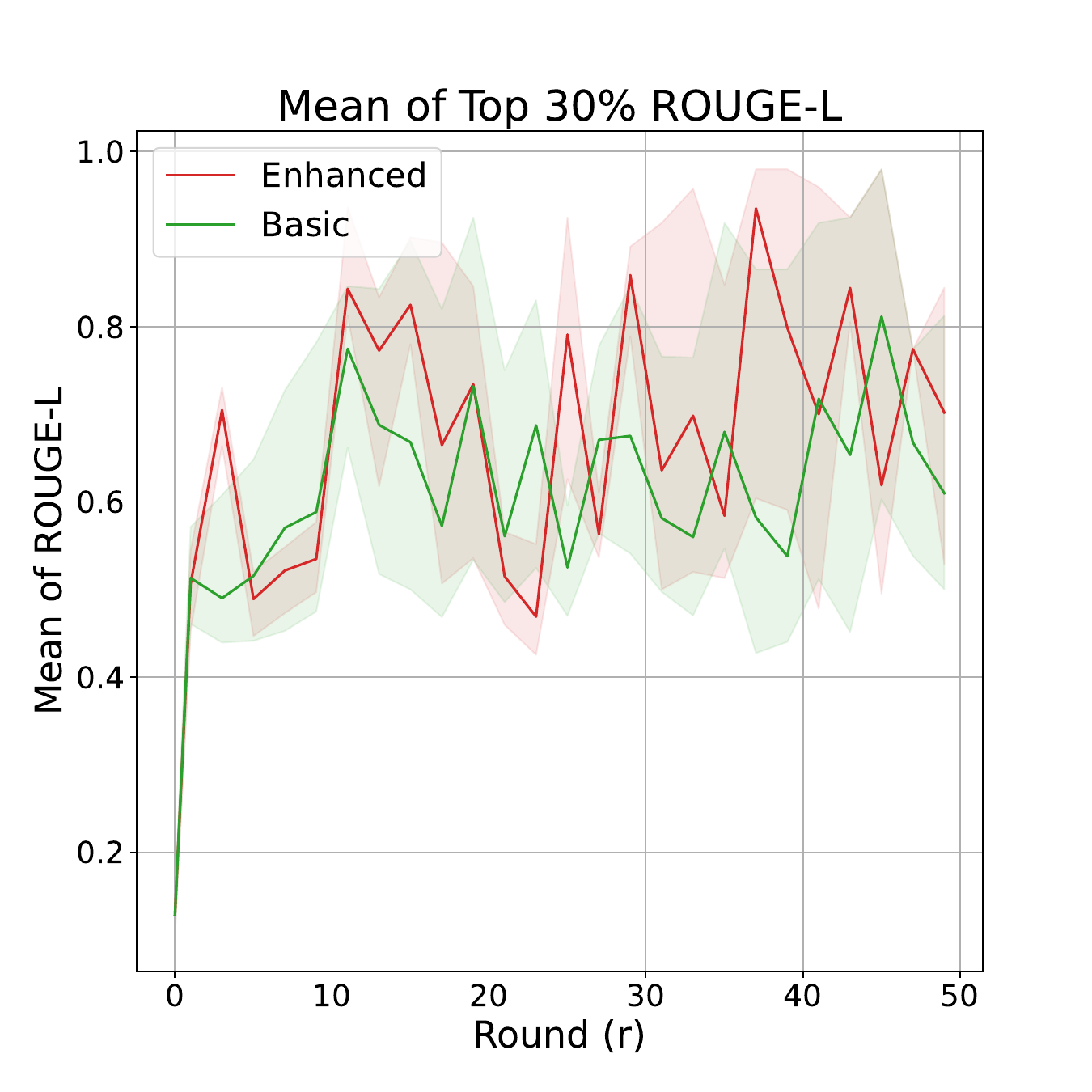}
}
\subfloat[Top 50\% on LLaMA-13B.]{
    \includegraphics[width=0.23\linewidth]{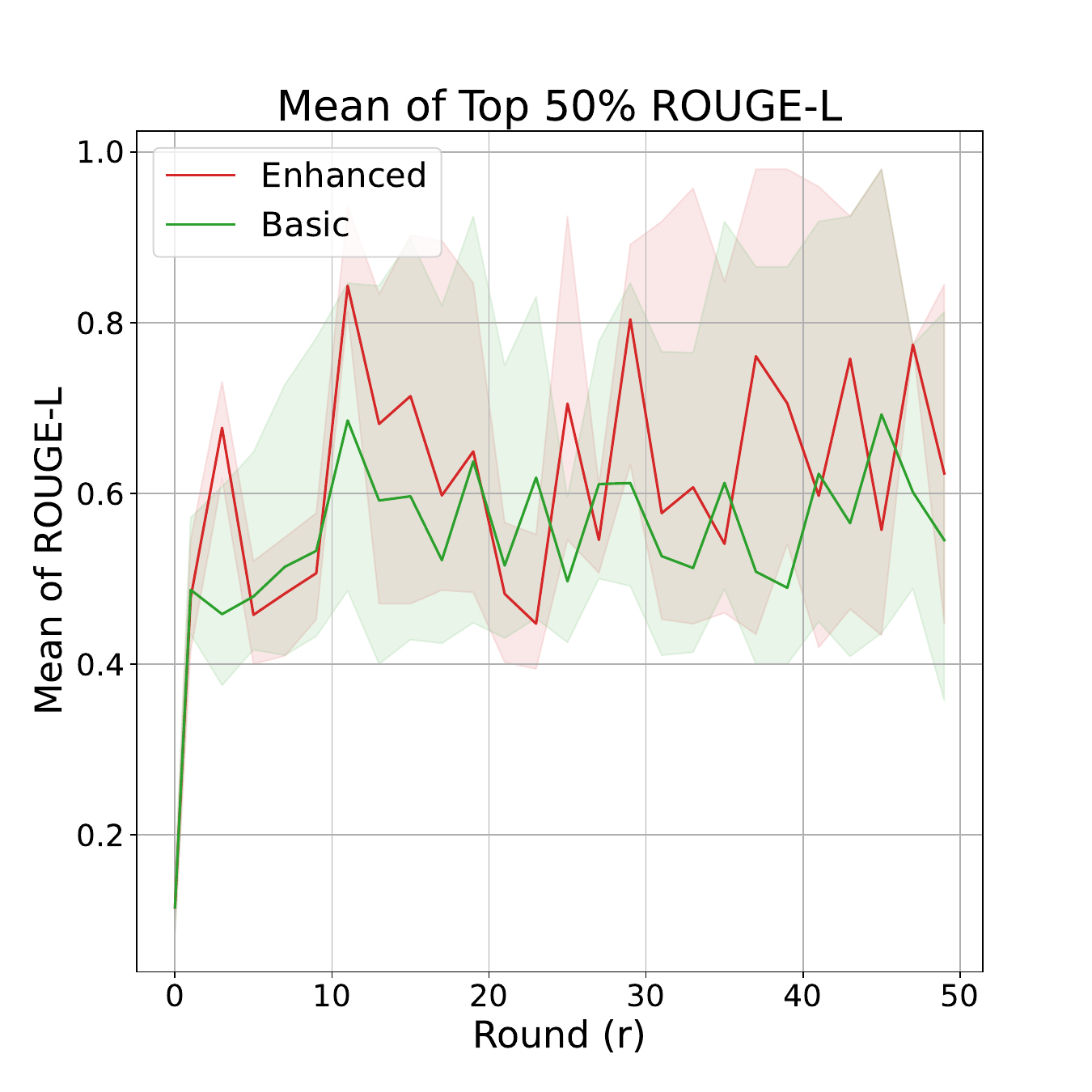}
}
\subfloat[Top 100\% on LLaMA-13B.]{
    \includegraphics[width=0.23\linewidth]{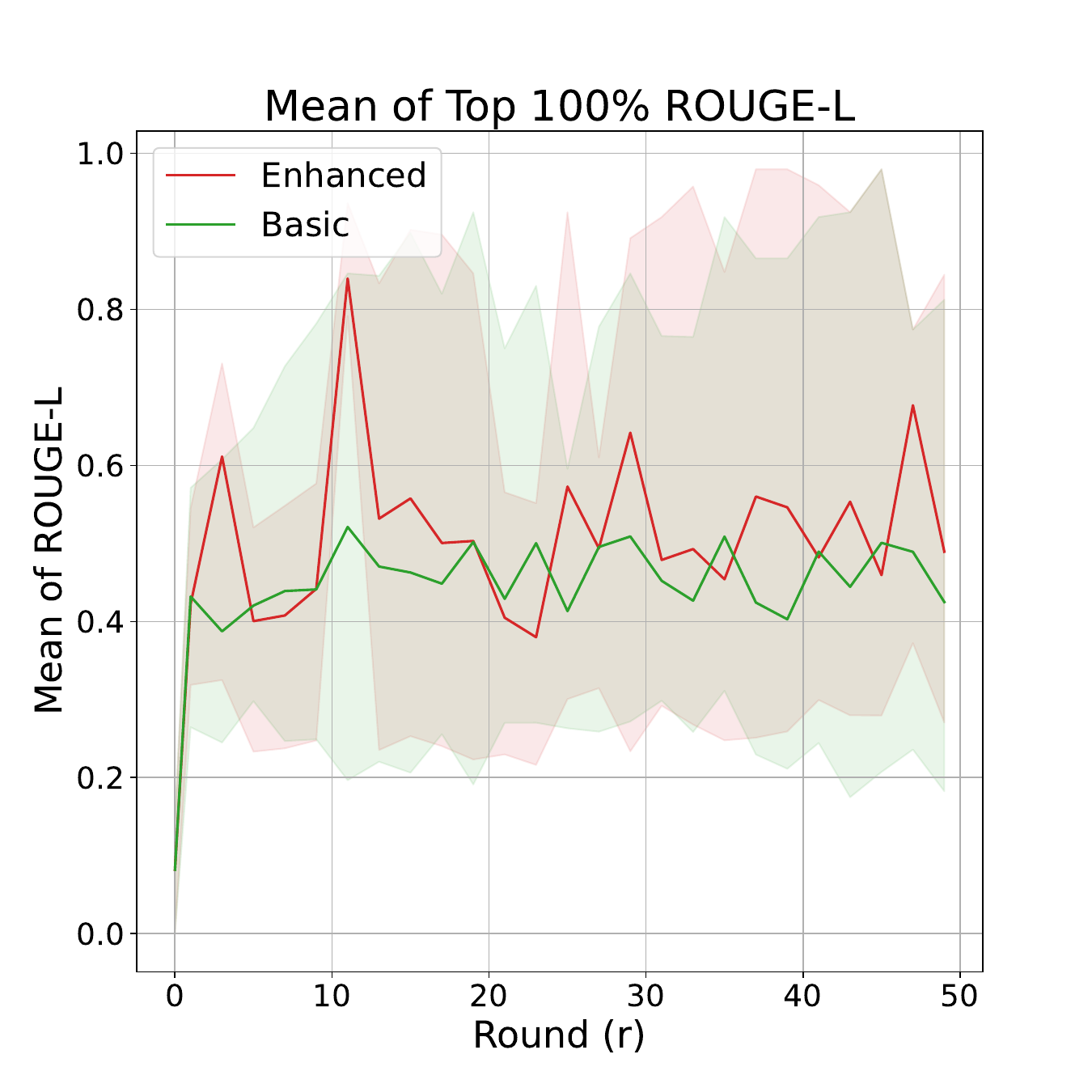}
}
\caption{Comparison of two hacking schemes' results for Zero-Input Generation. Experiments are repeated using LLaMA-8B and LLaMA-13B using Enron Email dataset.}
\label{fig:threshold2}
\vspace{-15pt}
\end{figure*}

\begin{figure*}[!htbp]
\setlength{\abovecaptionskip}{6pt}
\centering
\subfloat[Top 10\% ROUGE values.]{
    \includegraphics[width=0.45\linewidth]{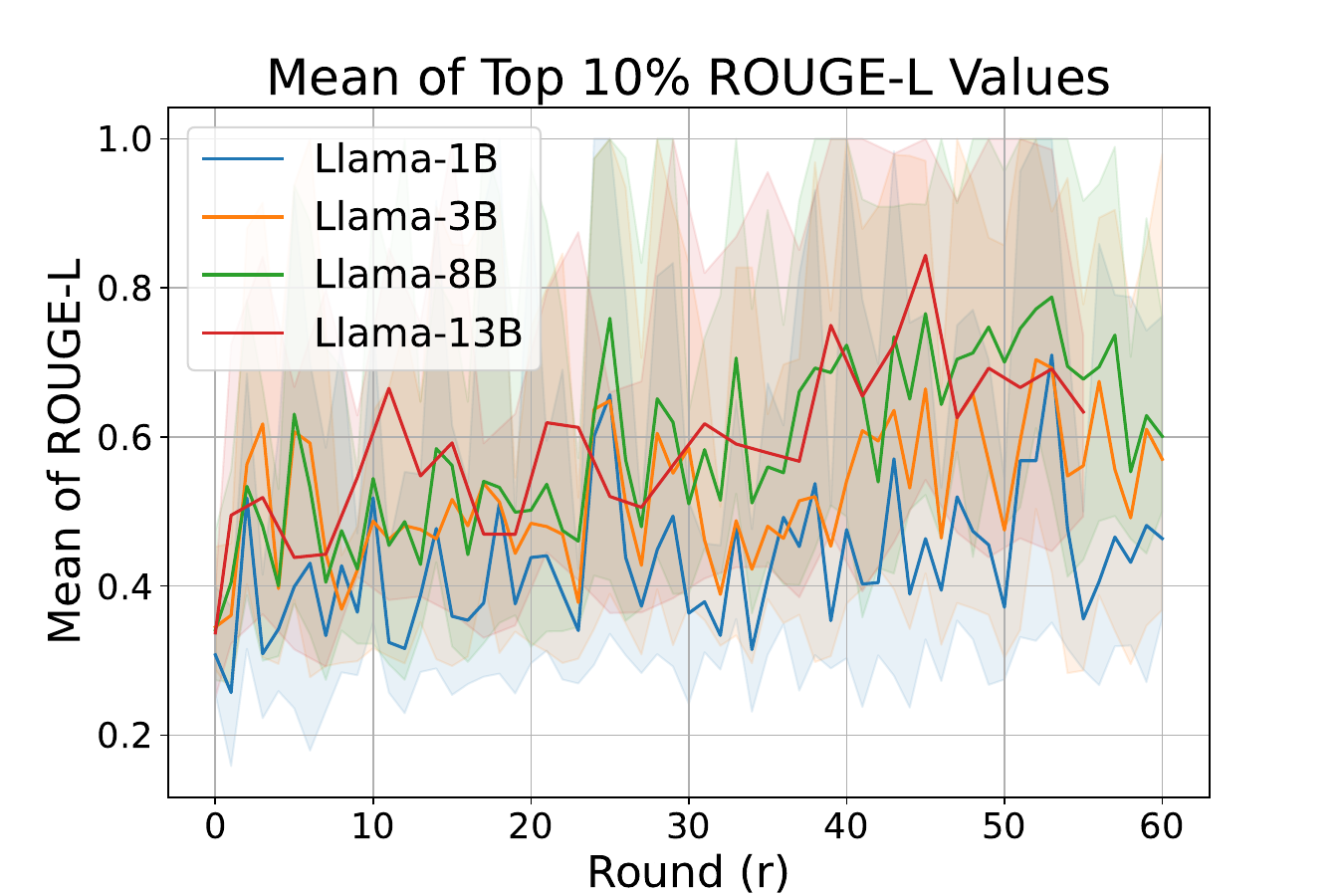}
}
\subfloat[Top 30\% ROUGE values.]{
    \includegraphics[width=0.45\linewidth]{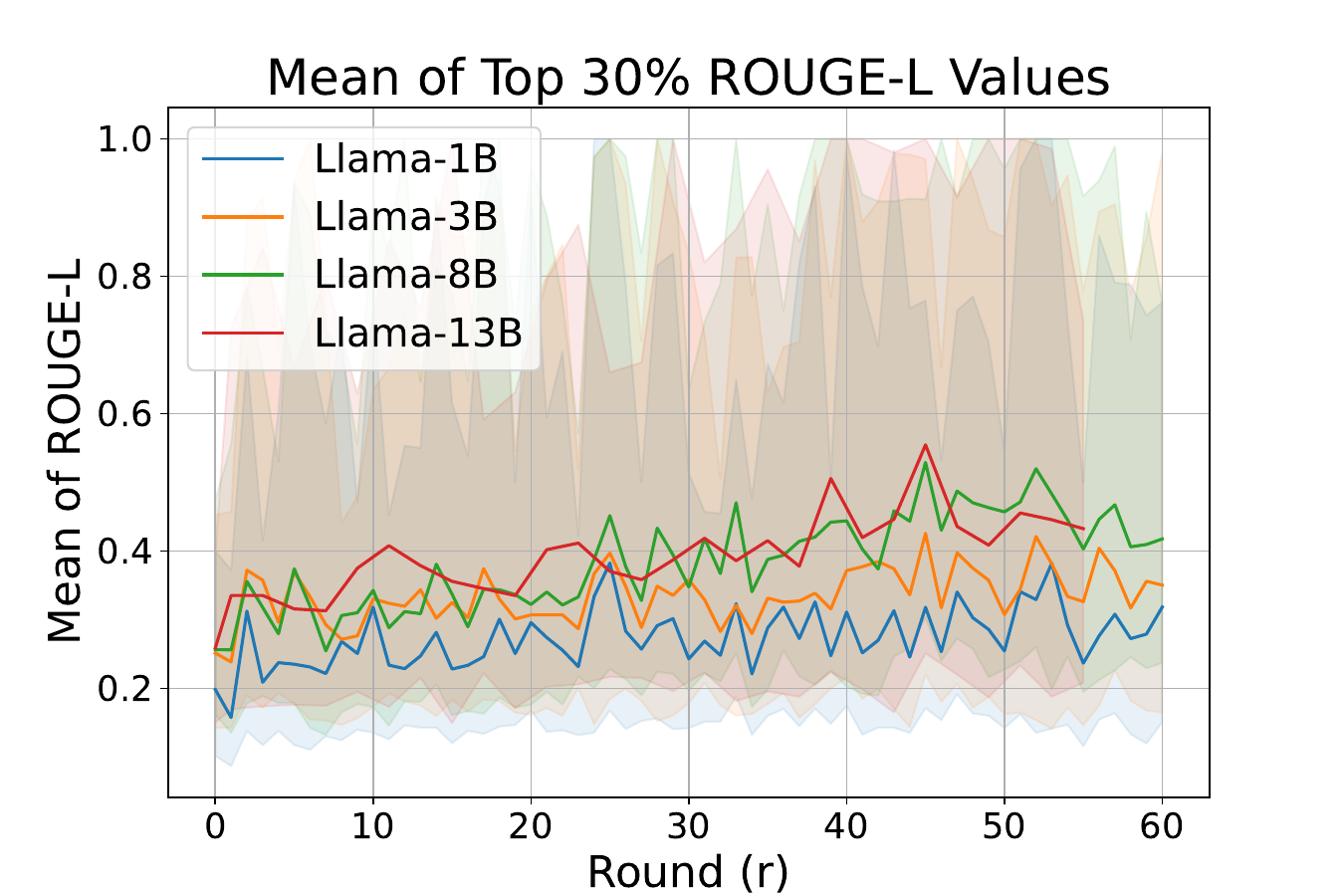}
}
\\
\subfloat[Top 50\% ROUGE values.]{
    \includegraphics[width=0.45\linewidth]{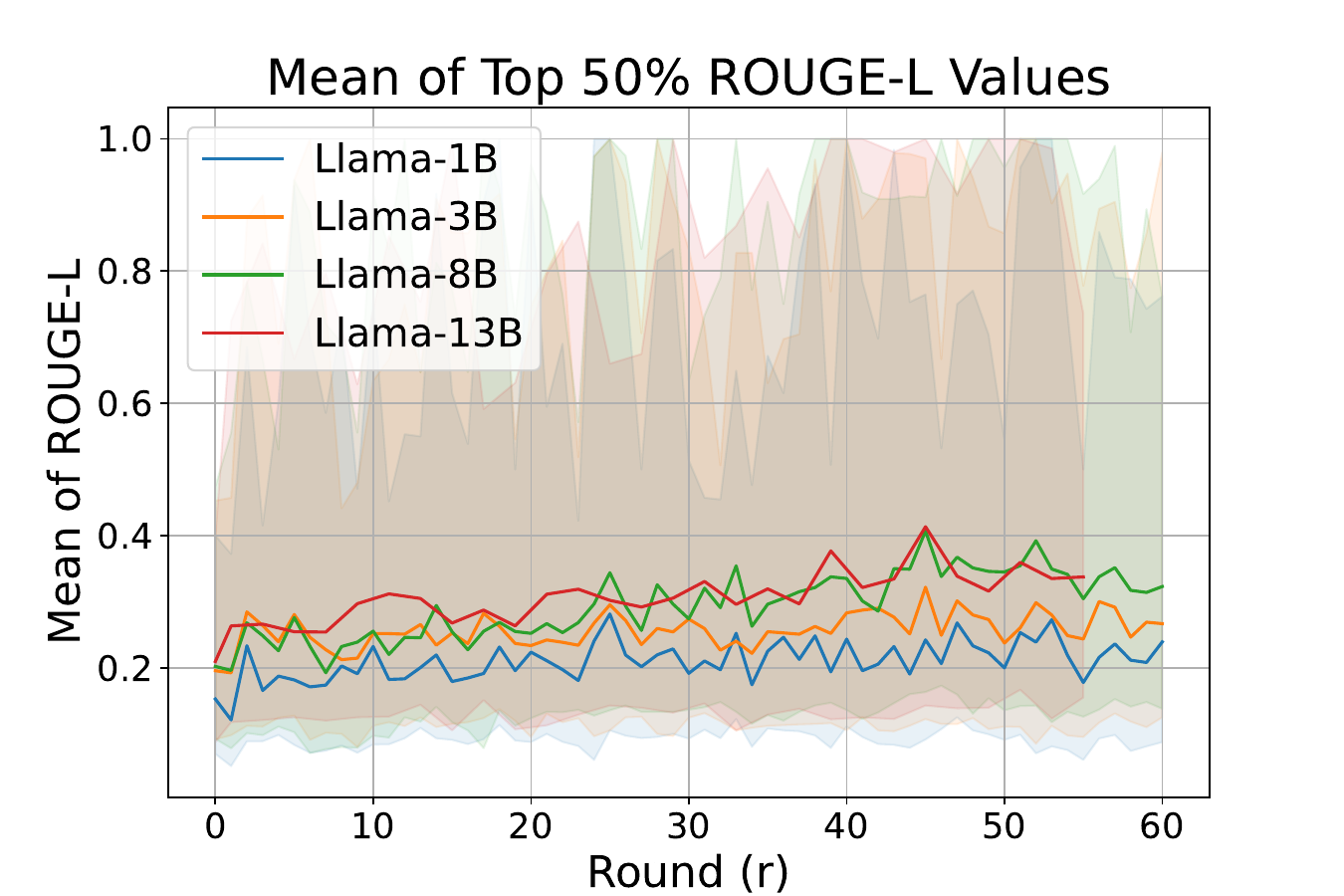}
}
\subfloat[Top 100\% ROUGE values.]{
    \includegraphics[width=0.45\linewidth]{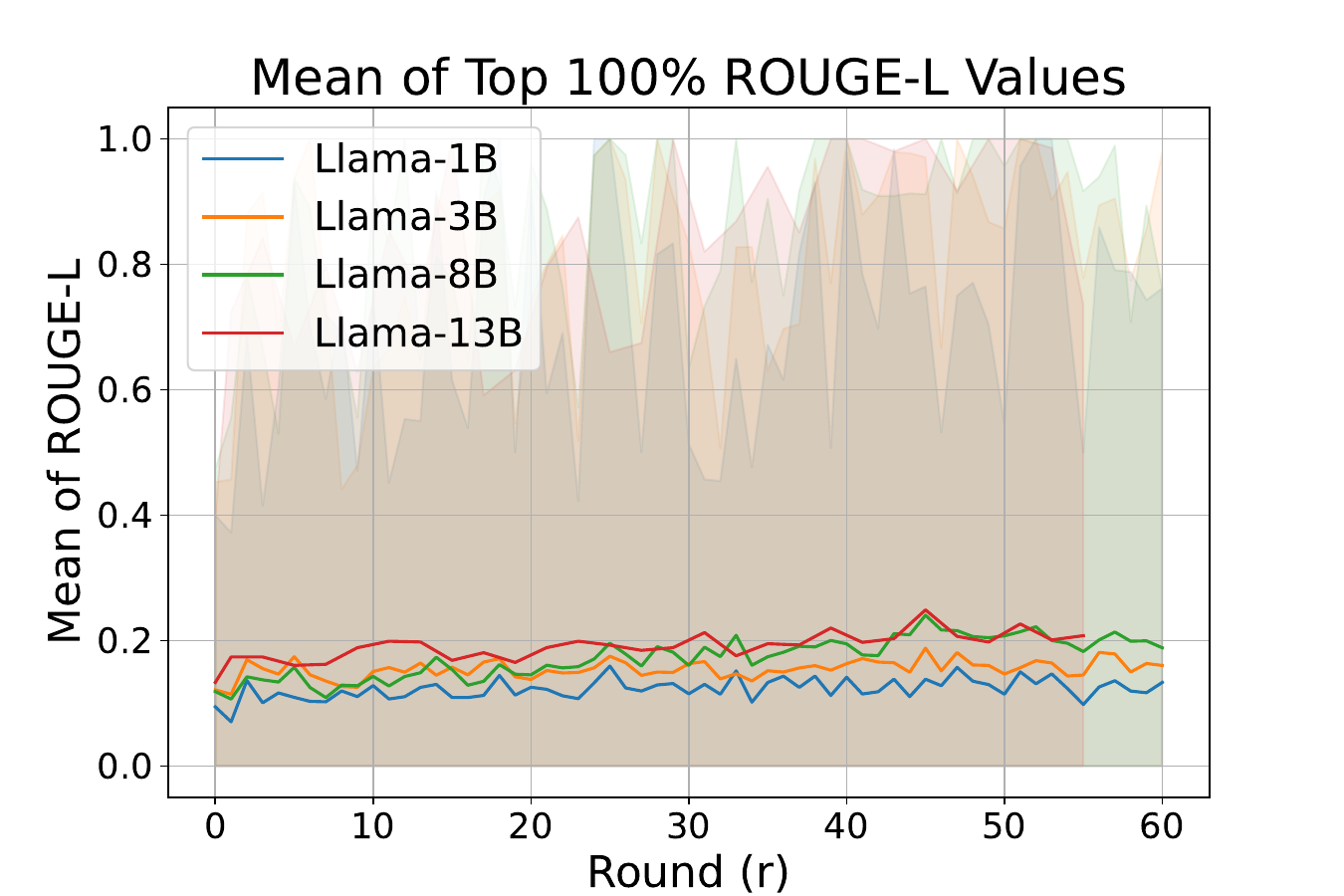}
}
\caption{Basic Hacking Scheme results of LLaMA models for Partial-Input Completion. Experiments are repeated using LLaMA-1B, LLaMA-3B, LLaMA-8B and LLaMA-13B using Enron Email dataset.}
\label{fig:threshold3}
\vspace{-15pt}
\end{figure*}

\begin{figure*}[!htbp]
\setlength{\abovecaptionskip}{6pt}
\centering
\subfloat[Top 10\% on LLaMA-8B.]{
    \includegraphics[width=0.23\linewidth]{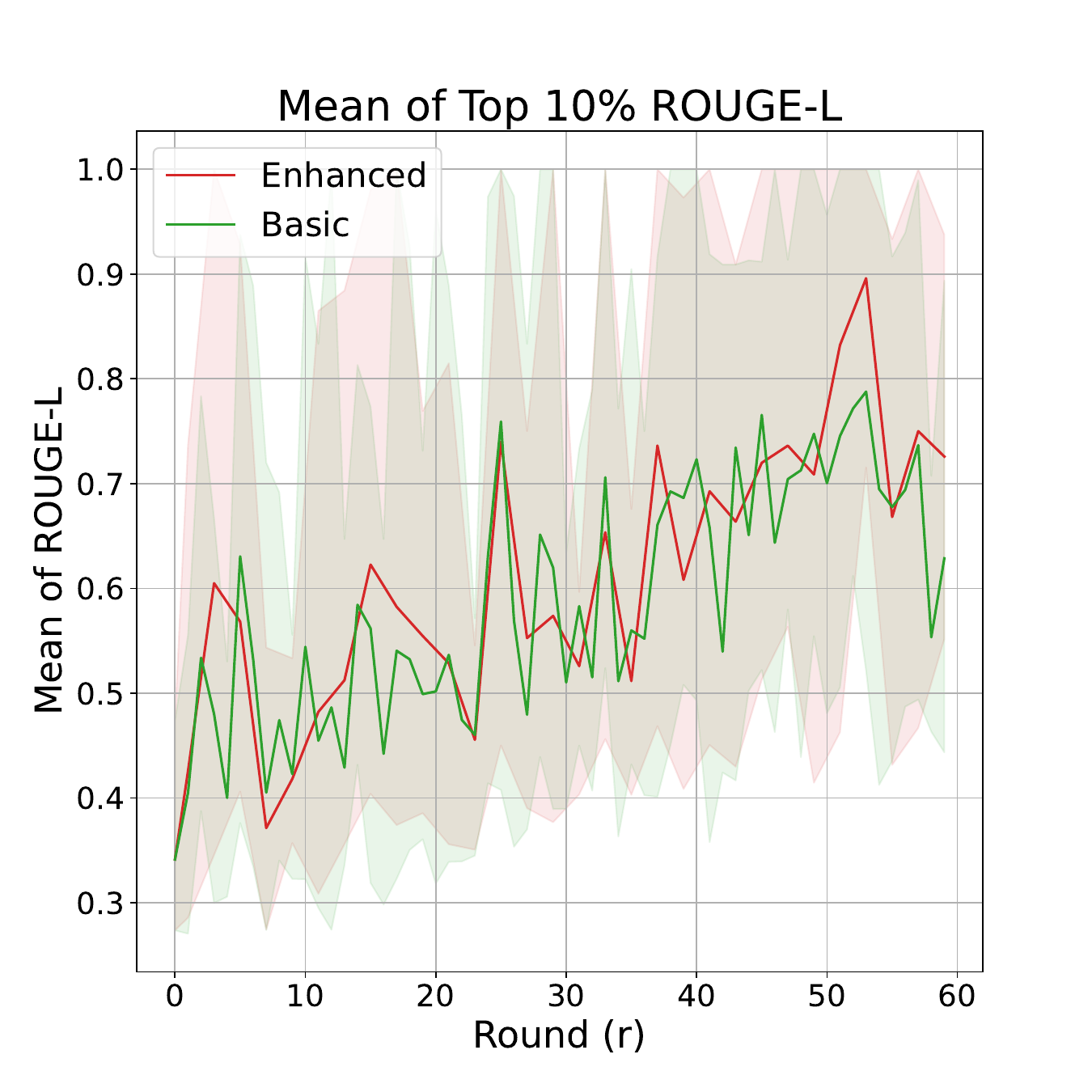}
}
\subfloat[Top 30\% on LLaMA-8B.]{
    \includegraphics[width=0.23\linewidth]{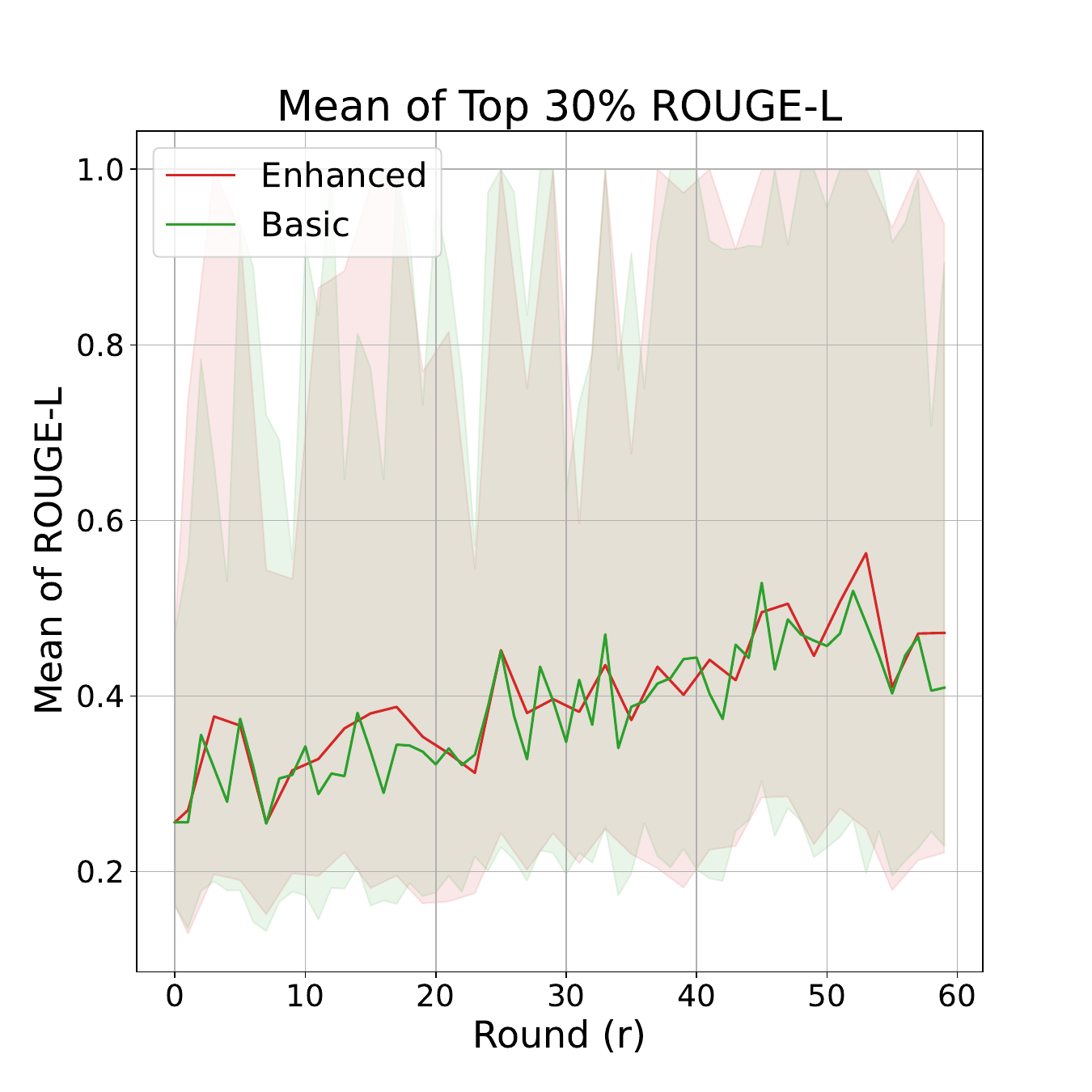}
}
\subfloat[Top 50\% on LLaMA-8B.]{
    \includegraphics[width=0.23\linewidth]{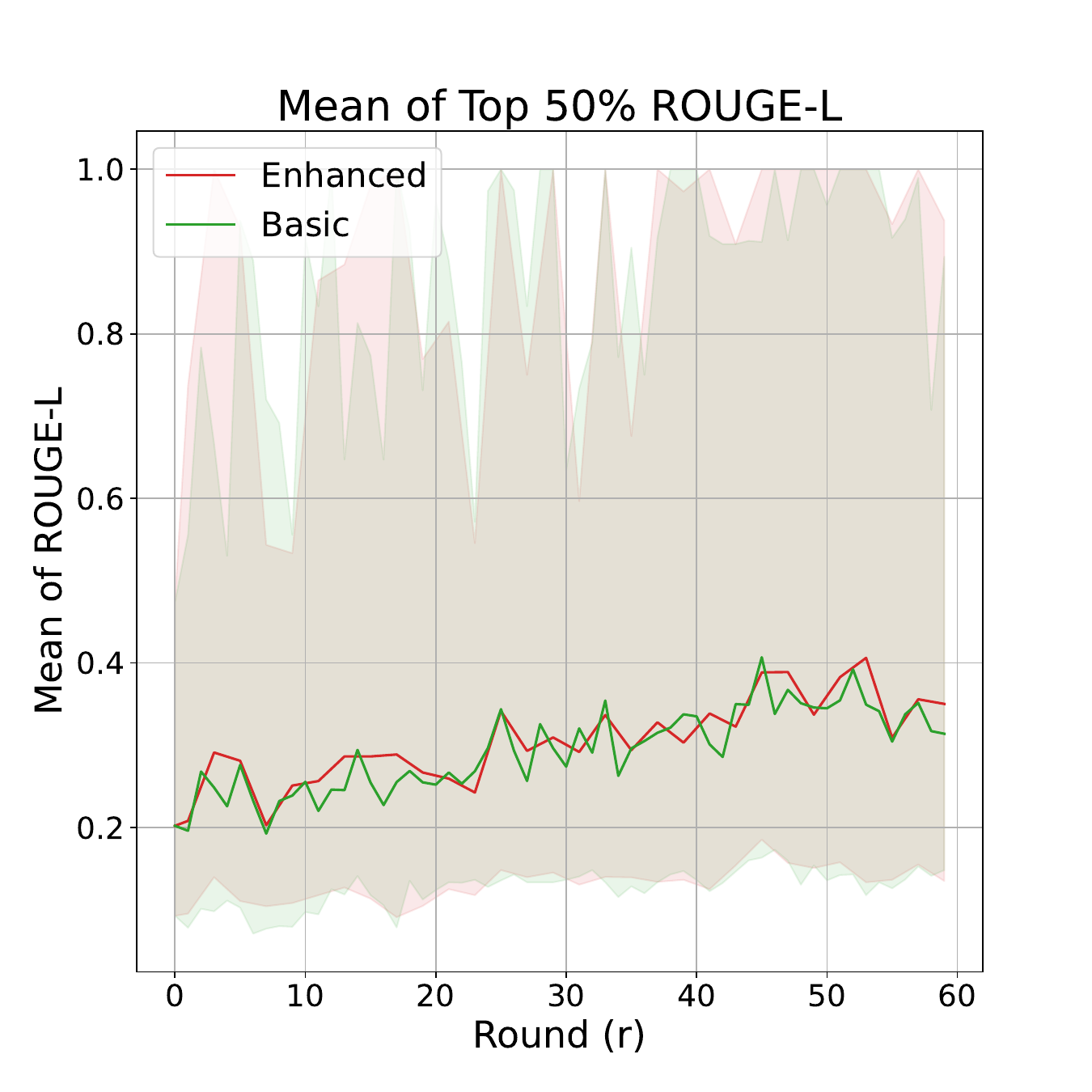}
}
\subfloat[Top 100\% on LLaMA-8B.]{
    \includegraphics[width=0.23\linewidth]{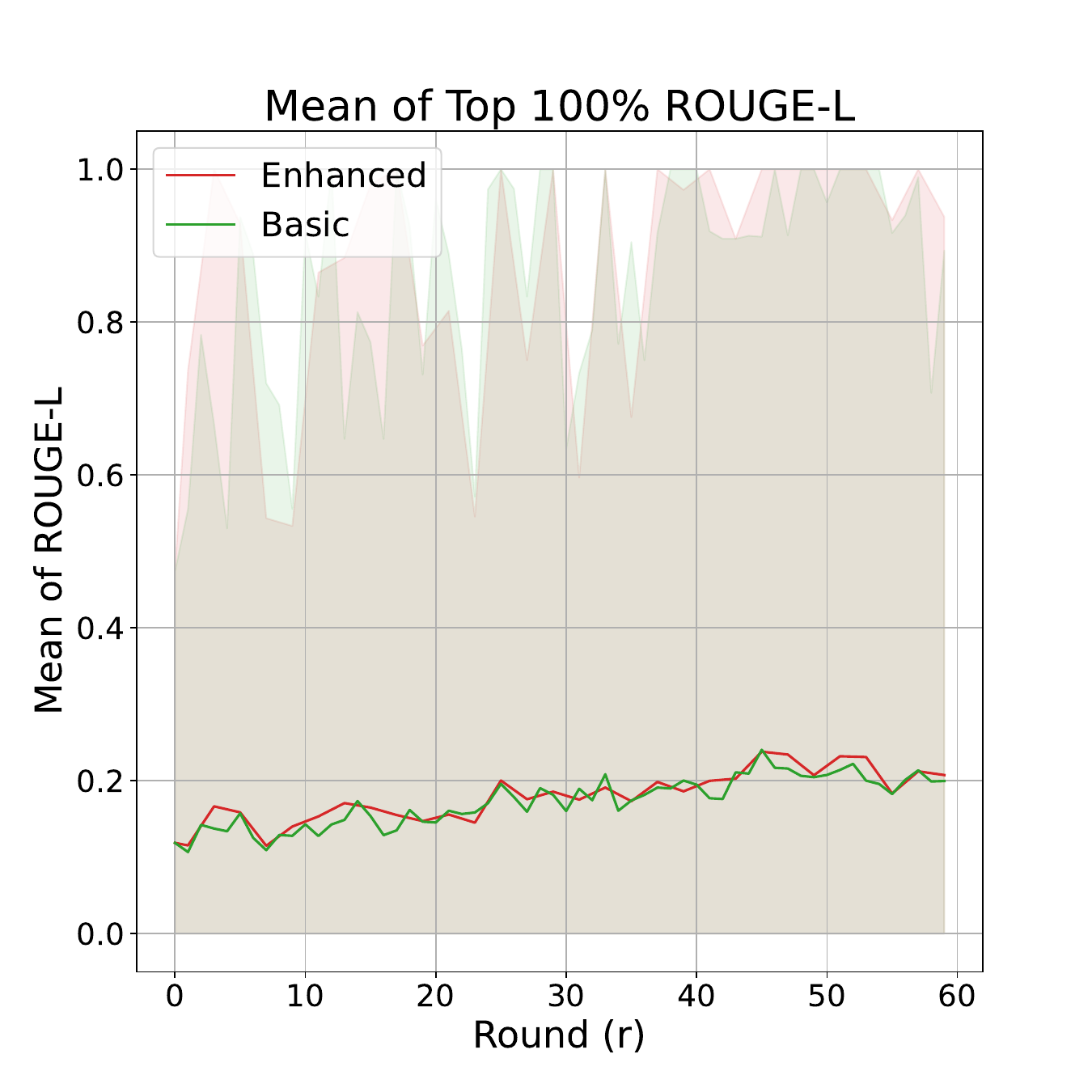}
}
\\
\subfloat[Top 10\% on LLaMA-13B.]{
    \includegraphics[width=0.23\linewidth]{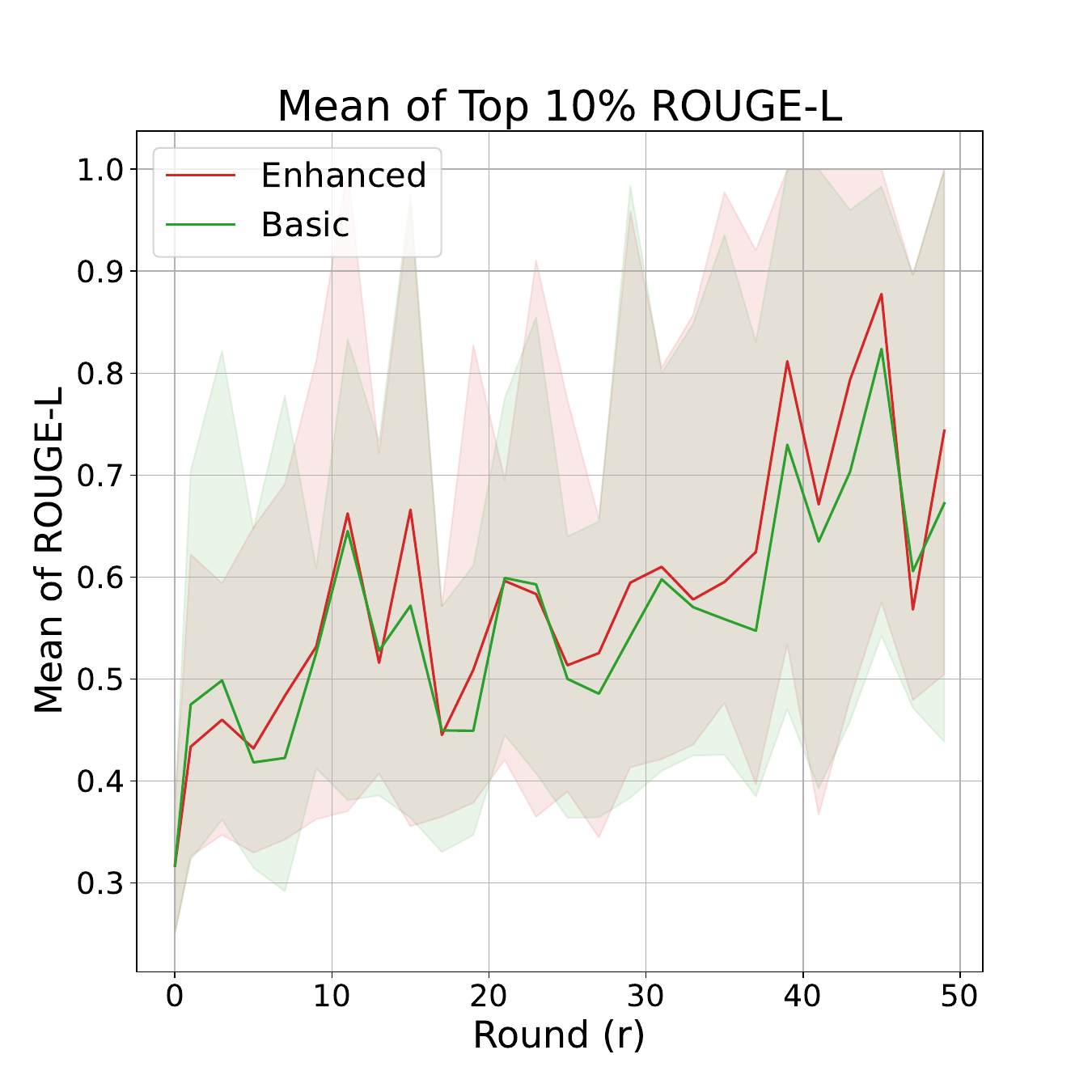}
}
\subfloat[Top 30\% on LLaMA-13B.]{
    \includegraphics[width=0.23\linewidth]{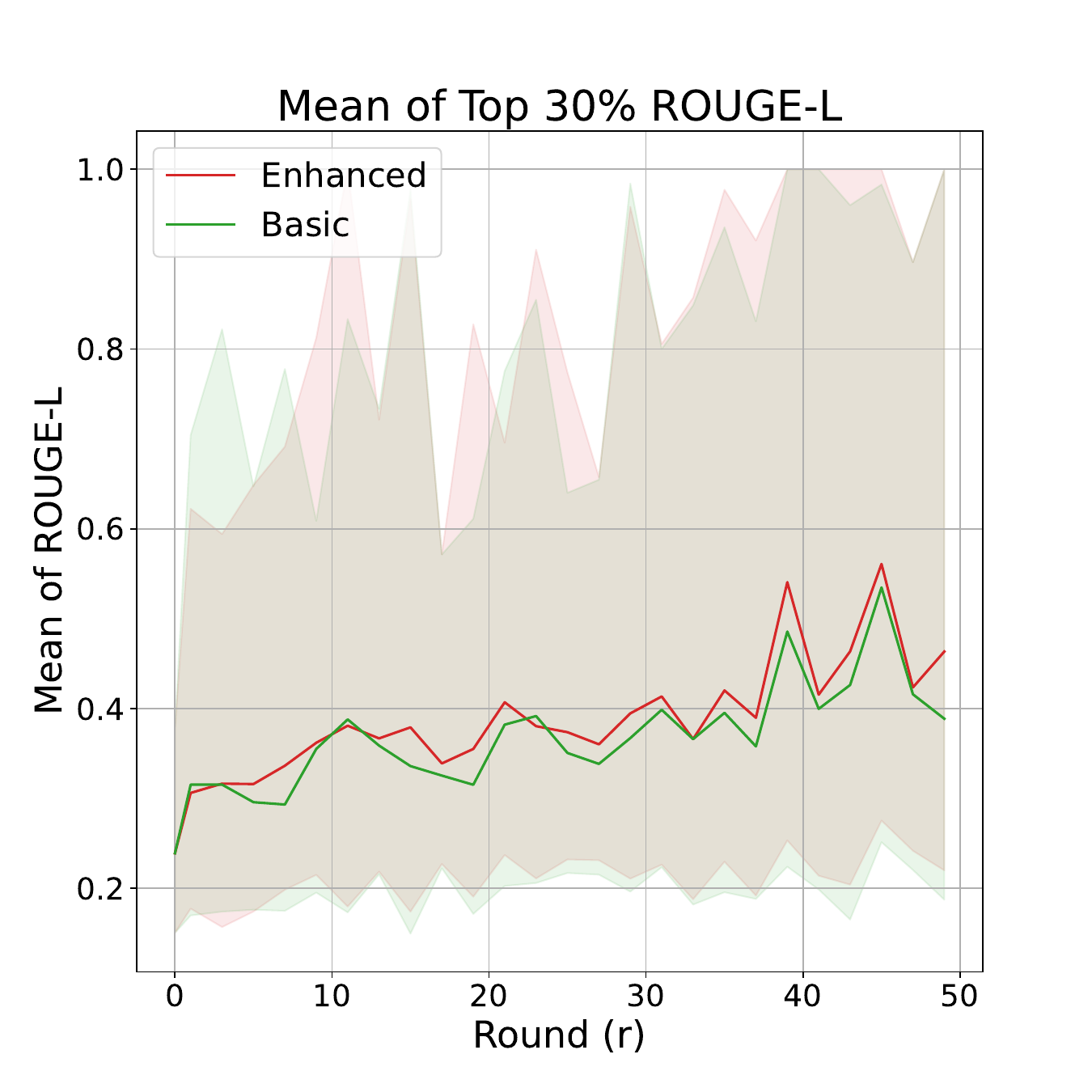}
}
\subfloat[Top 50\% on LLaMA-13B.]{
    \includegraphics[width=0.23\linewidth]{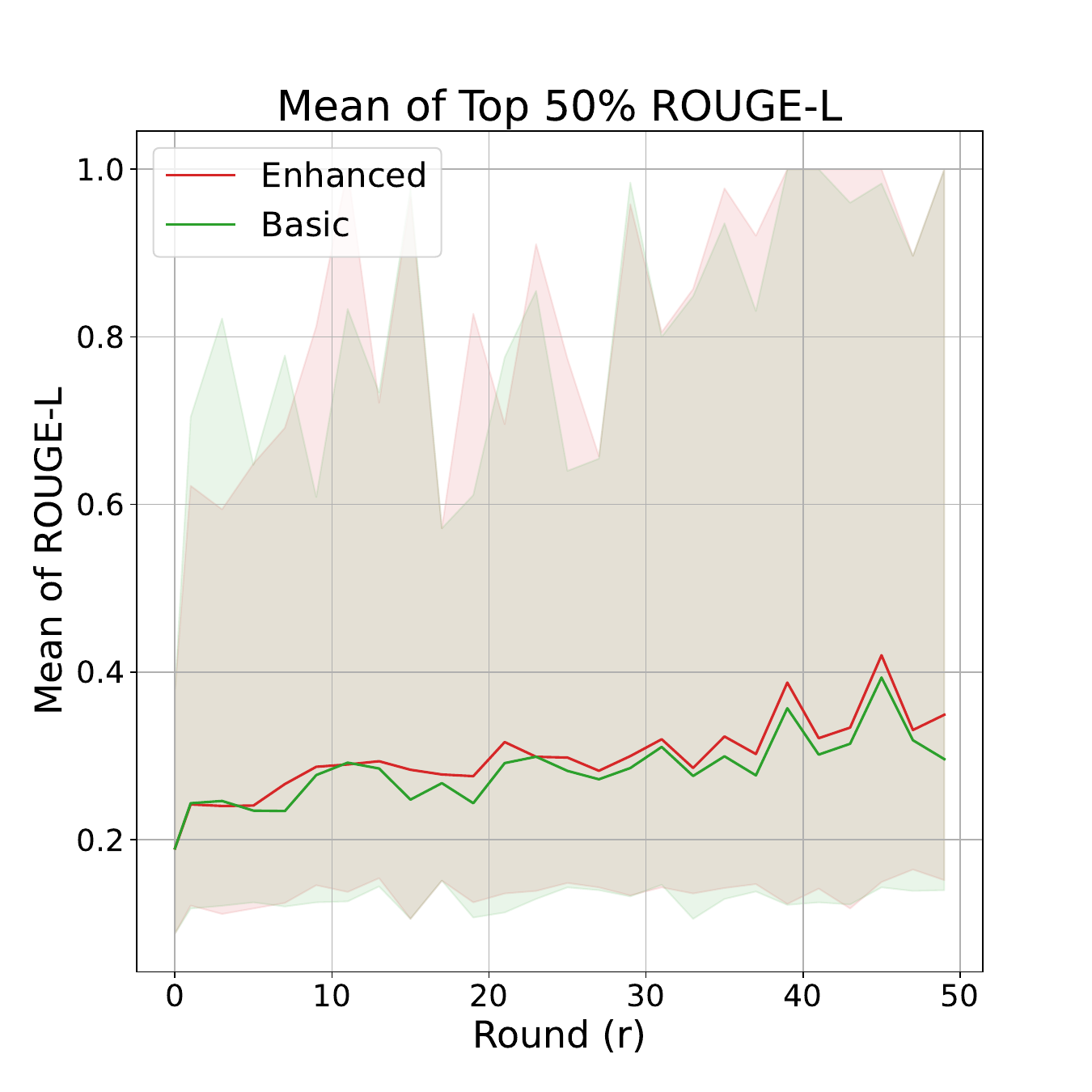}
}
\subfloat[Top 100\% on LLaMA-13B.]{
    \includegraphics[width=0.23\linewidth]{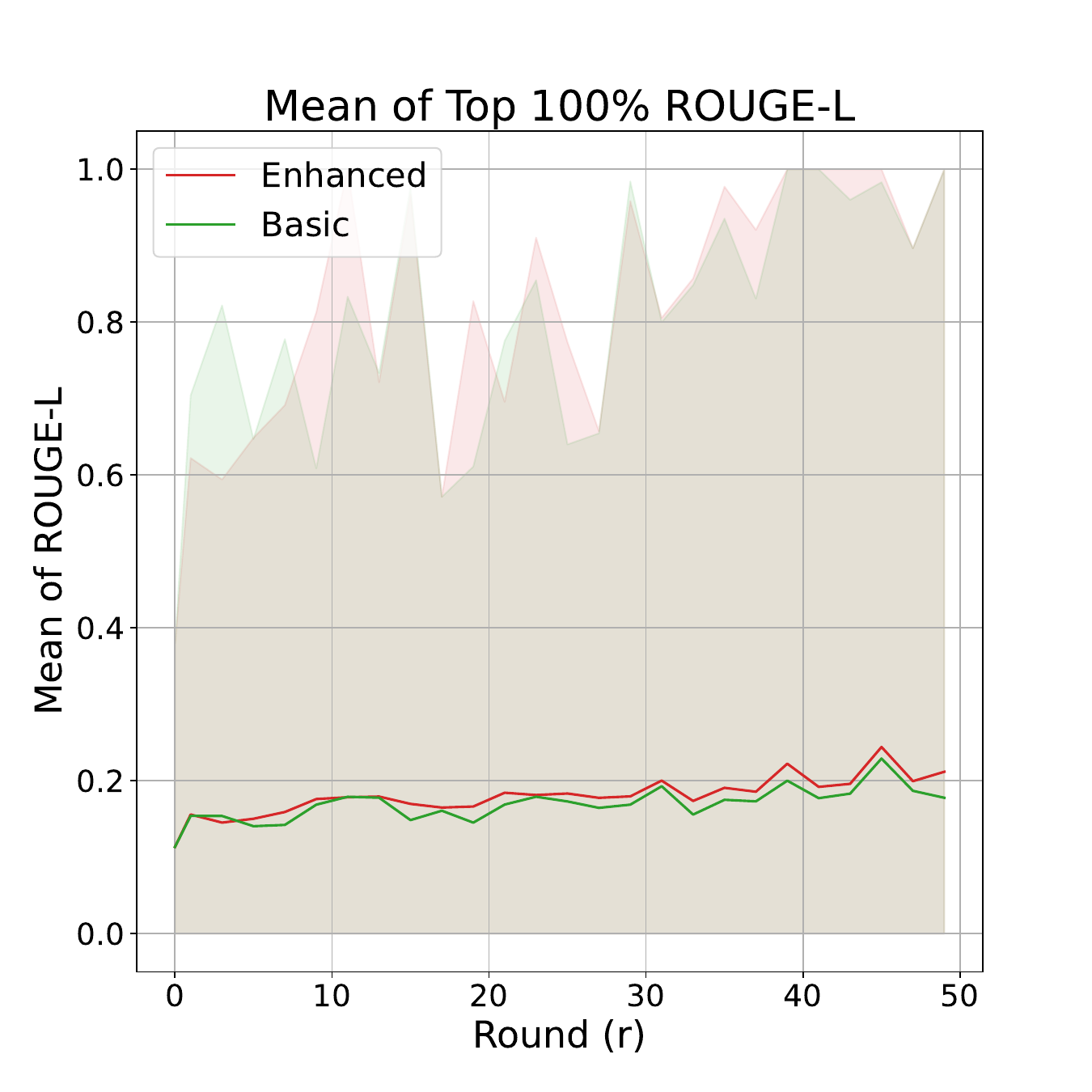}
}
\caption{Comparison of two hacking schemes' results for Partial-Input Completion. Experiments are repeated using LLaMA-8B and LLaMA-13B using Enron Email dataset.}
\label{fig:threshold4}

\end{figure*}

In this section, we provide additional results for various thresholds (e.g., top 10\%, top 30\%, top 50\%, 100\%) corresponding to Section~\ref{section:hack_result}. These are shown in Fig.~\ref{fig:threshold1}, ~\ref{fig:threshold2}, ~\ref{fig:threshold3} and ~\ref{fig:threshold4}.
From the results, we generally observe that for any given threshold, the hacked model completions have a substantially higher similarity compared to ground truths. This suggests a serious data leakage problem.

\section{Evaluation on Different Metrics}
\label{appendix:metrics}

\subsection{BERTScore}
\begin{figure*}[!htbp]
\setlength{\abovecaptionskip}{6pt}
\centering
\subfloat[Top 10\% ROUGE-L for Zero-Input Generation.]{
    \includegraphics[width=0.48\linewidth]{./figures/appendix/threshold/sim_rougeL_10.pdf}
}
\subfloat[Top 10\% BERTScore for Zero-Input Generation.]{
    \includegraphics[width=0.48\linewidth]{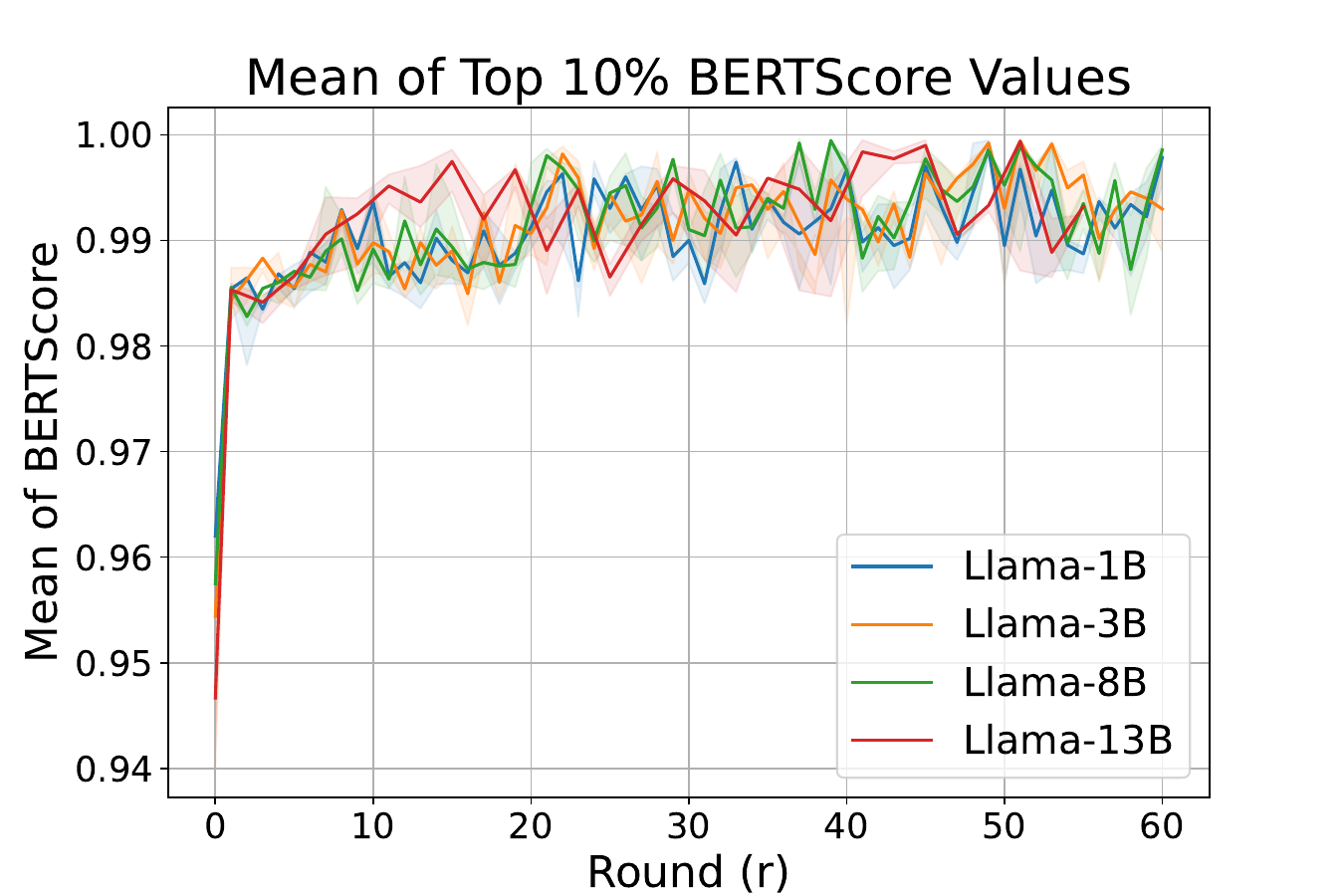}
}
\\
\subfloat[Top 10\% ROUGE-L for Partial-Input Completion.]{
    \includegraphics[width=0.48\linewidth]{./figures/appendix/threshold/com_rougeL_10.pdf}
}
\subfloat[Top 10\% BERTScore for Partial-Input Completion.]{
    \includegraphics[width=0.48\linewidth]{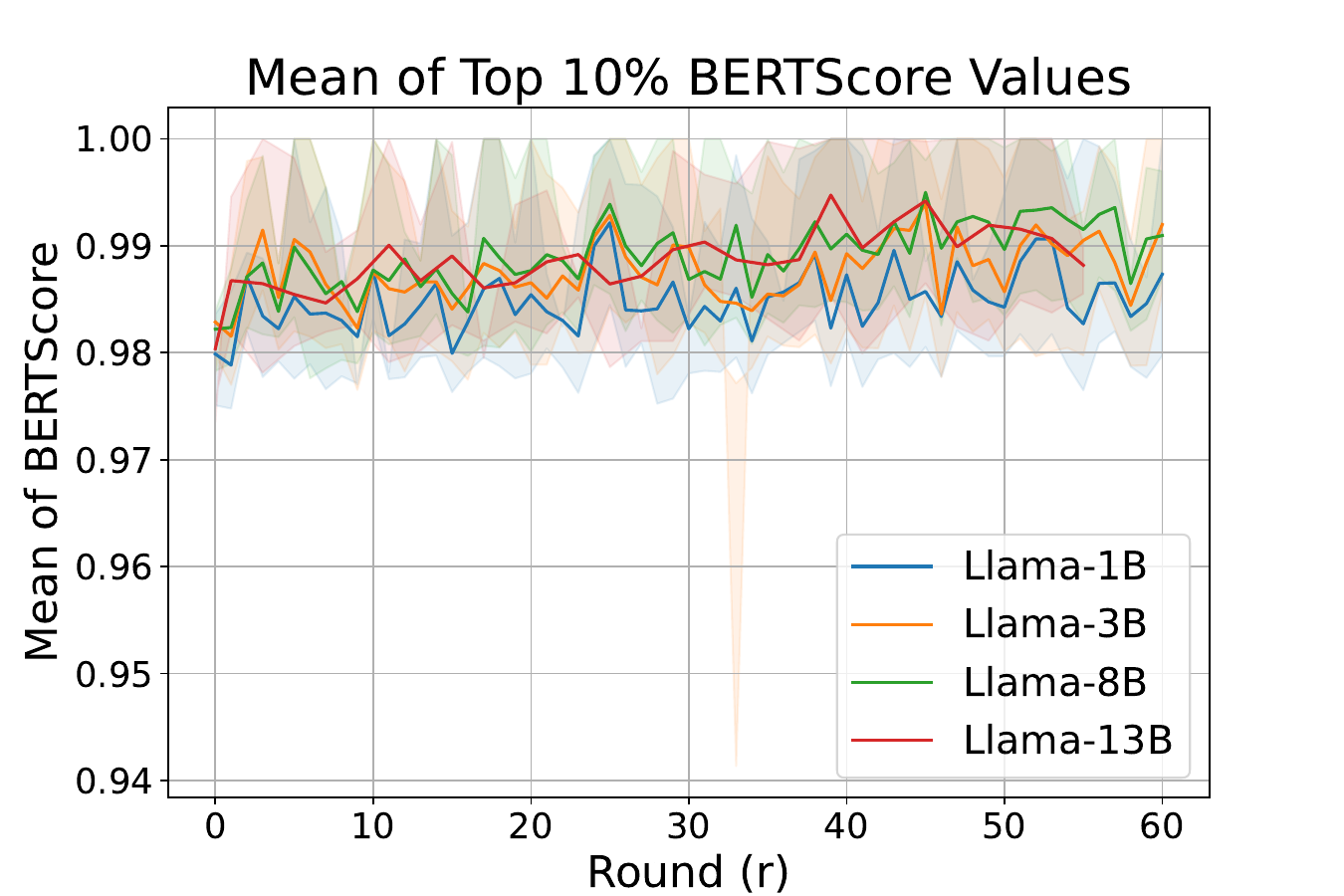}
}
\caption{Basic Hacking Scheme results of ROUGE-L and BertScore for Zero-Input/Partial-Input Generation. Experiments are repeated using LLaMA-1B, LLaMA-3B, LLaMA-8B and LLaMA-13B using Enron Email dataset.}
\label{fig:bert_basic}
\end{figure*}

\begin{figure*}[!htbp]
\setlength{\abovecaptionskip}{6pt}
\centering
\subfloat[ROUGE-L LLaMA-8B.]{
    \includegraphics[width=0.23\linewidth]{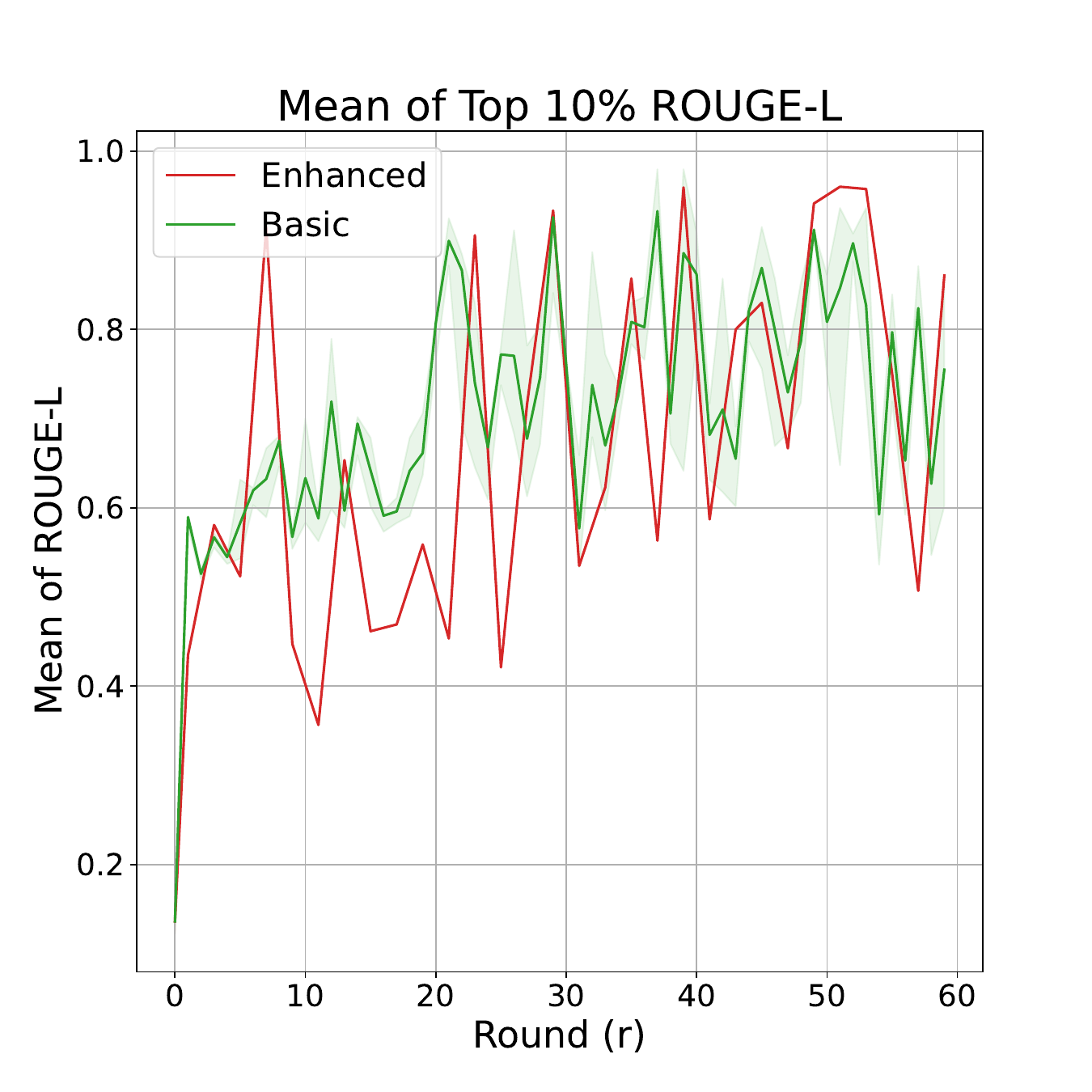}
}
\subfloat[BERTScore LLaMA-8B.]{
    \includegraphics[width=0.23\linewidth]{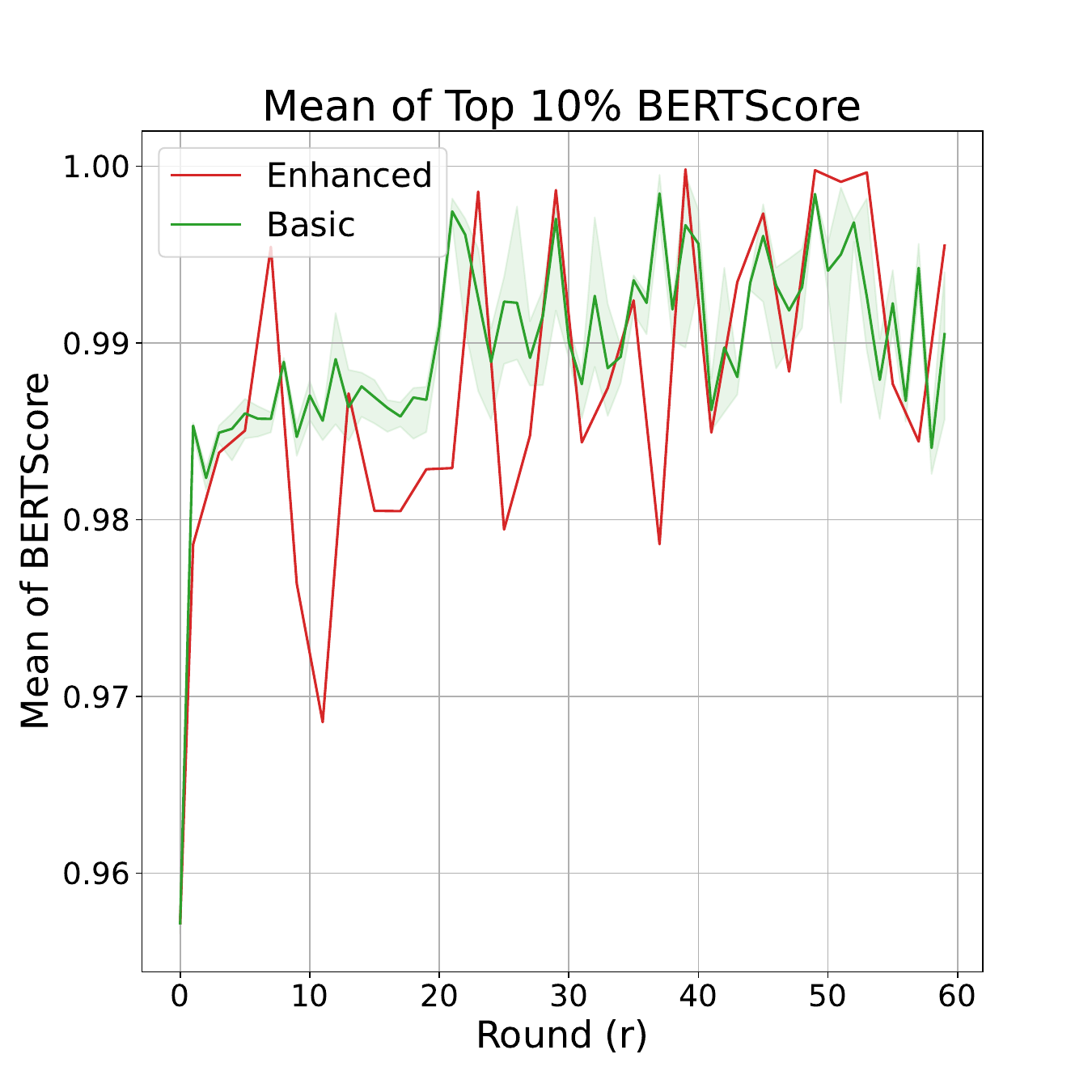}
}
\subfloat[ROUGE-L LLaMA-13B.]{
    \includegraphics[width=0.23\linewidth]{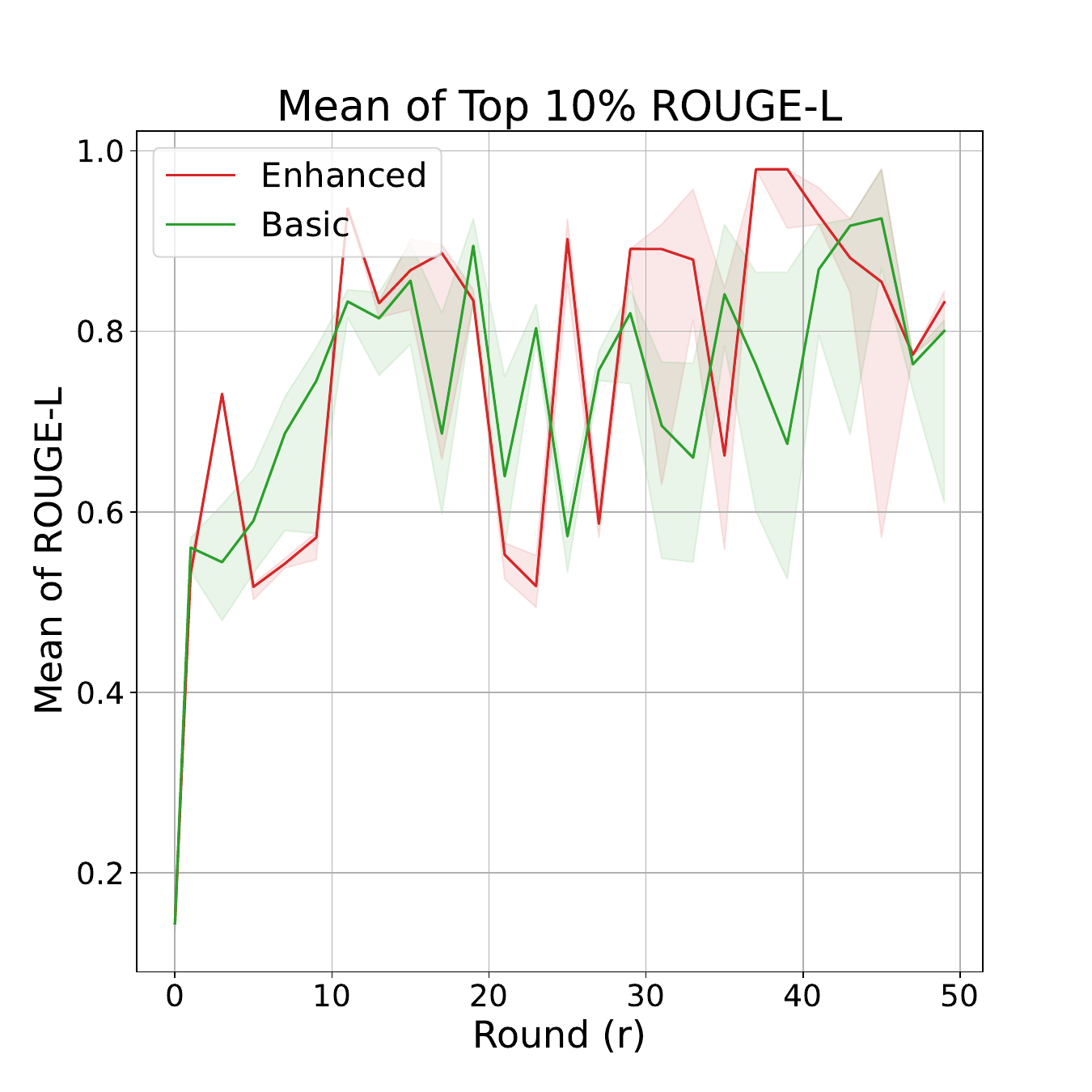}
}
\subfloat[BERTScore LLaMA-13B.]{
    \includegraphics[width=0.23\linewidth]{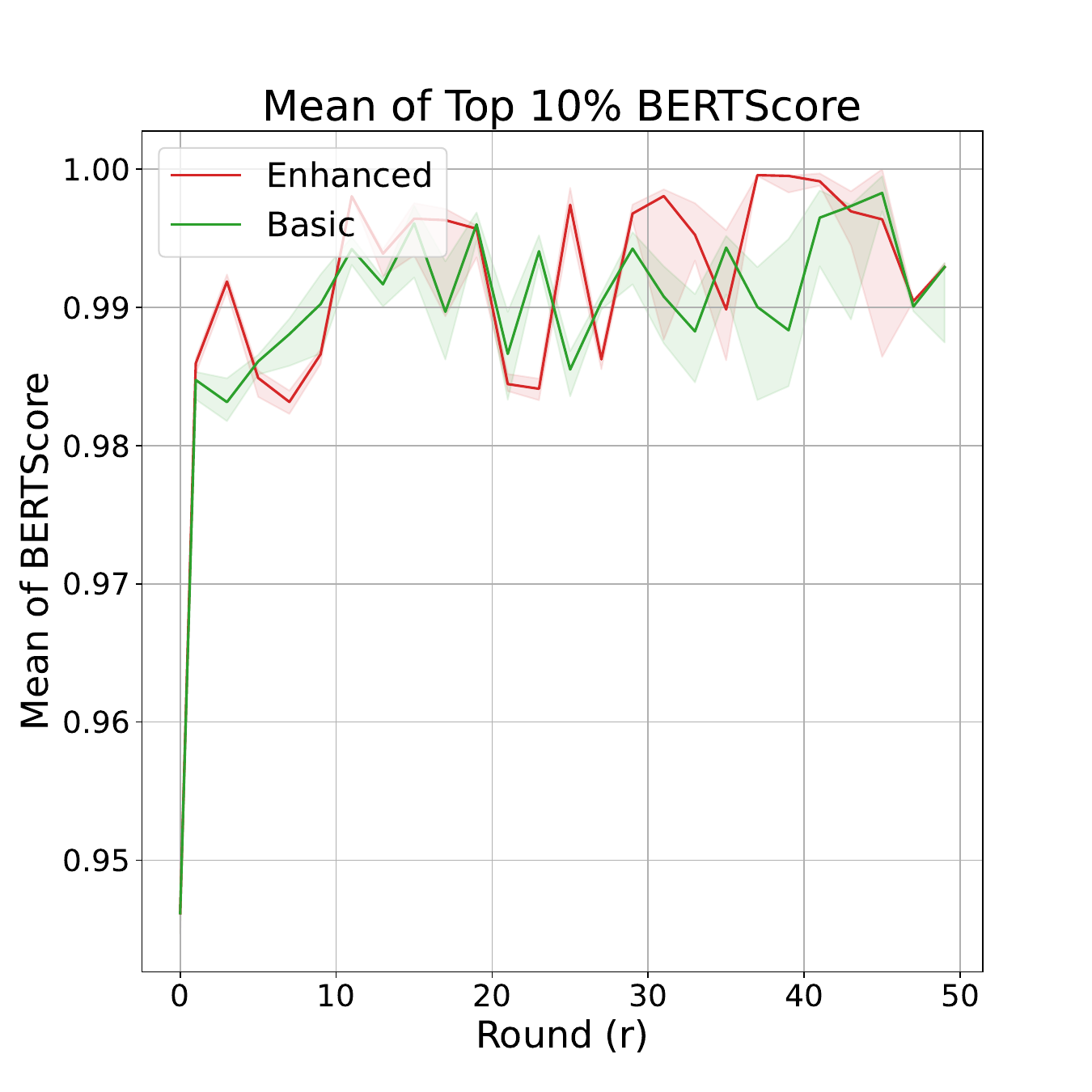}
}
\caption{Enhanced Hacking Scheme results of ROUGE and BertScore for Zero-Input Generation. Experiments are repeated using LLaMA-8B and LLaMA-13B using Enron Email dataset.}
\label{fig:bert_enhanced_sim}
\end{figure*}

\begin{figure*}[!htbp]
\setlength{\abovecaptionskip}{6pt}
\centering
\subfloat[ROUGE-L LLaMA-8B.]{
    \includegraphics[width=0.23\linewidth]{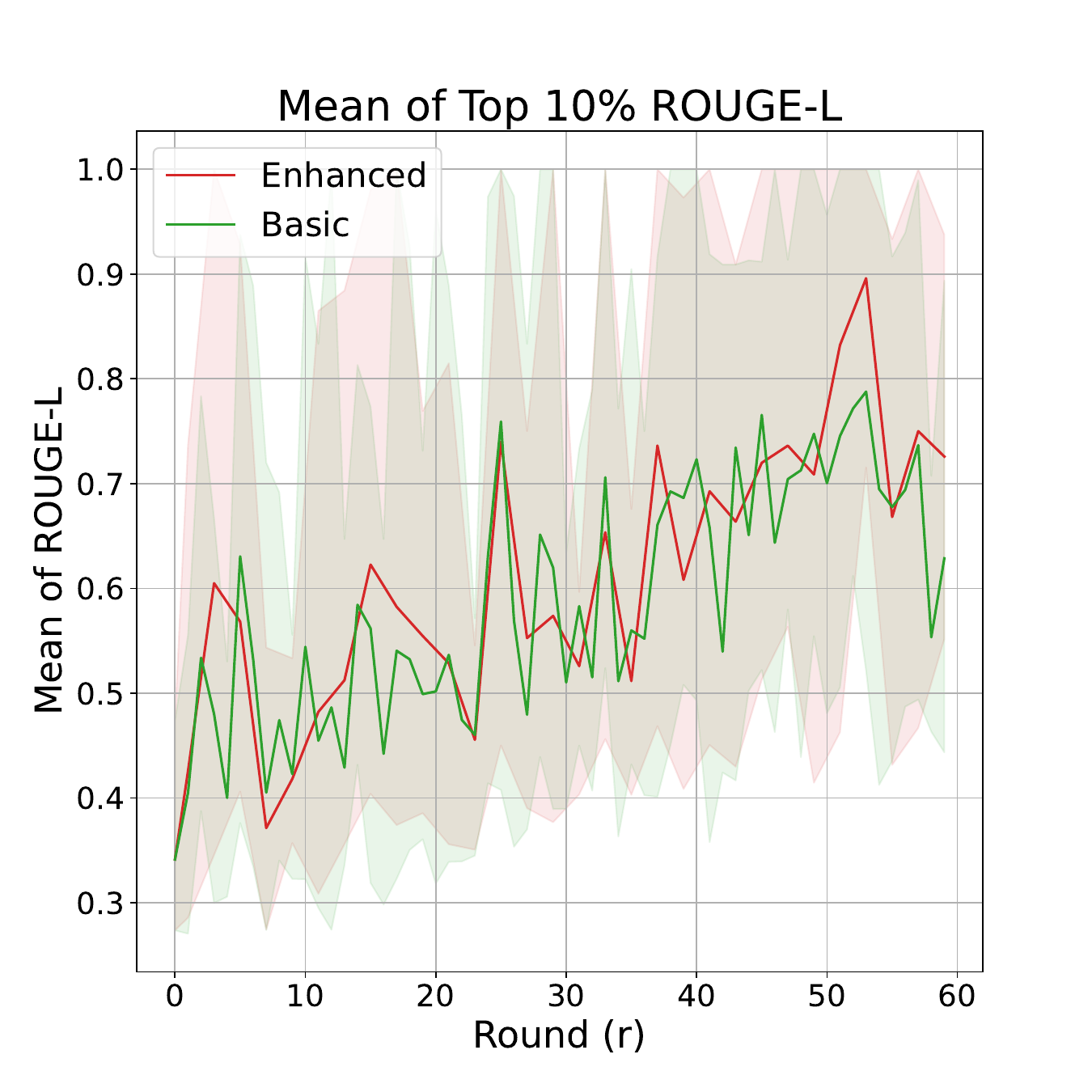}
}
\subfloat[BERTScore LLaMA-8B.]{
    \includegraphics[width=0.23\linewidth]{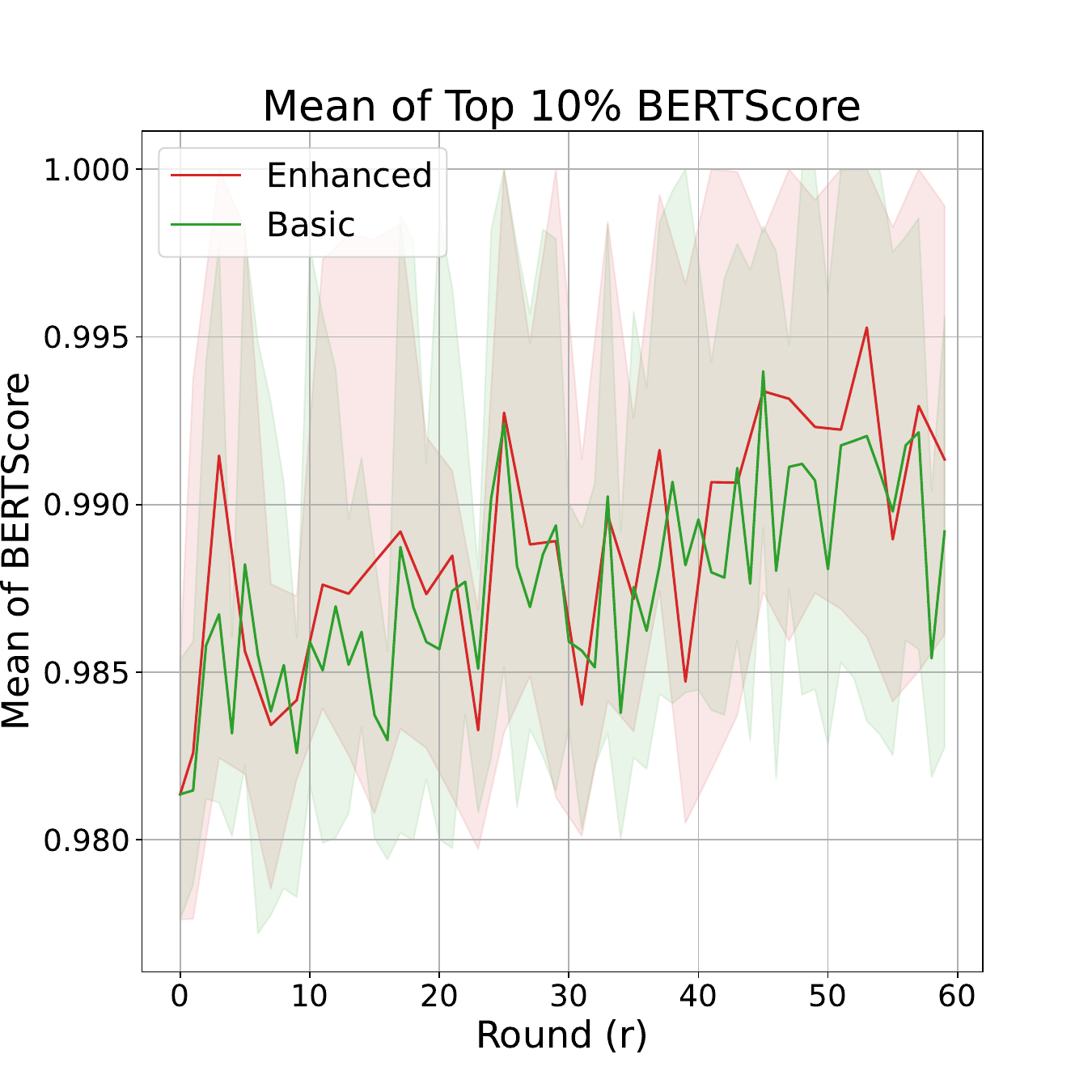}
}
\subfloat[ROUGE-L LLaMA-13B.]{
    \includegraphics[width=0.23\linewidth]{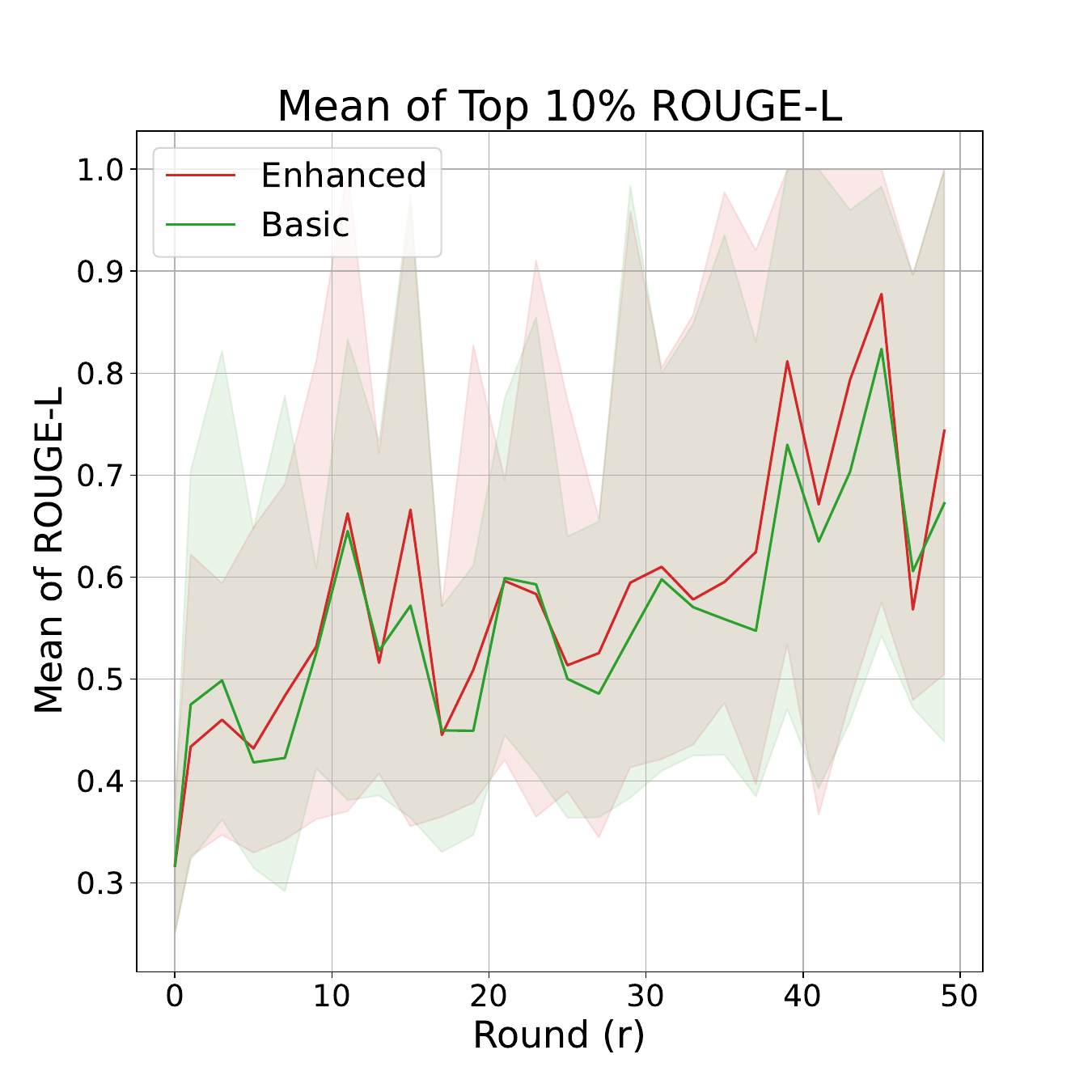}
}
\subfloat[BERTScore LLaMA-13B.]{
    \includegraphics[width=0.23\linewidth]{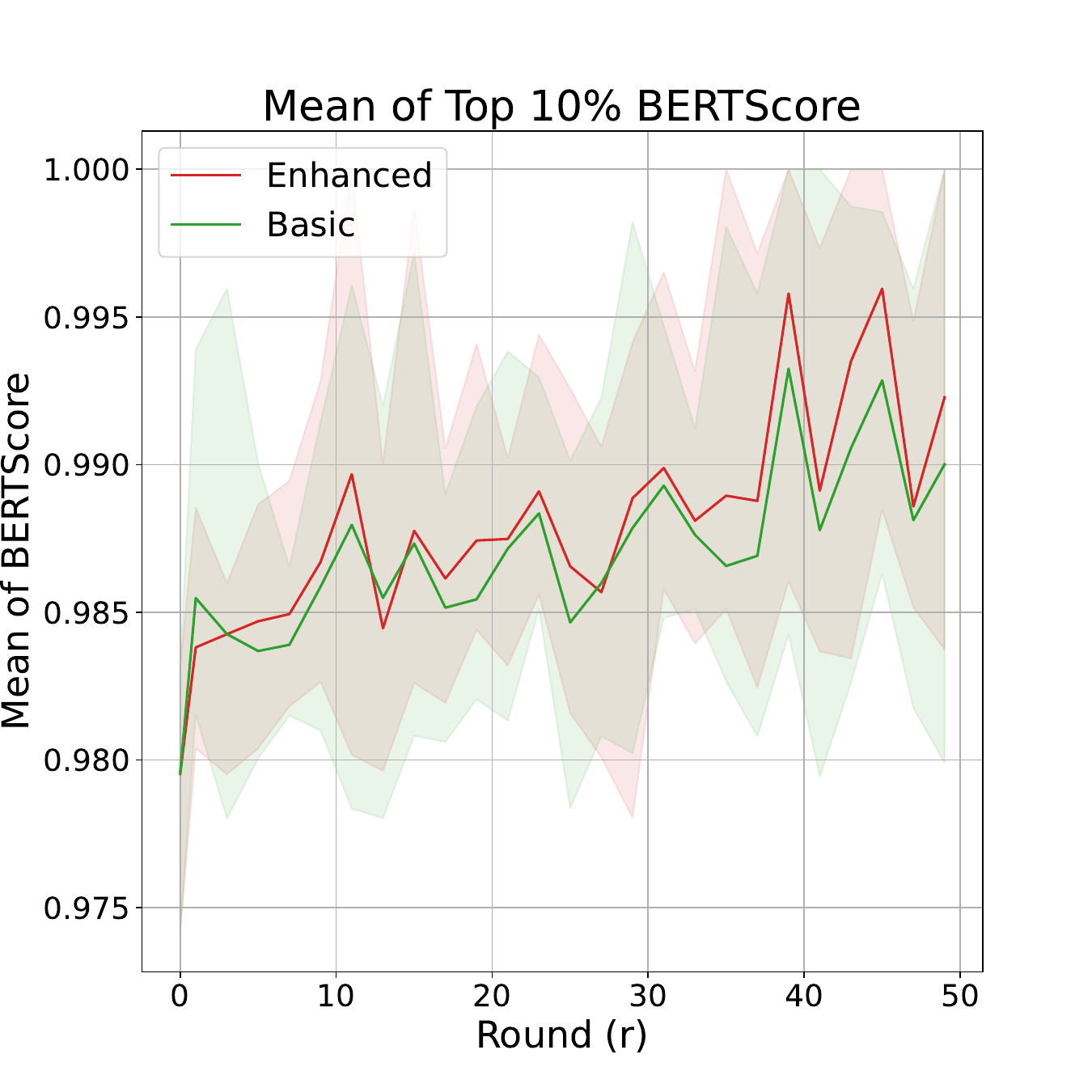}
}
\caption{Enhanced Hacking Scheme results of ROUGE and BertScore for Partial-Input Completion. Experiments are repeated using LLaMA-8B and LLaMA-13B using Enron Email dataset.}
\label{fig:bert_enhanced_com}
\end{figure*}

For evaluation, we primarily adopt ROUGE-L in the main paper, given its widespread use in recent studies for measuring text similarity. To provide a more comprehensive assessment, we additionally report other metrics such as BERTScore \citep{bertscore}, which capture similarity from different aspects. Detailed results are presented in Fig.~\ref{fig:bert_basic}, ~\ref{fig:bert_enhanced_sim} and ~\ref{fig:bert_enhanced_com}.

Here, we carefully discuss the differences between the chosen metrics and their suitability for hacking evaluation. BERTScore and ROUGE differ in their approach to measuring text similarity. BERTScore calculates \textit{semantic similarity} based on the BERT model’s output, which considers word meanings and context. In contrast, ROUGE evaluates \textit{word-level similarities} by measuring the overlap of unigrams, bigrams, and n-grams.

Given that the goal of our attack is to reconstruct sensitive data (e.g., emails, phone numbers) from the global model, word-level similarity (as measured by ROUGE) becomes more precise. We provide an example in Table~\ref{tab:simple_example} below, where we observe different results for ROUGE-L and BERTScore on generated texts. The lower ROUGE-L score for the generated text suggests less accurate preservation of specific privacy information. However, the BERTScore remains high because of similar semantics.

\begin{table*}[h]
    {\fontsize{8.7pt}{9.6pt}\selectfont
    \centering
    \setlength{\tabcolsep}{0.8mm}
    \renewcommand{\arraystretch}{0.85}
    \begin{tabular}{m{5.4cm}<{\centering}|m{5.4cm}<{\centering}|m{1.6cm}<{\centering}|m{1.6cm}<{\centering}}
        \toprule
        \textbf{Original Text} & \textbf{Generated Text} & \textbf{ROUGE-L} & \textbf{BERTScore} \\
        \midrule
        \shortstack{
            Request ID: 000000000021442\\
            Request Create Date: \textbf{3/2/01 8:27:00 AM}\\
            Requested For: mike.grigsby@enron.com\\
            Resource Name: \textbf{Market Data Bloomberg}\\
            Resource Type: \textbf{Applications}
        }
        & 
        \shortstack{
            Request ID: 000000000021442\\
            Request Create Date: \textbf{6/13/01 10:11:04 AM}\\
            Requested For: mike.grigsby@enron.com\\
            Resource Name: \textbf{\textbackslash nahoutrd\textbackslash houston\textbackslash pwr}\\
            \textbf{\textbackslash common\textbackslash Electric - [Read]}\\
            Resource Type: \textbf{Directory}
        }
        & 
        \shortstack{0.667} & \shortstack{0.989} \\
        \bottomrule
    \end{tabular}
    }
    \caption{An example of generated text and evaluation scores.}
    \label{tab:simple_example}
\end{table*}

\begin{table*}[h]
    \centering
    \begin{tabular}{c|c|c}
        \toprule
         & \textbf{Total Extraction Instances} & \textbf{Exactly Matched Instances} \\
        \midrule
        \textbf{Zero-Input Generation} & 30 & 4 \\
        \textbf{Partial-Input Completion} & 100 & 10 \\
        \bottomrule
    \end{tabular}
    \caption{Statistics of extracted and exactly matched instances for two hacking tasks.}
    \label{tab:sample_analysis}
\end{table*}

\subsection{ROUGE-1 \& ROUGE-2}
\begin{figure*}[!htbp]
\setlength{\abovecaptionskip}{6pt}
\centering
\subfloat[Top 10\% ROUGE-1 for Zero-Input Generation.]{
    \includegraphics[width=0.48\linewidth]{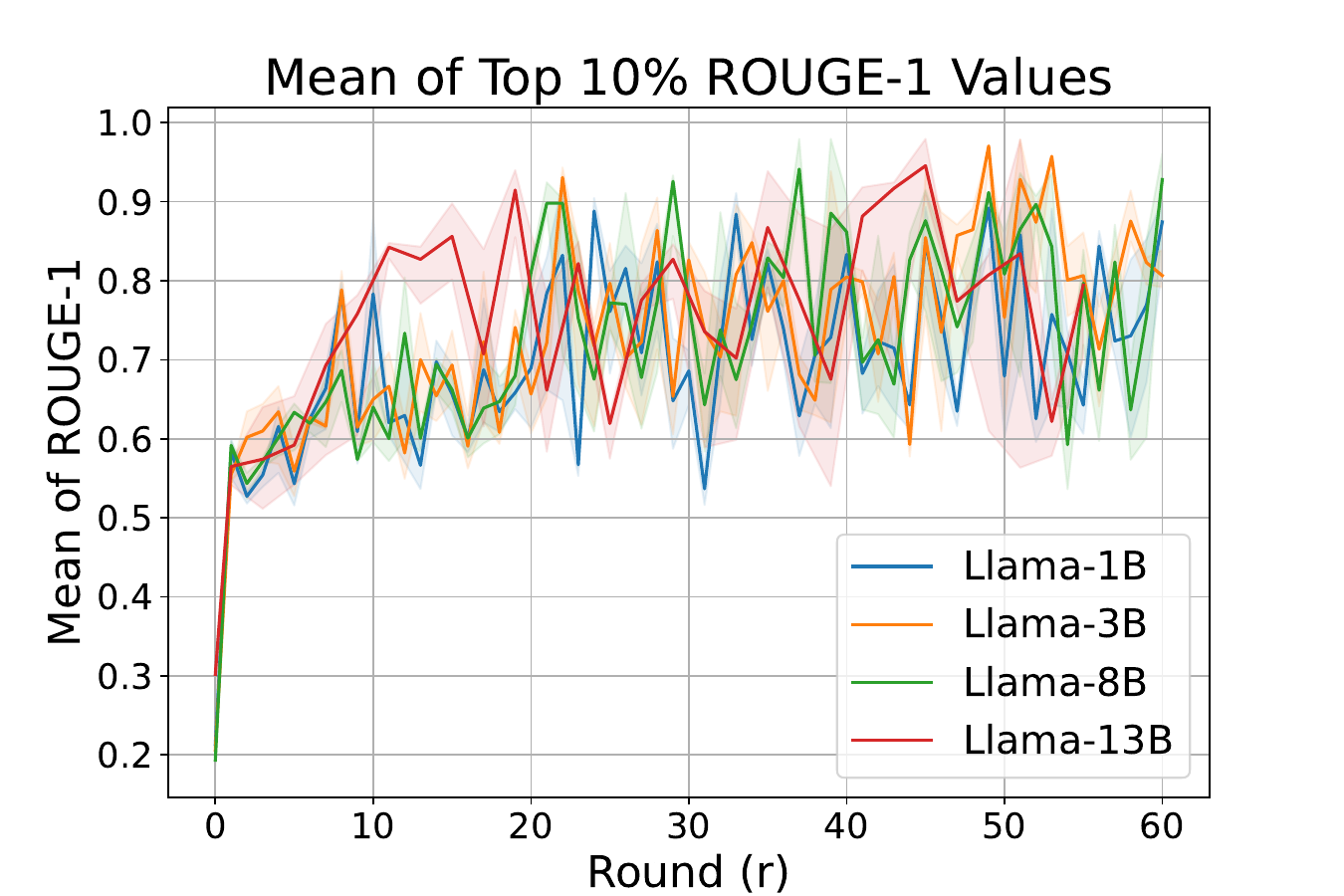}
}
\subfloat[Top 10\% ROUGE-2 for Zero-Input Generation.]{
    \includegraphics[width=0.48\linewidth]{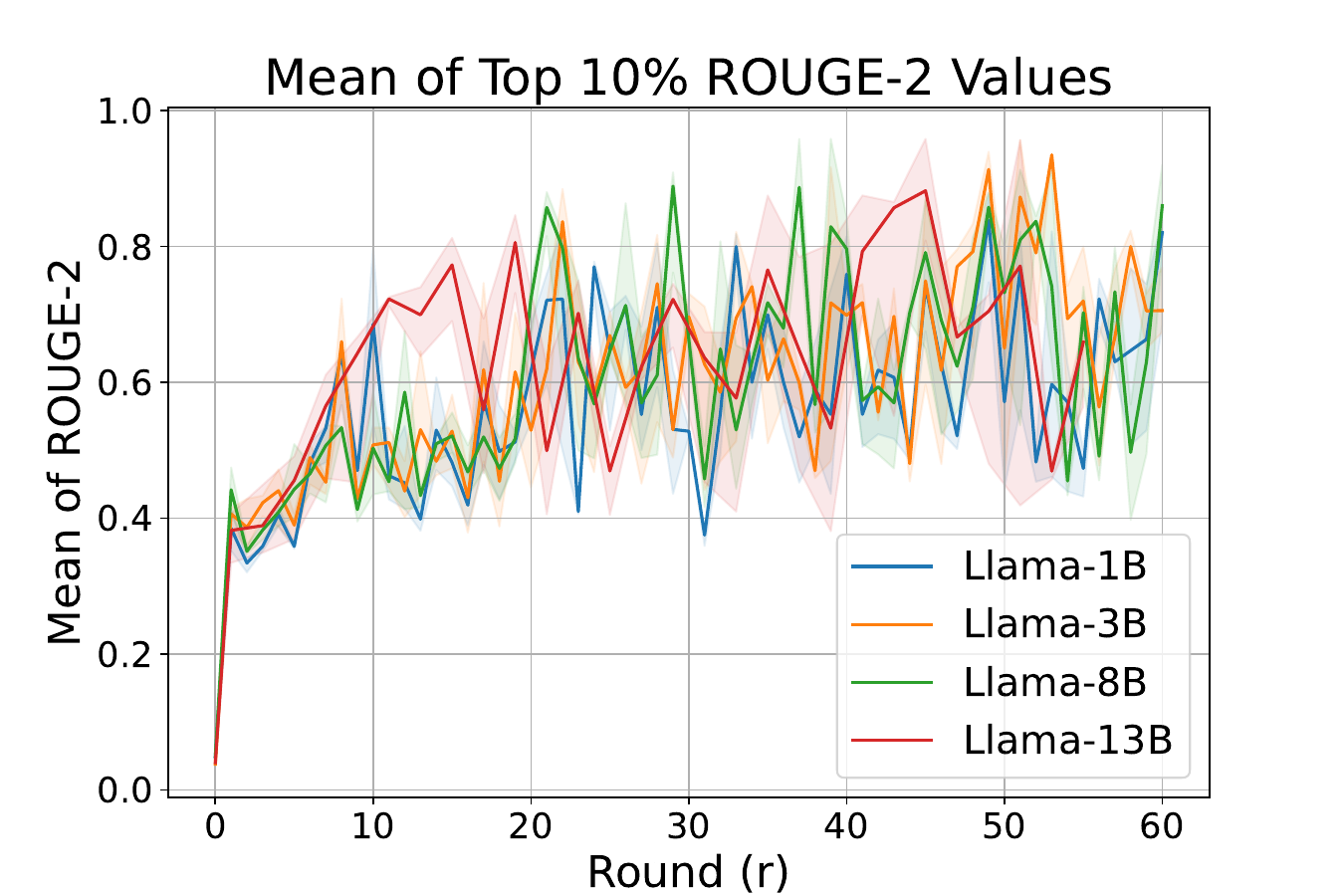}
}
\\
\subfloat[Top 10\% ROUGE-1 for Partial-Input Completion.]{
    \includegraphics[width=0.48\linewidth]{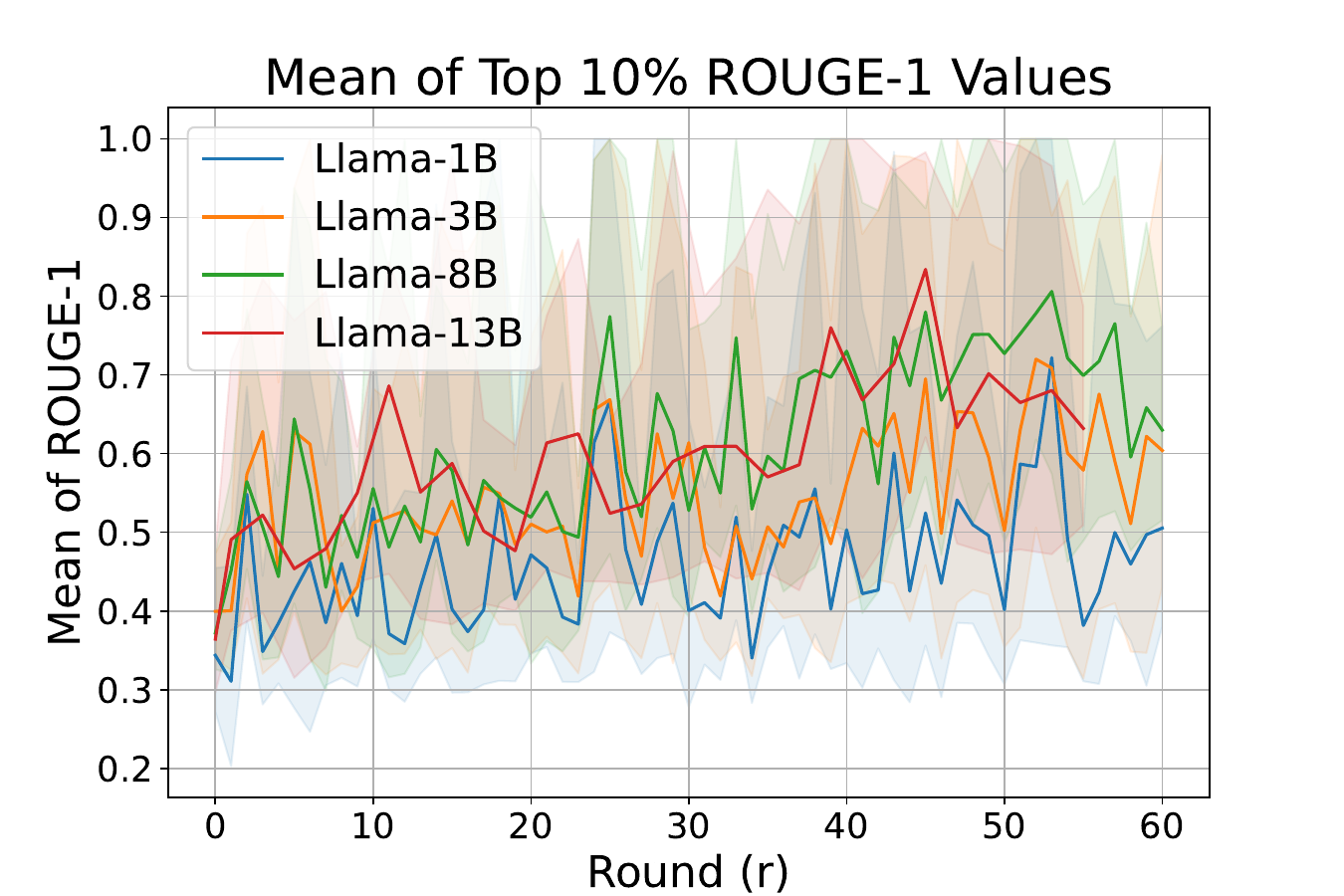}
}
\subfloat[Top 10\% ROUGE-2 for Partial-Input Completion.]{
    \includegraphics[width=0.48\linewidth]{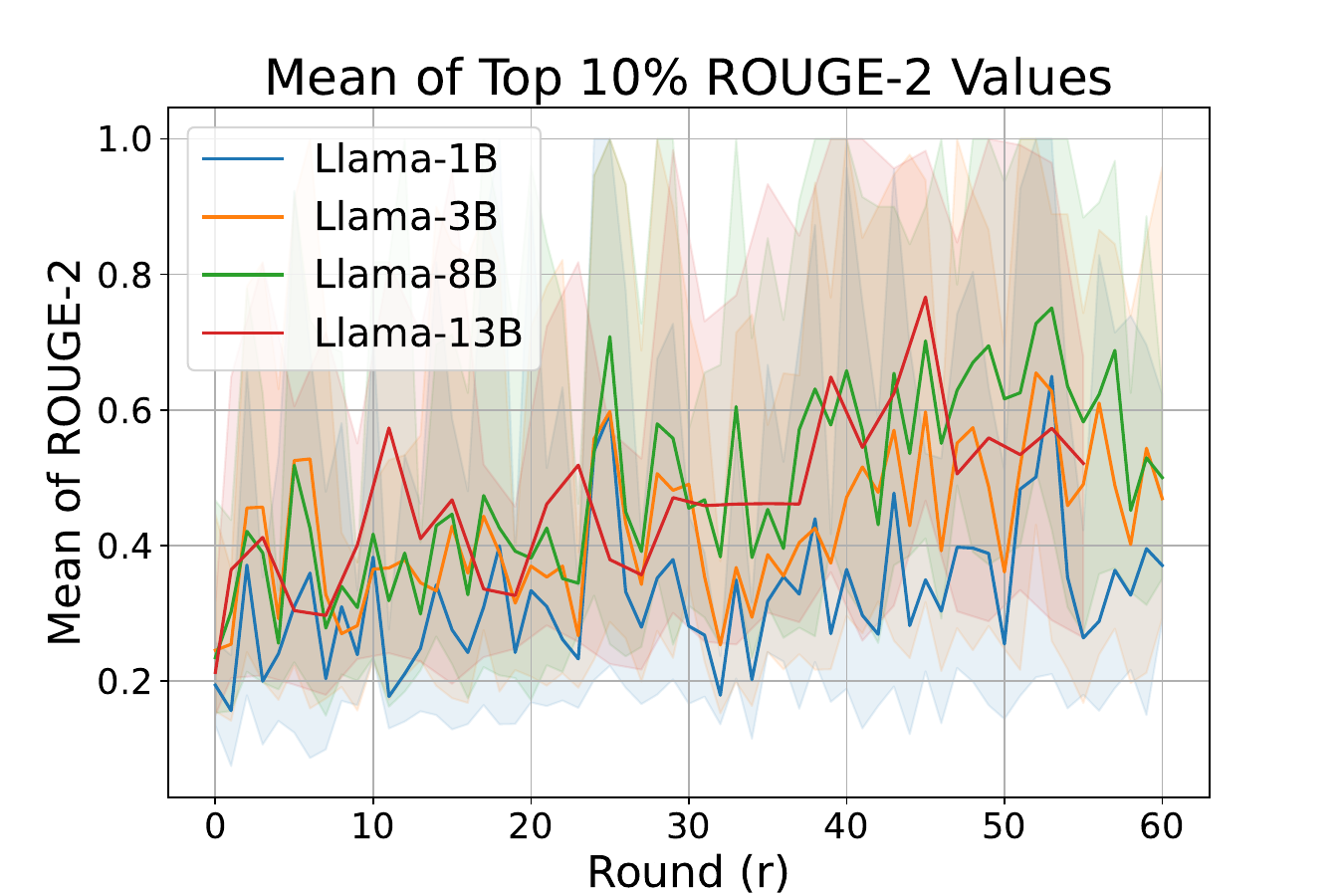}
}
\caption{Basic Hacking Scheme results of ROUGE-1 and ROUGE-2 for two hacking tasks. Experiments are repeated using LLaMA-1B and LLaMA-3B, LLaMA-8B and LLaMA-13B using Enron Email dataset.}
\label{fig:r12_basic}
\vspace{-30pt}
\end{figure*}

\begin{figure*}[!htbp]
\setlength{\abovecaptionskip}{6pt}
\centering
\subfloat[ROUGE-1 on LLaMA-8B.]{
    \includegraphics[width=0.23\linewidth]{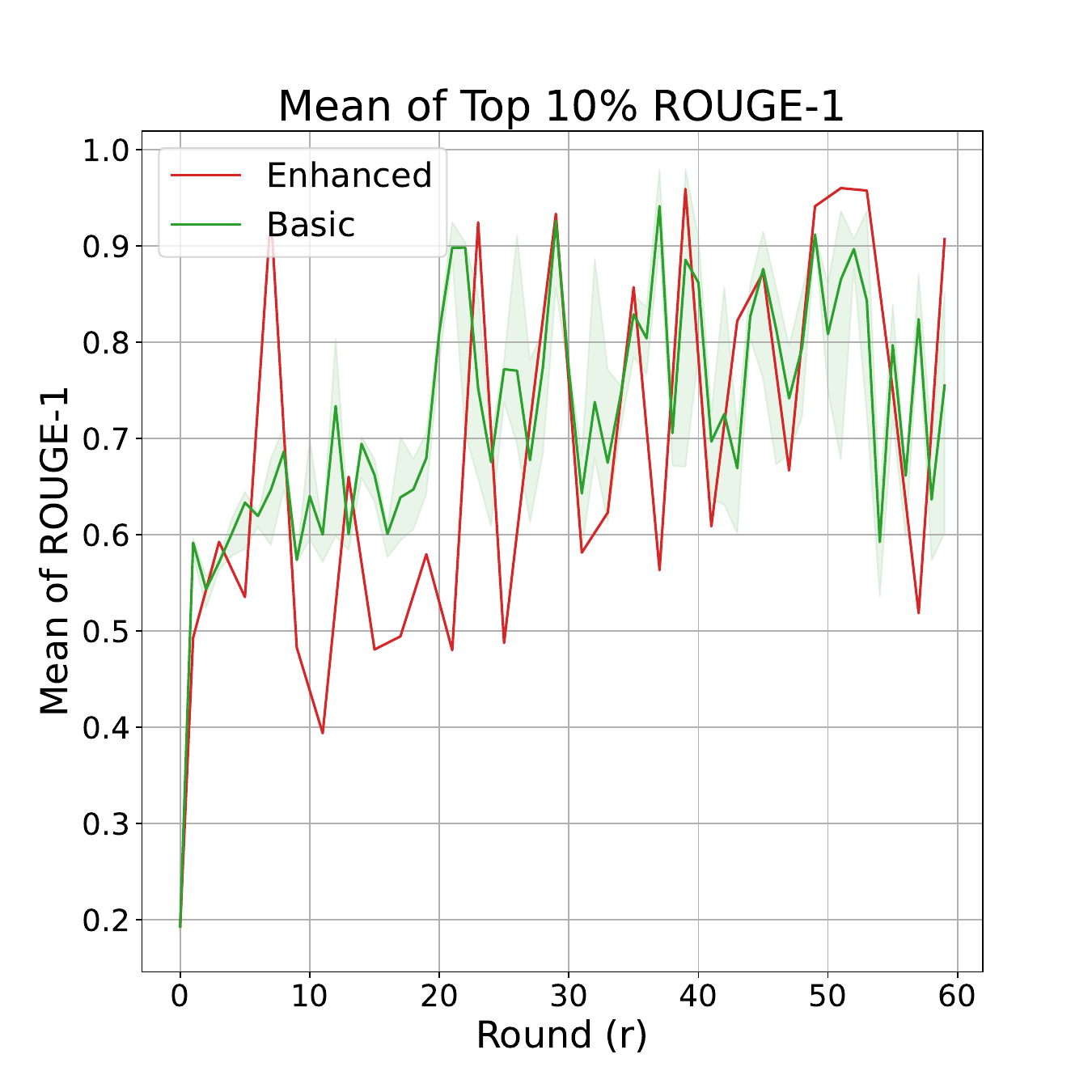}
}
\subfloat[ROUGE-2 on LLaMA-8B.]{
    \includegraphics[width=0.23\linewidth]{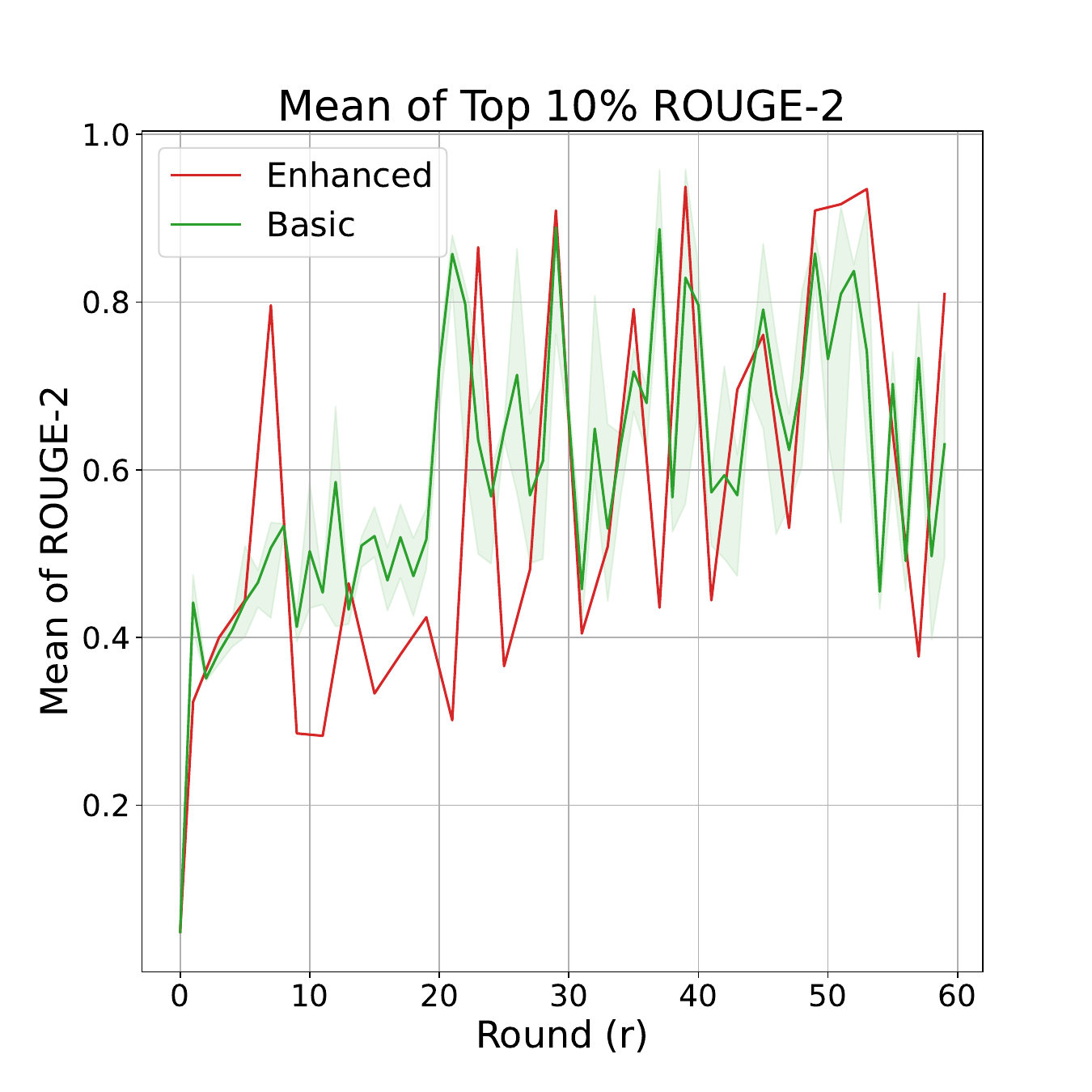}
}
\subfloat[ROUGE-1 on LLaMA-13B.]{
    \includegraphics[width=0.23\linewidth]{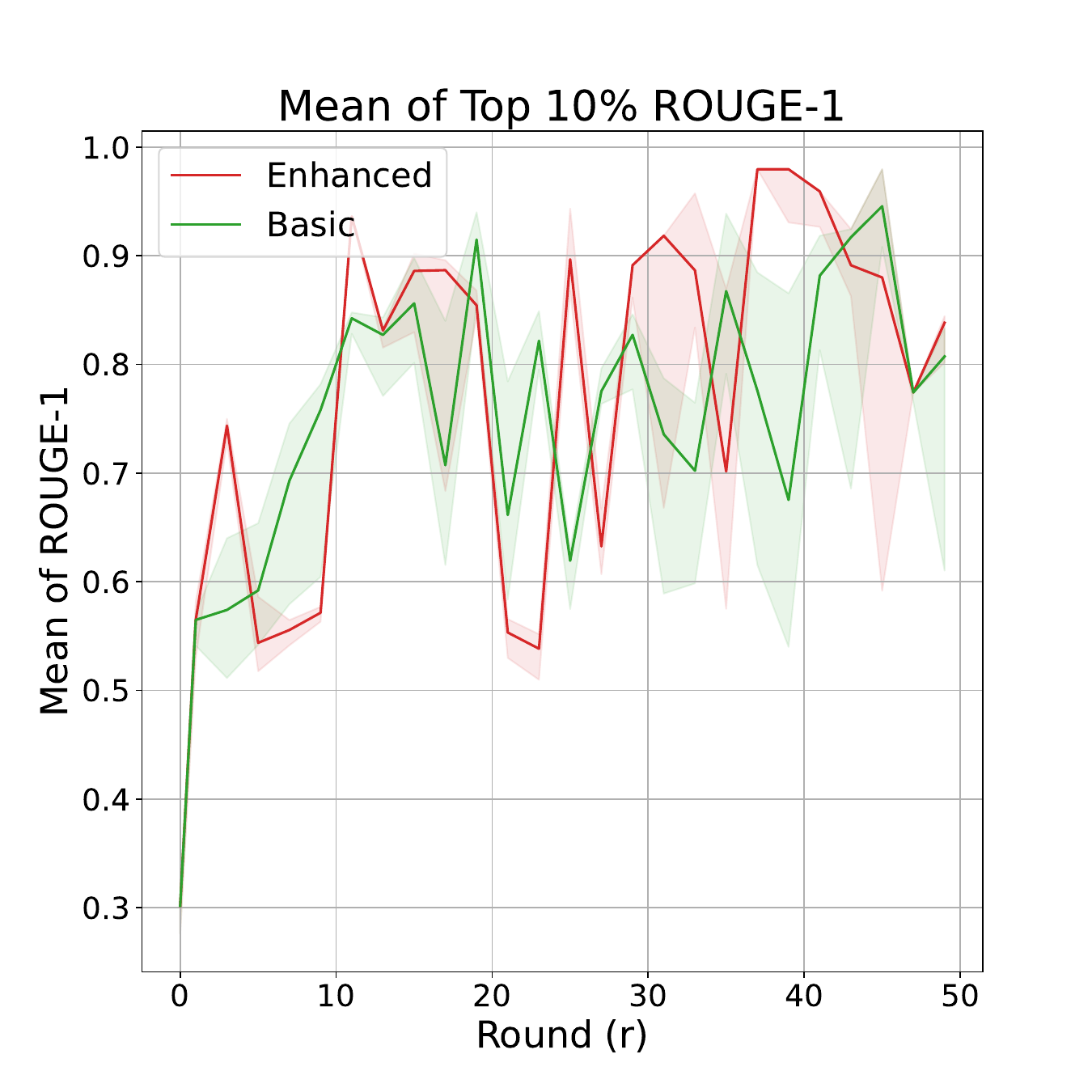}
}
\subfloat[ROUGE-2 on LLaMA-13B.]{
    \includegraphics[width=0.23\linewidth]{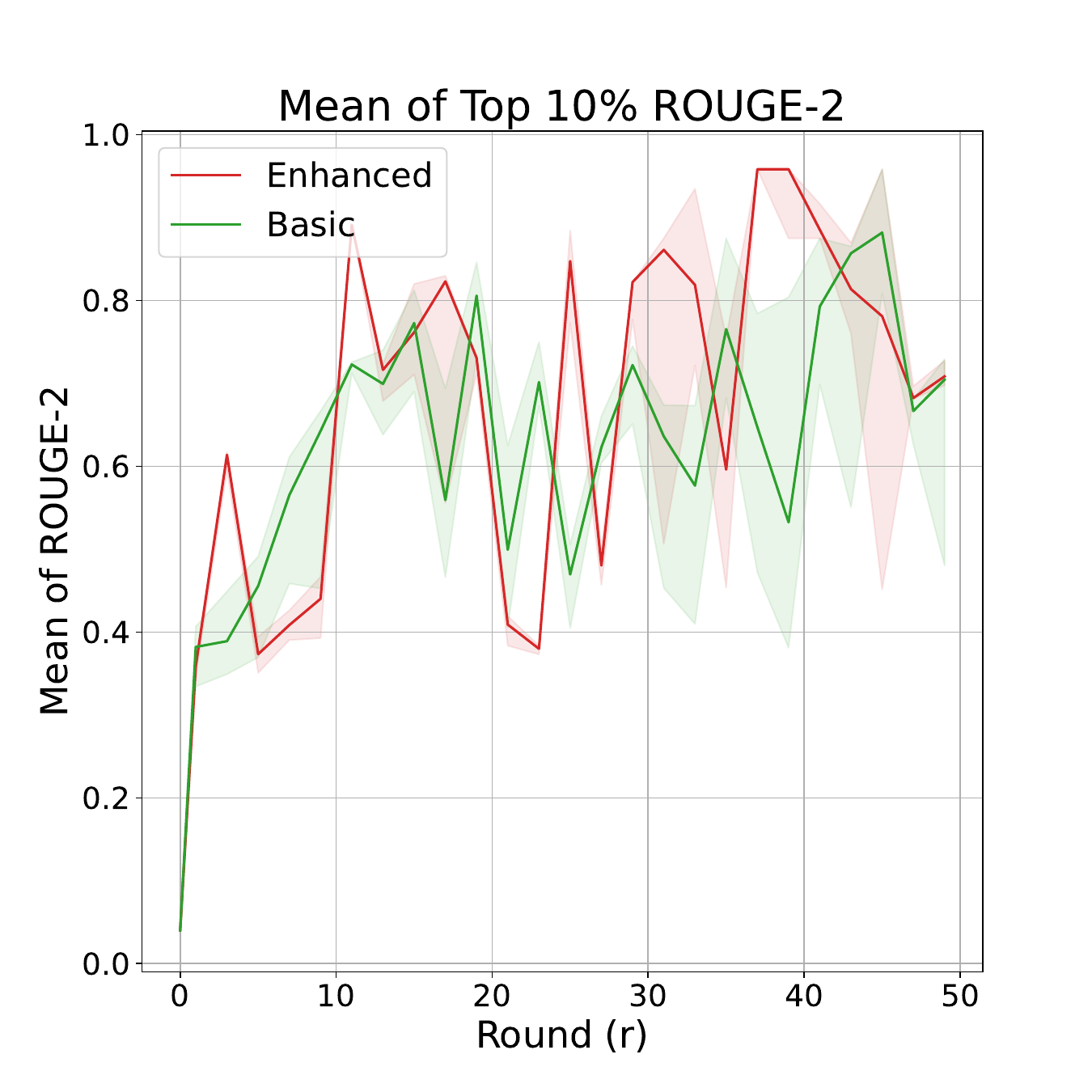}
}
\caption{Enhanced Hacking Scheme results of ROUGE-1 and ROUGE-2 for Zero-Input Generation. Experiments are repeated using LLaMA-1B and LLaMA-3B, LLaMA-8B and LLaMA-13B using Enron Email dataset.}
\label{fig:r12_enhanced_sim}
\vspace{-30pt}
\end{figure*}

\begin{figure*}[!htbp]
\setlength{\abovecaptionskip}{6pt}
\centering
\subfloat[ROUGE-1 on LLaMA-8B.]{
    \includegraphics[width=0.23\linewidth]{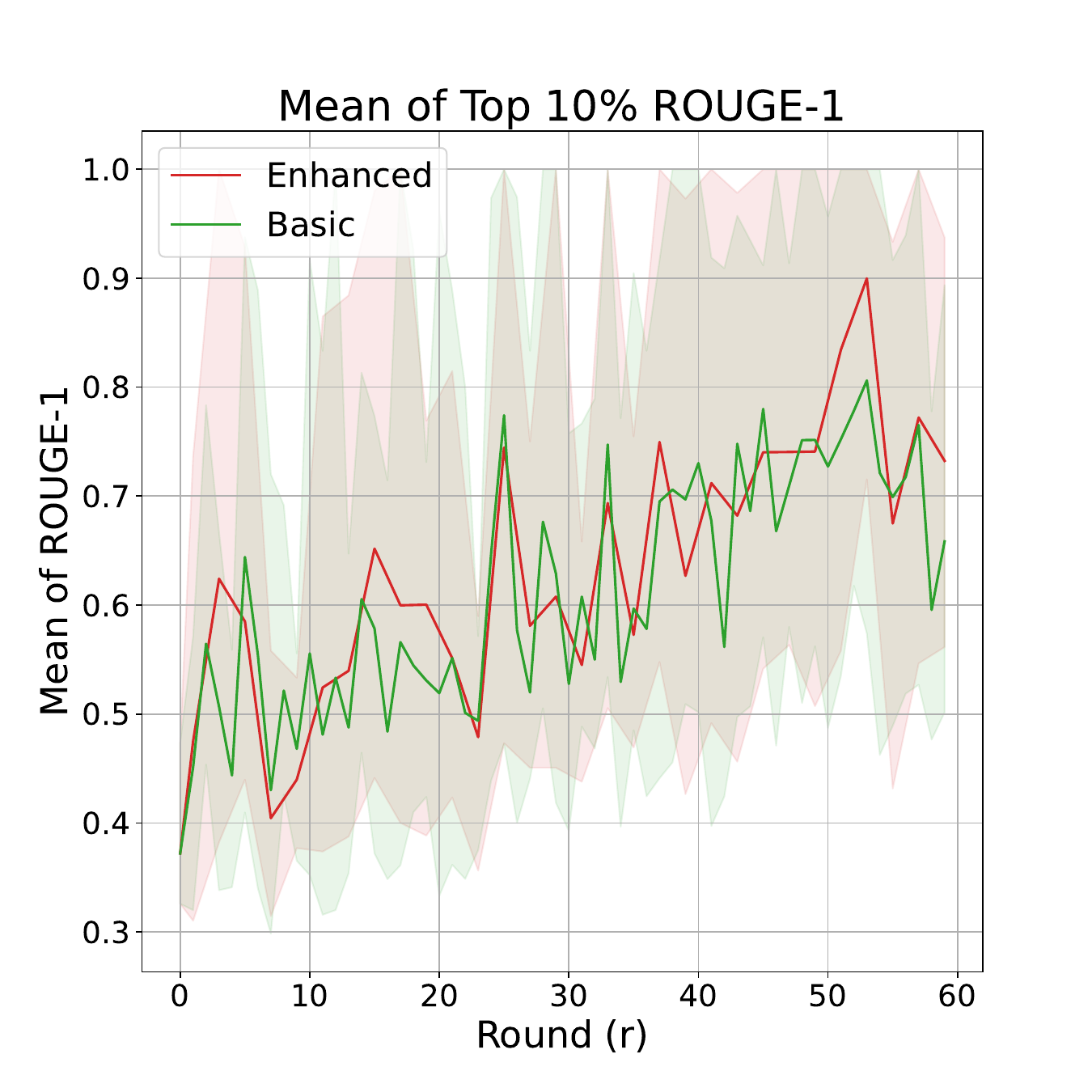}
}
\subfloat[ROUGE-2 on LLaMA-8B.]{
    \includegraphics[width=0.23\linewidth]{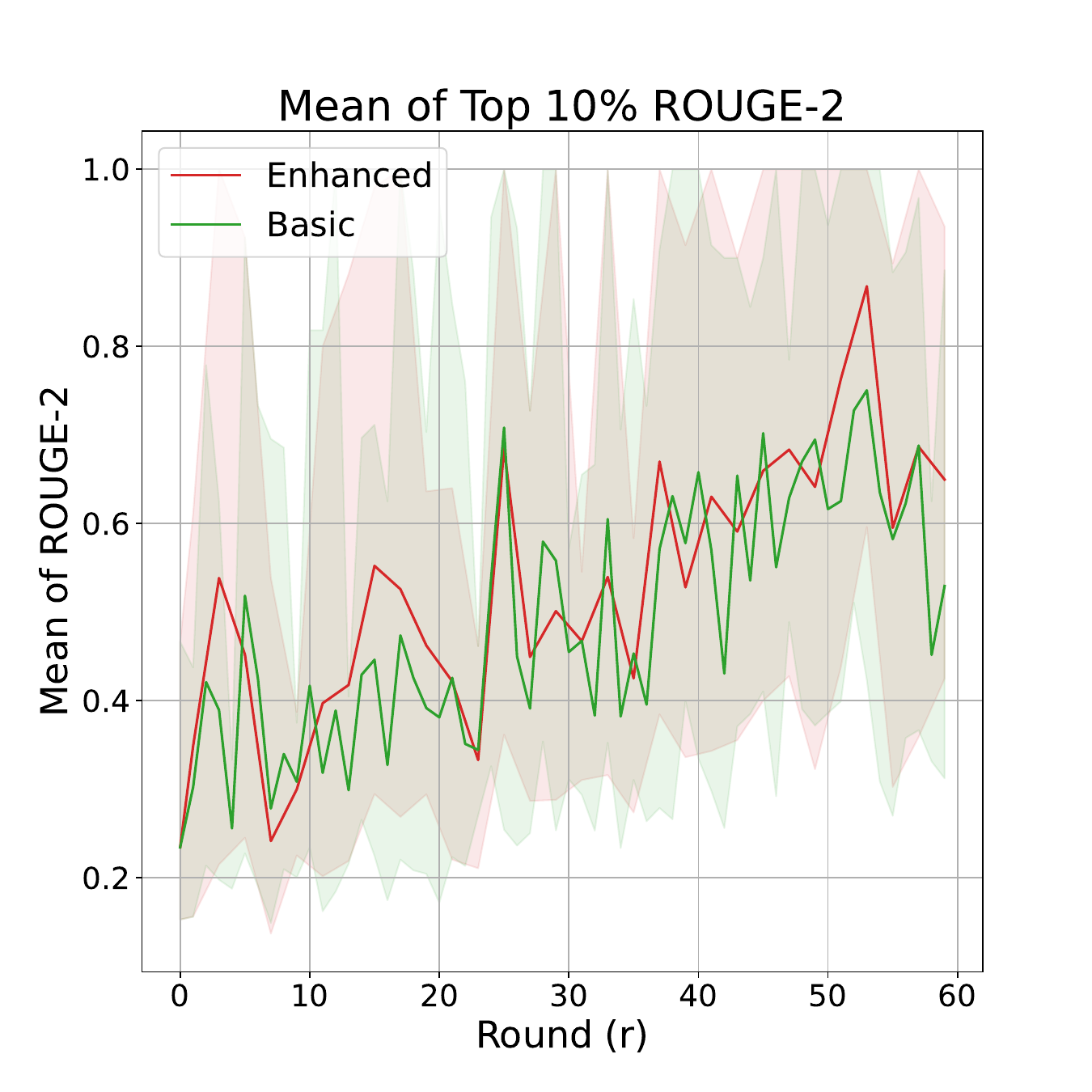}
}
\subfloat[ROUGE-1 on LLaMA-13B.]{
    \includegraphics[width=0.23\linewidth]{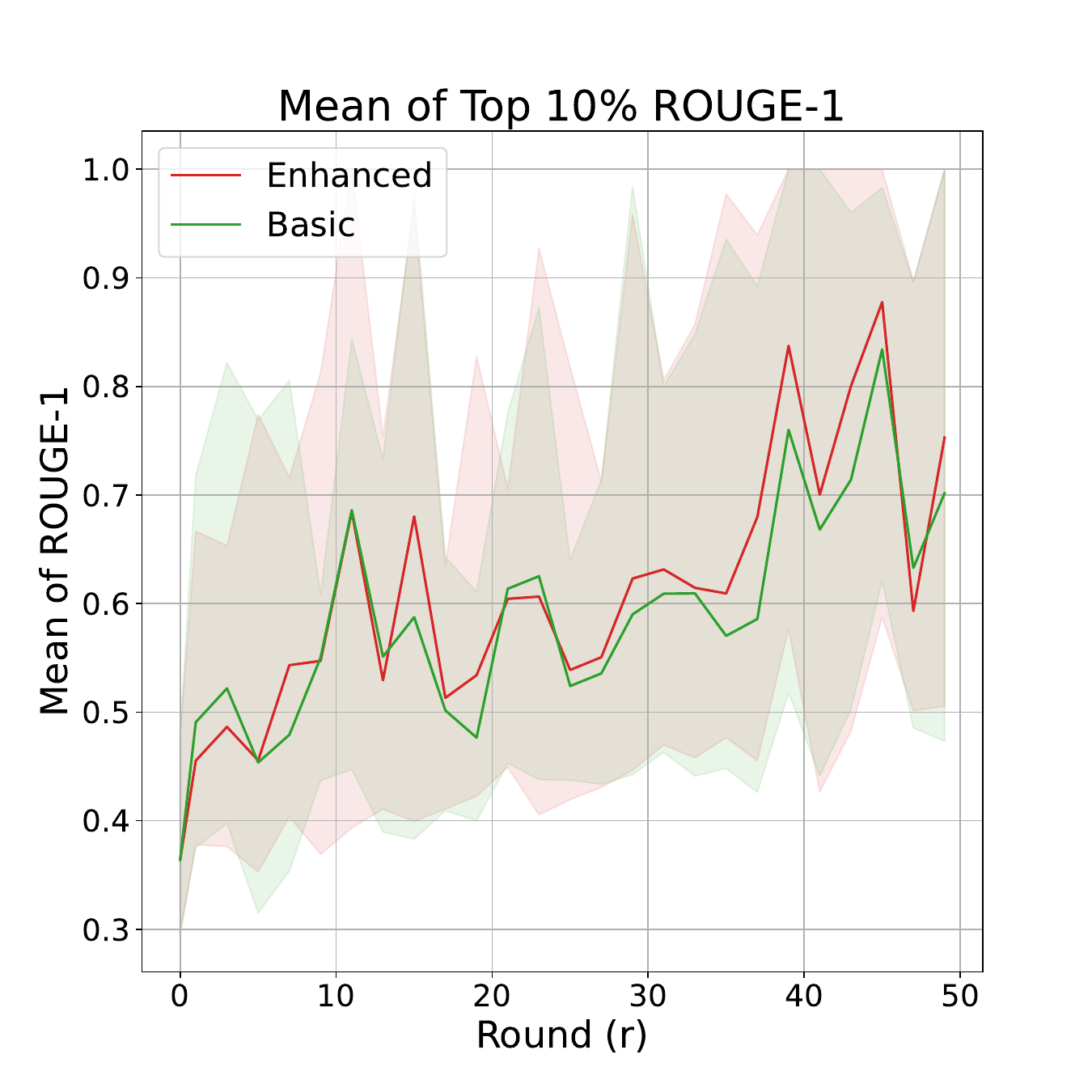}
}
\subfloat[ROUGE-2 on LLaMA-13B.]{
    \includegraphics[width=0.23\linewidth]{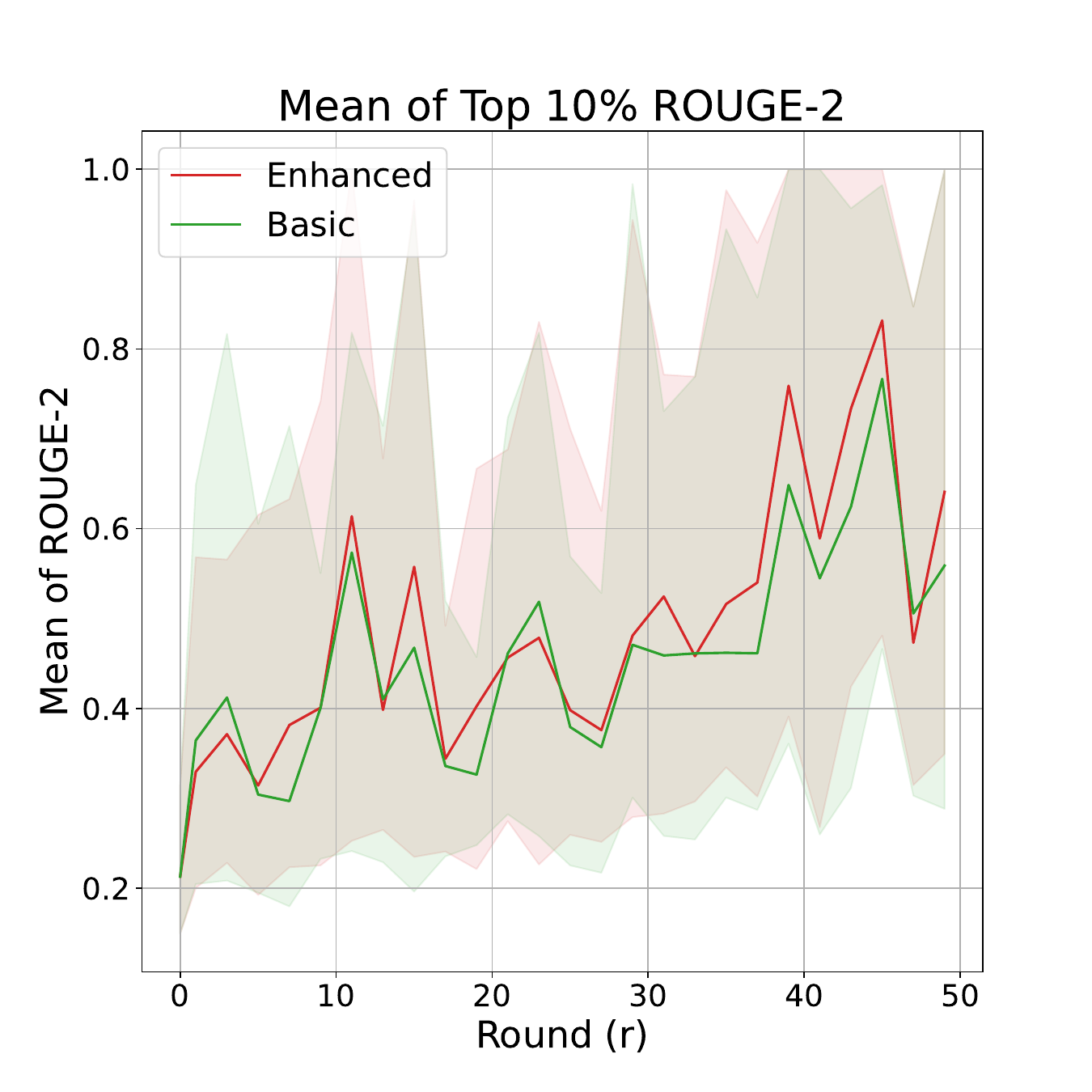}
}
\caption{Enhanced Hacking Scheme results of ROUGE-1 and ROUGE-2 for Partial-Input Completion. Experiments are repeated using LLaMA-8B and LLaMA-13B using Enron Email dataset}
\label{fig:r12_enhanced_com}
\vspace{-20pt}
\end{figure*}

Among the ROUGE family of metrics, ROUGE-1 and ROUGE-2 focus on unigram and bigram overlaps, respectively, while ROUGE-L measures the longest common subsequence between the generated and reference texts. This makes ROUGE-L more effective in capturing the overall structure and fluency of the output, which is particularly important for tasks that require coherent and well-ordered text generation, such as summarization or open-ended generation. Nevertheless, ROUGE-1 and ROUGE-2 remain valuable for evaluating lexical overlap and local consistency. In addition to ROUGE-L, we report ROUGE-1 and ROUGE-2 scores to provide a more comprehensive assessment of the similarity between generated text and private data. The detailed results are shown in Fig.~\ref{fig:r12_basic}, ~\ref{fig:r12_enhanced_sim} and ~\ref{fig:r12_enhanced_com}. As shown, the performance and trends of ROUGE-1, ROUGE-2, and ROUGE-L are highly consistent, further supporting the conclusions drawn from the experiments in Section~\ref{section:hack_result} regarding privacy leakage.

\section{Evaluation of More Models and Datasets}
\label{appendix:more}
The main text reports results for the LLaMA family of LLMs on the Enron Email Dataset. This appendix broadens the scope by evaluating two base models, Gemma‑2‑2B and Qwen2.5‑7B, and by adding two domain‑distinct corpora, the Reddit Comments Dataset and the CLERC Dataset.

\subsection{Enron Email Dataset}
We benchmark Gemma‑2‑2B and Qwen2.5‑7B on the Enron Email Dataset, comparing the two hacking tasks under both hacking schemes.

\begin{figure*}[!htbp]
\setlength{\abovecaptionskip}{6pt}
\centering
\subfloat[Top 30\% on Gemma-2-2B.]{
    \includegraphics[height=7cm, width=0.48\linewidth]{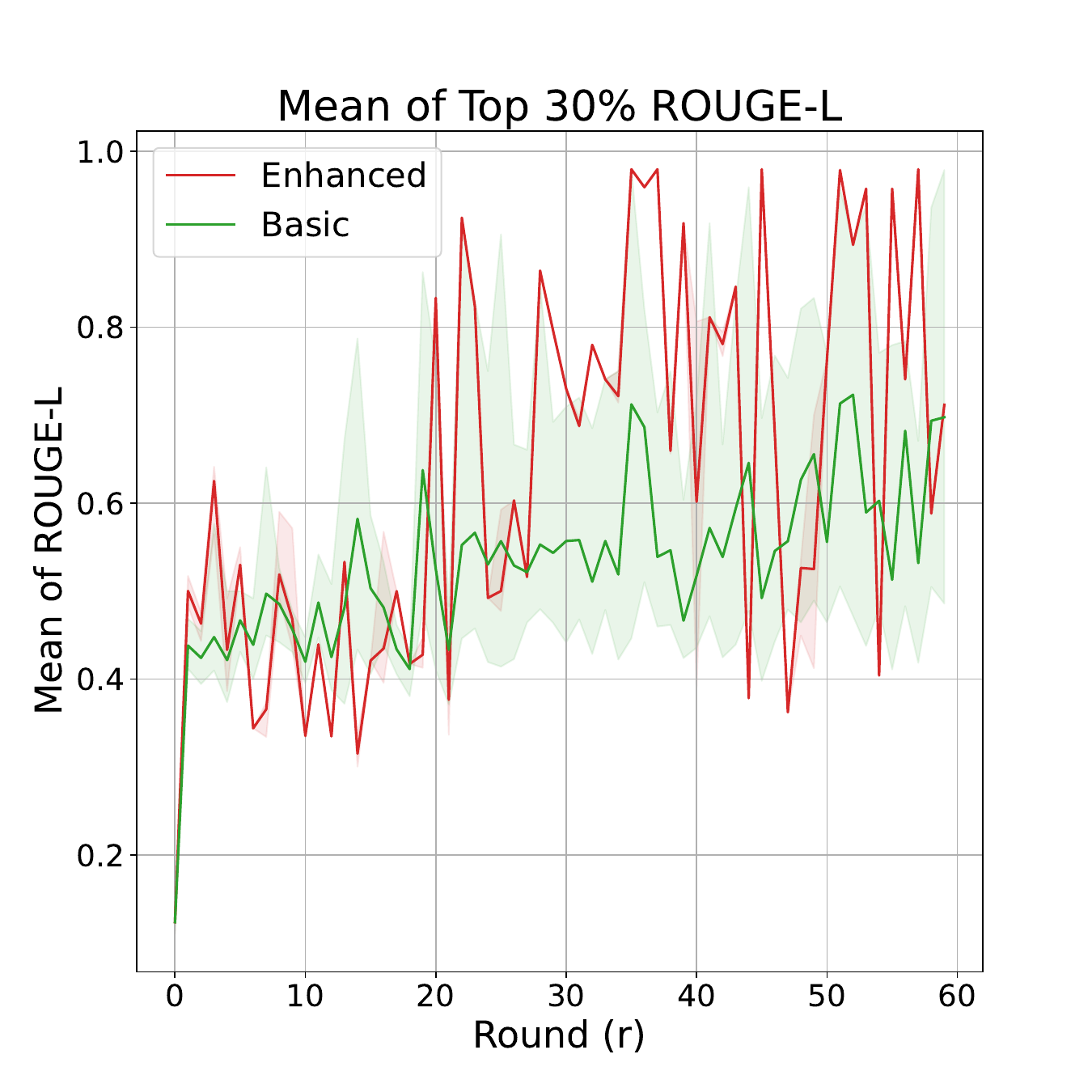}
}
\subfloat[Top 50\% on Gemma-2-2B.]{
    \includegraphics[height=7cm, width=0.48\linewidth]{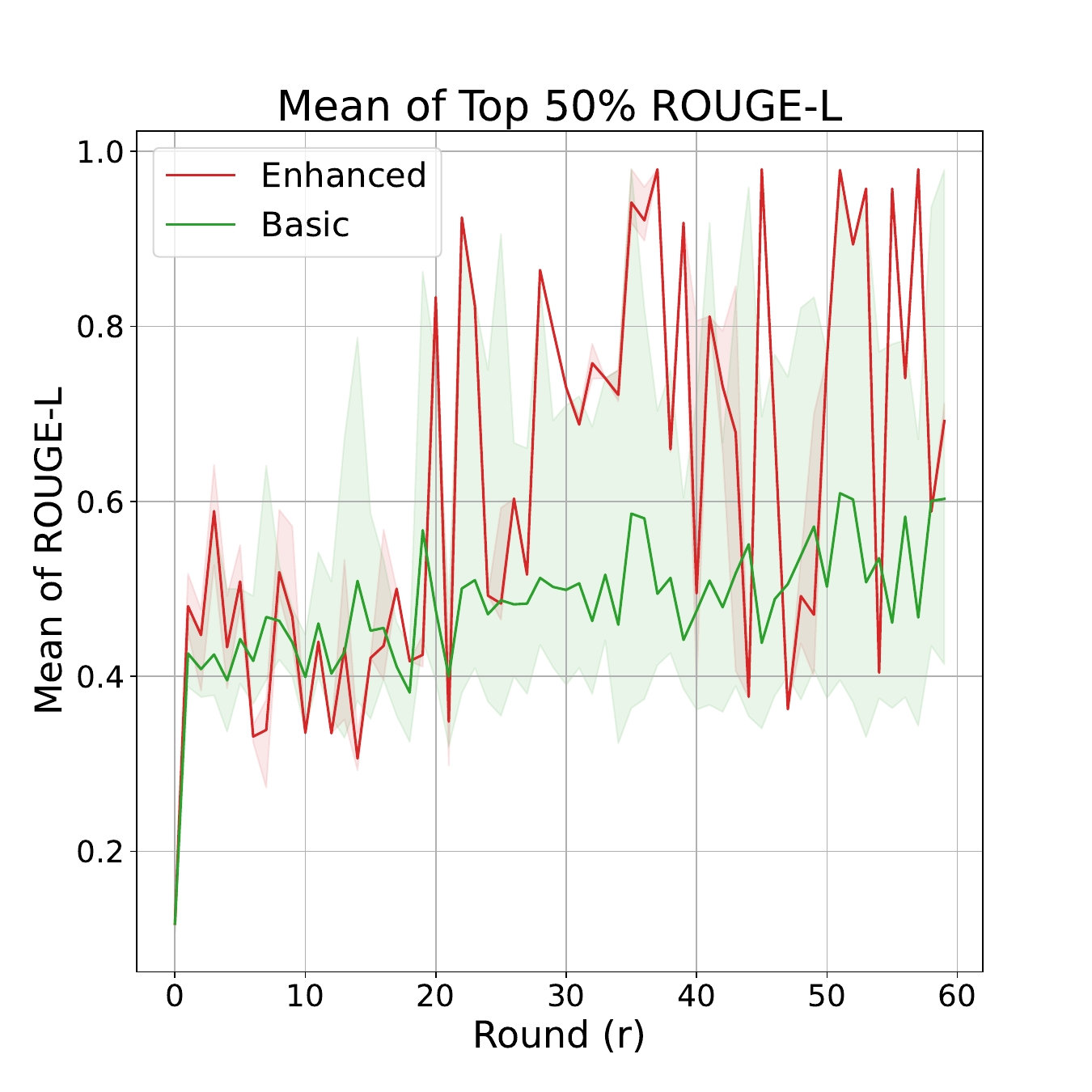}
}
\\
\subfloat[Top 30\% on Qwen2.5-7B.]{
    \includegraphics[height=7cm, width=0.48\linewidth]{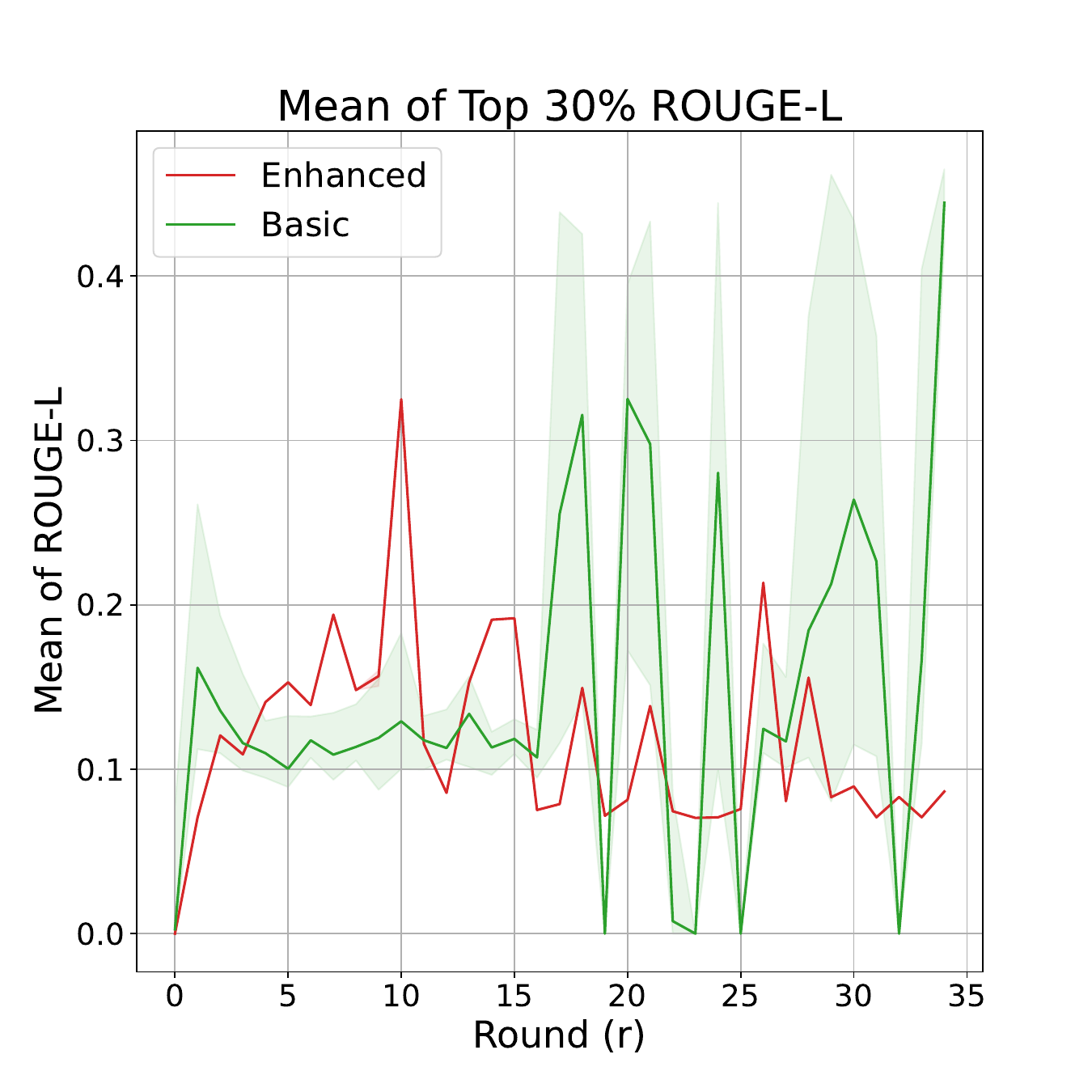}
}
\subfloat[Top 50\% on Qwen2.5-7B.]{
    \includegraphics[height=7cm, width=0.48\linewidth]{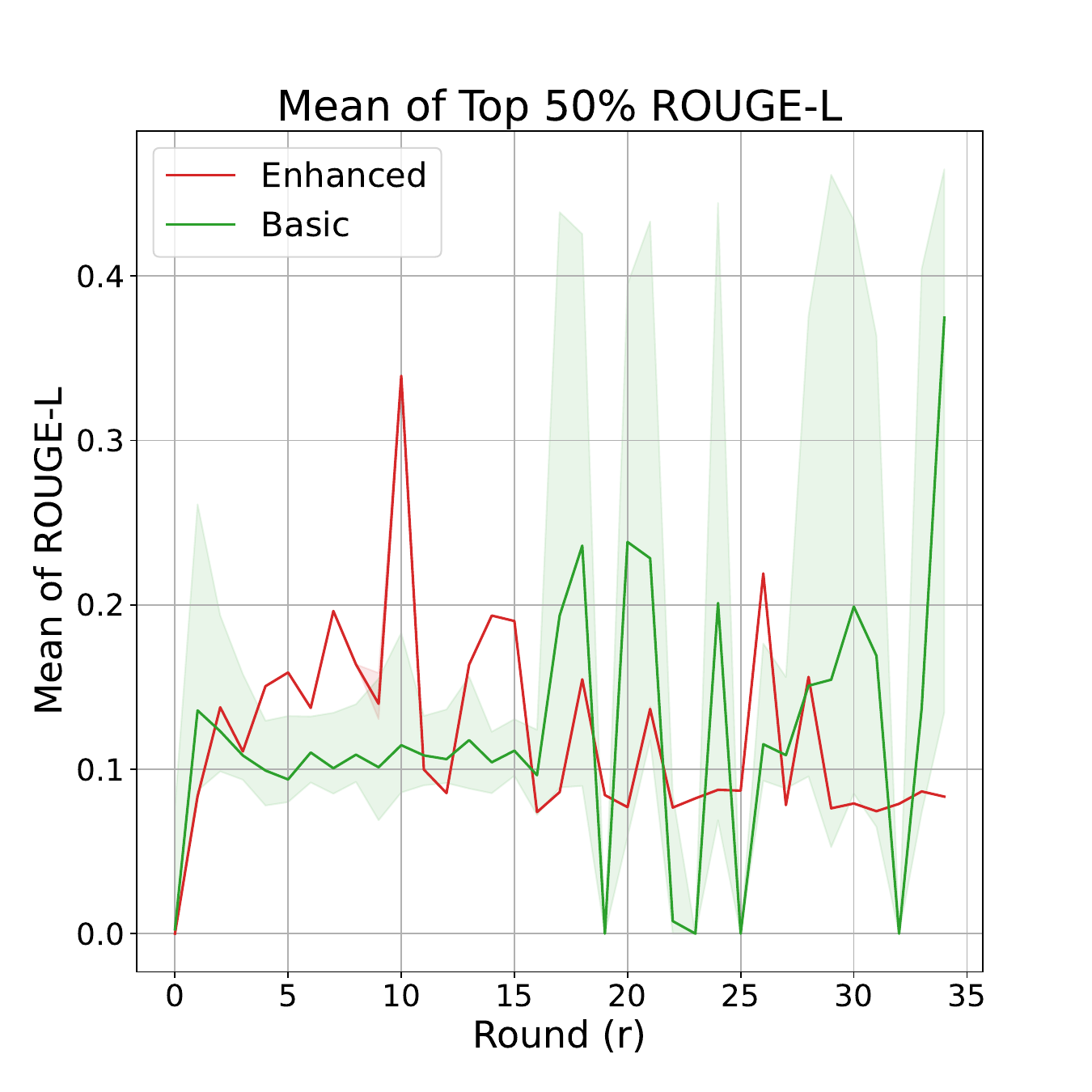}
}
\caption{Comparison of two hacking schemes' results for Zero-Input Generation on Enron Email Dataset. Experiments are repeated using Qwen2.5-7B and Gemma-2-2B.}
\label{fig:more_enron_sim}
\vspace{-20pt}
\end{figure*}

\begin{figure*}[!htbp]
\setlength{\abovecaptionskip}{6pt}
\centering
\subfloat[Top 30\% on Gemma-2-2B.]{
    \includegraphics[height=7cm, width=0.48\linewidth]{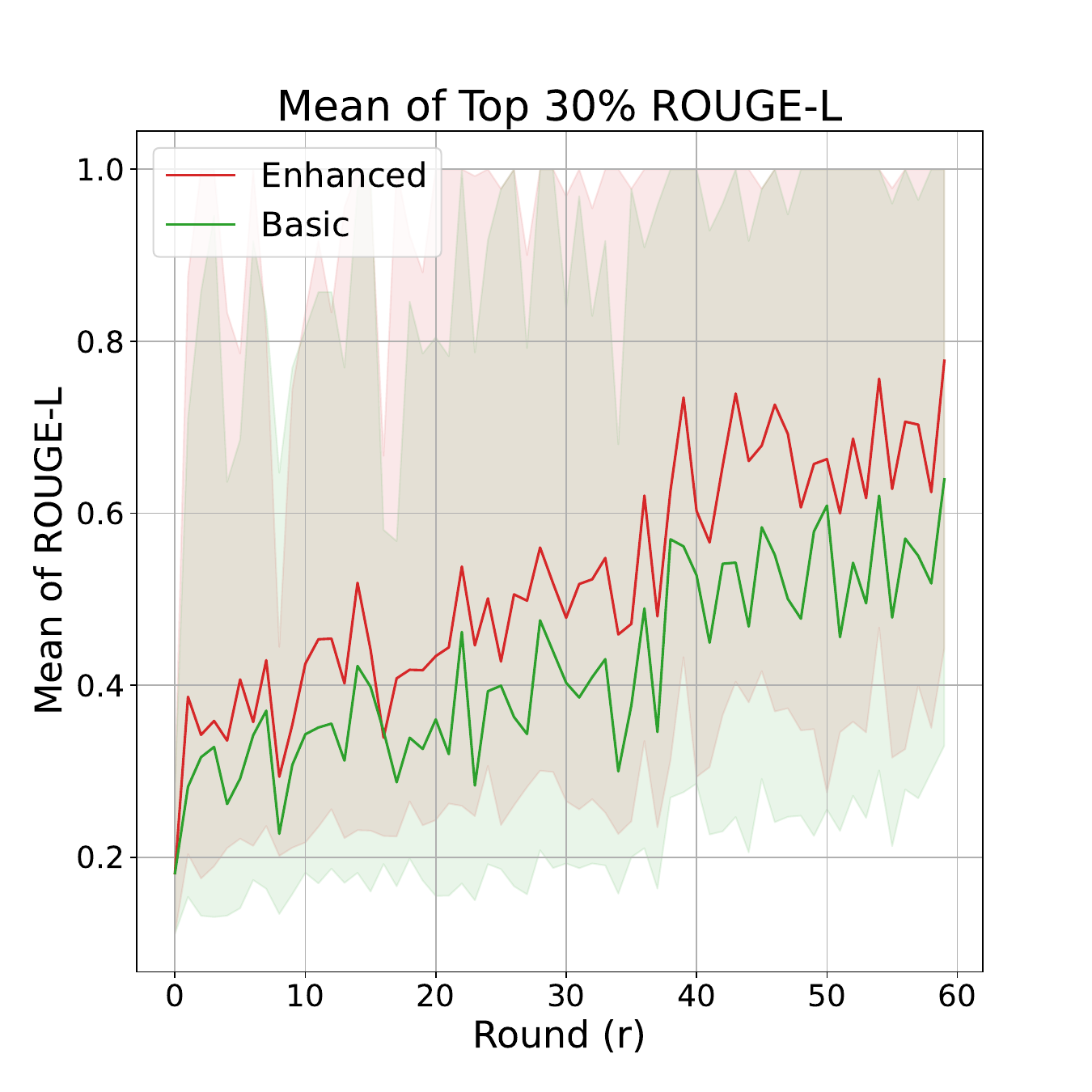}
}
\subfloat[Top 50\% on Gemma-2-2B.]{
    \includegraphics[height=7cm, width=0.48\linewidth]{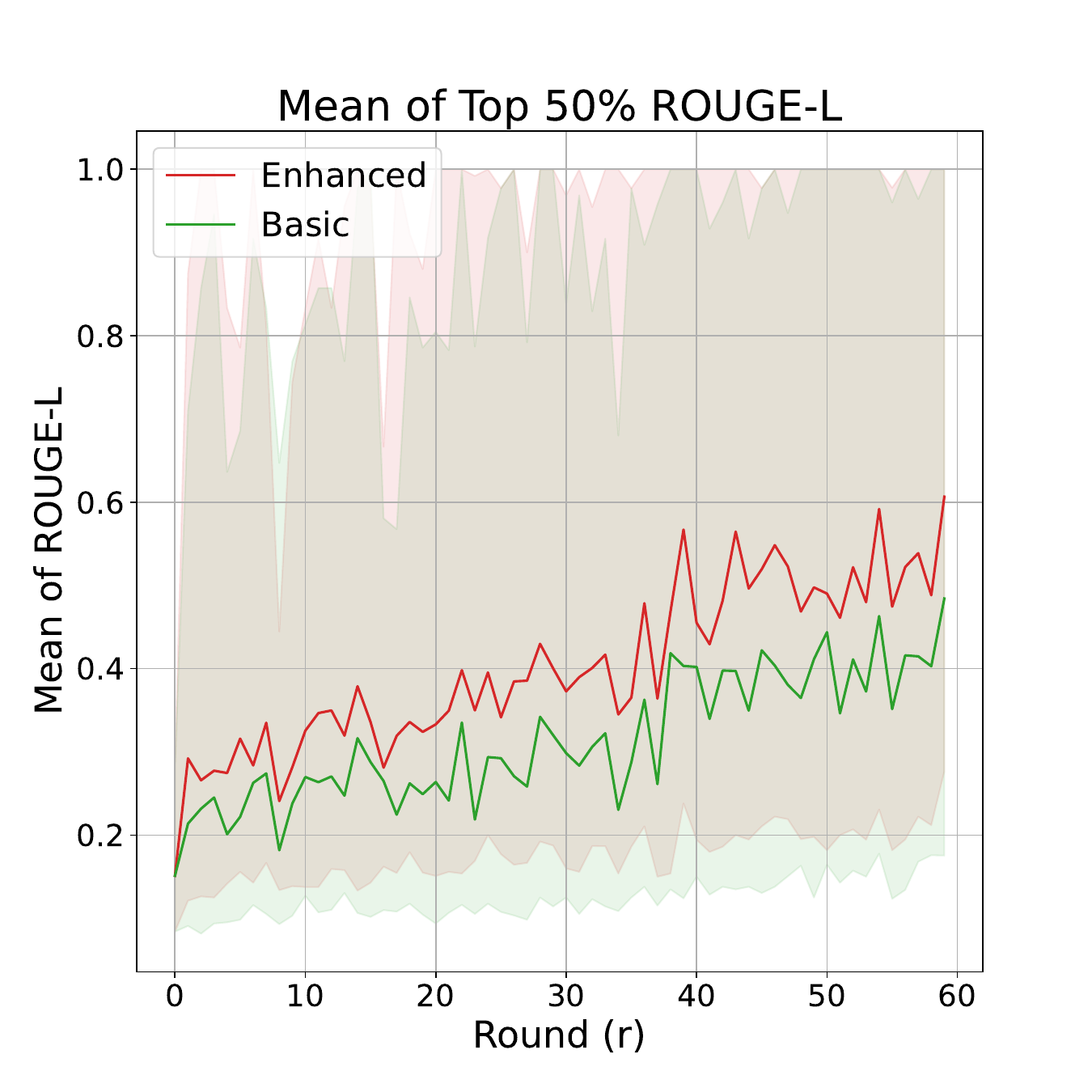}
}
\\
\subfloat[Top 30\% on Qwen2.5-7B.]{
    \includegraphics[height=7cm, width=0.48\linewidth]{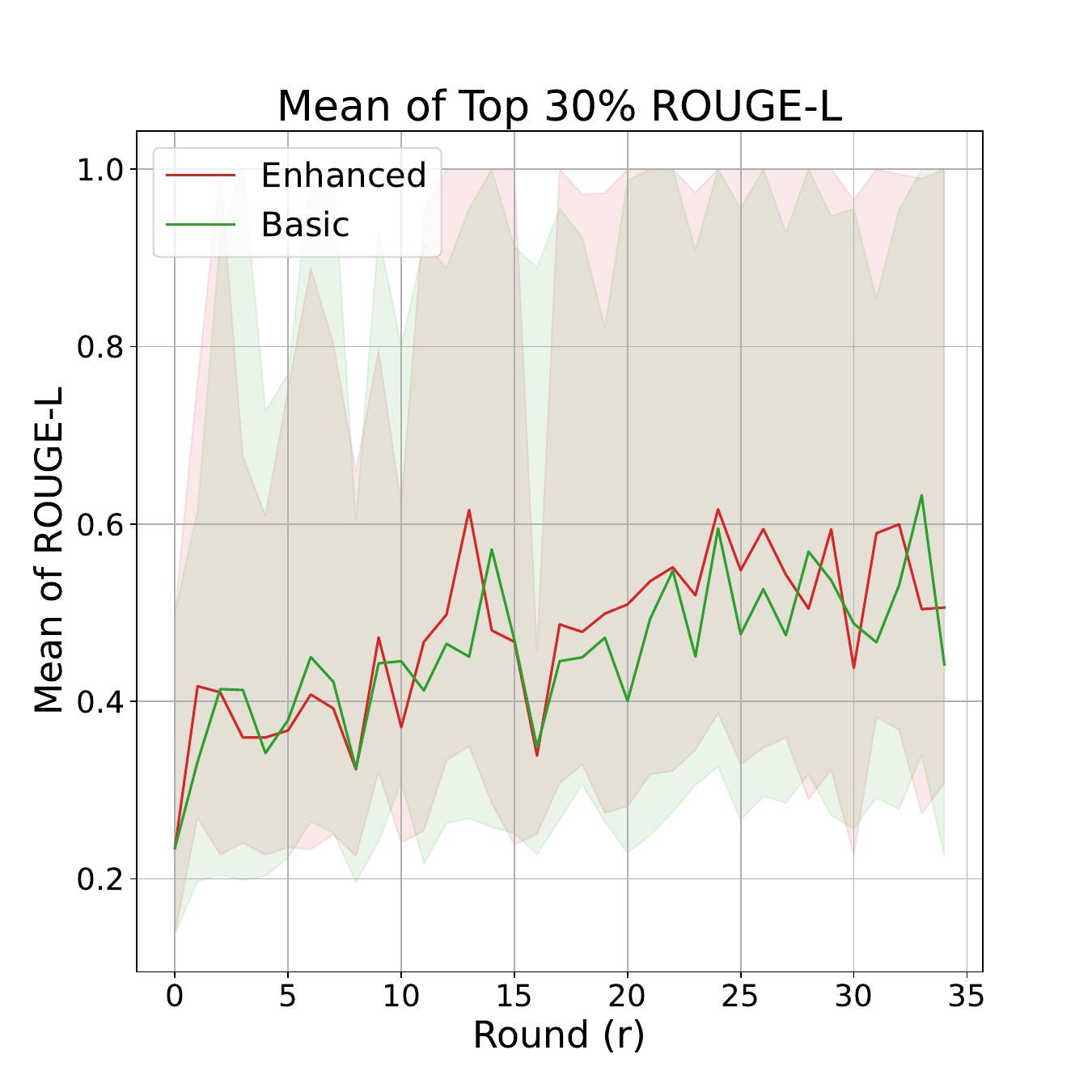}
}
\subfloat[Top 50\% on Qwen2.5-7B.]{
    \includegraphics[height=7cm, width=0.48\linewidth]{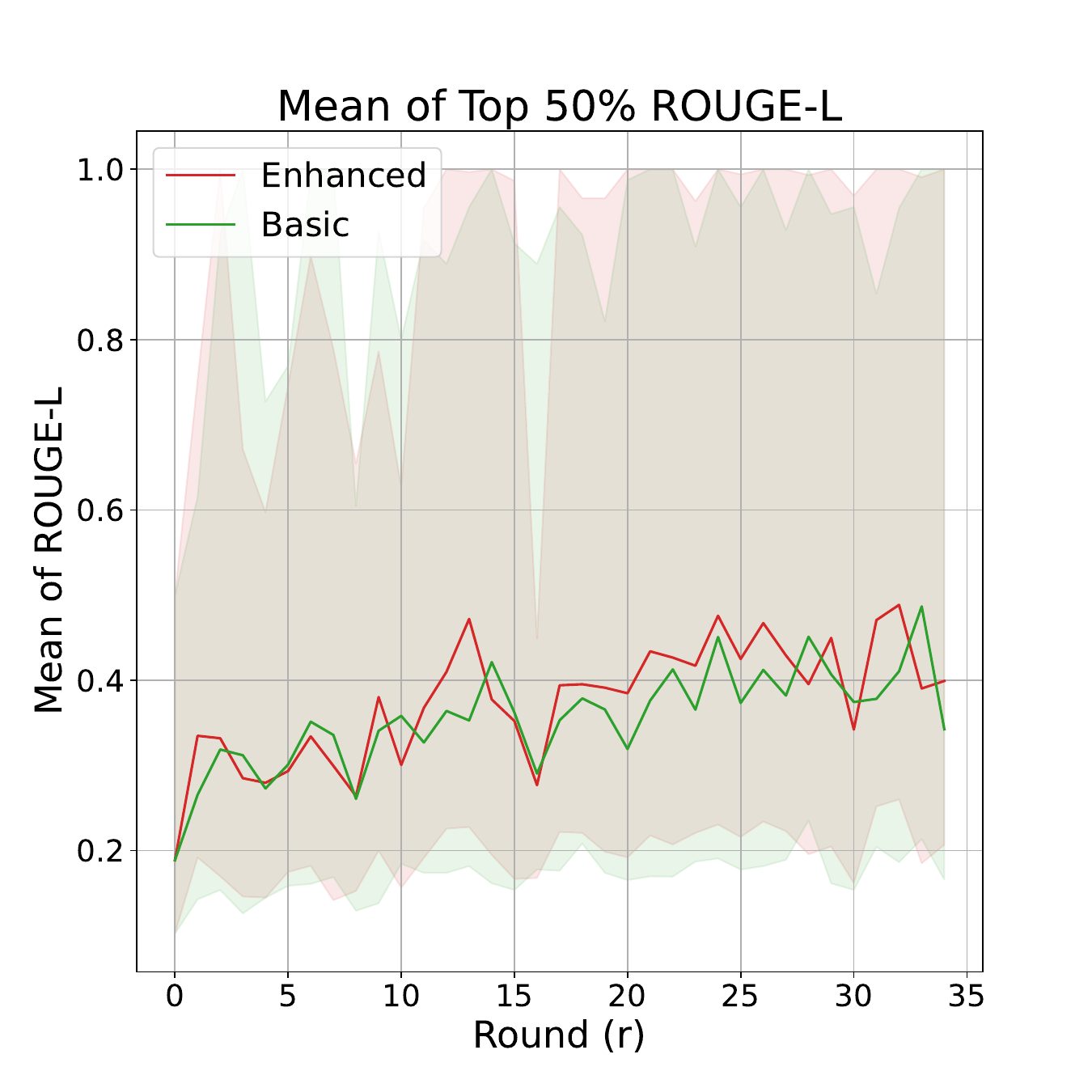}
}
\caption{Comparison of two hacking schemes' results for Partial-Input Completion on Enron Email Dataset. Experiments are repeated using Qwen2.5-7B and Gemma-2-2B.}
\label{fig:more_enron_com}
\vspace{-20pt}
\end{figure*}

As shown in Fig.~\ref{fig:more_enron_sim} and ~\ref{fig:more_enron_com}, both models already reveal substantial privacy leakage in the basic hacking scheme. And the enhanced hacking scheme amplifies this leakage even further.

\subsection{Reddit Comment Dataset}
Next, we evaluate LLaMA‑3.1‑8B, Gemma‑2‑2B, and Qwen2.5‑7B on the Reddit Comments Dataset, contrasting the two hacking tasks under both schemes.

\begin{figure*}[!htbp]
\setlength{\abovecaptionskip}{6pt}
\centering
\subfloat[Top 30\% on Gemma-2-2B.]{
    \includegraphics[height=7cm, width=0.48\linewidth]{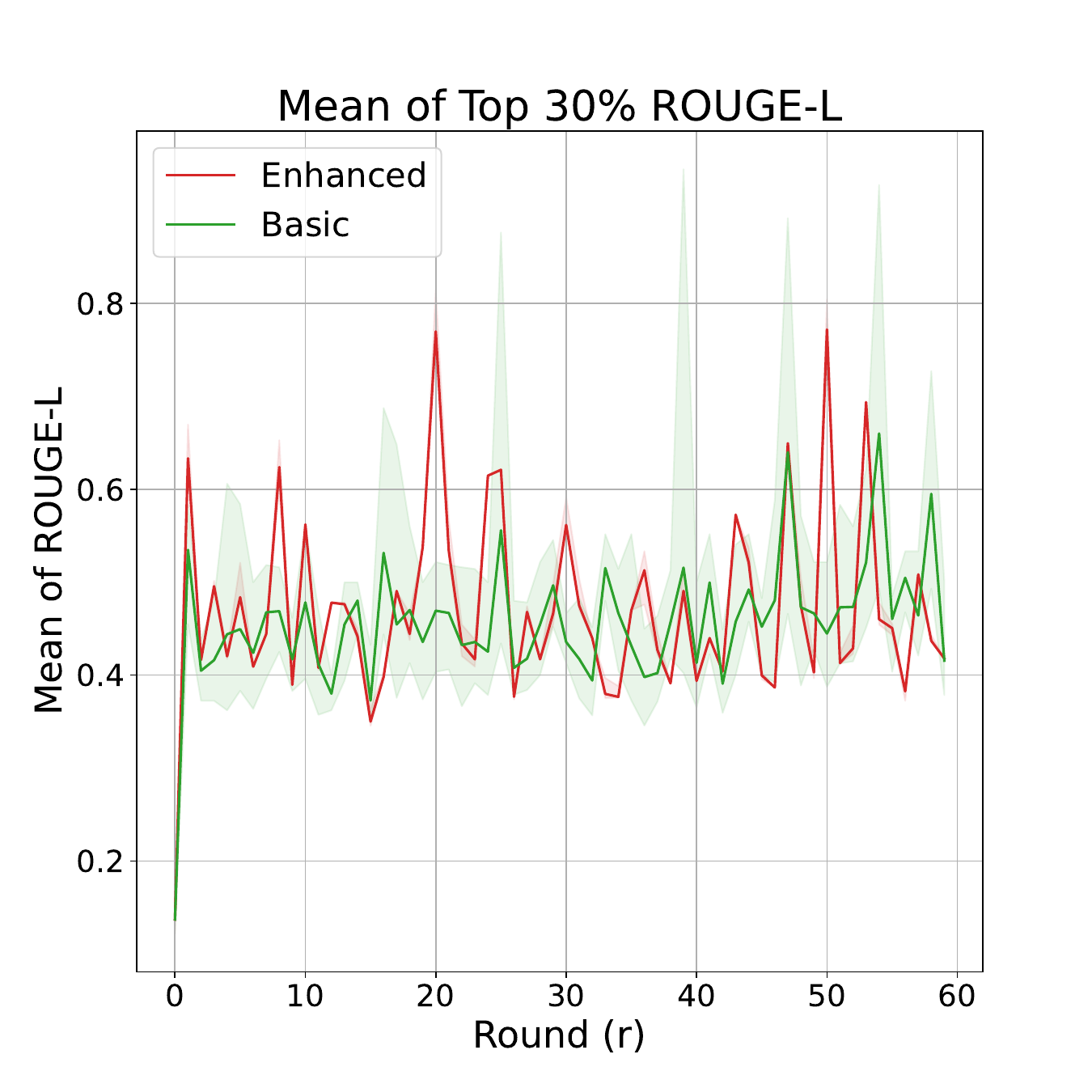}
}
\subfloat[Top 50\% on Gemma-2-2B.]{
    \includegraphics[height=7cm, width=0.48\linewidth]{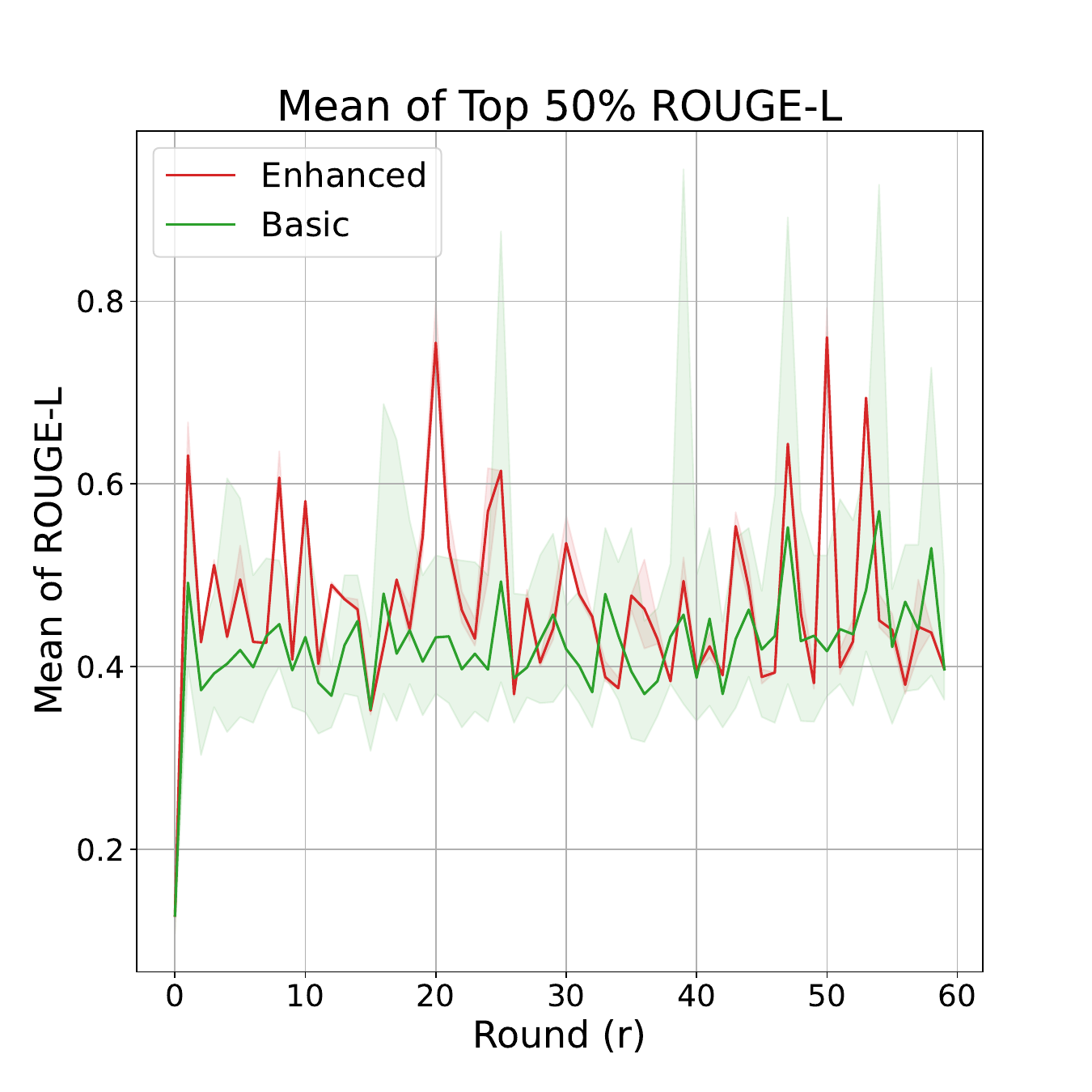}
}
\\
\vspace{-13pt}
\subfloat[Top 30\% on Qwen2.5-7B.]{
    \includegraphics[height=7cm, width=0.48\linewidth]{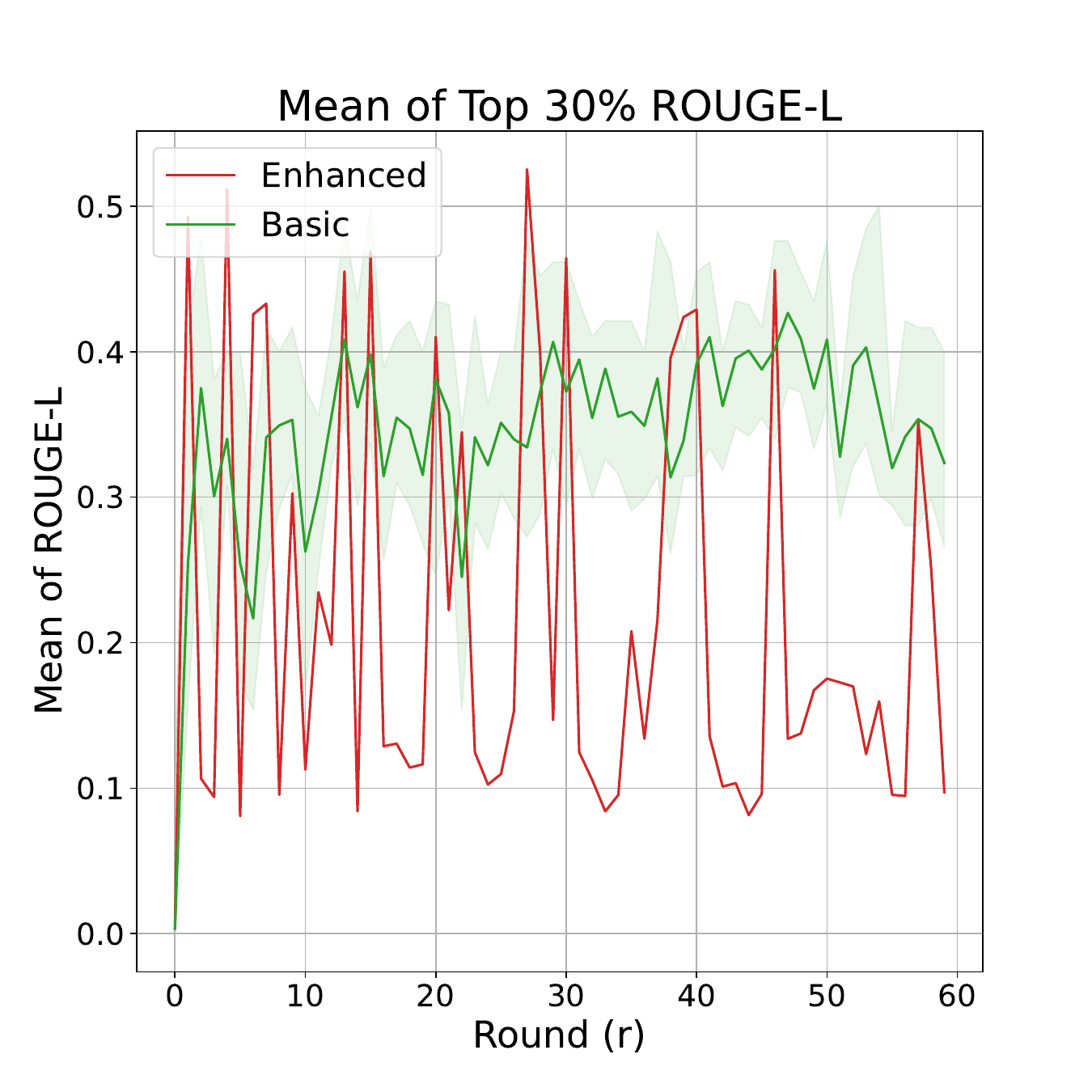}
}
\subfloat[Top 50\% on Qwen2.5-7B.]{
    \includegraphics[height=7cm, width=0.48\linewidth]{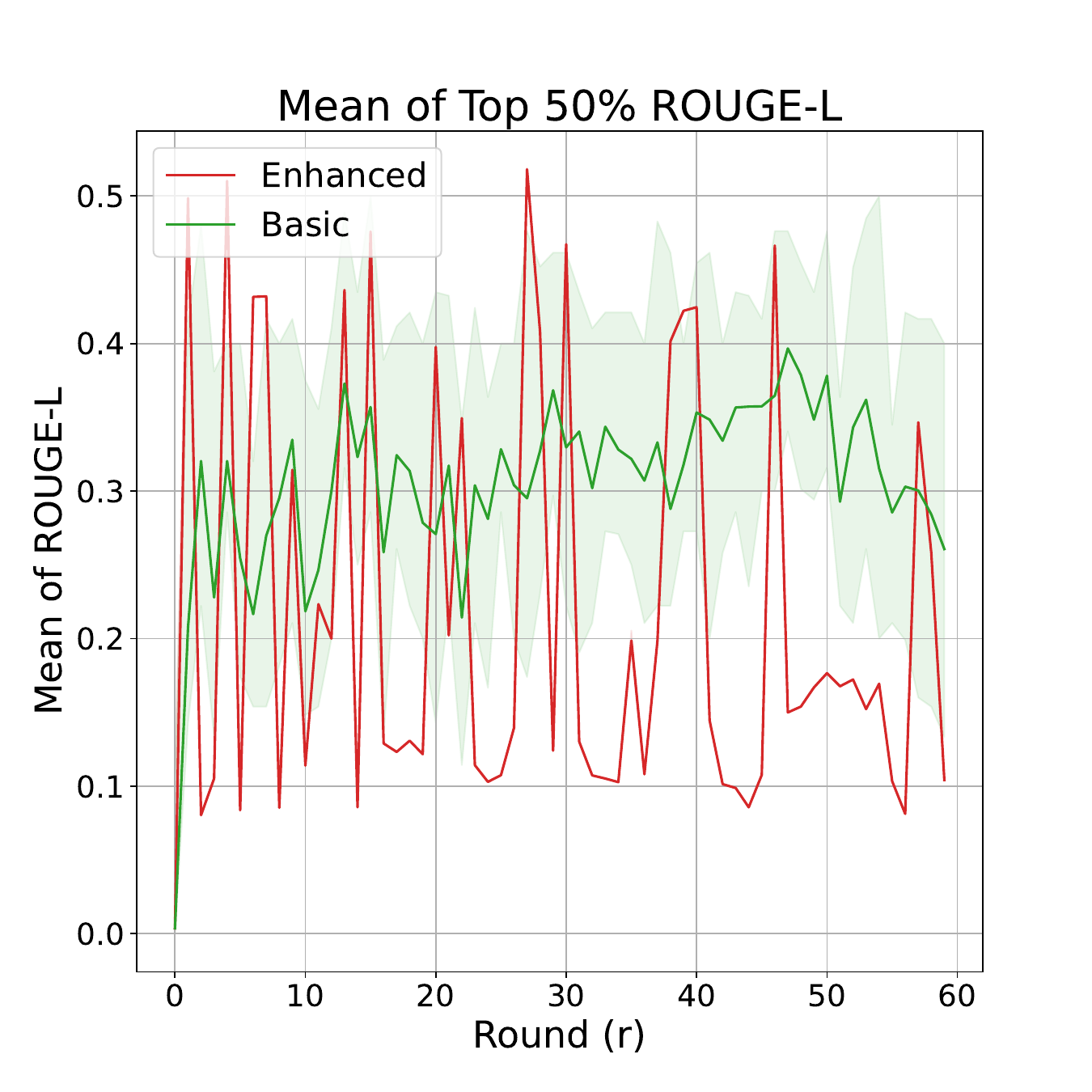}
}
\\
\vspace{-13pt}
\subfloat[Top 30\% on LLaMA-3.1-8B.]{
    \includegraphics[height=7cm, width=0.48\linewidth]{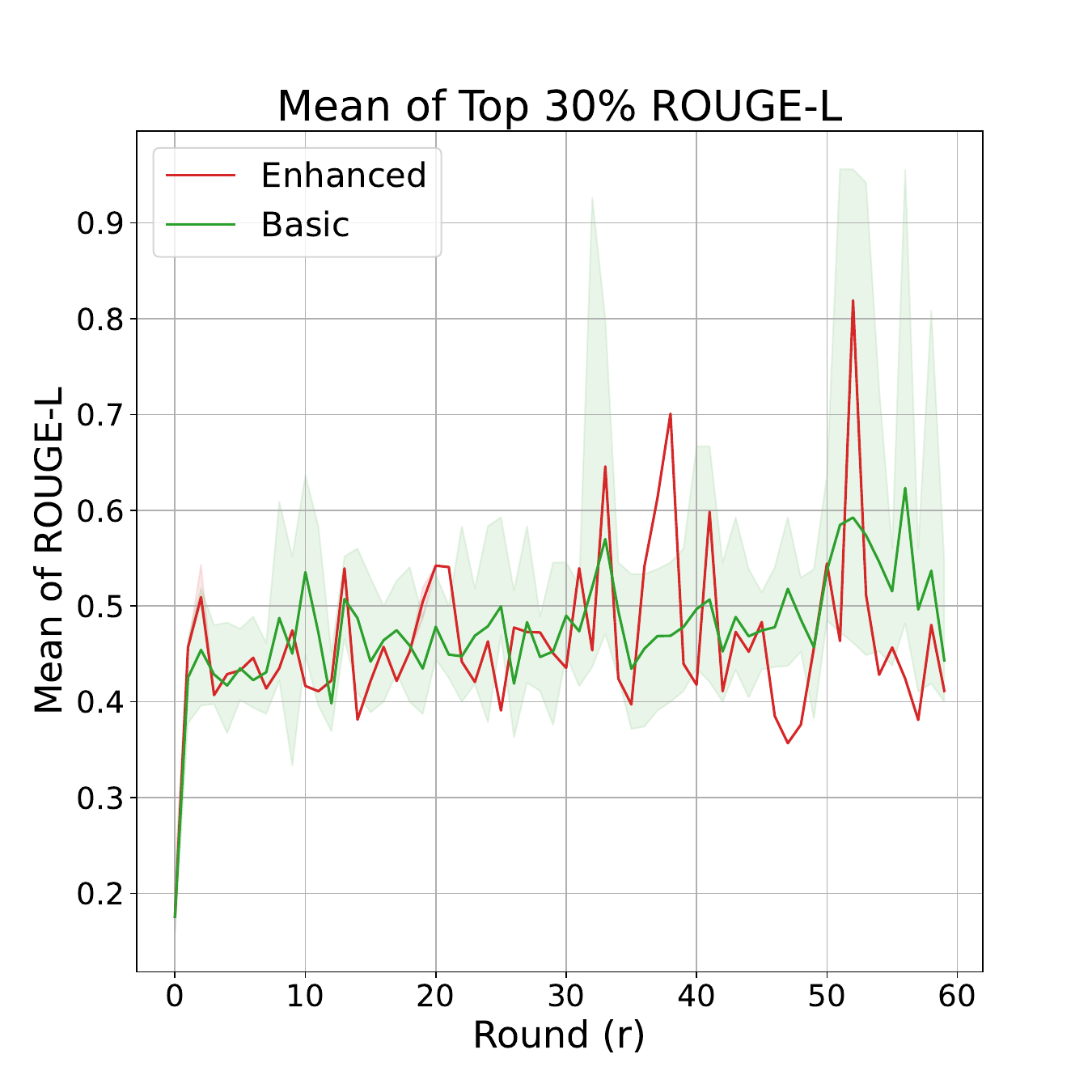}
}
\subfloat[Top 50\% on LLaMA-3.1-8B.]{
    \includegraphics[height=7cm, width=0.48\linewidth]{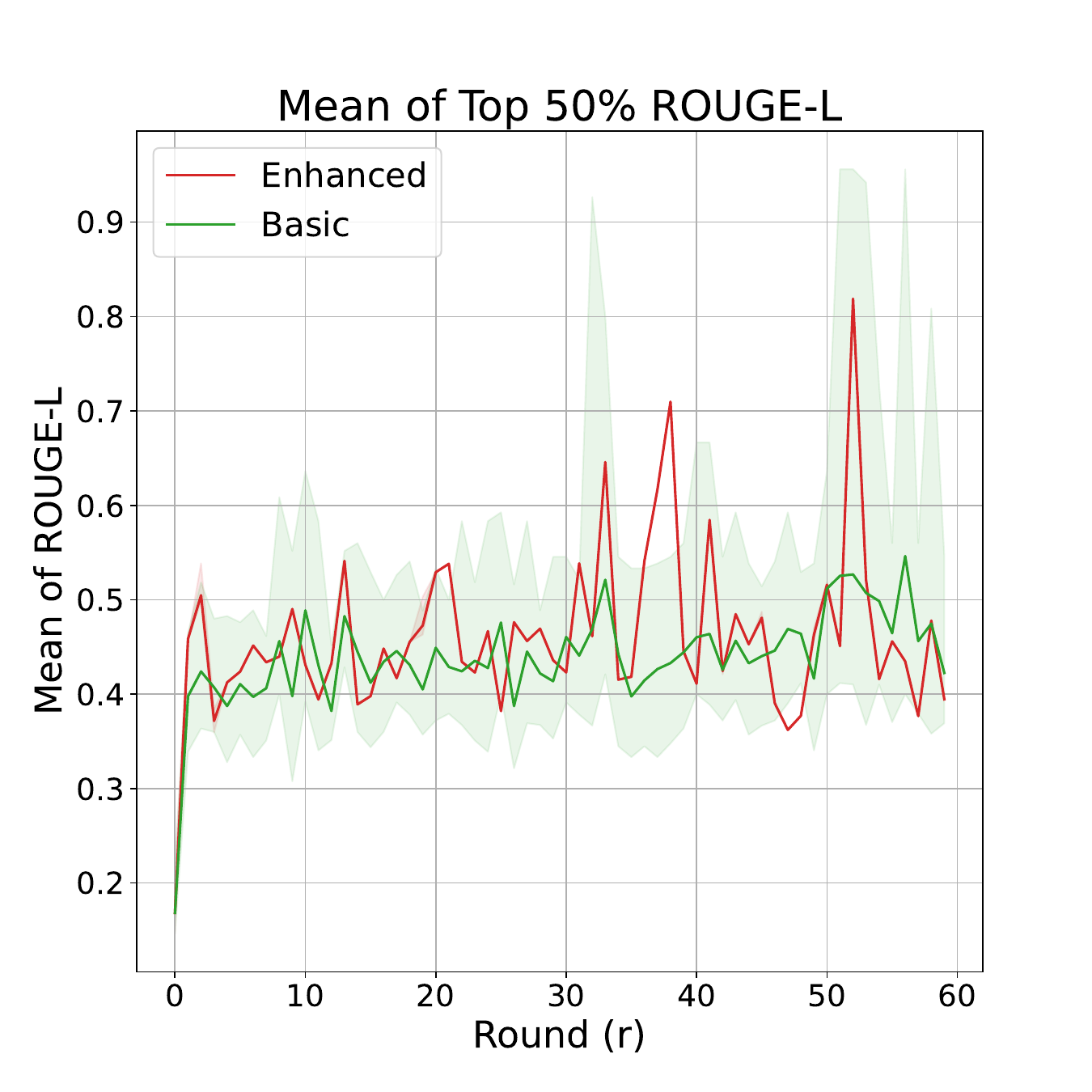}
}
\caption{Comparison of two hacking schemes' results for Zero-Input Generation on Reddit Comment Dataset. Experiments are repeated using Qwen2.5-7B, Gemma-2-2B and LLaMA-3.1-8B.}
\label{fig:more_reddit_sim}
\end{figure*}

\begin{figure*}[!htbp]
\setlength{\abovecaptionskip}{6pt}
\centering
\subfloat[Top 30\% on Gemma-2-2B.]{
    \includegraphics[height=7cm, width=0.48\linewidth]{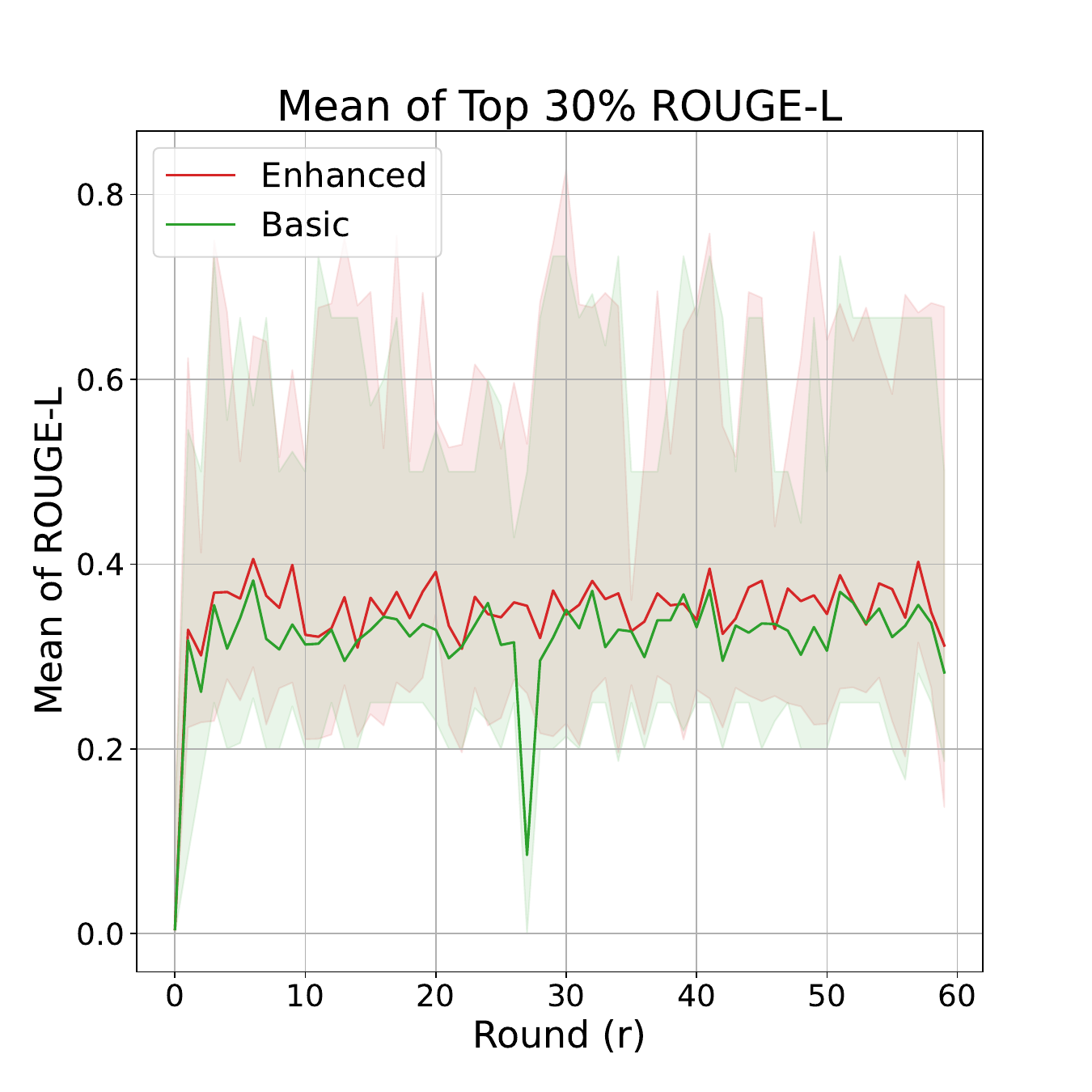}
}
\subfloat[Top 50\% on Gemma-2-2B.]{
    \includegraphics[height=7cm, width=0.48\linewidth]{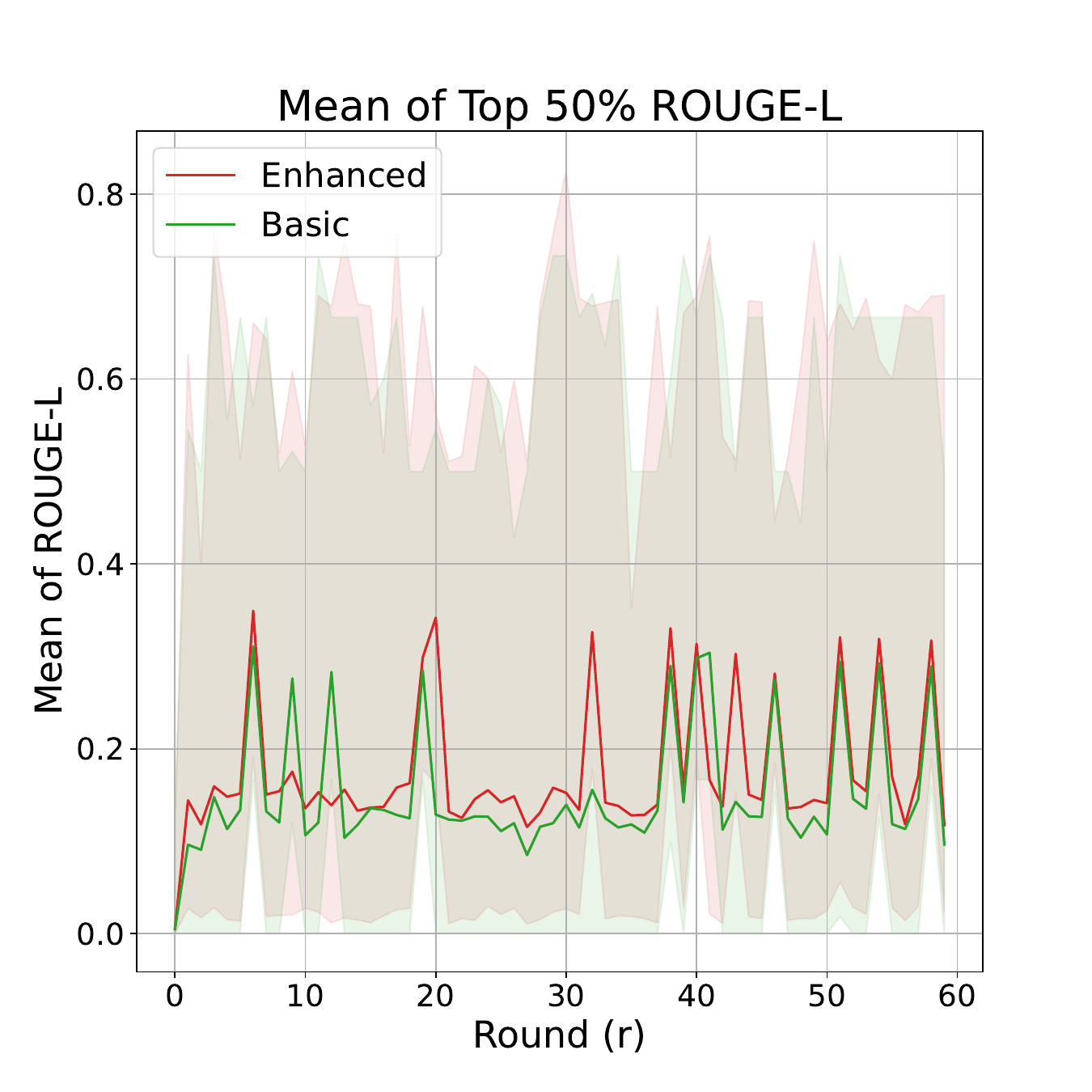}
}
\\
\vspace{-13pt}
\subfloat[Top 30\% on Qwen2.5-7B.]{
    \includegraphics[height=7cm, width=0.48\linewidth]{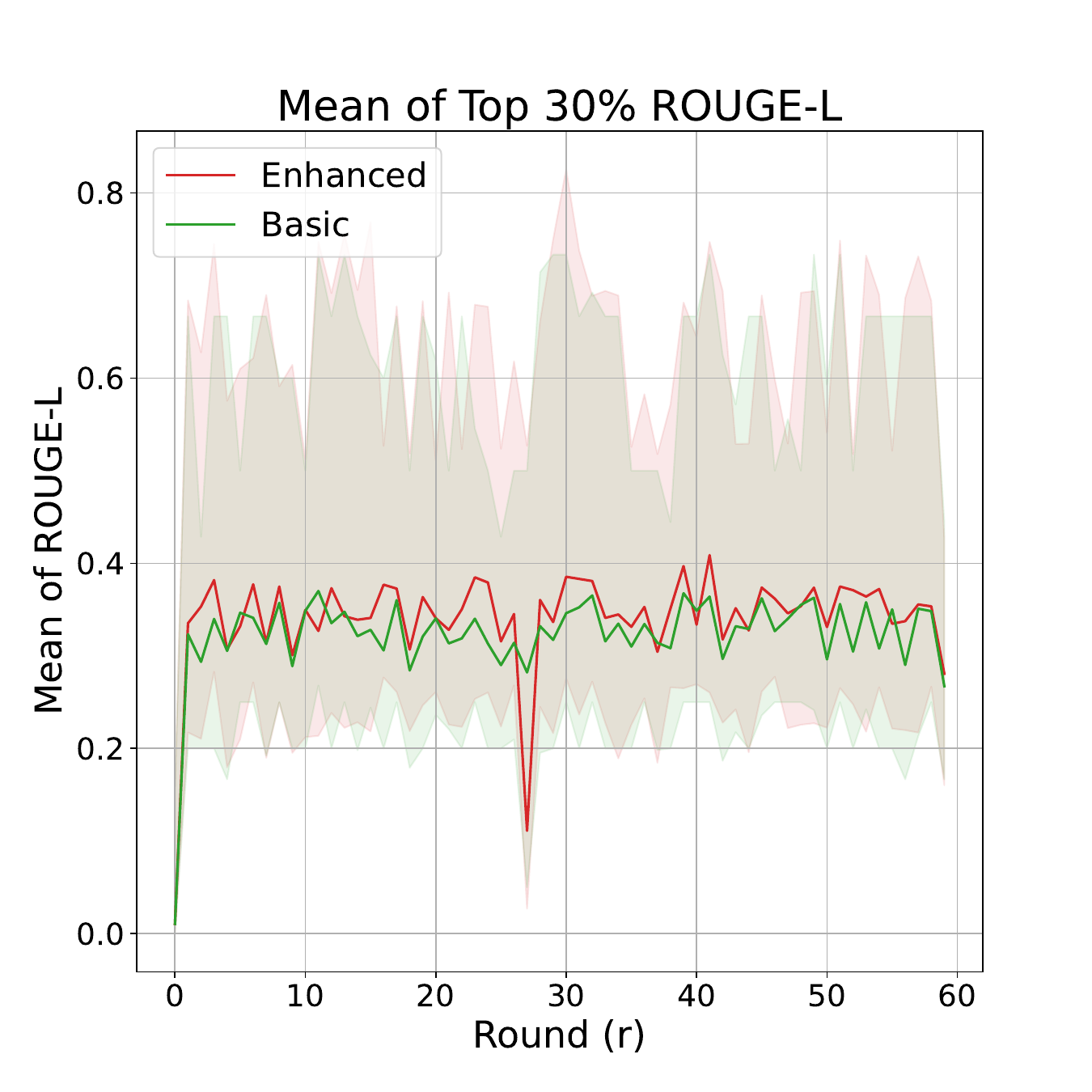}
}
\subfloat[Top 50\% on Qwen2.5-7B.]{
    \includegraphics[height=7cm, width=0.48\linewidth]{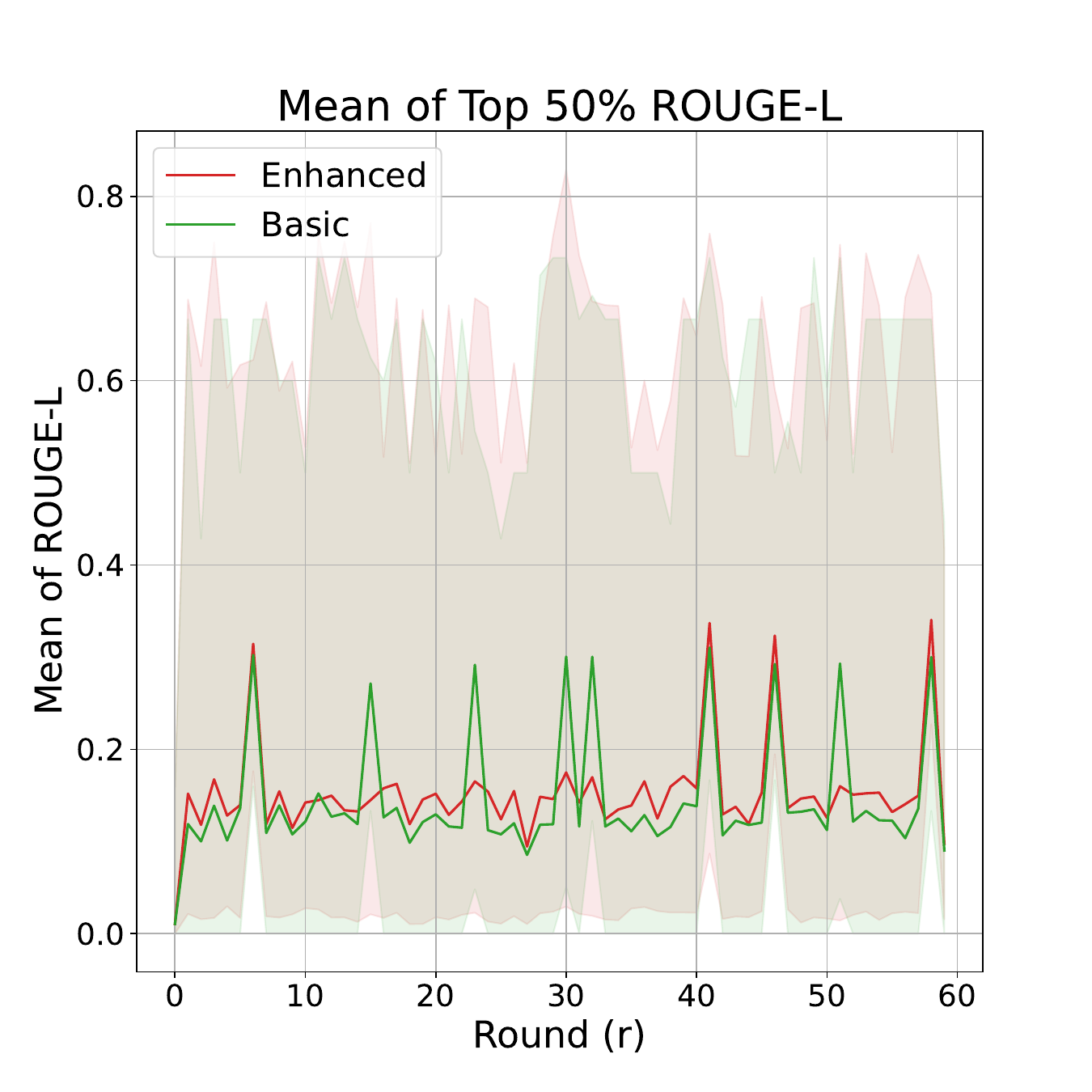}
}
\\
\vspace{-13pt}
\subfloat[Top 30\% on LLaMA-3.1-8B.]{
    \includegraphics[height=7cm, width=0.48\linewidth]{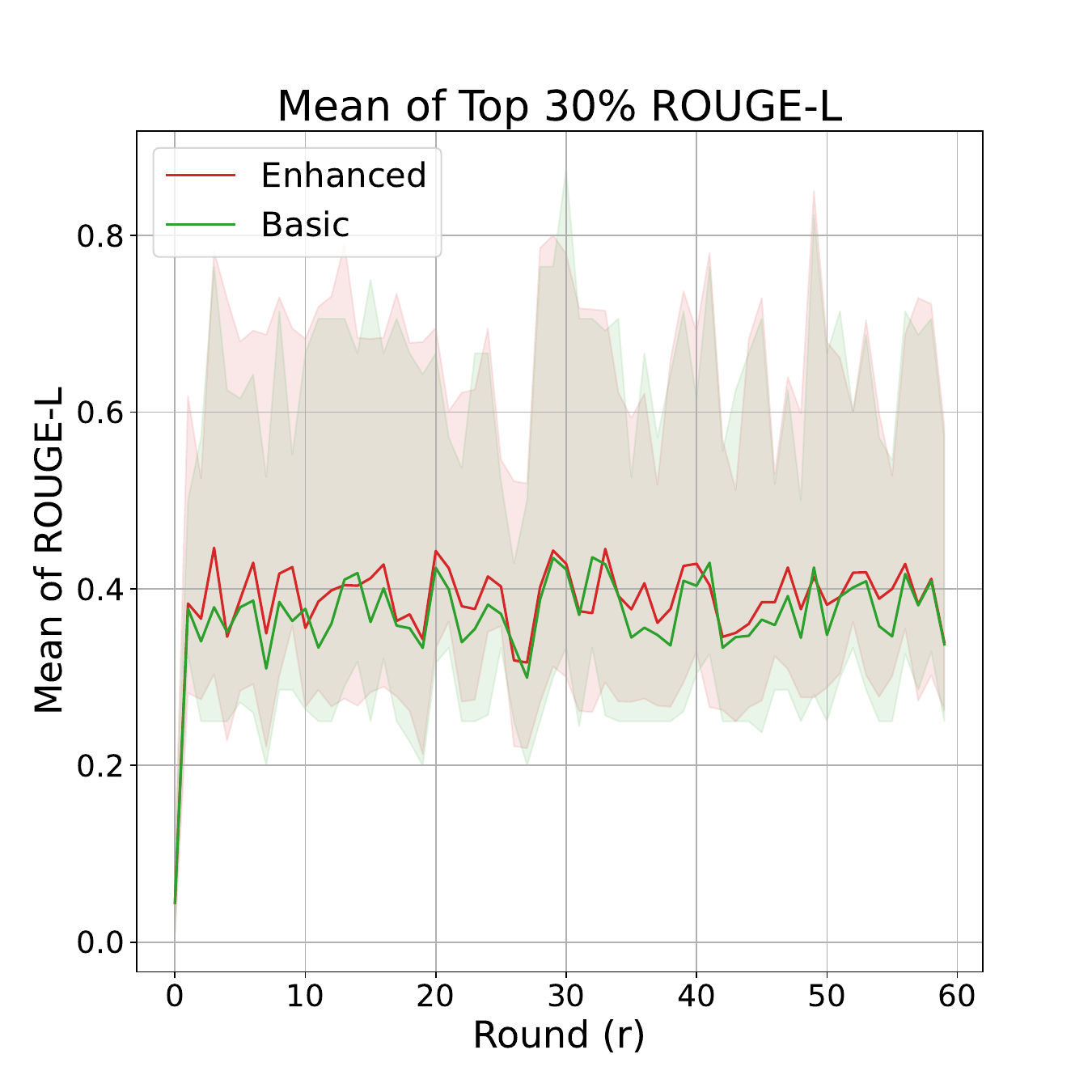}
}
\subfloat[Top 50\% on LLaMA-3.1-8B.]{
    \includegraphics[height=7cm, width=0.48\linewidth]{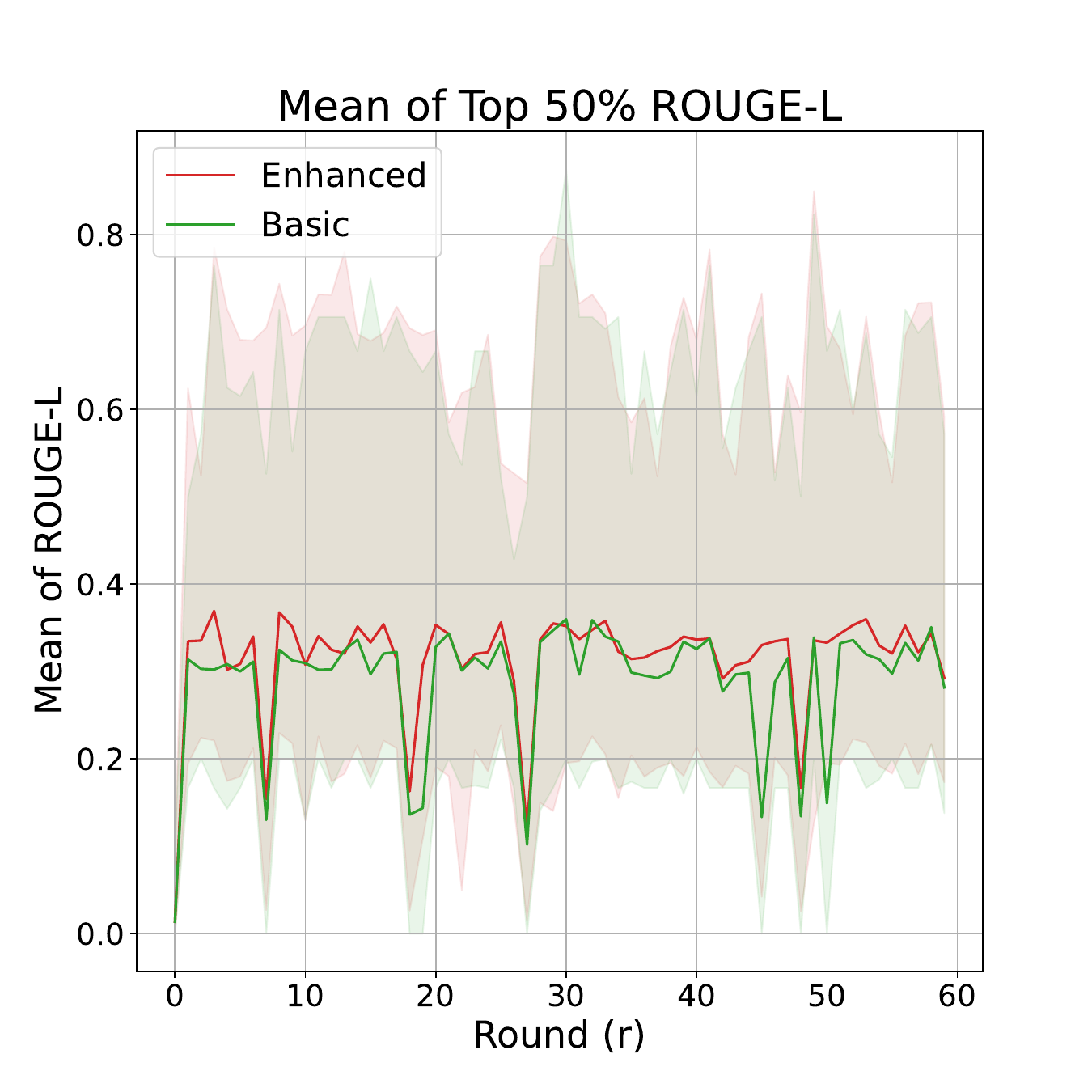}
}
\caption{Comparison of two hacking schemes' results for Partial-Input Completion on Reddit Comment Dataset. Experiments are repeated using Qwen2.5-7B, Gemma-2-2B and LLaMA-3.1-8B.}
\label{fig:more_reddit_com}
\end{figure*}

As shown in Fig.~\ref{fig:more_reddit_sim} and ~\ref{fig:more_reddit_com}, both models exhibit noticeable privacy leakage under the basic hacking scheme. However, the performance gain from the enhanced hacking scheme is relatively limited compared to the other datasets. We hypothesize that this is due to the characteristics of the Reddit Comments Dataset: the comments are typically short, semantically diverse, and contain relatively few meaningful named entities or content-specific terms. These properties make it more difficult for our difference-based enhanced hacking method to effectively extract target tokens.

\subsection{CLERC Dataset}
Finally, we present results for LLaMA‑3.1‑8B, Gemma‑2‑2B, and Qwen2.5‑7B on the CLERC Dataset, following the identical evaluation protocol with two hacking tasks and two hacking schemes.

\begin{figure*}[!htbp]
\setlength{\abovecaptionskip}{6pt}
\centering
\subfloat[Top 30\% on Gemma-2-2B.]{
    \includegraphics[height=7cm, width=0.48\linewidth]{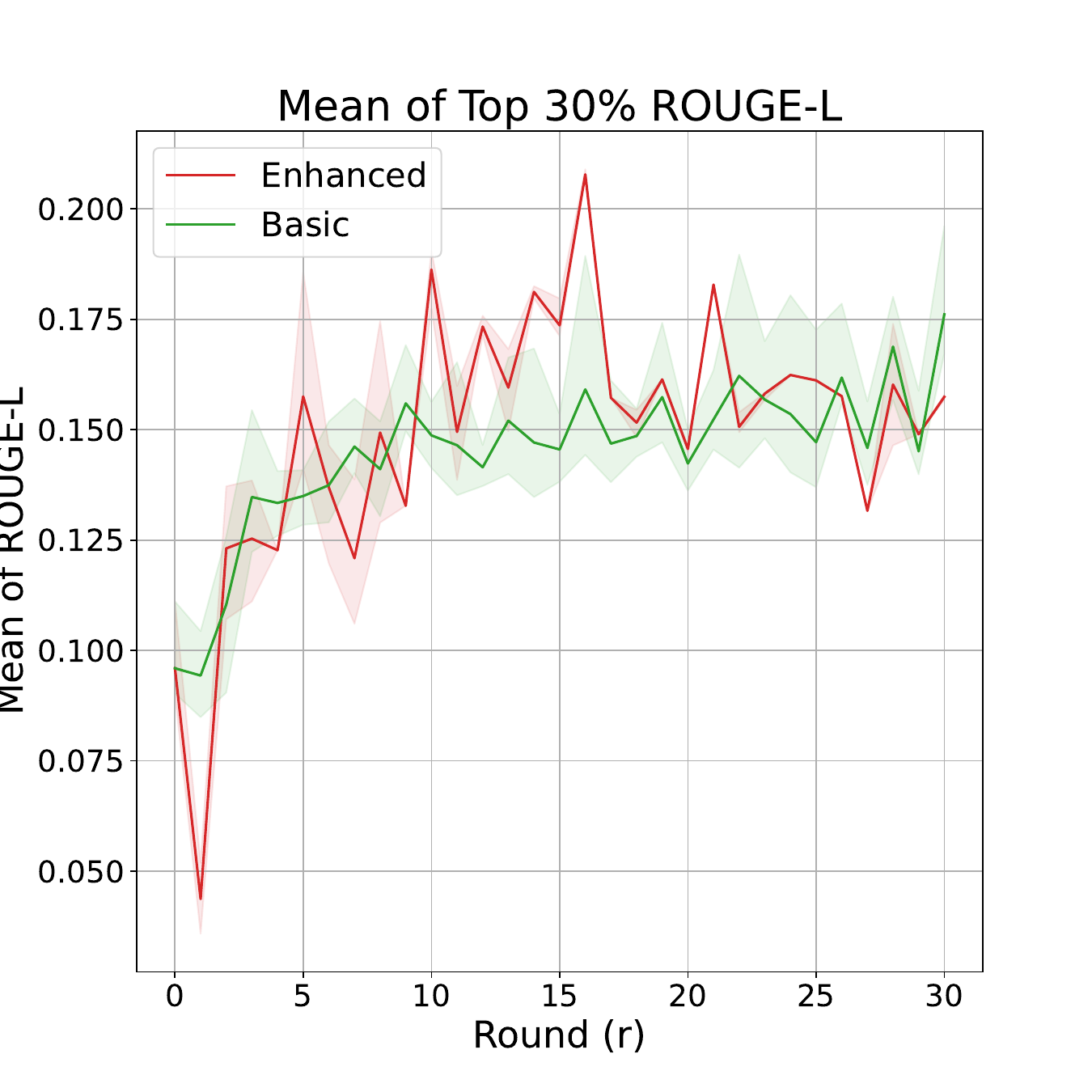}
}
\subfloat[Top 50\% on Gemma-2-2B.]{
    \includegraphics[height=7cm, width=0.48\linewidth]{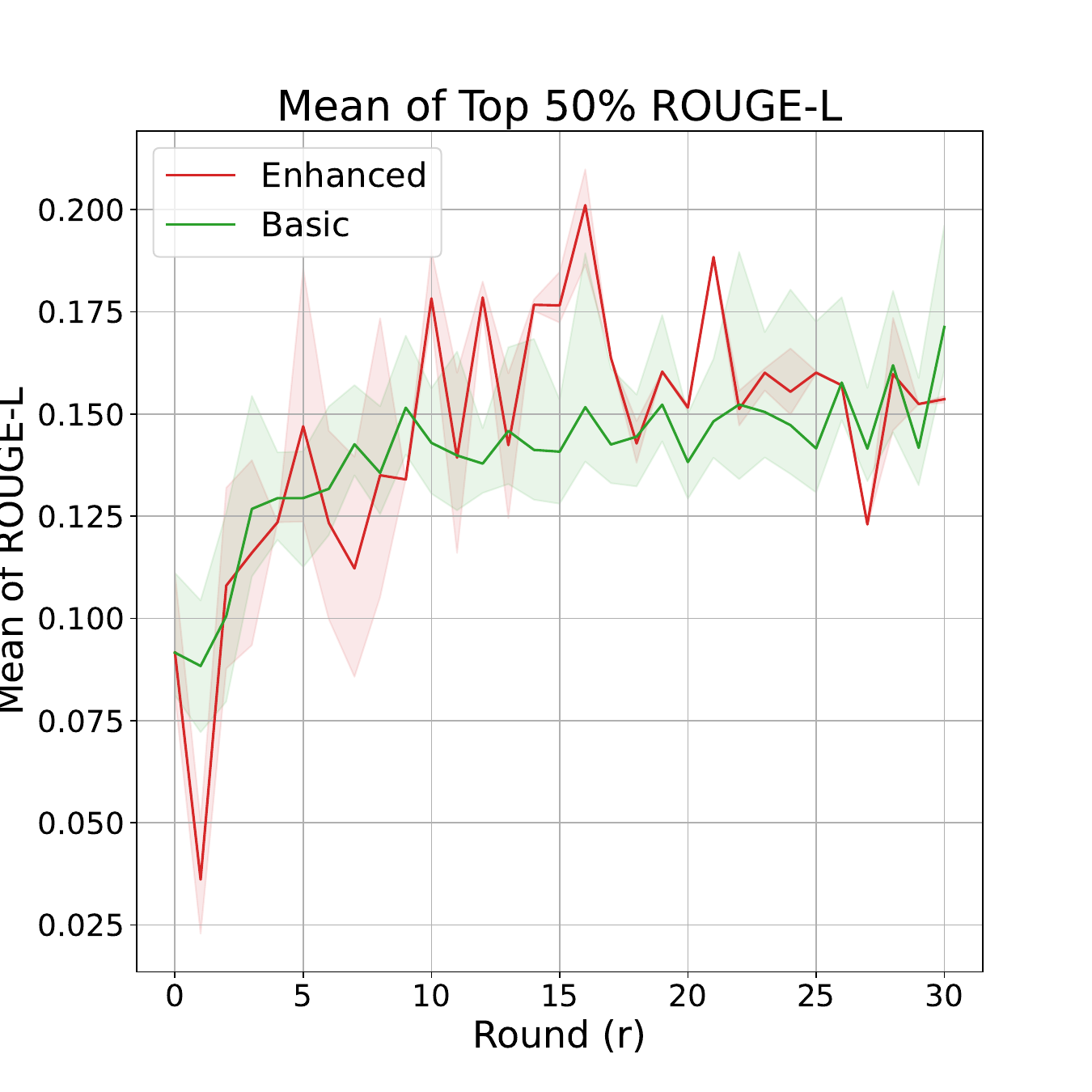}
}
\\
\vspace{-13pt}
\subfloat[Top 30\% on Qwen2.5-7B.]{
    \includegraphics[height=7cm, width=0.48\linewidth]{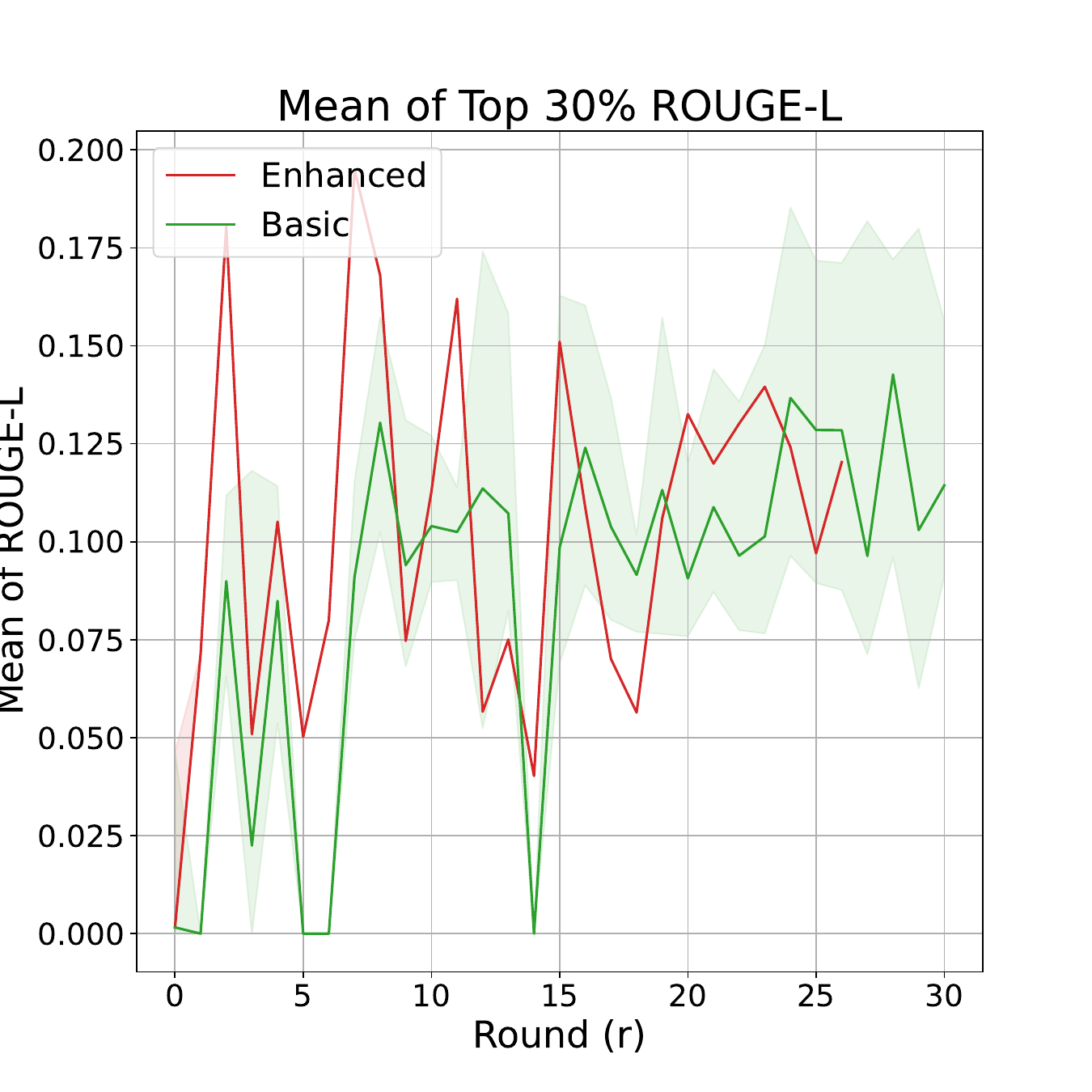}
}
\subfloat[Top 50\% on Qwen2.5-7B.]{
    \includegraphics[height=7cm, width=0.48\linewidth]{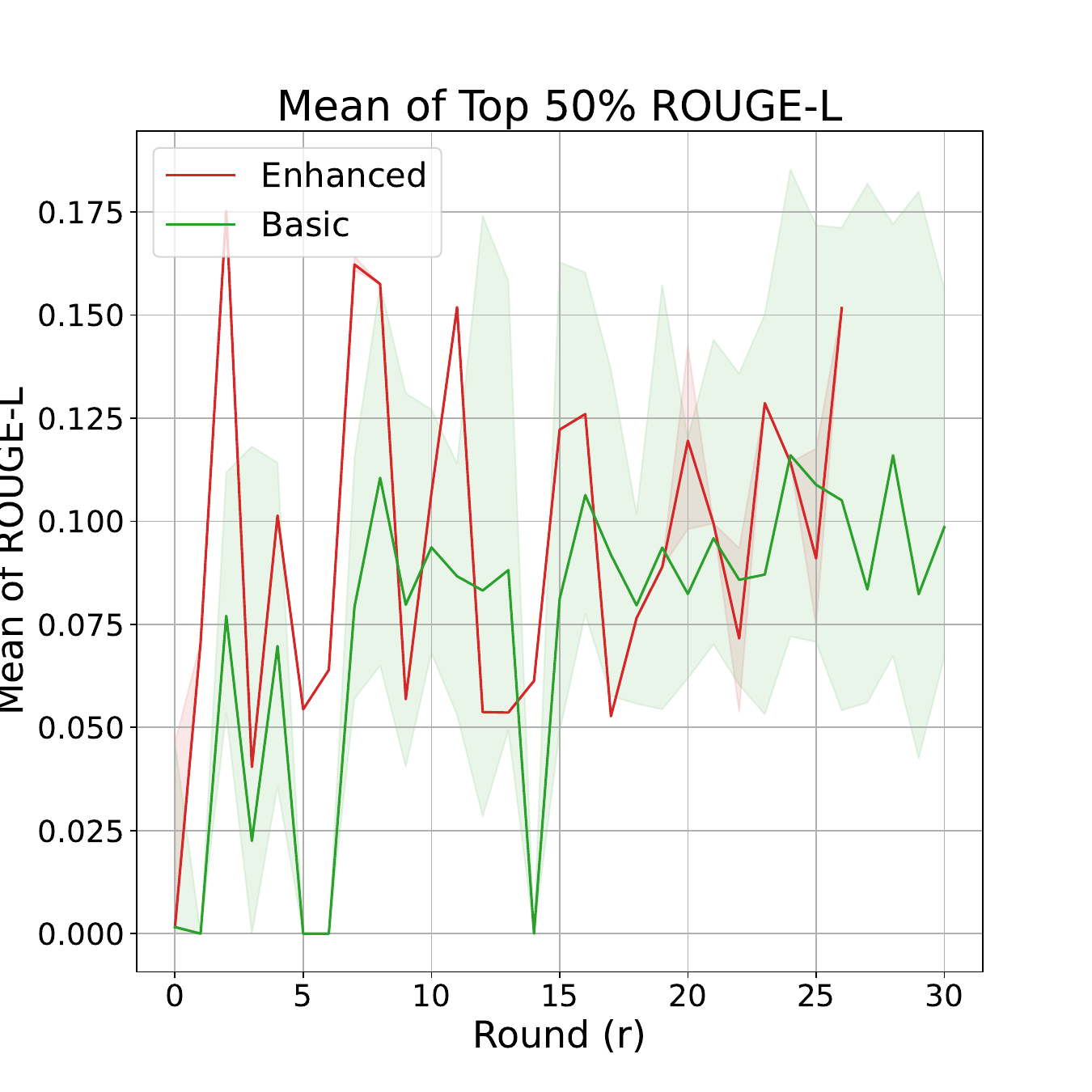}
}
\\
\vspace{-13pt}
\subfloat[Top 30\% on LLaMA-3.1-8B.]{
    \includegraphics[height=7cm, width=0.48\linewidth]{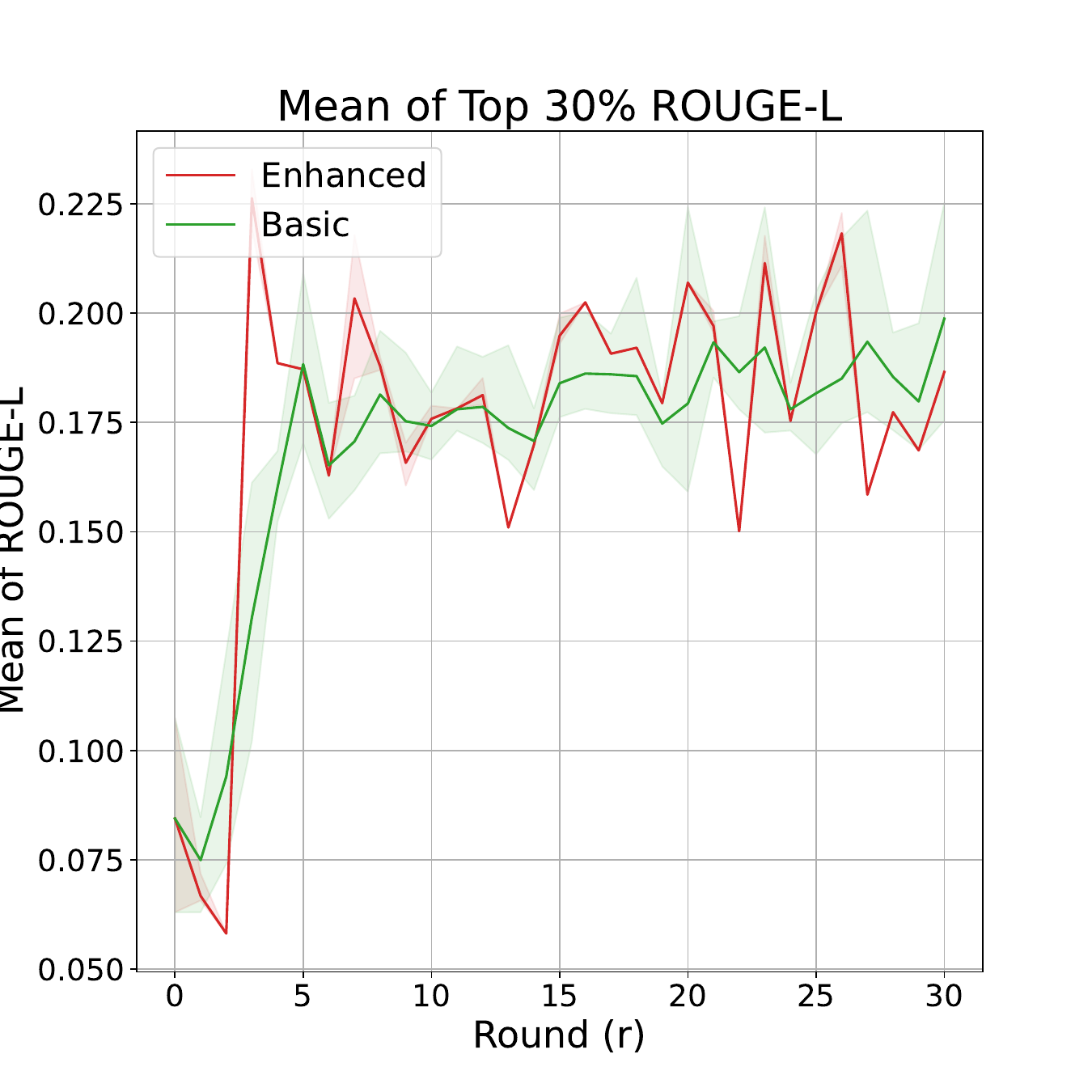}
}
\subfloat[Top 50\% on LLaMA-3.1-8B.]{
    \includegraphics[height=7cm, width=0.48\linewidth]{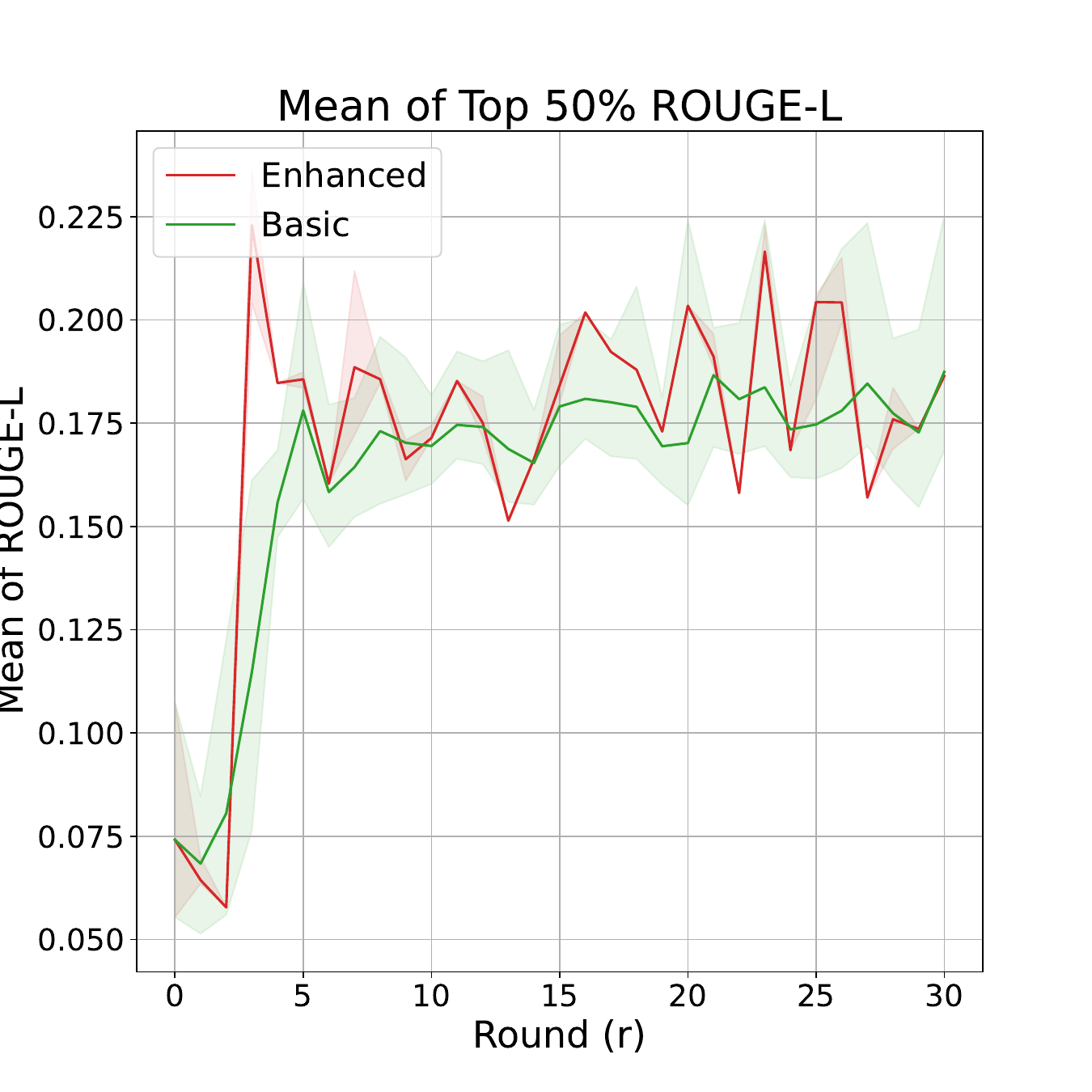}
}
\caption{Comparison of two hacking schemes' results for Zero-Input Generation on CLERC Dataset. Experiments are repeated using Qwen2.5-7B, Gemma-2-2B and LLaMA-3.1-8B.}
\label{fig:more_clerc_sim}
\end{figure*}

\begin{figure*}[!htbp]
\setlength{\abovecaptionskip}{6pt}
\centering
\subfloat[Top 30\% on Gemma-2-2B.]{
    \includegraphics[height=7cm, width=0.48\linewidth]{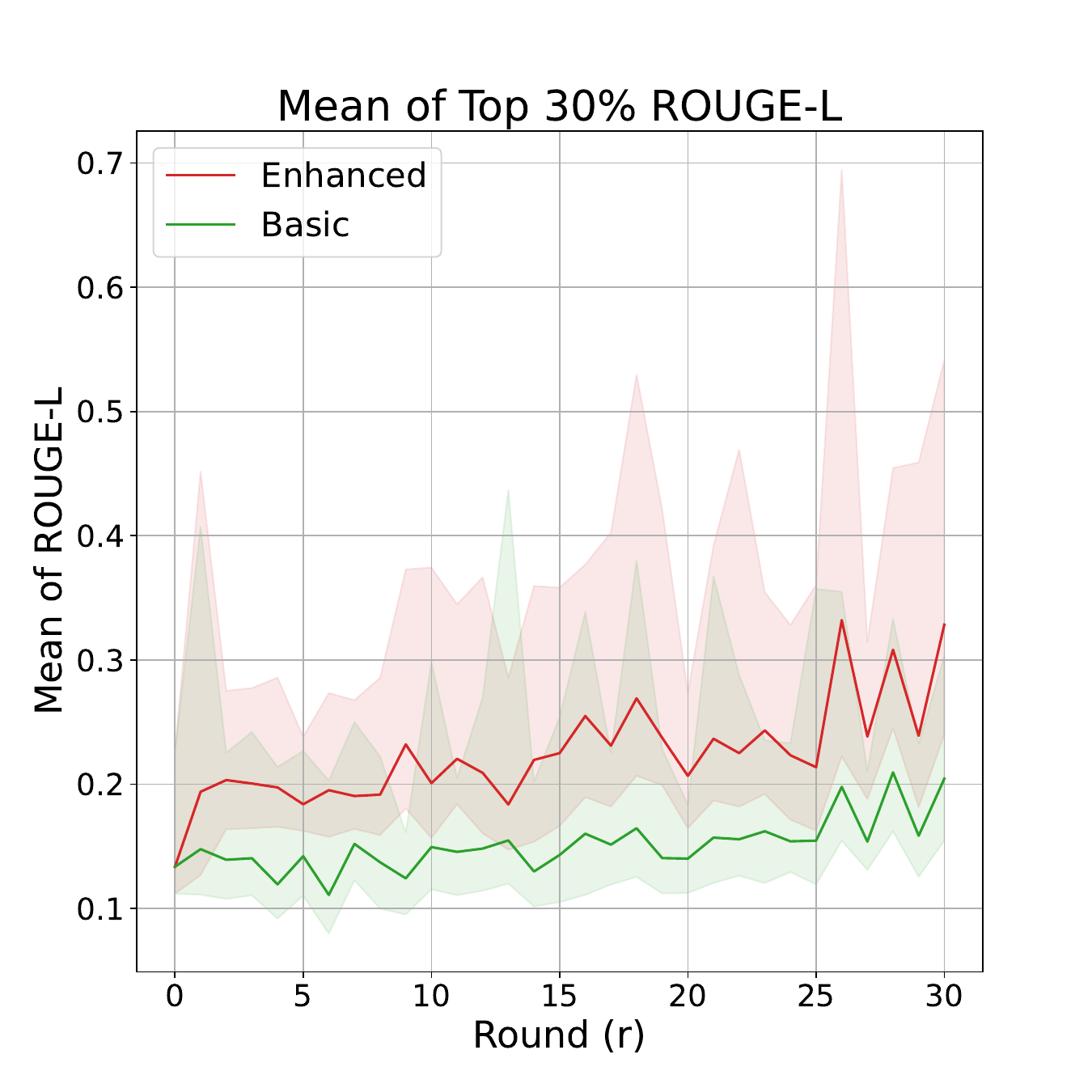}
}
\subfloat[Top 50\% on Gemma-2-2B.]{
    \includegraphics[height=7cm, width=0.48\linewidth]{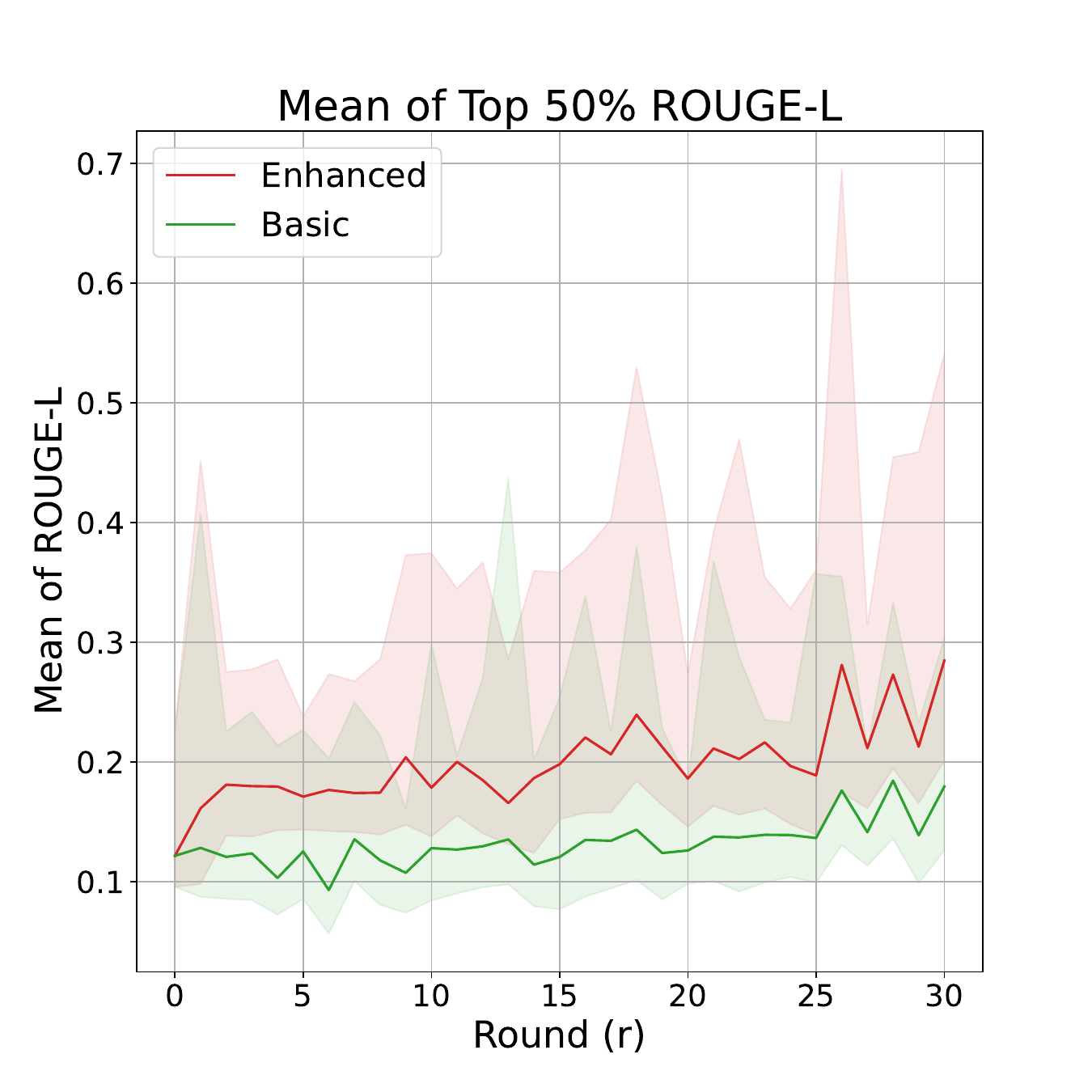}
}
\\
\vspace{-13pt}
\subfloat[Top 30\% on Qwen2.5-7B.]{
    \includegraphics[height=7cm, width=0.48\linewidth]{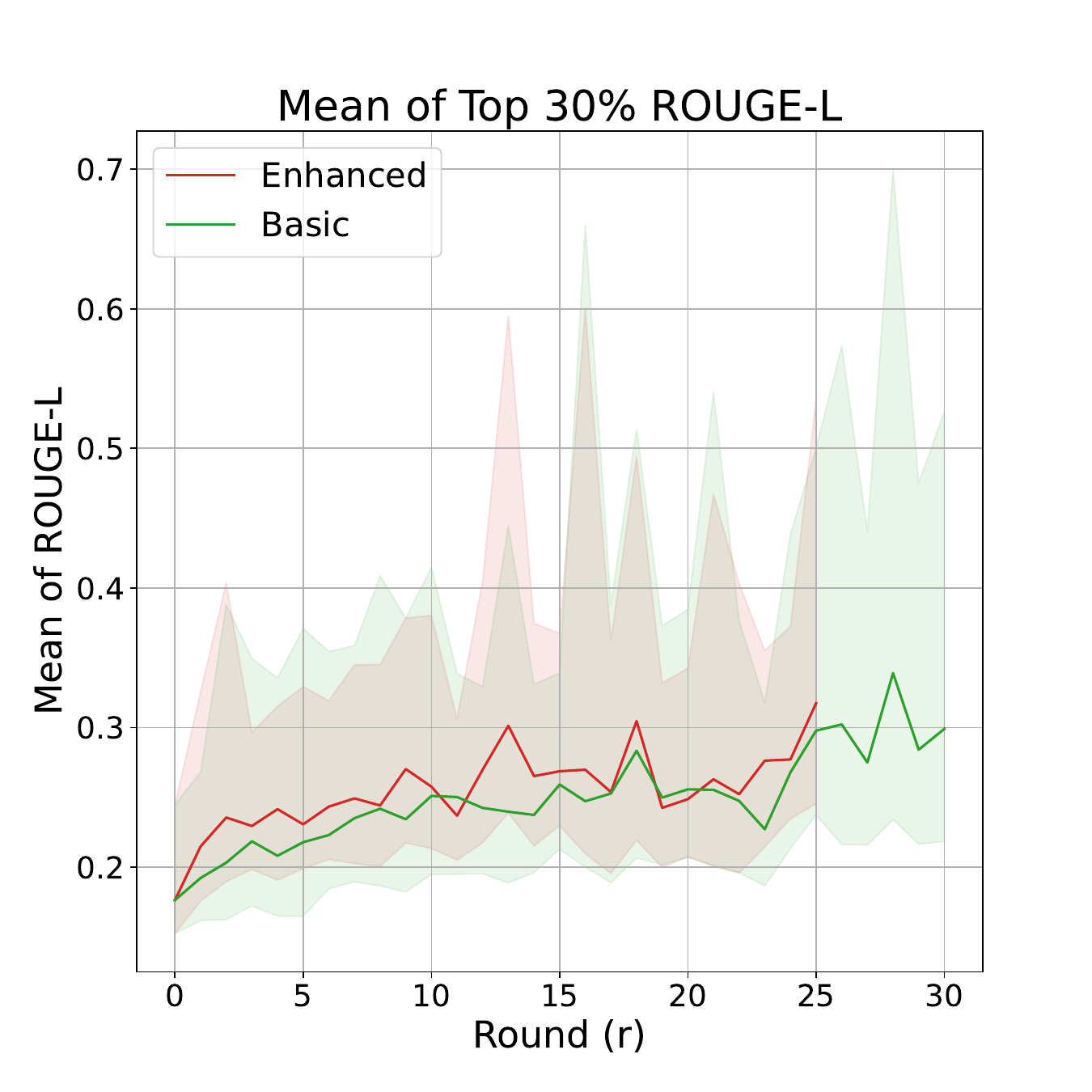}
}
\subfloat[Top 50\% on Qwen2.5-7B.]{
    \includegraphics[height=7cm, width=0.48\linewidth]{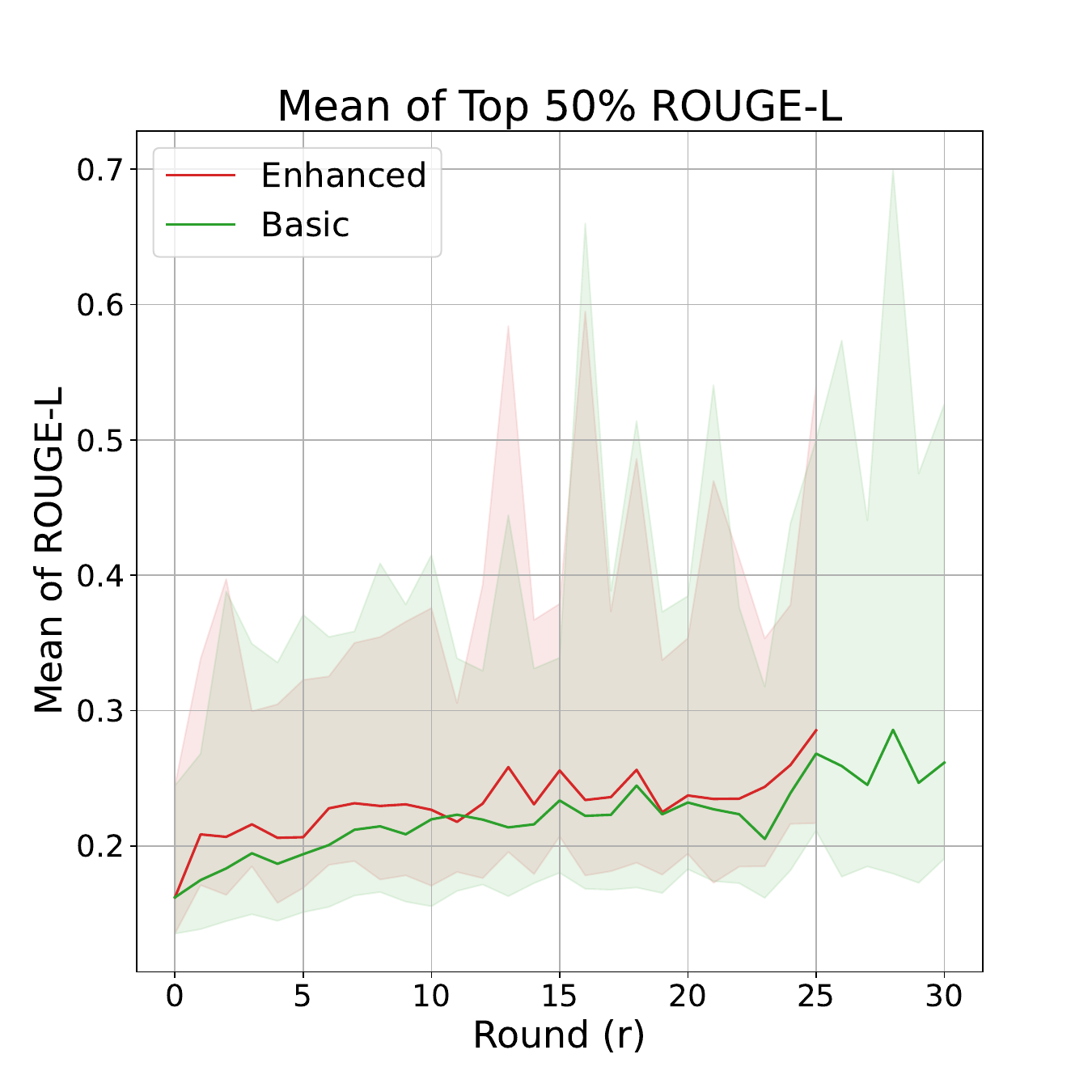}
}
\\
\vspace{-13pt}
\subfloat[Top 30\% on LLaMA-3.1-8B.]{
    \includegraphics[height=7cm, width=0.48\linewidth]{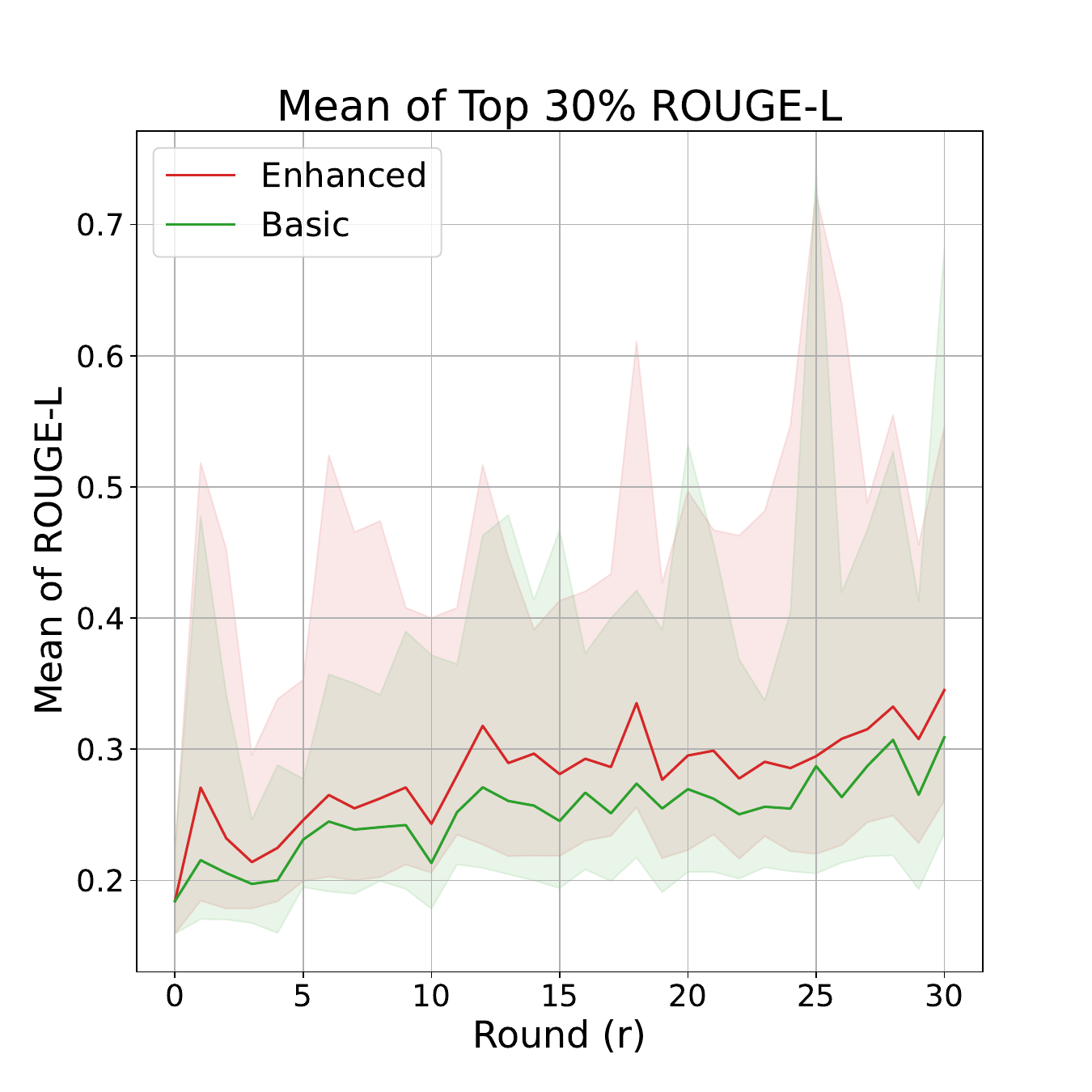}
}
\subfloat[Top 50\% on LLaMA-3.1-8B.]{
    \includegraphics[height=7cm, width=0.48\linewidth]{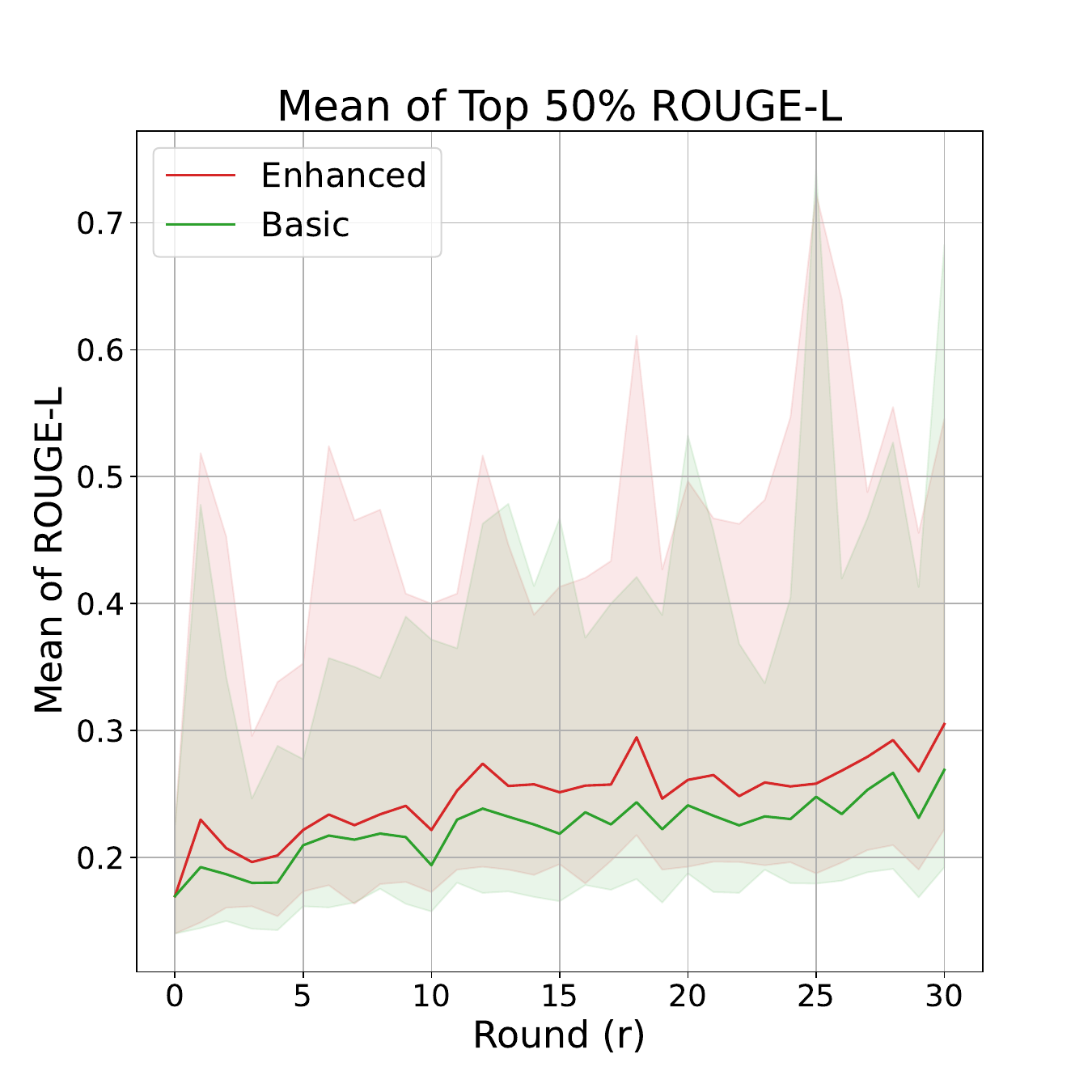}
}
\caption{Comparison of two hacking schemes' results for Partial-Input Completion on CLERC Dataset. Experiments are repeated using Qwen2.5-7B, Gemma-2-2B and LLaMA-3.1-8B.}
\label{fig:more_clerc_com}
\end{figure*}

As shown in Fig.~\ref{fig:more_clerc_sim} and ~\ref{fig:more_clerc_com}, both models already reveal substantial privacy leakage in the basic hacking scheme. And the enhanced hacking scheme amplifies this leakage even further.

\section{Different Percents of Input for the Partial-Input Completion}
\label{appendix:input}
\begin{figure*}[!htbp]
\setlength{\abovecaptionskip}{6pt}
\centering
\subfloat[Top 10\% ROUGE-L score with 30\% Input.]{
    \includegraphics[height=7cm, width=0.48\linewidth]{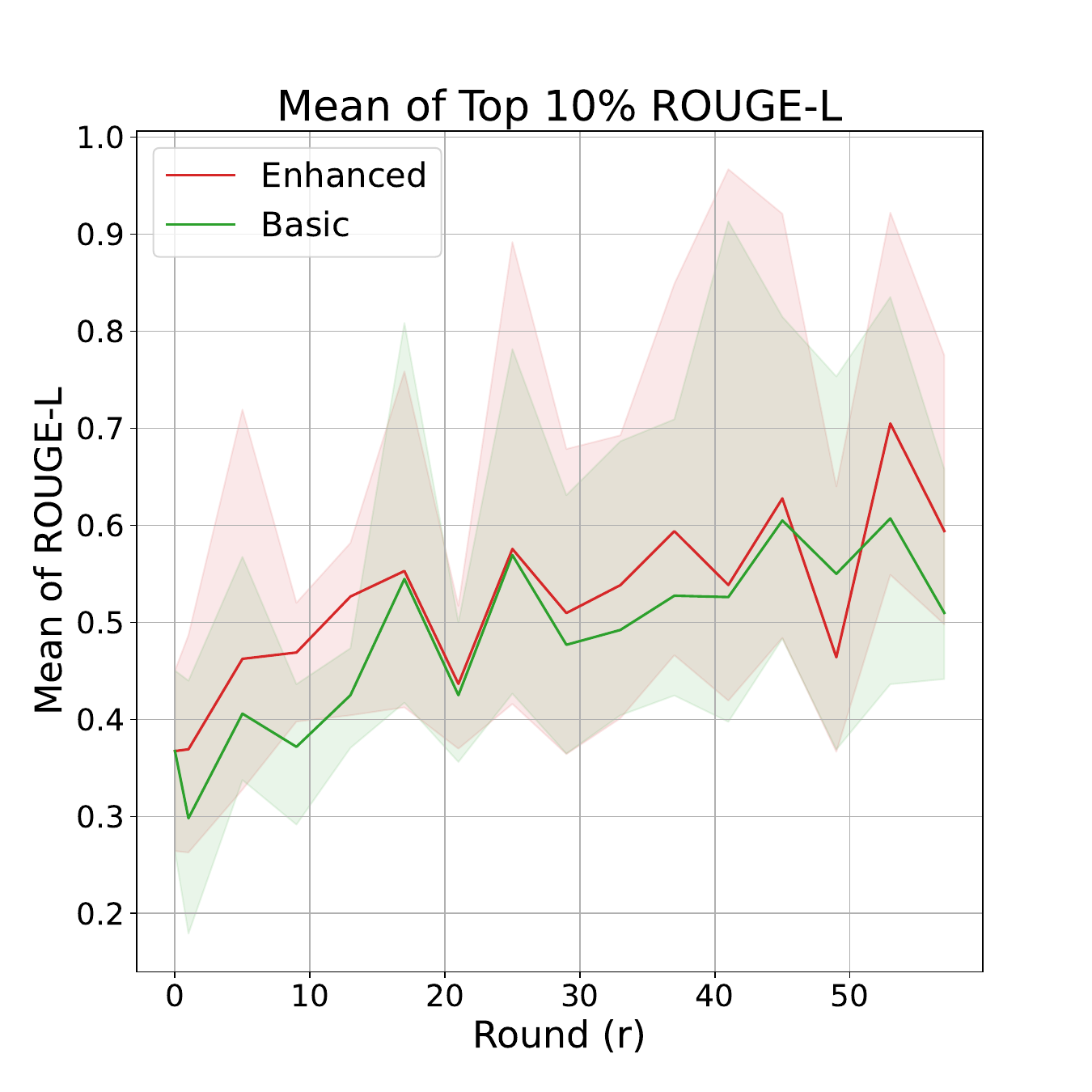}
}
\subfloat[Top 10\% ROUGE-L score with 80\% Input.]{
    \includegraphics[height=7cm, width=0.48\linewidth]{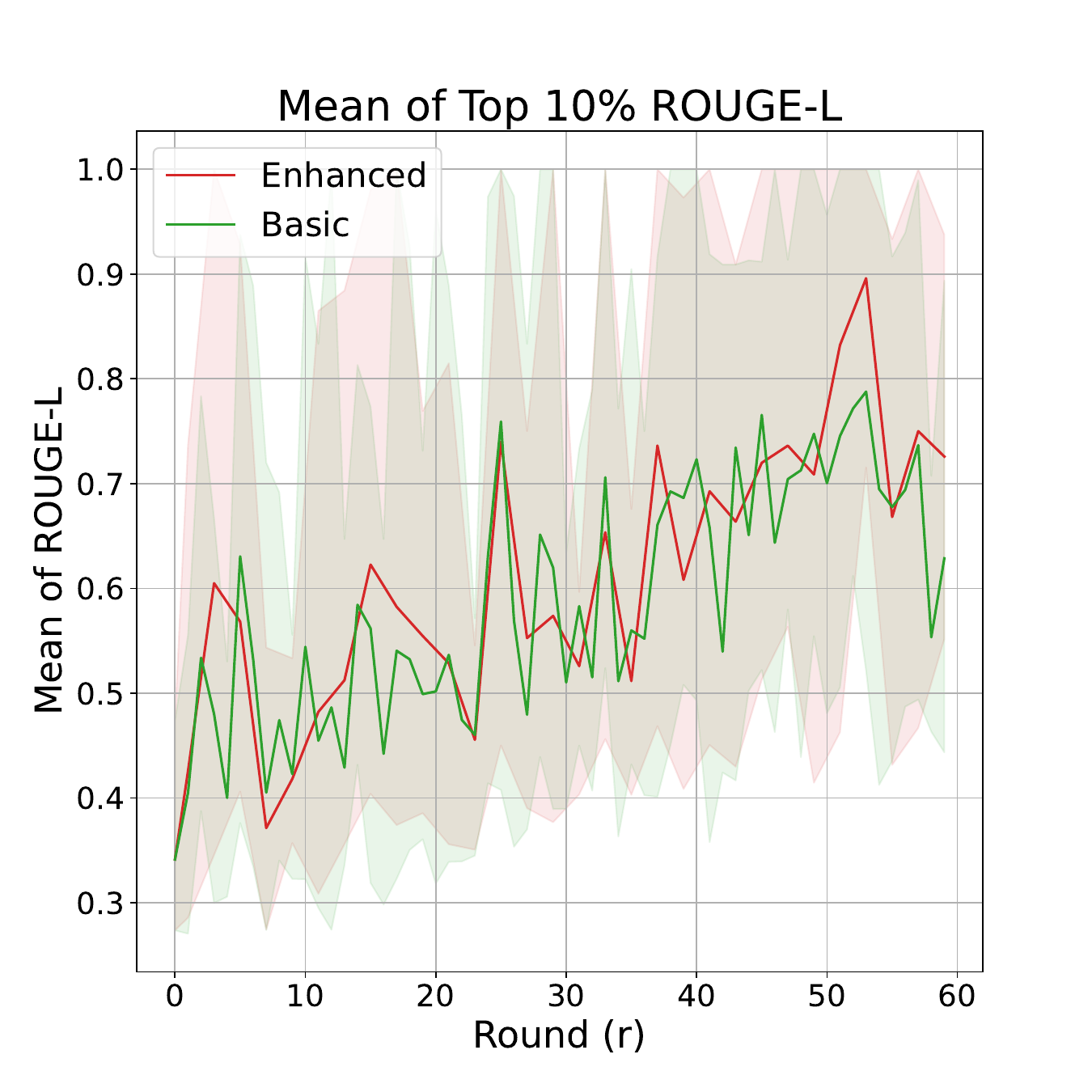}
}
\\
\subfloat[Top 30\% ROUGE-L score with 30\% Input.]{
    \includegraphics[height=7cm, width=0.48\linewidth]{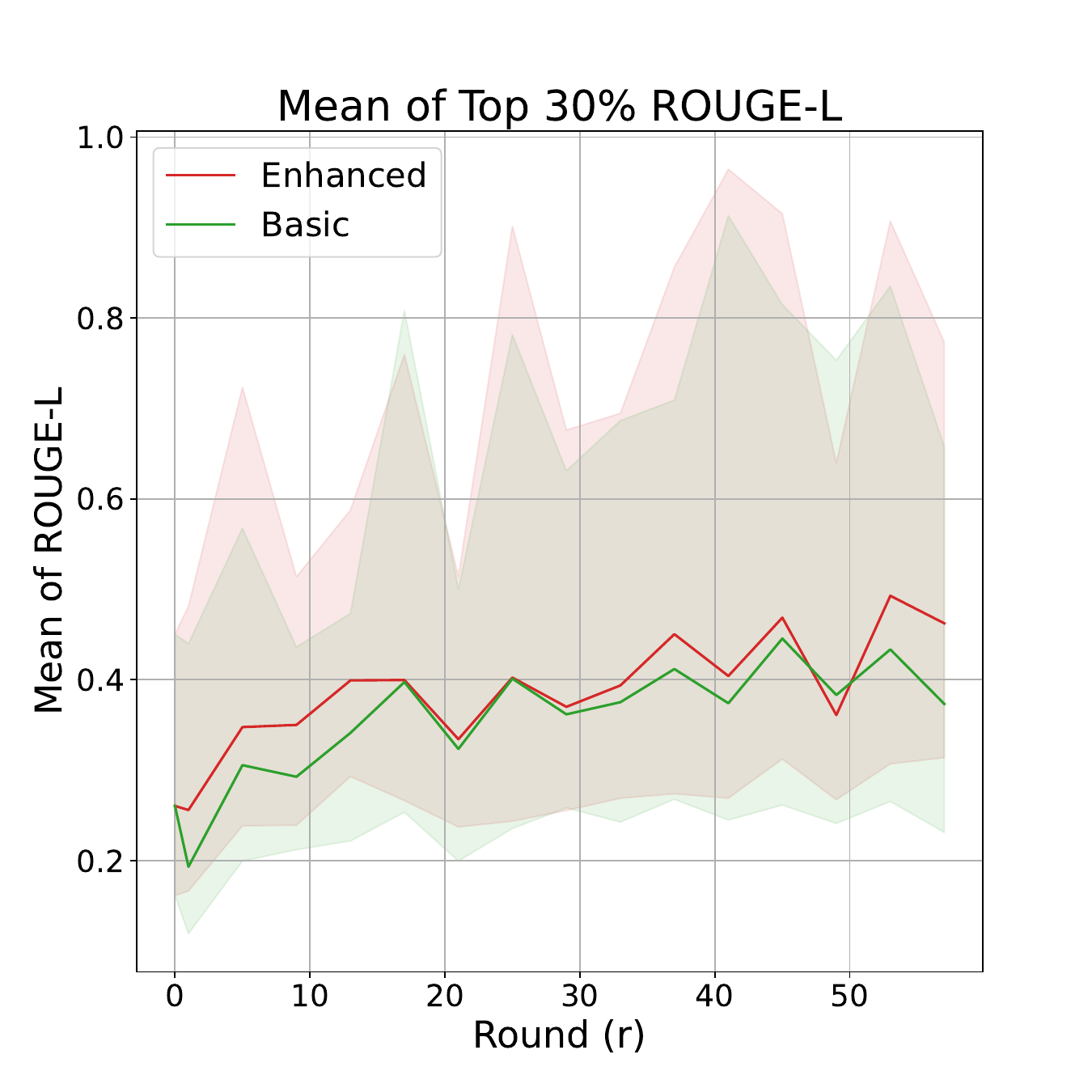}
}
\subfloat[Top 30\% ROUGE-L score with 80\% Input.]{
    \includegraphics[height=7cm, width=0.48\linewidth]{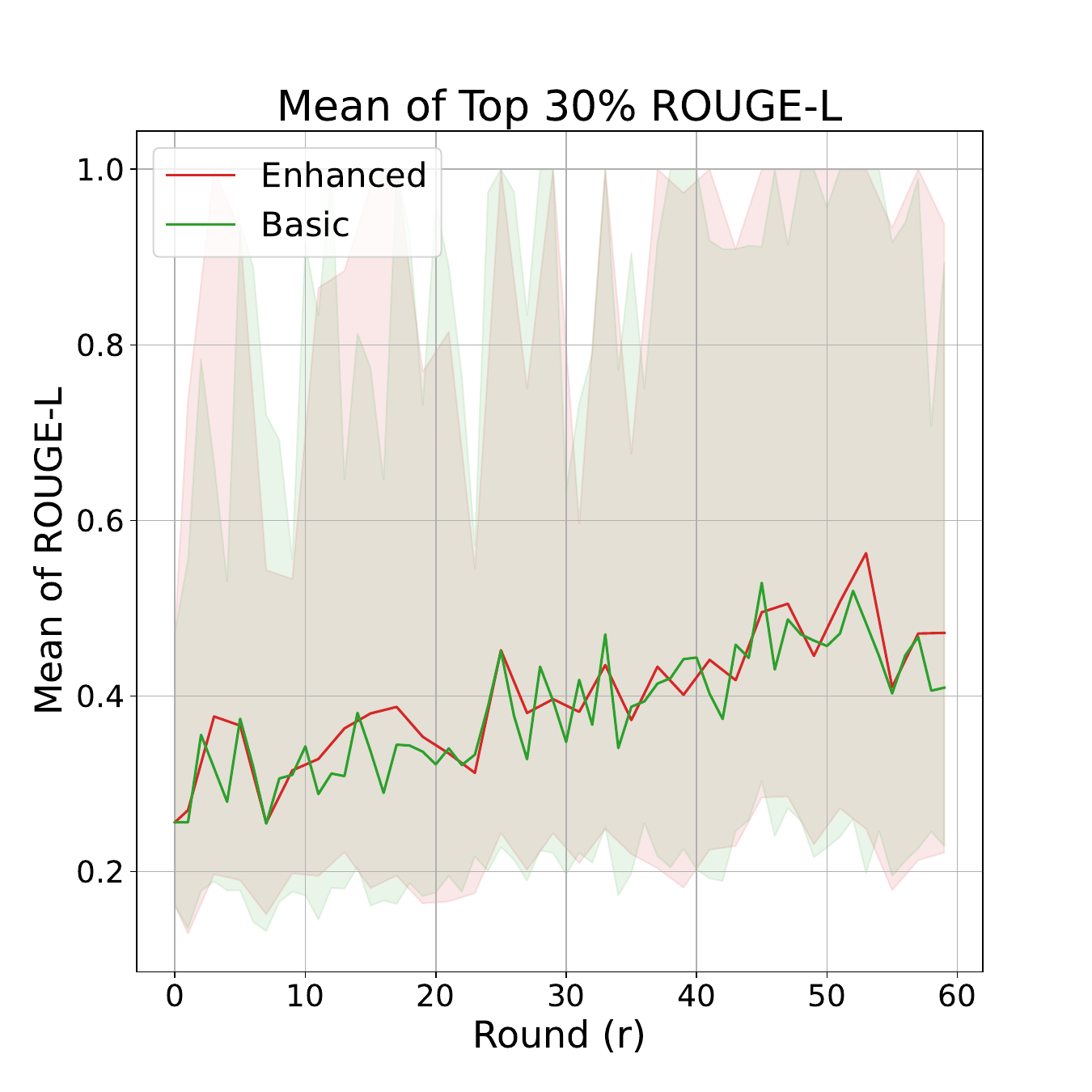}
}
\caption{Comparison of different input amount results for Partial-Input Completion on LLaMA-3.1-8B model and Enron Email Dataset.}
\label{fig:input}
\vspace{-20pt}
\end{figure*}

In section~\ref{section:com}, we provide the first 80\% as input and let the model complete the remaining part for each original training sample in Partial-Input Completion. In this section, we also conduct experiments on LLaMA-3.1-8B model using 30\% of the input in the Partial-Input Completion scenario to better understand the impact of varying the input amount. The experimental results are shown in Fig.~\ref{fig:input}, indicating that even with 30\% input, the attack remains effective, though worse than with 80\% input. The enhanced scheme improves reconstruction quality in both cases.

\section{Use of AI Assistants}
In this paper, we use AI assistants only for language polishing purposes. No original content was generated by the AI, and all code implementations were completed entirely by human.

\end{document}